\def\RevisionMode{0}
\documentclass[a4paper,fleqn]{cas-dc}

\usepackage[utf8]{inputenc}
\usepackage[T1]{fontenc}
\usepackage{microtype}
\usepackage{amsmath}
\usepackage{amssymb}
\usepackage{amsfonts}

\usepackage[switch,mathlines]{lineno}

\usepackage{graphicx}
\usepackage{tikz}
\usetikzlibrary{arrows.meta,positioning,calc}
\tikzset{sciencebox/.style={draw=black!65,rounded corners=2pt,align=center,font=\normalfont\footnotesize,inner sep=5pt},sciencearrow/.style={-{Latex[length=2mm]},thick,draw=black!70}}
\usepackage{booktabs}
\usepackage{multirow}
\usepackage{array}
\usepackage{tabularx}
\usepackage{makecell}
\usepackage{siunitx}
\usepackage{subcaption}
\providecommand{\RevisionMode}{1}
\usepackage{xcolor}
\definecolor{reviewblue}{RGB}{0,51,170}
\definecolor{reviewred}{RGB}{190,0,0}
\ifnum\RevisionMode=1\relax
  \colorlet{revisionblue}{reviewblue}
  \colorlet{secondrevision}{reviewblue}
\else
  \colorlet{revisionblue}{black}
  \colorlet{secondrevision}{black}
  \colorlet{blue}{black}
\fi
\DeclareRobustCommand{\rev}[1]{{\color{revisionblue}#1}}
\newenvironment{revblock}{\begingroup\color{revisionblue}}{\par\endgroup}
\newcommand{\ValidationRevisionColor}{\color{revisionblue}}
\DeclareRobustCommand{\revsecond}[1]{{\color{secondrevision}#1}}

\definecolor{reviewgreen}{RGB}{0,110,55}
\ifnum\RevisionMode=1\relax
  \colorlet{thirdrevision}{reviewblue}
\else
  \colorlet{thirdrevision}{black}
\fi
\DeclareRobustCommand{\revgreen}[1]{{\color{thirdrevision}#1}}

\usepackage{etoolbox}
\newcommand{\AgriSenseBibColor}[1]{\ifcsname AgriSenseGreenBib@#1\endcsname\color{thirdrevision}\else\color{black}\fi}
\AtBeginEnvironment{thebibliography}{%
  \let\AgriSenseOriginalBibitem\bibitem
  \renewcommand{\bibitem}[2][]{%
    \if\relax\detokenize{#1}\relax
      \AgriSenseOriginalBibitem{#2}%
    \else
      \AgriSenseOriginalBibitem[#1]{#2}%
    \fi
    \AgriSenseBibColor{#2}}}
\expandafter\def\csname AgriSenseGreenBib@abanay2022agrieco\endcsname{1}
\expandafter\def\csname AgriSenseGreenBib@awais2024agrogpt\endcsname{1}
\expandafter\def\csname AgriSenseGreenBib@brohan2023rt2\endcsname{1}
\expandafter\def\csname AgriSenseGreenBib@cuaran2026horticulturalplanning\endcsname{1}
\expandafter\def\csname AgriSenseGreenBib@devlin2019bert\endcsname{1}
\expandafter\def\csname AgriSenseGreenBib@dosovitskiy2021vit\endcsname{1}
\expandafter\def\csname AgriSenseGreenBib@elamraoui2024strawberry_detection\endcsname{1}
\expandafter\def\csname AgriSenseGreenBib@gund2025digital_twin_agriculture\endcsname{1}
\expandafter\def\csname AgriSenseGreenBib@he2016resnet\endcsname{1}
\expandafter\def\csname AgriSenseGreenBib@he2023pointlio\endcsname{1}
\expandafter\def\csname AgriSenseGreenBib@isaacsimofficial\endcsname{1}
\expandafter\def\csname AgriSenseGreenBib@jiang2022greenhouse_navigation\endcsname{1}
\expandafter\def\csname AgriSenseGreenBib@kim2025openvla\endcsname{1}
\expandafter\def\csname AgriSenseGreenBib@mohanty2016plantvillage\endcsname{1}
\expandafter\def\csname AgriSenseGreenBib@oquab2023dinov2\endcsname{1}
\expandafter\def\csname AgriSenseGreenBib@plantseg2025\endcsname{1}
\expandafter\def\csname AgriSenseGreenBib@radford2021clip\endcsname{1}
\expandafter\def\csname AgriSenseGreenBib@salman2026agribt\endcsname{1}
\expandafter\def\csname AgriSenseGreenBib@tan2019efficientnet\endcsname{1}
\expandafter\def\csname AgriSenseGreenBib@tectonic2026release0170\endcsname{1}
\expandafter\def\csname AgriSenseGreenBib@zhao2025agrivln\endcsname{1}

\usepackage{capt-of}
\usepackage{etoolbox,placeins}

\ExplSyntaxOn
\cs_new_eq:NN \revision_original_figcaption:nn \__make_fig_caption:nn
\cs_set_protected:Npn \__make_fig_caption:nn #1#2
  { \group_begin: \cs_set_eq:NN \sffamily \rmfamily
    \revision_original_figcaption:nn {#1}{#2} \group_end: }
\patchcmd{\__make_tbl_caption:nn}{\sffamily}{\rmfamily}{\typeout{FONTFIX|table|1}}{\PackageError{revision}{Table caption font patch failed}{}}
\ExplSyntaxOff
\AtBeginEnvironment{tabularx}{\rmfamily}
\AtBeginEnvironment{tabular}{\rmfamily}
\usepackage{longtable,placeins,pdflscape,needspace,fancyhdr}
\usepackage{algorithm}
\usepackage{algpseudocode}

\usepackage{enumitem}
\usepackage{xurl}
\usepackage{refcount}

\usepackage[authoryear,round]{natbib}
\setcitestyle{authoryear,round,semicolon,aysep={,},yysep={;}}

\newcommand{\supppanel}[2]{%
\begin{minipage}[t]{0.47\textwidth}
    \centering\rmfamily
    \includegraphics[width=\linewidth]{Figures/#1}
\end{minipage}%
}

\newcommand{\AgriSense}{AgriSense}
\newcommand{\VLDA}{VLDA}

\newcommand{\vect}[1]{\boldsymbol{#1}}

\providecommand{\URLprefix}{\space URL:\space}
\hypersetup{colorlinks=true,linkcolor=black,citecolor=black,urlcolor=black}
\newcolumntype{Y}{>{\raggedright\arraybackslash}X}
\newcolumntype{Z}{>{\centering\rmfamily\arraybackslash}X}
\newcommand{\Supplement}[1]{Supplementary Material, Section~S#1}
\newcommand{\SuppFig}[1]{Supplementary Figure~\csname SuppFig@#1\endcsname}
\NewDocumentCommand{\HawaiiInstitution}{}{University of Hawai\textquoteleft{}i at M\={a}noa}

\expandafter\def\csname SuppFig@fig:vlda_zone_response\endcsname{S13}
\expandafter\def\csname SuppFig@fig:trajectory_error_results\endcsname{S15}
\expandafter\def\csname SuppFig@fig:environment_delta_map\endcsname{S16}
\expandafter\def\csname SuppFig@fig:plant_level_intervention\endcsname{S17}
\expandafter\def\csname SuppFig@fig:app_ui\endcsname{S18}

\let\AgriOriginalIncludeGraphics\includegraphics
\renewcommand{\includegraphics}[2][]{%
  \IfFileExists{#2}{\AgriOriginalIncludeGraphics[#1]{#2}}{%
    \begingroup\setlength{\fboxsep}{8pt}%
    \fbox{\parbox{0.80\linewidth}{\centering\footnotesize
      Missing figure asset: \texttt{\detokenize{#2}}}}%
    \endgroup
  }%
}

\begin{document}
\raggedbottom
\shorttitle{AgriSense}
\shortauthors{Saad Saoud et al.}

\title[mode=title]{AgriSense: \rev{Vision--Language} \rev{Decision} \rev{Support} \rev{and} Robotic \rev{Intervention in a Virtual} Greenhouse}

\author{Lyes {Saad Saoud}}

\author[2]{Oualid Doukhi}
\author[3]{Reza Ghorbani}
\author[4]{Moussa Ayyash}
\affiliation[2]{
    organization={Jeonbuk National University},
    addressline={Deokjin-gu},
    city={Jeonju-si},
    state={Jeollabuk-do},
    country={Republic of Korea}
}
\affiliation[3]{
    organization={\HawaiiInstitution},
    city={Honolulu},
    state={Hawaii},
    country={United States}
}
\affiliation[4]{
    organization={Chicago State University},
    city={Chicago},
    state={Illinois},
    country={United States}
}


\begin{abstract}
Autonomous greenhouse management is moving beyond static sensing and conventional control toward embodied systems that can perceive, navigate, regulate, and intervene at plant level. However, coupling \rev{multimodal} \rev{predictions} \rev{to robotic} and \rev{environmental-control} \rev{interfaces requires explicit output definitions} and \rev{task-specific} \rev{evaluation}. This paper presents \AgriSense{}, a framework \rev{integrating} an Isaac Sim \rev{virtual} greenhouse, a quadruped mobile manipulator, RGB and \rev{LiDAR--inertial} sensing, \rev{environmental-reference} execution, and a supervised vision--language decision-to-action (\VLDA{}) module. \rev{Frozen} \rev{DINOv2} \rev{and} \rev{BERT} \rev{encoders} \rev{and} \rev{a} \rev{structured} environmental \rev{encoder feed a concatenation-based network that jointly proposes twelve environmental references, a target zone}, and \rev{a} \rev{binary} \rev{intervention} \rev{flag.} \rev{Checkpoint replay reproduces 41/43 target-zone decisions on the original split. Five-seed retraining on a source-grouped test set gives 43/44 correct decisions \rev{in each seed}} for \rev{the} \rev{full}, \rev{image-removed,} \rev{and language-removed configurations; thus an incremental multimodal zone-accuracy benefit is not established. The full model's} temperature, humidity, \rev{and} soil-moisture mean absolute errors \rev{are} 0.\rev{249~$^\circ$C, 2.030} and 0.\rev{997 percentage points, respectively. The sensor-only condition also matches full zone accuracy, while language-only reaches 68.64\%. Separate real-image benchmarks reveal a substantial distribution-shift limitation. Five offline Point-LIO routes give rigidly aligned trajectory errors of 5.22--17.32}~\rev{mm; 32 accepted virtual-arm targets meet geometric criteria, with eight boundary rejections. Six long-horizon climate cases and distinct fault protocols further characterize integration limits. Navigation and integrated treatment use simulator-state inputs; physical greenhouse accuracy and agronomic treatment efficacy are not established}. The \rev{contribution} \rev{is} \rev{a} \rev{reproducible} multimodal \rev{decision-support} \rev{interface} and \rev{its} \rev{component-level} \rev{characterization} \rev{in} a \rev{virtual} greenhouse\rev{,} \rev{rather} \rev{than a field-validated autonomous treatment system}.

\end{abstract}


\begin{keywords}
Agricultural robotics \sep
\rev{Virtual greenhouse} \sep
\rev{Multimodal decision support} \sep
LiDAR--IMU odometry \sep
\rev{Experimental evaluation}
\end{keywords}
\maketitle

\section{Introduction}
\label{sec:introduction}

Protected agriculture seeks to improve crop productivity while reducing water, energy, labor, and chemical inputs. Modern smart greenhouses already employ fixed sensors and automated controllers to regulate temperature, humidity, soil moisture, lighting, ventilation, and irrigation. {Recent reviews in greenhouse AI, sensory-data integration, and greenhouse temperature prediction/control further emphasize plant-centric intelligence, multimodal sensing, and reliable embedded execution as active research directions \citep{alqudah2025smartgreenhouses,yu2025greenhousetemp}.} However, most existing systems remain limited to greenhouse- or zone-level monitoring and regulation. \rev{Zone-level monitoring does not by itself provide a spatially localized inspection and intervention workflow. Such a workflow must associate observations with management zones and separate environmental references from plant-level treatment proposals.}

Mobile robots have therefore been investigated for greenhouse mapping, row following, crop inspection, fruit detection, and pesticide spraying
{\citep{sanchezmolina2024robotics,jiang2022greenhouse_navigation,abanay2022agrieco,canadas2024collaborative,cheng2024greenhouse_fusion,elamraoui2024strawberry_detection,tan2025greenhousesprayingnav}}.
These studies demonstrate the potential of agricultural robotics, but many existing systems remain restricted to wheeled platforms, predefined task logic, isolated perception modules, or navigation-only demonstrations. Perception, spatial grounding, environmental regulation, and physical intervention are often evaluated independently rather than as components of a unified autonomy pipeline.

Agricultural digital twins are also increasingly used for greenhouse monitoring, prediction, visualization, and decision support
{\citep{ariesen2022digitaltwins,gund2025digital_twin_agriculture,lang2025digitaltwinharvesting}}.
However, many of these platforms represent the greenhouse as a passive monitored process rather than an embodied robotic environment. In contrast, \rev{an embodied virtual greenhouse must represent} robot motion, RGB and LiDAR observations, inertial sensing, local map construction, greenhouse-zone association, actuator feedback, and {plant-level spray-intervention workflow.} Isaac Sim is particularly suitable for this purpose because it supports physically based robotics simulation, multimodal sensor generation, robot--environment interaction, and compatibility with robot-learning workflows
\citep{nvidia_isaac_sim,nvidia_isaac_lab,isaac_sdg_docs}.

Recent embodied multimodal models have shown that visual, textual, and continuous state inputs can support robot decision-making. PaLM-E demonstrated embodied multimodal reasoning from language, vision, and state-estimation inputs
\citep{driess2023palme}, while RT-2 and OpenVLA connected vision--language representations with robotic actions
\citep{brohan2023rt2,kim2025openvla}.
Nevertheless, directly mapping an unconstrained semantic-model output to greenhouse actuators would be unsafe. Navigation goals must be consistent with the local map, environmental setpoints must satisfy agronomic and actuator bounds, and plant-level interventions must satisfy confidence, freshness, reachability, and safety requirements. This motivates a hierarchical architecture in which multimodal reasoning generates structured decision proposals, while {deterministic navigation, PI/PID reference-tracking controllers, and predefined manipulator primitives remain responsible for physical execution.}

\label{subsec:motivation_gap}

\rev{The research question is how to specify an image--text--environment proposal interface for greenhouse tasks and determine which inputs contribute to its predictions. AgriSense addresses this question through a three-output supervised model and grouped modality comparisons, with separate virtual experiments characterizing localization, environmental response, targeting and command admission. The model's role is decision support; the supporting interfaces are studied within a common spatial setting.}

\subsection{\rev{Related work and positioning}}
\label{sec:related}
Smart greenhouse systems commonly combine distributed sensors with rule-based, PI/PID, model-predictive, or learning-based controllers to regulate temperature, humidity, soil moisture, lighting, ventilation, and irrigation {\citep{alqudah2025smartgreenhouses,yu2025greenhousetemp,ding2018mpcagriculture}.} These systems are effective for maintaining greenhouse- or zone-level microclimate conditions, but their sensing and actuation are generally spatially coarse. Measurements are typically obtained from fixed sensor nodes, while control actions are applied to an entire greenhouse compartment or management zone.

This architecture is suitable for broad environmental regulation but is less effective for localized conditions such as leaf disease, pest infestation, fruit-level anomalies, uneven water stress, or isolated crop damage. Addressing these conditions requires more than accurate sensing: the system must acquire close inspection views, associate each observation with a physical location, determine the appropriate response, and execute a localized action. In \AgriSense{}, zone-level environmental regulation is therefore complemented by {mobile robotic inspection and plant-level spray-intervention workflow.}

Mobile agricultural robots have been investigated for greenhouse mapping, crop inspection, row following, fruit detection, harvesting, and pesticide spraying {\citep{sanchezmolina2024robotics,canadas2024collaborative,fujinaga2025highbed,tan2025greenhousesprayingnav}.} Representative systems combine 3D LiDAR perception with 2D LiDAR SLAM for greenhouse navigation \citep{jiang2022greenhouse_navigation}, use LiDAR-based row traversal to support strawberry-greenhouse spraying \citep{abanay2022agrieco}, and apply multi-sensor fusion to greenhouse positioning and navigation \citep{cheng2024greenhouse_fusion}. Datasets such as GREENBOT further highlight the importance of multimodal sensing and localization for greenhouse robots \citep{canadas2024greenbot}.

These studies establish important foundations for greenhouse navigation. Nevertheless, localization and motion planning are often evaluated independently from environmental regulation, multimodal semantic reasoning, and localized intervention. Moreover, many greenhouse platforms are wheeled robots designed for relatively uniform aisles. Legged robots provide an alternative for environments containing narrow passages, uneven terrain, irrigation infrastructure, soil patches, and obstacles, although they introduce additional locomotion, stability, and manipulation constraints.

\rev{A local pose and map provide the spatial reference for associating observations and proposals with management zones.}

Embodied multimodal models provide a mechanism through which robots can interpret visual observations, language instructions, and continuous state information. Prior work has demonstrated embodied reasoning from language, visual, and state-estimation inputs \citep{driess2023palme}, connected vision--language representations with robotic actions \citep{brohan2023rt2}, and introduced an open vision--language--action architecture for robotic manipulation \citep{kim2025openvla}.

Directly applying an unconstrained vision--language output to greenhouse actuators, however, presents substantial safety and reliability risks. A predicted navigation goal may be inconsistent with the current local map, an environmental setpoint may fall outside agronomic or actuator limits, and a requested spray action may be unnecessary, stale, or physically unreachable. Consequently, \AgriSense{} does not use the multimodal model as an end-to-end low-level controller.

\rev{The VLDA formulation generates environmental-reference, target-zone and binary intervention proposals.} Deterministic navigation, PI/PID environmental controllers, and predefined manipulation primitives remain responsible for physical execution. The distinction between semantic proposal generation and validated control execution is central to the proposed architecture.

Agricultural digital twins are increasingly used for crop monitoring, environmental prediction, resource planning, visualization, and decision support
{\citep{ariesen2022digitaltwins,gund2025digital_twin_agriculture,lang2025digitaltwinharvesting}}. Many existing agricultural twins emphasize process-level representations of crop growth or greenhouse climate. Although valuable for planning and monitoring, they do not necessarily model an embodied robot moving through the greenhouse, generating viewpoint-dependent RGB and LiDAR observations, updating a local map, and executing physical intervention primitives.

High-fidelity robotics simulators provide complementary capabilities. Isaac Sim supports physically based rendering, robot and sensor simulation, synthetic data generation, robot--environment interaction, and integration with robot-learning workflows
\citep{nvidia_isaac_sim,nvidia_isaac_lab}. These capabilities make it suitable for constructing \rev{an embodied virtual greenhouse} in which sensing, localization, navigation, environmental monitoring, and manipulation can be evaluated within a shared spatial environment.

In \AgriSense{}, \rev{the virtual greenhouse is not used} only as a visualization interface. \rev{It provides a virtual environment for sensing, management-zone association, multimodal inputs, and separate localization/navigation tests, together with} plant-level inspection and targeted intervention workflows.

\label{subsec:agrisense_positioning}

The individual components used in \AgriSense{}---LiDAR--IMU localization, PI/PID environmental regulation, multimodal representation learning, quadruped locomotion, and \rev{virtual-greenhouse simulation---}are not independently new. \rev{The contribution is the explicit learned-proposal interface and component-level evaluation of its supporting workflow.}

\rev{Table~\ref{tab:comparison} compares named studies by task and output contract.}

\rev{Agricultural studies provide more task-specific context. AgriVLN addresses instruction-following navigation in agricultural scenes \citep{zhao2025agrivln}; visual-language-guided horticultural planning connects RGB observations and semantic maps to robot primitives \citep{cuaran2026horticulturalplanning}; AgroGPT adapts conversational vision--language models to agricultural concepts \rev{\citep{awais2024agrogpt}}; and AGRI-BT combines language-generated behaviour trees with \rev{CLIP-based perception in simulated greenhouse tasks} \citep{salman2026agribt}. These approaches motivate an explicit distinction between a supervised predictor, a generative planner, and a low-level action policy. The present \VLDA{} belongs to the first category: BERT encodes supplied text, and three supervised heads produce fixed-format proposals. \rev{Table~\ref{tab:comparison} compares the tasks and output definitions of these systems.}}

\begin{table*}[!tb]
\centering\rmfamily\footnotesize\color{revisionblue}
\caption{\rev{Positioning by specific task and output contract, rather than universal capability claims.}}
\label{tab:comparison}
\setlength{\tabcolsep}{3.6pt}
\renewcommand{\arraystretch}{1.14}
\begin{tabularx}{\textwidth}{p{.23\textwidth}Y Y}
\toprule
Study & Primary task/output & Relation to the present study \\
\midrule
AgriVLN \citep{zhao2025agrivln} & Instruction-following agricultural navigation & \rev{Evaluates navigation rather than environmental-reference, target-zone, and intervention prediction.} \\
Horticultural planning \citep{cuaran2026horticulturalplanning} & Visual-language planning over robot primitives & A planner and a fixed-format predictor have different evaluation targets. \\
AgroGPT \rev{\citep{awais2024agrogpt}} & Conversational agricultural visual interpretation & \rev{Evaluates agricultural visual question answering rather than the reference/zone/intervention outputs studied here.} \\
AGRI-BT \citep{salman2026agribt} & Language-generated behaviour trees and visual perception & Represents task-planning integration rather than this supervised fusion model. \\
\AgriSense{} & Image--text--environment prediction plus separate robotic/control interfaces & Evaluated in a virtual greenhouse; no claim of new encoders, a new LIO algorithm, or field-validated treatment. \\
\bottomrule
\end{tabularx}
\par\smallskip\begin{minipage}{\textwidth}\footnotesize Descriptions summarize the cited task settings, not a common experimental leaderboard or proof of superiority. A matched-input agricultural planning comparison remains outside the available experiments.\end{minipage}
\end{table*}

\rev{The central contribution is the implemented multimodal proposal interface, specified through its input schema, three nonlinear prediction heads, training objective, and decoding rules. \rev{The evaluation combines checkpoint reproduction, five-seed modality ablations, explicit source-group splits, and separate image-transfer diagnostics.} Navigation, offline localization, environmental regulation, manipulation, and fault tests assess \rev{the supporting system interfaces.} This focus does not require claiming that ordinary feature concatenation is a new fusion mechanism. Instead, the experiments identify both useful component-level behaviour and limits on the role of the visual and language branches.}

\begin{figure*}[!tb]
\centering
\includegraphics[width=.97\textwidth,height=9.5cm,keepaspectratio]{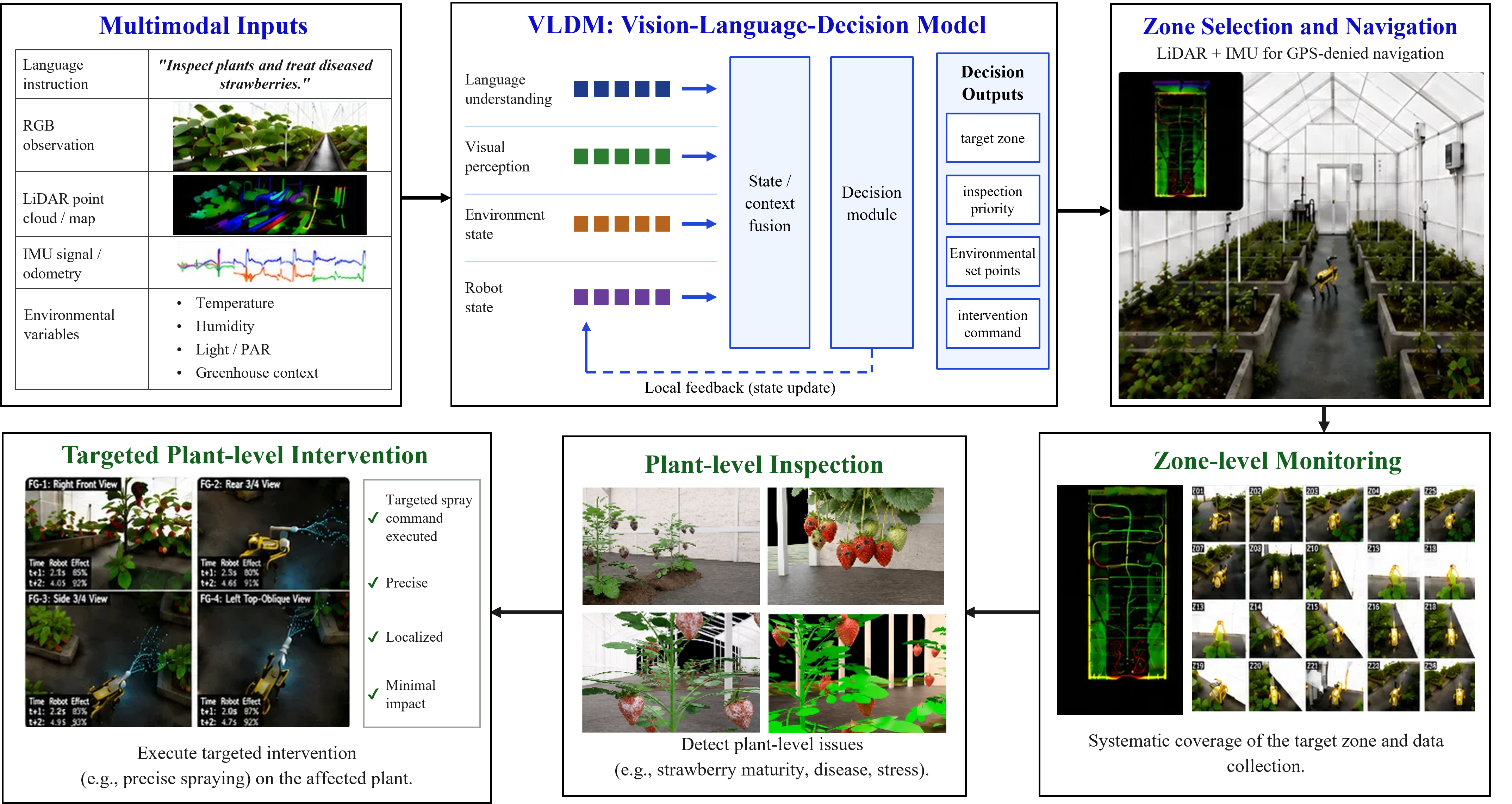}
\caption{\rev{\AgriSense{} multimodal decision-to-action pipeline. The original system-level illustration is retained to show the proposed sensing, decision, navigation, and intervention interfaces. The learned implementation accepts RGB, text, and 19 structured environmental/context values; LiDAR--IMU processing and continuous robot state belong to separate modules. ``Inspection priority,'' ``precise,'' and ``minimal impact'' in the illustration are workflow labels or design objectives, not additional learned heads or measured agronomic outcomes. The component-specific evidence and execution boundaries are given in Sections~\ref{sec:learned_model}--\ref{sec:limitations}.}}
\label{fig:main_architecture}
\end{figure*}
The remainder of this paper is organized as follows. Section~\rev{\ref{sec:methods}} presents the \rev{system} \rev{interfaces,} \rev{virtual} greenhouse\rev{,} \VLDA{} \rev{implementation}, \rev{datasets}, and \rev{evaluation} \rev{protocols}. Section~\ref{sec:results} reports \rev{and} discusses the \rev{prediction}, \rev{localization, regulation, intervention}, and \rev{fault-handling} \rev{results}. Section~\rev{\ref{sec:conclusions}} concludes the paper. \rev{Detailed protocols, retained historical plots, and extended results are provided in a separate Supplementary Material file.}

\section{\rev{Materials and Methods}}
\label{sec:methods}
\subsection{\rev{System architecture and evaluation scope}}
\label{sec:architecture}
\label{subsec:overall_framework}
\rev{AgriSense combines a supervised decision-support model with a virtual-greenhouse testbed. The learned component receives RGB, prompt text and 19 structured environmental/current-zone inputs, and produces twelve environmental references, a target zone and a binary intervention proposal. Supporting experiments evaluate offline LiDAR--inertial localization, simulator-pose navigation, environmental responses, geometric targeting and command admission. These are component protocols: the navigation and integrated missions use simulator pose, diagnosis and target positions rather than an estimated-pose, learned-diagnosis feedback chain. No physical greenhouse is used. This scope applies throughout Section~\ref{sec:methods}; outstanding validation requirements are consolidated in Section~\ref{sec:limitations}.}

\rev{The four management zones share a local greenhouse frame. Each zone has temperature, relative-humidity and soil-moisture channels, and the model also receives their greenhouse-level aggregates. Reference prediction is separated from actuator execution. Robot-mounted RGB, LiDAR and IMU support observation and spatial measurements, while the supervisory application displays maps, telemetry and mission events. Figure~\ref{fig:main_architecture} summarizes the evaluated pathways. The extended activity-state and dispatch design is retained in Supplementary Section S9 as a conceptual extension.}

\paragraph{\rev{System Modules and Interfaces}}
\label{subsec:system_modules}

Table~\ref{tab:modules} summarizes the main modules and the information exchanged across the \AgriSense{} architecture. \rev{The inventory separates prediction, spatial state, command admission and execution.}

\begin{figure*}[!tb]
\centering
\includegraphics[width=.97\textwidth]{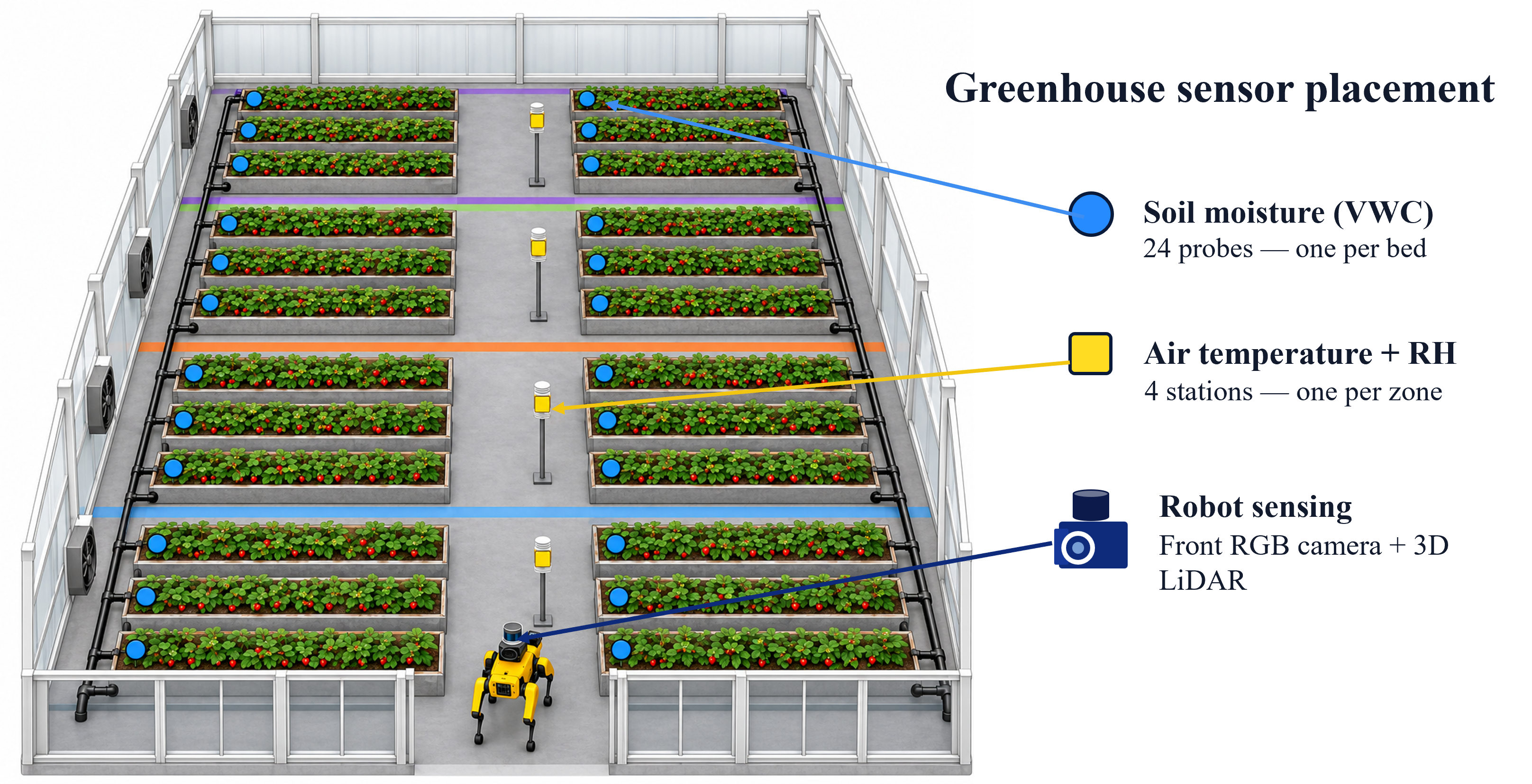}
\caption{\rev{Sensor and actuator placements in the virtual greenhouse. Blue arrows identify sensing and orange arrows actuation in the supplied cutaway. Fan rotation and water particles visualize model actuator fractions, not computational fluid dynamics or calibrated deposition. Full bed/zone identifiers and coordinates are supplied separately in the supplementary sensor-layout view.}}
\label{fig:sensor_layout}
\end{figure*}

\begin{table*}[!tb]
\centering\rmfamily
\caption{\rev{System modules and interface inventory. The evaluated pathways are specified in the accompanying protocols; extended dispatch requirements are given in Supplementary Section S9.}}
\label{tab:modules}
\small
\renewcommand{\arraystretch}{1.16}
\setlength{\tabcolsep}{4pt}

\begin{tabularx}{\textwidth}{
>{\raggedright\arraybackslash}p{2.6cm}
>{\raggedright\arraybackslash}p{3.5cm}
>{\raggedright\arraybackslash}p{3.8cm}
>{\raggedright\arraybackslash}X}
\toprule
\textbf{Module} &
\textbf{Inputs} &
\textbf{Outputs} &
\textbf{Representative interface or artifact} \\
\midrule

Robot sensing
& RGB, LiDAR, IMU, proprioception
& Images, point clouds, inertial measurements
& Robot-view image stream, LiDAR cloud, IMU and robot-state topics \\

Zone sensing
& Fixed environmental sensor nodes
& Temperature, humidity, and soil-moisture states
& Global and per-zone ROS~2 messages and synchronized CSV logs \\

LiDAR--IMU spatial grounding
& Point clouds, IMU, local map
& Robot pose, trajectory, current zone, map update
& Local odometry, reconstructed point cloud, zone-association output \\

Multimodal dataset recorder
& Image, prompt, zone, pose, and environmental state
& Aligned multimodal samples and target labels
& Dataset CSV, image directory, preprocessing metadata \\

\VLDA{} inference
& Prompt, RGB image, environmental vector, current zone
& {Setpoint proposals, target zone, spray-intervention probability}
& Prediction packet and offline evaluation outputs \\

Validation and command dispatch
& Model outputs, timestamps, pose, map, actuator limits
& Accepted command, rejected proposal, or fallback request
& Validated command buffer, event log, execution-mode state \\

Environmental control
& Validated references and measured zone states
& Irrigation, ventilation, heating/cooling, and humidity commands
& {PI/PID reference-tracking commands and bounded actuator signals} \\

Navigation and intervention
& Validated target zone, local map, target region
& Motion command, inspection pose, or spray primitive
& Navigation-goal queue and manipulator-action interface \\

Supervision
& Maps, telemetry, decisions, diagnostics, and alerts
& Review displays and high-level operator requests
& Non-real-time application and remote dashboard \\

\bottomrule
\end{tabularx}
\end{table*}
\rev{Figure~\ref{fig:sensor_layout} identifies the modeled sensor and actuator placements; the full coordinate view is retained in the supplement.}

\subsection{\rev{Virtual greenhouse, visual data and scenario specification}}
\label{sec:isaac}
\paragraph{\rev{Embodied Virtual-Greenhouse Environment}}
\label{subsec:digital_twin_environment}

\rev{The \AgriSense{} virtual greenhouse is implemented} in Isaac Sim as a simulation-centric embodied environment for evaluating robot sensing, spatial grounding, navigation, inspection, environmental regulation, and {plant-level spray-intervention workflow.} Unlike greenhouse digital twins used primarily for visualization or process monitoring, the proposed environment explicitly represents the robot, its sensors, its local spatial context, and its interaction with the greenhouse.

The simulated scene contains the greenhouse structure, cultivation beds, aisle corridors, management-zone boundaries, structural obstacles, sensing locations, strawberry-like crop assets, and the quadruped mobile manipulator. The environment supports viewpoint-dependent RGB observations, 3D LiDAR returns, IMU measurements, robot-state feedback, and manipulator motion. Management zones are defined within the local map so that robot poses, environmental measurements, inspection observations, and intervention decisions can be associated with a consistent spatial reference.

\begin{figure*}[p]
\centering\rmfamily
\begin{subfigure}[t]{0.98\textwidth}
\centering\rmfamily
\includegraphics[width=\linewidth,height=8cm,keepaspectratio]{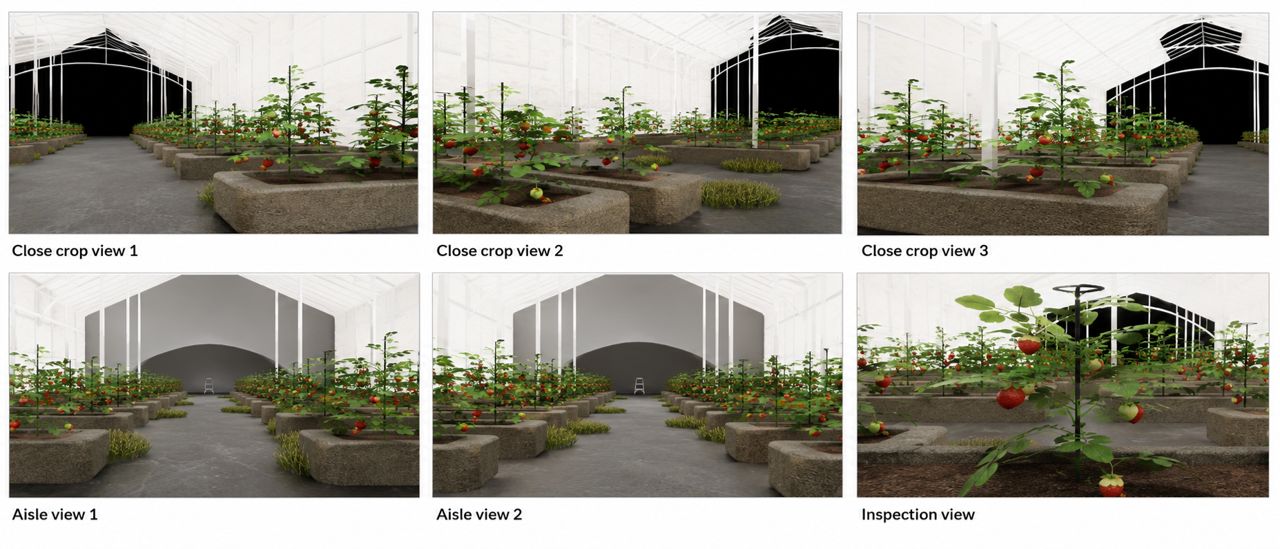}
\caption{Representative RGB observations.}
\end{subfigure}

\vspace{0.8em}

\begin{subfigure}[t]{0.32\textwidth}
\centering\rmfamily
\includegraphics[width=\linewidth,height=5.7cm,keepaspectratio]{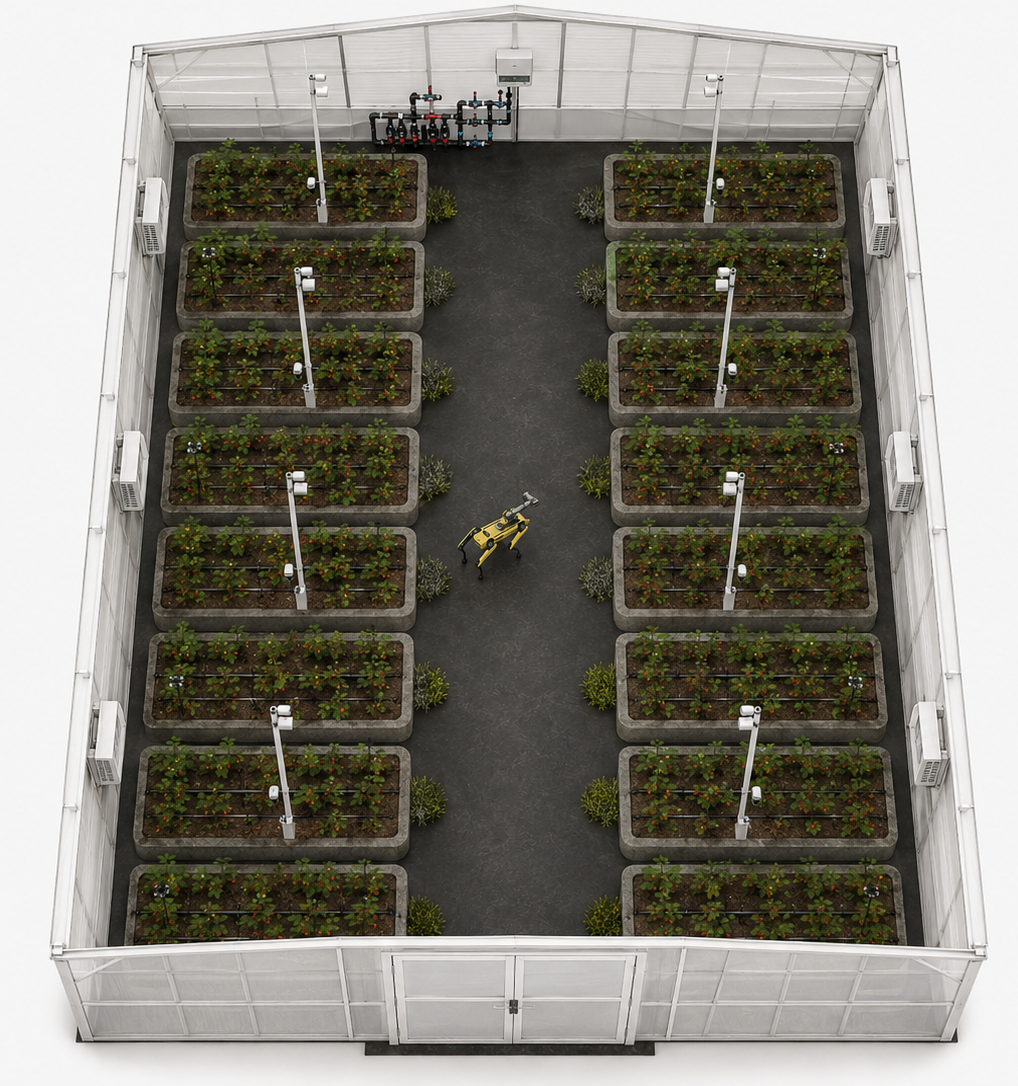}
\caption{Top-down greenhouse layout.}
\label{fig:dt_top_view}
\end{subfigure}\hfill
\begin{subfigure}[t]{0.32\textwidth}
\centering\rmfamily
\includegraphics[width=\linewidth,height=5.7cm,keepaspectratio]{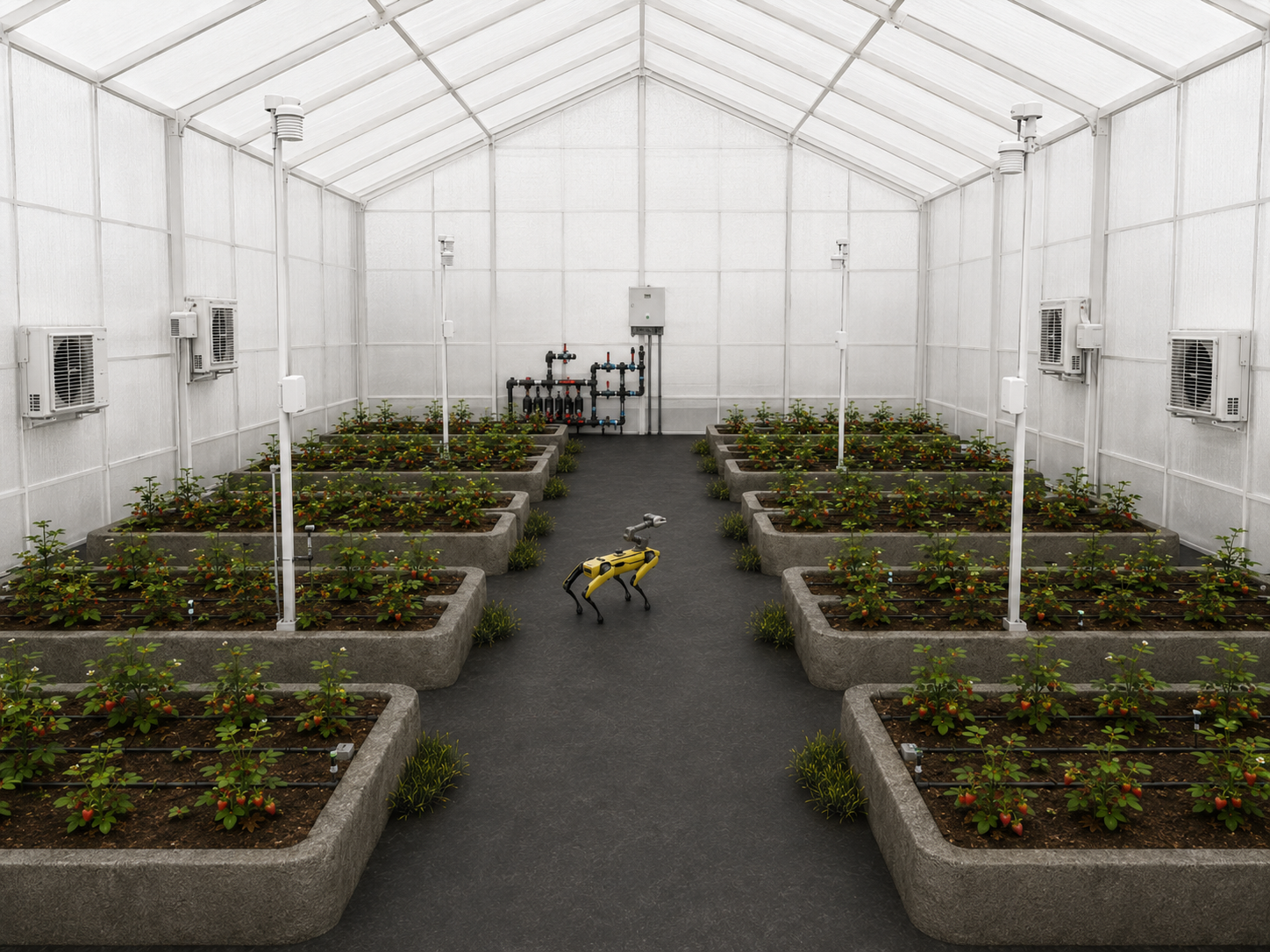}
\caption{Interior aisle view.}
\label{fig:dt_front_view}
\end{subfigure}\hfill
\begin{subfigure}[t]{0.32\textwidth}
\centering\rmfamily
\includegraphics[width=\linewidth,height=5.7cm,keepaspectratio]{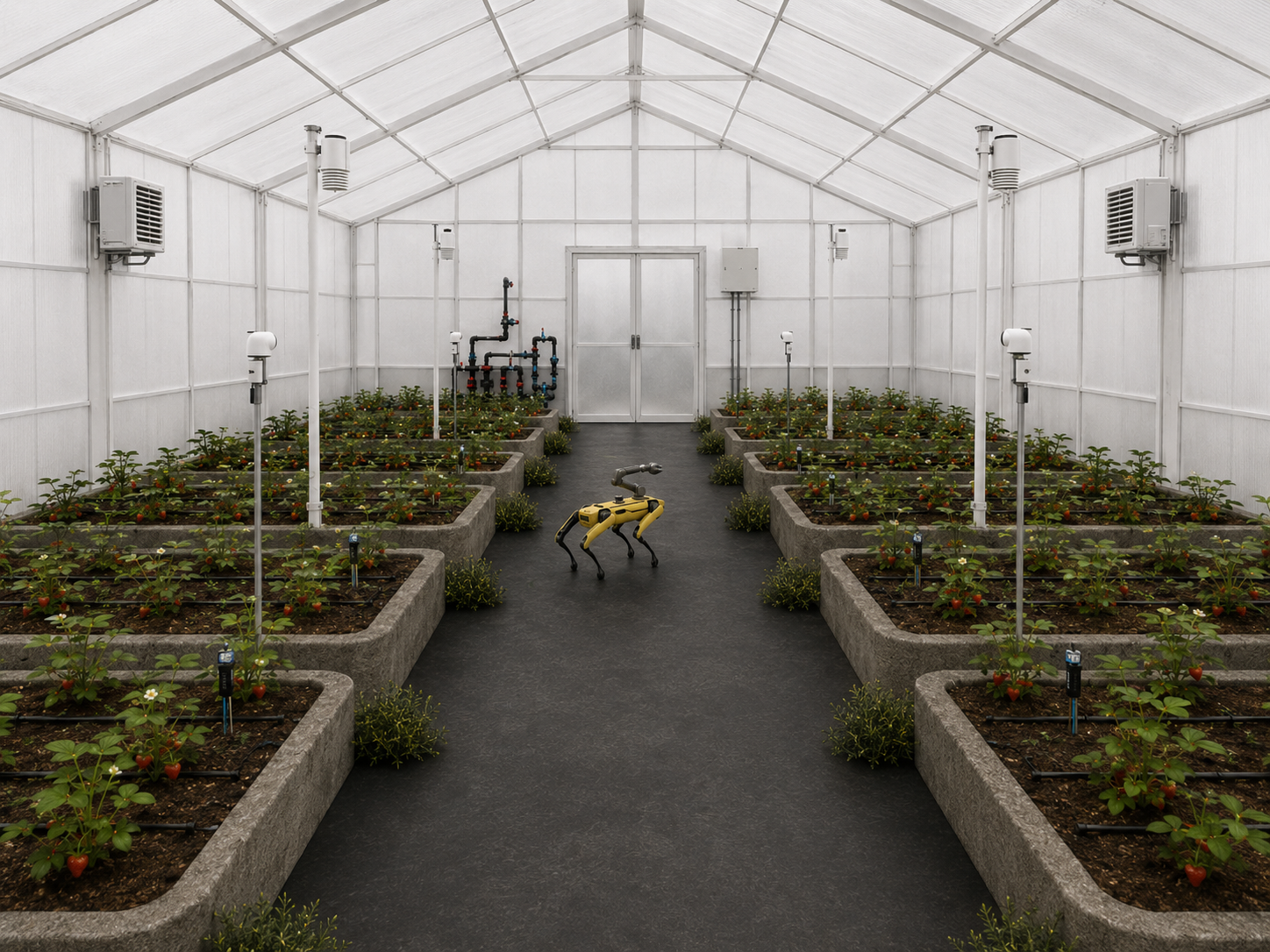}
\caption{Complementary interior view.}
\label{fig:dt_rear_view}
\end{subfigure}

\caption{\rev{Original virtual-greenhouse views combined to retain spatial context alongside the added experimental figures.} Representative RGB observations generated in \rev{the Isaac Sim virtual greenhouse}. The samples include aisle-level robot views, crop-row observations, greenhouse spatial context, and close crop-inspection viewpoints used for multimodal dataset construction and simulation-centric pipeline evaluation. The top-down view provides the global spatial organization, whereas the interior views show the robot operating context, cultivation-bed arrangement, sensing infrastructure, and GNSS-denied navigation corridors. \rev{The source images are unchanged; these qualitative views do not measure plant-target registration accuracy.}}
\label{fig:digital_twin_rgb}
\label{fig:local_digital_twin}
\end{figure*}

Figure~\ref{fig:digital_twin_rgb} shows representative RGB observations generated from the robot and greenhouse viewpoints. These images are used for multimodal dataset construction, perception-pipeline verification, and scenario replay. 

\paragraph{\rev{Quadruped, Manipulator, and Multimodal Sensor Modeling}}
\label{subsec:robot_sensor_simulation}

The simulated mobile platform consists of a quadruped base equipped with a manipulator, robot-mounted RGB camera, 3D LiDAR, IMU, local computing interface, and spray end-effector. The quadruped traverses the greenhouse aisles, acquires crop and environmental observations, and moves toward validated inspection or intervention locations.

The RGB camera provides visual information for crop-condition interpretation and plant-level inspection. The 3D LiDAR and IMU provide the measurements required for GNSS-denied localization, local map reconstruction, motion estimation, and greenhouse-zone association. Sensor measurements are timestamped and synchronized with the robot pose and zone-level environmental state before being recorded or passed to the \VLDA{} module.

The manipulator provides the physical interface for {localized spray treatment in simulation.} \rev{The evaluated intervention uses scenario targets and the geometric/dwell admission criteria specified below.} The primitive represents a bounded sequence consisting of target approach, directed spray execution, and return to a safe manipulator configuration.

\paragraph{\rev{Zone-State Representation and Environmental-Control Emulation}}
\label{subsec:environmental_emulation}

In addition to robot and sensor simulation, \rev{the virtual-greenhouse workflow} maintains synchronized environmental states for each greenhouse management zone. The zone-state channels include temperature, relative humidity, and soil moisture. These values are associated with the corresponding greenhouse zones and are recorded together with the robot pose, RGB observations, mission prompt, and model outputs.

Environmental disturbances are introduced through controlled scenario profiles, including elevated temperature, excessive humidity, low soil moisture, and mixed multi-zone conditions. The \VLDA{} module processes these states together with the visual and spatial inputs and proposes environmental reference values. The proposed references are not treated as direct actuator commands. 

\rev{The environmental execution protocol is described in Section~\ref{sec:execution}.}

\paragraph{\rev{Multimodal Dataset Construction and Controlled Visual Augmentation}}
\label{subsec:dataset_generation}
\label{sec:data}

The \AgriSense{} multimodal dataset was constructed in two stages. The first
stage generated the base simulation dataset directly from the embodied
greenhouse environment in NVIDIA Isaac Sim. The second stage introduced
controlled disease-like visual conditions through post-simulation
image-to-image augmentation. This distinction is important because the
disease appearances were not physically modeled as plant pathologies inside
Isaac Sim.

During the first stage, the quadruped robot was operated from multiple
greenhouse locations and viewpoints, including aisle-level traversal,
crop-row observation, and close strawberry-inspection poses. Robot-view RGB
images were recorded together with the corresponding greenhouse and task
context. Each visual observation was associated with its management-zone
identifier, global and zone-specific environmental states, and the
corresponding task record. The environmental state contained temperature,
relative humidity, and soil-moisture measurements for the four management
zones.

For a recorded sample \(i\), the multimodal dataset entry can be summarized
as

\begin{equation}
\mathcal{D}_i =
\left\{
I_i,\,
\vect{q}_i,\,
z_i,\,
\vect{e}_i,\,
\vect{s}^{*}_i,\,
z^{*}_i,\,
y^{p}_i
\right\},
\label{eq:multimodal_dataset_sample}
\end{equation}

where \(I_i\) is the robot-view RGB observation, \(\vect{q}_i\) is the
associated task or mission prompt, \(z_i\) is the current management zone,
\(\vect{e}_i\) contains the global and per-zone environmental measurements,
\(\vect{s}^{*}_i\) contains the target environmental references,
\(z^{*}_i\) is the target-zone label, and \(y^{p}_i\) is the binary
plant-level spray-intervention label.

Controlled diversity was introduced during simulation through changes in
robot viewpoint, crop-observation distance, partial occlusion, illumination,
and environmental disturbances. Temperature, humidity, and soil-moisture
conditions were varied across the four management zones so that the dataset
contained both spatial and environmental variation rather than repeated
images associated with a single greenhouse state.

The Isaac Sim crop assets primarily represented healthy strawberry
appearances. To expand the visual-condition space without changing the
underlying spatial and environmental context, selected robot-view images were
therefore processed in a separate post-simulation augmentation stage. \rev{The augmentation scripts specify an OpenAI Image API workflow. Per-image execution settings are unavailable; script defaults are recorded separately in Supplementary Section S1.} The editing instructions restricted modifications to visible
strawberry fruits while preserving the original camera pose, greenhouse
geometry, foliage, illumination, shadows, depth relationships, and overall
scene composition.

Three controlled disease-like visual classes were introduced:
powdery-mildew-like, Botrytis-like, and anthracnose-like appearances. \rev{The original script-level sampling specification was}

\begin{equation}
P(c)=
\begin{cases}
0.35, & c=\text{powdery mildew},\\
0.35, & c=\text{Botrytis},\\
0.30, & c=\text{anthracnose}.
\end{cases}
\label{eq:disease_sampling}
\end{equation}

The corresponding editing constraints were designed to produce subtle white
powdery regions for the powdery-mildew-like condition, localized
grey--brown mould and fruit degradation for the Botrytis-like condition, and
dark sunken lesions for the anthracnose-like condition. These labels describe
controlled visual proxies and should not be interpreted as biologically
confirmed strawberry diseases.

\rev{The final dataset is evaluated using its stored prompts and labels. Source reconciliation identifies which base/edited pairs preserve environmental context and which labels change, as quantified below and detailed in Supplementary Section S1. Script-level sampling and metadata design are retained in Supplementary Section S9.}

Figure~\ref{fig:condition_samples} presents representative visual conditions
from this process. The examples illustrate healthy simulated observations,
disease-like visual proxies, environmental-stress conditions, occlusion, and
illumination variation.

\begin{figure*}[!tb]
\centering\rmfamily
\includegraphics[width=.97\textwidth,height=5.9cm,keepaspectratio]{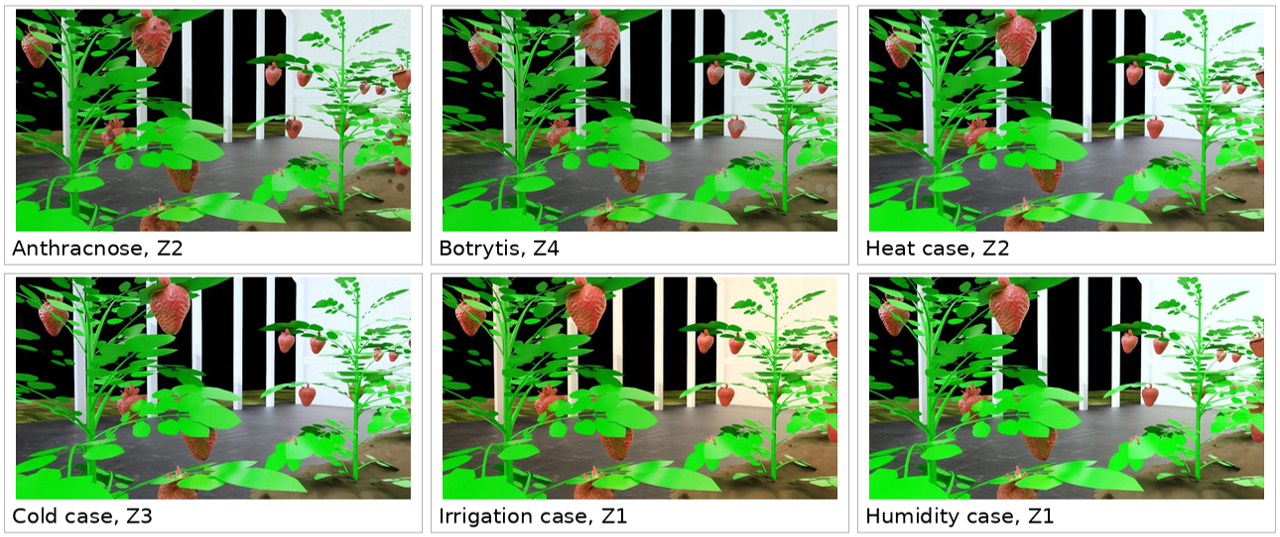}
\caption{Representative visual conditions used in the \AgriSense{} multimodal dataset. Healthy robot-view observations originate from the Isaac Sim greenhouse environment, whereas disease-like strawberry appearances are generated through controlled post-simulation image editing. The powdery-mildew-, Botrytis-, and anthracnose-like examples are synthetic visual proxies\rev{,} not biologically confirmed diagnoses. \rev{Labels attached to these examples describe the generated dataset and do not establish disease sensitivity or treatment efficacy.}}
\label{fig:condition_samples}
\end{figure*}

This construction preserves the controlled spatial and environmental context
\rev{of the virtual greenhouse while} expanding the visual appearance space available to
the multimodal model. The augmented samples are used to evaluate target-zone,
environmental-reference, inspection, and spray-intervention proposal behavior
under synthetic visual stress. 
\rev{The final supervised CSV contains 283 decodable images from one episode: 150 base frames at $1280\times720$ and 133 edited frames at $1536\times1024$. The 2,917 raw-directory images are not all labelled examples. Prompts and reference outputs were generated offline; the offline teacher-labeling script requests both in one response and sanitizes output ranges to temperature 16--24~$^\circ$C, humidity 45--65\%, soil moisture 55--75\%, and zone IDs 1--4. These are historical labeling rules, not expert-validated treatment thresholds. The final CSV omits the teacher request/response history, so the exact generation provenance remains incomplete.}

\rev{A source reconciliation identifies six base-to-augmentation pairs within the 150 supervised base records. All six preserve the current-zone code, environmental values, and twelve reference targets, but change the final prompt, target-zone label, and intervention flag. \rev{The source parents of the remaining 127 augmented records are not represented within that base subset.} Accordingly, complete prompt/label inheritance cannot be assumed. Moreover, all base rows have negative intervention labels and all augmented rows positive labels; a case-insensitive rule matching ``anthracnose,'' ``powdery mildew,'' or ``botrytis'' in the prompt reproduces all 283 intervention labels. The rule is an observed shortcut diagnostic, not proof of which input route the network actually uses.}

\begin{table*}[!tb]
\centering\rmfamily\footnotesize\color{revisionblue}
\caption{\rev{Original dataset composition and evaluation split definitions. Zone counts list target-zone labels, not current-zone inputs.}}
\label{tab:data_composition}
\setlength{\tabcolsep}{3.6pt}
\renewcommand{\arraystretch}{1.14}
\begin{tabularx}{\textwidth}{p{.18\textwidth}ZZp{.24\textwidth}Z}
\toprule
Subset & Base/edited & Total & Target zones 1/2/3/4 & Negative/positive \\
\midrule
Historical training & 105/93 & 198 & 27/26/125/20 & 105/93 \\
Historical validation & 22/20 & 42 & 6/6/26/4 & 22/20 \\
Historical test & 23/20 & 43 & 6/5/28/4 & 23/20 \\
All supervised rows & 150/133 & 283 & 39/37/179/28 & 150/133 \\
\midrule
Grouped split & \multicolumn{4}{l}{197 training / 42 validation / 44 test rows; 265 source/similarity components} \\
\bottomrule
\end{tabularx}
\end{table*}

\rev{The historical seed-42 row split is reproducible, but disjoint row indices do not ensure independence. Five cross-split pairs share a source-frame key, sensor vector, and all twelve references; three test records have training relatives. A separate dHash64-distance-at-most-four screen identifies twenty cross-split candidates, which are not necessarily disjoint from the source-lineage pairs. Before \rev{retraining}, shared source keys and the same similarity screen define 265 connected components, \rev{each assigned to a single subset of the 197/42/44 training/validation/test split.} This removes the identified cross-split links without creating a second acquisition episode or independent physical-plant evidence. The original 198/42/43 results are retained as historical reproduction, not pooled with the grouped evaluation.}

\subsection{\rev{VLDA inputs, architecture, and training objective}}
\label{sec:learned_model}

The \VLDA{} module converts multimodal greenhouse observations into structured high-level proposals. At decision time \(t\), the model receives a robot-view RGB image \(I_t\), a user or mission prompt \(\vect{q}_t\), and a structured sensor-and-zone vector. The latter is defined as

\begin{equation}
\vect{v}^{\mathrm{sen}}_t =
\left[
\mathrm{onehot}(z_t)^{\top},
T_{g,t},
H_{g,t},
S_{g,t},
\vect{e}_{1,t}^{\top},
\ldots,
\vect{e}_{K,t}^{\top}
\right]^{\top},
\label{eq:sensor_vector}
\end{equation}

where \(z_t\) is the current management zone; \(T_{g,t}\), \(H_{g,t}\), and \(S_{g,t}\) are greenhouse-level temperature, humidity, and soil-moisture values; and

\begin{equation}
\vect{e}_{k,t}
=
\begin{bmatrix}
T_{k,t} & H_{k,t} & S_{k,t}
\end{bmatrix}^{\top}
\end{equation}

contains the corresponding measurements for zone \(k\). For the four-zone greenhouse considered in this work, \(\vect{v}^{\mathrm{sen}}_t\) contains a four-dimensional zone code, three global measurements, and twelve zone-specific measurements.

\rev{All 19} sensor\rev{/context} channels\rev{, including the current-zone one-hot code,} are standardized using \rev{means and standard deviations} computed from the training split\rev{. Standard deviations below $10^{-6}$ are replaced by one}. The current-zone code provides coarse spatial context to the learned model, whereas the continuous robot pose and local map are retained by the downstream execution layers.

The model predicts three output types:

\begin{equation}
\begin{aligned}
\hat{\vect{s}}_t &\in \mathbb{R}^{3K},
&
\boldsymbol{\pi}^{z}_t &\in [0,1]^K,\\
\hat{p}_t &\in [0,1].
\end{aligned}
\label{eq:vlda_output_spaces}
\end{equation}

{Here, \(\hat{\vect{s}}_t\) contains the proposed per-zone environmental references, \(\boldsymbol{\pi}^{z}_t\) contains the target-zone class probabilities, and \(\hat{p}_t\) is the plant-level spray-intervention proposal probability.}

The target-zone proposal is obtained as

\begin{equation}
\hat{z}_t
=
\arg\max_{k\in\{1,\ldots,K\}}
\pi^{z}_{k,t}.
\label{eq:target_zone_argmax}
\end{equation}

For \(K=4\), the regression head produces twelve values corresponding to temperature, humidity, and soil-moisture references for Zones~1--4. These values are interpreted as high-level reference proposals and are not applied directly to greenhouse actuators.

\rev{The image processor converts RGB input to $3\times224\times224$ tensors. The evaluated model uses pretrained DINOv2-small token embeddings of width 384 and BERT-base-uncased embeddings of width 768 \citep{oquab2023dinov2,devlin2019bert}; BERT tokenization uses a maximum length of 128 and an attention mask. Both backbones are frozen, and \rev{the final-layer CLS embedding is selected from each.} Projection blocks are Linear(384,256)--ReLU--Dropout(0.1) and Linear(768,128)--ReLU--Dropout(0.1). The environmental encoder is Linear(19,128)--ReLU--Dropout(0.1)--Linear(128,128)--ReLU. The resulting 256, 128, and 128 features are concatenated, not cross-attended, to form a 512-dimensional vector.}

Let \(f_I(\cdot)\), \(f_Q(\cdot)\), and \(f_E(\cdot)\) denote the visual, textual, and structured-state encoders, respectively. Their latent representations are

\begin{align}
\vect{h}^{I}_t &= f_I(I_t),\\
\vect{h}^{Q}_t &= f_Q(\vect{q}_t),\\
\vect{h}^{E}_t &= f_E(\vect{v}^{\mathrm{sen}}_t).
\end{align}

The evaluated implementation uses a DINOv2-small visual encoder, a BERT-base-uncased text encoder, and a feed-forward network for the structured sensor-and-zone vector. The three \rev{projected} representations are concatenated and passed through a \rev{shared} fusion network\rev{, Linear(512,256)--ReLU--Dropout(0.2)--Linear(256,256)--ReLU}:

\begin{equation}
\vect{h}_t
=
\phi
\left(
\left[
\vect{h}^{I}_t;
\vect{h}^{Q}_t;
\vect{h}^{E}_t
\right]
\right),
\label{eq:vlda_fusion}
\end{equation}

\rev{Each task head has its own Linear(256,128)--ReLU--Dropout(0.1) hidden block followed by a final linear map of width 12, 4, or 1. Thus the heads are nonlinear functions of the shared representation, not single affine maps. With $g_s$, $g_z$, and $g_p$ denoting these three MLPs, the actual decoding is}

\begin{revblock}
\begin{align}
\widehat{\widetilde{\vect{s}}}_t &= g_s(\vect{h}_t), &
\hat{\vect{s}}_t &= \boldsymbol{\mu}_{s}+\boldsymbol{\sigma}_{s}\odot\widehat{\widetilde{\vect{s}}}_t,\label{eq:setpoint_head}\\
\boldsymbol{\pi}^{z}_t &= \operatorname{softmax}(g_z(\vect{h}_t)),\label{eq:zone_head}\\
\hat p_t &= \sigma(g_p(\vect{h}_t)), &
\hat y^{p}_t&=\mathbb{I}[\hat p_t>0.5].\label{eq:pesticide_head}
\end{align}
\end{revblock}

\rev{The regression head has no final activation, and $\boldsymbol\mu_s,\boldsymbol\sigma_s$ are fitted only to training targets. The target order is temperature, humidity, and soil moisture for Zone~1 followed by Zones~2--4. Proposals are not \rev{guaranteed to be optimal setpoints.} The binary output is an intervention proposal, not a disease class, lesion mask, dose, or automatic authorization to spray. The model contains 132,052,881 parameters, of which 131,538,816 are frozen and 514,065 trainable.}

\begin{figure*}[!tb]
\centering
\includegraphics[width=.97\textwidth,height=7.6cm,keepaspectratio]{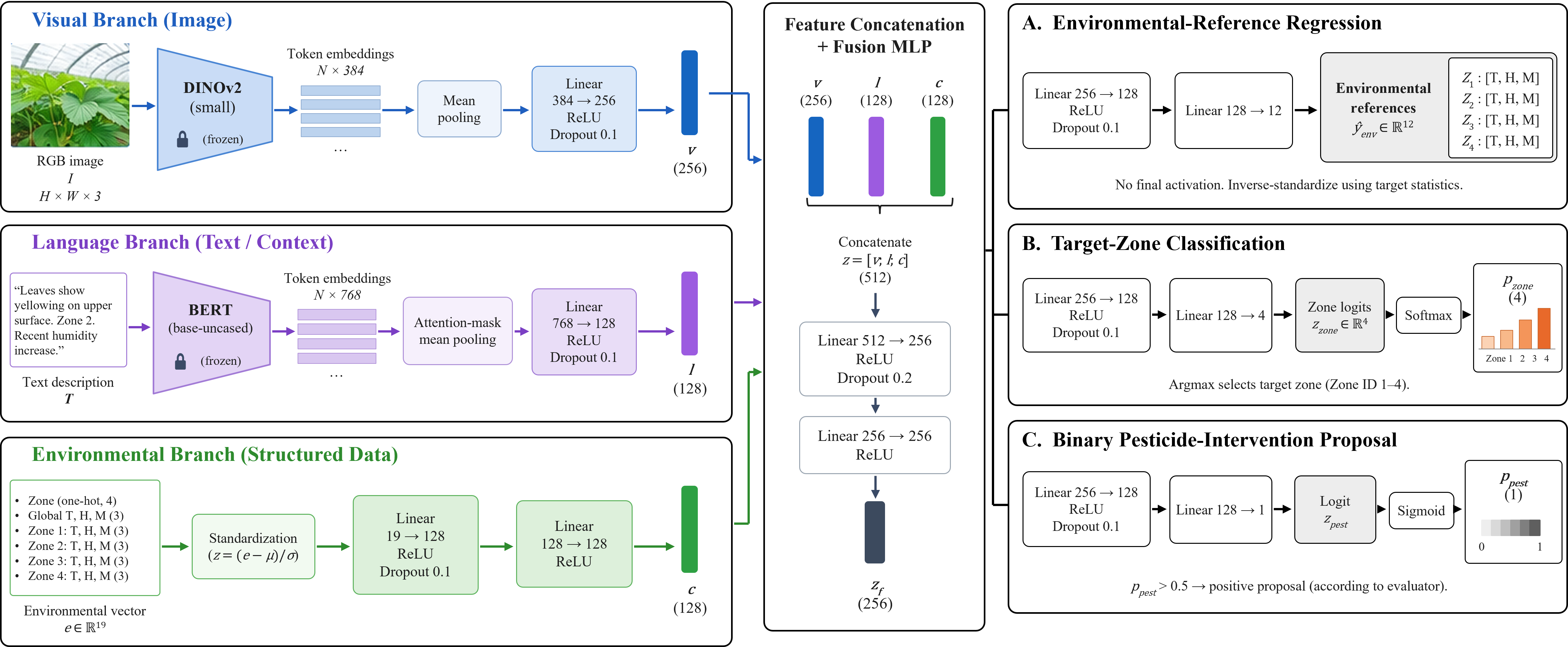}
\caption{\rev{Internal \VLDA{} architecture for the evaluated checkpoint. Frozen DINOv2-small and BERT-base-uncased encoders provide CLS features, which are projected and concatenated with the encoded 19-channel environmental/context vector. A shared fusion MLP feeds environmental-reference, target-zone, and intervention-proposal heads. Images are processed to 3 \ensuremath{\times} 224 \ensuremath{\times} 224 tensors, and structured inputs and regression targets use training-derived statistics. Input readiness and execution supervision are external to the network. The alternative mean-pooling trainer is distinguished in Table~\ref{tab:training}.}}
\label{fig:vlda_internal}
\end{figure*}

The \VLDA{} model is trained jointly for environmental-reference regression, target-zone classification, and plant-level spray-intervention proposal prediction. The total objective is

\begin{equation}
\mathcal{L}
=
\lambda_s\mathcal{L}_{\mathrm{reg}}
+
\lambda_z\mathcal{L}_{\mathrm{zone}}
+
\lambda_p\mathcal{L}_{\mathrm{pest}},
\label{eq:multitask_loss}
\end{equation}

where \rev{$\lambda_s=\lambda_z=\lambda_p=1$ in the saved run. The original trainer uses MSE on standardized targets}, \rev{training-frequency-weighted cross-entropy}, and \rev{positive-weighted} \rev{BCE-with-logits.} \rev{For} \rev{a} \rev{batch $\mathcal B$, these terms are}

\begin{revblock}
\begin{align}
\mathcal L_{\mathrm{reg}}&=\frac{1}{12|\mathcal B|}\sum_{i\in\mathcal B}\sum_{j=1}^{12}(\widehat{\widetilde s}_{ij}-\widetilde s_{ij})^2,\label{eq:regression_loss}\\
\mathcal L_{\mathrm{zone}}&=\frac{-\sum_{i\in\mathcal B}w_{y_i}\log\pi^z_{y_i,i}}{\sum_{i\in\mathcal B}w_{y_i}},\quad w_c=\frac{N}{4n_c},\label{eq:zone_loss}\\
\mathcal L_{\mathrm{pest}}&=-\frac{1}{|\mathcal B|}\sum_{i\in\mathcal B}\left[q y_i^p\log\hat p_i+(1-y_i^p)\log(1-\hat p_i)\right],\label{eq:pesticide_loss}
\end{align}
\end{revblock}

\rev{where $q=\max(1,N_{\mathrm{negative}}/N_{\mathrm{positive}})$ and all counts come from training labels. \rev{Here, $N$ is the number of training samples and $n_c$ is the number with target-zone class $c$.} Checkpoints are selected by minimum total validation loss. Alternative-trainer settings are documented separately in Supplementary Section S1.}

\begin{table*}[!tb]
\centering\rmfamily\footnotesize\color{revisionblue}
\caption{\rev{Model and training configurations. The historical checkpoint and the new retrainings are separate protocols.}}
\label{tab:training}
\setlength{\tabcolsep}{3.6pt}
\renewcommand{\arraystretch}{1.14}
\begin{tabularx}{\textwidth}{p{.25\textwidth}Y Y}
\toprule
Item & Historical checkpoint & Repeated branch evaluation \\
\midrule
Backbones/pooling & Frozen DINOv2-small and BERT; CLS/CLS & Same frozen weights; cached evaluation-mode CLS features \\
Trainable capacity & 514,065 parameters & Same stored projection/encoder/fusion/head topology; removed features zeroed \\
Optimizer & AdamW; learning rate $10^{-4}$, weight decay $10^{-4}$ & AdamW; same rates \\
Batch/epoch policy & 16; maximum 300; patience 10 & 16; maximum 100; patience 10 \\
Recorded history & 50 epochs; selected epoch 40 & Validation-selected per run; per-run records supplied \\
Seeds & 42 & 17, 29, 43, 71, 101 \\
Objective & MSE + weighted CE + weighted BCE; weights 1/1/1 & Same objective; training-only standardization \\
Decision thresholds & Zone argmax; intervention probability $>0.5$ & Same decoding; no test-threshold selection \\
Forward latency & Median 7.1006~ms; p95 7.4684~ms & Not a matched timing comparison across training protocols \\
\bottomrule
\end{tabularx}
\par\smallskip\begin{minipage}{\textwidth}\footnotesize Original training calls model.train(), so freezing weights does not itself disable BERT dropout. Cached-feature retraining disables backbone dropout; it is not an exact replay of stochastic online training. Batch-one FP32 latency was measured on an RTX 5090 Laptop GPU with five warm-ups and 30 timed repeats; it excludes decoding, preprocessing, transfer, I/O, and ROS. Historical training wall time and peak GPU memory are unavailable.\end{minipage}
\end{table*}

\subsection{\rev{Input readiness and command admission}}
\label{subsec:safe_dispatch}
\rev{Two distinct checks are evaluated. The neural-interface readiness function requires the input fields to be present, receipt age at most 2 s, and timestamp spread at most 1 s. It does not test numerical ranges or nonfinite values. The virtual-controller command guard is evaluated separately with stale sensor/diagnosis, nonfinite command, out-of-range command, unreachable target, unsafe clearance, disconnect, invalid-state and abort fixtures. Safe dwell requires upright posture, planar speed below 0.035 m/s and yaw rate below 0.06 rad/s for 0.5 s. A numerical upright-angle threshold is not available in the retained protocol. Results use the specific fixture counts in Section~\ref{sec:safety_results}. The broader confidence, reference-rate and localization-consistency admission rules are conceptual extensions in Supplementary Section S9.}

\paragraph{\rev{Dataset schema and recorded outputs}}
\label{subsec:vlda_dataset_schema}

Table~\ref{tab:vlda_schema} summarizes the multimodal inputs, prediction targets, and evaluation artifacts used by the \VLDA{} pipeline. \rev{Image, language and sensor/context fields define the supervised records; dispatch context belongs to the separate execution protocols.}

\begin{table}[!t]
\centering\rmfamily
\caption{\VLDA{} Dataset Schema and Prediction Targets}
\label{tab:vlda_schema}
\footnotesize
\renewcommand{\arraystretch}{1.13}
\setlength{\tabcolsep}{4pt}

\begin{tabularx}{\columnwidth}{
>{\raggedright\arraybackslash}p{2.05cm}
>{\raggedright\arraybackslash}X}
\toprule
\textbf{Field group} & \textbf{Contents} \\
\midrule

Image
& Robot-view RGB image or controlled augmented crop-view sample \\

Language
& User request, mission prompt, or task instruction \\

Sensor and zone context
& Current-zone code and global/per-zone temperature, humidity, and soil-moisture measurements \\

Regression targets
& Twelve environmental references: temperature, humidity, and soil moisture for Zones~1--4 \\

Classification target
& Target management zone \(z^{*}\in\{1,2,3,4\}\) \\

Binary-action target
& Apply or skip localized pesticide intervention \\

Dispatch context
& Timestamp, robot pose, local map status, safe-goal status, and manipulator reachability \\

Evaluation artifacts
& Sample identifier, image path, predicted references, class probabilities, binary-action probability, errors, confusion matrices, and summary metrics \\

\bottomrule
\end{tabularx}
\end{table}
\subsection{\rev{Localization, environmental response and robotic protocols}}
\label{sec:execution}
\label{subsec:zone_grounding}
\rev{Point-LIO \citep{he2023pointlio} is evaluated by offline replay on five simulated routes. Inputs comprise 16 rings with 200 azimuth samples per ring, 10 Hz rolling scans acquired over 0.1 s, a body-origin IMU at 100 Hz, 35 m range, and a body-aligned LiDAR translation of $[0,0,0.33]$ m. Range, dropout and inertial perturbations vary across routes. The estimator reads LiDAR and IMU streams; a separate evaluator reads reference pose. Collision geometry generates returns. Full-route rigid and initial-pose alignments are both reported, using the metrics in Section~\ref{sec:protocols}. Navigation uses simulator pose, as defined by the scope above. Standard estimation mathematics is given in Supplementary Section S7.}

\label{subsec:environmental_execution}
\rev{The environmental component is an Isaac Sim extension with zone temperature and relative-humidity states, bed water-balance states, and bounded exhaust, circulation and irrigation fractions. Six 12-hour scenarios use synthetic weather inputs and 1 Hz logging: dry substrate, high humidity, high temperature, hot/dry conditions, combined disturbance and actuator saturation. Evaluation measures reference error, recovery, sustained settling and actuator saturation. The protocol and complete outcomes, including failures, are reported in Section~\ref{sec:climate} and Supplementary Section S4. The present characterization is limited to these logged responses; the missing process parameterization is identified in Section~\ref{sec:limitations}.}

\label{subsec:quadruped_navigation}
\label{subsec:plant_intervention}
\rev{The geometric arm protocol comprises 32 accepted crop-edge targets and eight far-boundary requests. It uses 3 s settling, 6 s control, a 0.95 m standoff, target heights of 0.65--0.85 m and a target-disk radius of 30 mm. Pose acceptance requires position error below 30 mm and nozzle-axis error below 0.1 rad. The modeled nozzle is the finger-link origin. Integrated full-route, left-target and right-target missions use scenario-defined diagnosis and target positions to evaluate navigation, dwell, approach, treatment-command sequencing and stow. The stationary-base admission condition is stated in Section~\ref{subsec:safe_dispatch}. Geometric pose success, ray/disk intersection and modeled treatment units are reported as separate outcomes.}

\label{subsec:execution_logging}
\rev{Navigation/control cohorts comprise ten nominal repeats, six clear-layout robustness cases, two moving obstacle-stop cases and 20 valid arm-proxy controls. Arm requests, three integrated missions, 90 isolated forced faults and nine mission-injected faults have separate denominators. Command speed, simulated elapsed time, physics-step checks and dose-command checks retain their distinct units. These definitions prevent pooling unrelated checks into one success rate.}

\subsection{\rev{Experimental protocols and statistical reporting}}
\label{sec:protocols}

The \rev{original} \AgriSense{} \rev{implementation} was \rev{evaluated} under Ubuntu~22.04.5~LTS using Python~3.10.12 on a workstation with \rev{an} Intel Core i9-13980HX processor, an NVIDIA GeForce RTX~4060 Laptop GPU with 8~GB dedicated memory, \rev{and} 15.2~GB RAM\rev{. The added checkpoint replay and image-model campaign are separate executions; replay timing uses an RTX~5090 Laptop GPU, PyTorch~2.9.1+cu128, transformers~4.37.2}, and \rev{FP32} \rev{inference.} \rev{Hardware and timing scopes are not combined across experiments}.

\rev{The repeated \VLDA{} experiment trains the actual projection, environmental, fusion, and three-head modules with cached frozen-backbone features. Full, no-image, no-language, and no-sensor/current-zone conditions are each trained with five seeds on both the historical and grouped splits, giving 40 retrainings. Removed embeddings are zeroed during both training and \rev{evaluation}, preserving fusion width. Five-seed image-only zone probes use ResNet-50, EfficientNet-B0, ViT-B/16, and CLIP image features on the same grouped test. These probes receive less information and have different heads/budgets, so they are not a matched-information proof of fusion superiority. A separate twelve-condition experiment perturbs one fixed historical checkpoint; inference-time removal is not called retraining.}

\rev{An additional campaign trains five sensor/current-zone-only models, five language-only models and five matched full controls on the unchanged grouped split. All branches execute before post-projection masking, preserving dropout random-number consumption and the 514,065-parameter topology; masks apply during training, validation and test. Sensor-only retains all 19 structured inputs; language-only zeros the complete structured branch. The AdamW, normalization, loss, epoch limit, patience and validation-selection settings are unchanged. The current full controls exactly reproduce historical full predictions, tensors, normalization and histories. The new runs use CPU execution; timings are not compared with prior GPU latency. No historical/current seed pooling is performed.}

\rev{The real-image benchmark is deliberately separate from \VLDA{}'s three-output task. PlantVillage supplies 54,305 decoded colour images; conservative source/leaf/similarity grouping quarantines 5,207 cross-label component members and retains 49,098 images with 34,360/4,910/9,828 train/validation/test membership and 38 crop--condition labels. Quarantine may exclude valid hard lookalikes; 13,194 records lack resolved physical-leaf IDs. PlantSeg v7 supplies 7,774 decoded images/masks, of which 7,551 remain after 223 unique annotation/conflicting-label exclusions. The external classification subset contains 238 official-test images, 238 groups, and fifteen compatible true classes, while keeping all 38 model outputs and unrestricted argmax. No environmental references, greenhouse-zone targets, or intervention labels are fabricated for these datasets.}

\rev{Four frozen-image representations each have five trained linear heads, and CLIP also has one fixed image--text zero-shot evaluation using predeclared crop/disease templates. Nine further experiments fine-tune all parameters of ResNet-50, EfficientNet-B0, and ViT-B/16 over three seeds and at most ten epochs. Full-network AdamW backbone/head learning rates are $10^{-5}/10^{-4}$, weight decay $10^{-4}$, batch size 64, and checkpoint selection uses validation cross-entropy. Frozen probes use learning rate $10^{-3}$, batch 512, and at most 100 epochs with patience 10. Different augmentation, capacity, and budgets prevent interpreting the two cohorts as an isolated unfreezing ablation. Plant Pathology 2021 remained access-blocked (HTTP 401), and no PlantSeg segmentation model was trained.}

\rev{Continuous errors are reported separately in degrees Celsius and percentage points; the original mixed-unit MAE is retained only as a historical scalar. Classification uses accuracy, fixed-class macro-F1, per-class metrics, and confusion matrices. Seed means and sample standard deviations describe optimization variability, not test-sampling confidence. Source-group bootstrap intervals resample whole discovered groups; crossed seed/group intervals also resample fitted seeds. The available exports use 1,000 draws with declared missing-class handling, while balanced-accuracy intervals are conditional on class-complete draws. No resampling of one episode creates between-greenhouse evidence. Calibration, corruption, risk--coverage, and out-of-distribution diagnostics, together with their thresholds and prevalence, are retained in the supplement rather than merged into intervention-safety claims.}

\rev{Localization reports initial-pose-aligned and all-position rigid $\mathrm{SE}(3)$-aligned absolute trajectory error (ATE), with scale fixed to one, and one-second relative pose error (RPE). Ground-truth position is interpolated, orientation uses SLERP, and no extrapolation is performed; the RPE pairing tolerance is 0.12~s. A full-route rigid fit and an initial-pose transform answer different questions and are therefore both retained. Robot, arm, and fault cohorts are counted separately; neither physics-step counts nor image-model seeds are treated as independent integrated missions.}

\section{\rev{Results and Discussion}}
\label{sec:results}
\subsection{\rev{Original visual records and their evidentiary scope}}
\rev{The original visual sequence, historical tables and narrative are retained in Supplementary Material, Section S6. The main results figures now emphasize the added quantitative experiments. Original trajectory logs and maps are historical records whose estimator provenance is not equivalent to the separately verified offline Point-LIO experiment. Figure artwork is not recolored; revised captions state these boundaries.}

\subsection{\AgriSense{} Application and Remote Supervision}
\label{subsec:remote_supervision_results}

The \AgriSense{} application functions as a non-real-time supervision
layer. It visualizes edge-generated telemetry, greenhouse measurements,
perception outputs, local maps, mission coverage, diagnostics, and
validated action states. \rev{The interface design treats operator inputs as high-level requests rather than direct actuator commands. The screenshots do not establish implementation coverage of every proposed edge-side guard.}

For readability in the supplementary record, \SuppFig{fig:app_ui} includes
six representative interface views. \rev{These screenshots illustrate the supervision workflow.}

\subsection{\rev{Prediction performance and modality dependence}}
\label{sec:vlda_results}

The \VLDA{} model maps prompt text, robot-view RGB observations, greenhouse sensor states, and current-zone context into three types of structured proposal: four-zone environmental references, target-zone selection, and a binary pesticide-intervention \rev{proposal}. Table~\ref{tab:vlda_metrics} \rev{retains} the \rev{original} held-out \rev{results and places the new reproduction beside them}.

\begin{table*}[!tb]
\centering\rmfamily\footnotesize\color{revisionblue}
\caption{\rev{Historical checkpoint results and checkpoint replay on the original 43-row test set. Source-group dependence is addressed separately by the grouped experiment.}}
\label{tab:vlda_metrics}
\setlength{\tabcolsep}{3.6pt}
\renewcommand{\arraystretch}{1.14}
\begin{tabularx}{\textwidth}{YZZ}
\toprule
Metric & Original evaluation & Checkpoint replay \\
\midrule
Target-zone accuracy & 95.35\% (41/43) & 95.35\% (41/43) \\
Target-zone macro-F1 & 0.9311 & 0.9311 \\
Intervention-label agreement & 100\% (43/43) & 100\% (43/43) \\
Mixed-unit reference MAE & 1.266192 & 1.265588 \\
Temperature MAE ($^\circ$C) & Not separated & 0.323820 \\
Humidity MAE (percentage points) & Not separated & 2.234821 \\
Soil-moisture MAE (percentage points) & Not separated & 1.238123 \\
\bottomrule
\end{tabularx}
\par\smallskip\begin{minipage}{\textwidth}\footnotesize Both discrete outputs match the archive; the small regression difference is retained rather than described as bit-exact. Mixed-unit MAE is not a physically homogeneous accuracy measure. Generated intervention labels are not expert treatment decisions.\end{minipage}
\end{table*}

The target-zone metrics indicate that the model generally distinguishes the requested or inferred management region \rev{on} \rev{this} \rev{particular} \rev{split}. The spray-label result should be interpreted cautiously because it was obtained from \rev{only} 43 samples \rev{and is fully reproducible by a prompt-keyword rule}. It establishes \rev{agreement} \rev{with} the \rev{generated} dataset\rev{,} not field-level disease-diagnosis or treatment performance.

\rev{The historical checkpoint's source-group 95\% interval for zone accuracy is 87.80--100\%; the macro-F1 interval is 0.7185--1.0000. These intervals are conditional on the fitted model and observed one-episode groups, and do not repair training/test lineage leakage. Representative original decision-output panels are retained in \SuppFig{fig:vlda_zone_response}; \Supplement{6} retains the original loss summaries and full historical narrative.}

\begin{figure*}[!tb]
\centering\rmfamily
\includegraphics[width=.97\textwidth]{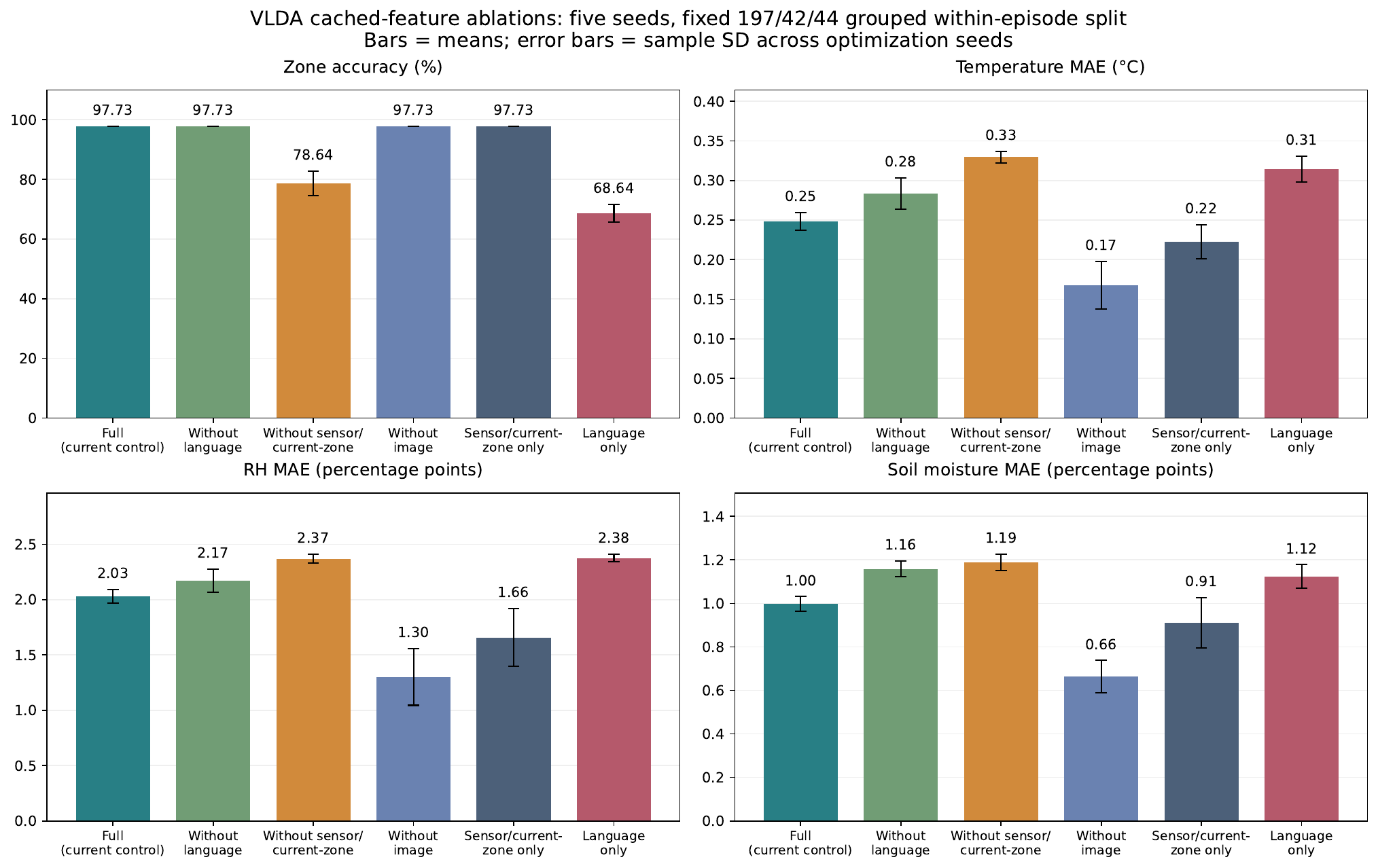}
\caption{\rev{Six-condition modality evaluation on the frozen grouped split. Error bars show sample SD across five optimization seeds. Full, sensor-only and language-only use current runs; the three earlier single-branch removals retain historical results. Matched current full controls reproduce the historical full predictions and model tensors. Sensor-only matches full zone accuracy with lower mean reference errors; language-only has weaker zone performance.}}
\label{fig:ablations}
\end{figure*}

\rev{The new grouped comparison reveals which inputs contribute under the available labels (Table~\ref{tab:ablations}, Fig.~\ref{fig:ablations}). The full model attains 43/44 correct target zones in every seed, but so do the no-image and no-language variants. Removing the complete sensor/current-zone branch lowers average zone accuracy to 78.636\%$\pm$4.129 percentage points. Because that removal excludes both current-zone indicators and continuous environmental values, the contrast does not isolate an abiotic causal effect.}

\begin{table*}[!tb]
\centering\rmfamily\footnotesize\color{revisionblue}
\caption{\rev{Five-seed VLDA retraining on the 197/42/44 grouped split. Values are means plus/minus sample SD; pp denotes percentage points.}}
\label{tab:ablations}
\setlength{\tabcolsep}{3.6pt}
\renewcommand{\arraystretch}{1.14}
\begin{tabularx}{\textwidth}{p{.24\textwidth}ZZZZZ}
\toprule
Condition & Zone accuracy (\%) & Macro-F1 & Temp. MAE ($^\circ$C) & RH MAE (pp) & Soil MAE (pp) \\
\midrule
Full model & $97.727\pm0.000$ & $0.9762\pm0.0000$ & $0.249\pm0.011$ & $2.030\pm0.061$ & $0.997\pm0.034$ \\
Without language & $97.727\pm0.000$ & $0.9762\pm0.0000$ & $0.283\pm0.020$ & $2.172\pm0.104$ & $1.158\pm0.036$ \\
Without image & $97.727\pm0.000$ & $0.9762\pm0.0000$ & $0.168\pm0.030$ & $1.302\pm0.257$ & $0.664\pm0.074$ \\
Without sensor/current zone & $78.636\pm4.129$ & $0.6685\pm0.0751$ & $0.329\pm0.007$ & $2.369\pm0.040$ & $1.188\pm0.037$ \\
Sensor/current-zone only & \(97.727\pm0.000\) & \(0.9762\pm0.0000\) & \(0.222\pm0.021\) & \(1.657\pm0.262\) & \(0.911\pm0.115\) \\
Language only & \(68.636\pm2.963\) & \(0.3988\pm0.0673\) & \(0.314\pm0.017\) & \(2.375\pm0.034\) & \(1.123\pm0.054\) \\

\bottomrule
\end{tabularx}
\par\smallskip\begin{minipage}{\textwidth}\footnotesize Full and language-only intervention-label agreement is 100\%; sensor-only agreement is 97.727\% with F1 0.9744 in each seed. These measure generated-label agreement. Identical accuracy does not imply identical probabilities, regression errors, or calibration; one source-grouped episode remains the sampling population. Full and the two joint-removal conditions use five current runs each; the three earlier single-branch removals use historical runs. Current full controls reproduce the historical full tensors, predictions, histories and normalization exactly; historical/current seeds are not pooled.\end{minipage}
\end{table*}

\rev{The environmental-reference results qualify the apparent zone performance. Removing language increases the mean temperature/RH/soil errors from 0.249/2.030/0.997 to 0.283/2.172/1.158 in their respective units, whereas removing images reduces them to 0.168/1.302/0.664. The lower no-image errors caution against attributing the full model's output quality to visual evidence. \rev{On this small teacher-labeled dataset, an additional modality does not necessarily provide independent information about the targets.} A future dataset must vary crop observations and intervention outcomes independently of prompt wording and editing format before the necessity of all three modalities can be established.}

\rev{Image-only grouped zone probes obtain mean accuracies of 65.00\% (ResNet-50), 64.55\% (EfficientNet-B0), 70.00\% (ViT-B/16), and 73.64\% (CLIP image probe). The full model's higher 97.73\% is a comparison across different information sets, not evidence that its fusion MLP is intrinsically superior. All twelve fixed-checkpoint perturbations retain perfect generated intervention-label agreement; changes in image brightness, blur, noise, or prompt content therefore do not resolve the label shortcut or establish robust disease recognition.}

\rev{The added sensor/current-zone-only condition matches the full model at 43/44 correct zones and macro-F1 0.9762 in every seed. Its temperature/RH/soil MAEs are $0.222\pm0.021$~$^\circ$C, $1.657\pm0.262$ and $0.911\pm0.115$ percentage points, lower in mean than the full model. Language-only reaches $68.636\pm2.963$\% zone accuracy and macro-F1 $0.3988\pm0.0673$, with respective reference MAEs $0.314\pm0.017$, $2.375\pm0.034$ and $1.123\pm0.054$. Sensor-only intervention-label agreement is 43/44 with F1 0.9744 in every seed; language-only retains 44/44 agreement. Thus the structured inputs suffice for the observed zone accuracy, while generated intervention labels remain predictable from language alone. These findings characterize this supervision and split, without identifying a unique causal feature pathway or establishing treatment validity.}

\subsection{\rev{Real-image benchmarks and external transfer}}
\label{sec:image_results}

\rev{The public-image experiments assess the limits of supporting visual representations, not the greenhouse reference/zone/intervention contract. Table~\ref{tab:image_results} reports clean PlantVillage and compatible external PlantSeg results. ViT-B/16 has the highest observed full-network clean accuracy, 99.342\%$\pm$0.149 percentage points, but external accuracy is 27.171\%$\pm$1.698. The corresponding approximate crossed seed/group intervals are 99.1097--99.5553\% and 21.8487--32.7801\%, respectively. Thus high within-dataset disease-label accuracy does not establish a reliable visual component for greenhouse deployment.}

\begin{table*}[!tb]
\centering\rmfamily\footnotesize\color{revisionblue}
\caption{\rev{Separate disease-image benchmarks. External F1 uses the fifteen supported true classes; all 38 predicted classes are retained.}}
\label{tab:image_results}
\setlength{\tabcolsep}{3.6pt}
\renewcommand{\arraystretch}{1.14}
\begin{tabularx}{\textwidth}{p{.27\textwidth}ZZZZ}
\toprule
Model/regime & Runs & PV accuracy (\%) & External accuracy (\%) & External macro-F1 \\
\midrule
ResNet-50 / frozen probe & 5 & $96.803\pm0.059$ & $20.168\pm1.188$ & $0.1764\pm0.0088$ \\
EfficientNet-B0 / frozen probe & 5 & $97.271\pm0.064$ & $24.622\pm0.703$ & $0.2305\pm0.0094$ \\
ViT-B/16 / frozen probe & 5 & $98.209\pm0.075$ & $21.681\pm0.967$ & $0.1611\pm0.0128$ \\
CLIP / image probe & 5 & $97.900\pm0.042$ & $24.874\pm0.690$ & $0.2198\pm0.0075$ \\
CLIP / fixed zero-shot & 1 & 12.566 & 30.252 & 0.2546 \\
ResNet-50 / full network & 3 & $98.911\pm0.097$ & $25.070\pm0.485$ & $0.2076\pm0.0172$ \\
EfficientNet-B0 / full network & 3 & $98.589\pm0.117$ & $25.770\pm0.875$ & $0.2149\pm0.0126$ \\
ViT-B/16 / full network & 3 & $99.342\pm0.149$ & $27.171\pm1.698$ & $0.2293\pm0.0177$ \\
\bottomrule
\end{tabularx}
\par\smallskip\begin{minipage}{\textwidth}\footnotesize PV test: 9,828 images/38 labels. External test: 238 images/15 supported true classes. Error terms are seed sample SD, not confidence intervals. The untuned zero-shot classifier has no optimization-seed SD. Public datasets do not supply the missing greenhouse action labels.\end{minipage}
\end{table*}

\rev{Filtering PlantVillage test targets to the same fifteen true class names retains 4,106 images in 1,948 groups. Under that post hoc comparison, full-network ViT has an external-minus-PV accuracy gap of $-72.212$ percentage points, with interval $[-77.662,-66.355]$. Class frequencies, organs, backgrounds, and acquisition conditions still differ; the comparison is more informative than directly equating the full 38-class and 15-class test mixtures, but it does not isolate a single shift mechanism. This is consistent with positioning agricultural conversational models and planning systems by their task contracts rather than transferring headline accuracy claims between tasks.}

\rev{Image degradation also matters: under the declared blur, full-network EfficientNet-B0 accuracy falls to 36.91\%, ResNet-50 to 59.14\%, and ViT-B/16 to 85.54\%. The supplement preserves complete frozen/full-network learning curves, per-class results, calibration and risk--coverage plots, and separate novel-host and known-host condition diagnostics. These analyses neither add independent greenhouse episodes nor measure the safety of the intervention proposal. In particular, class-prompt CLIP is an image--text baseline, not a matched greenhouse action planner.}

\begin{figure*}[!tb]
\centering\rmfamily
\includegraphics[width=.94\textwidth,height=15.3cm,keepaspectratio]{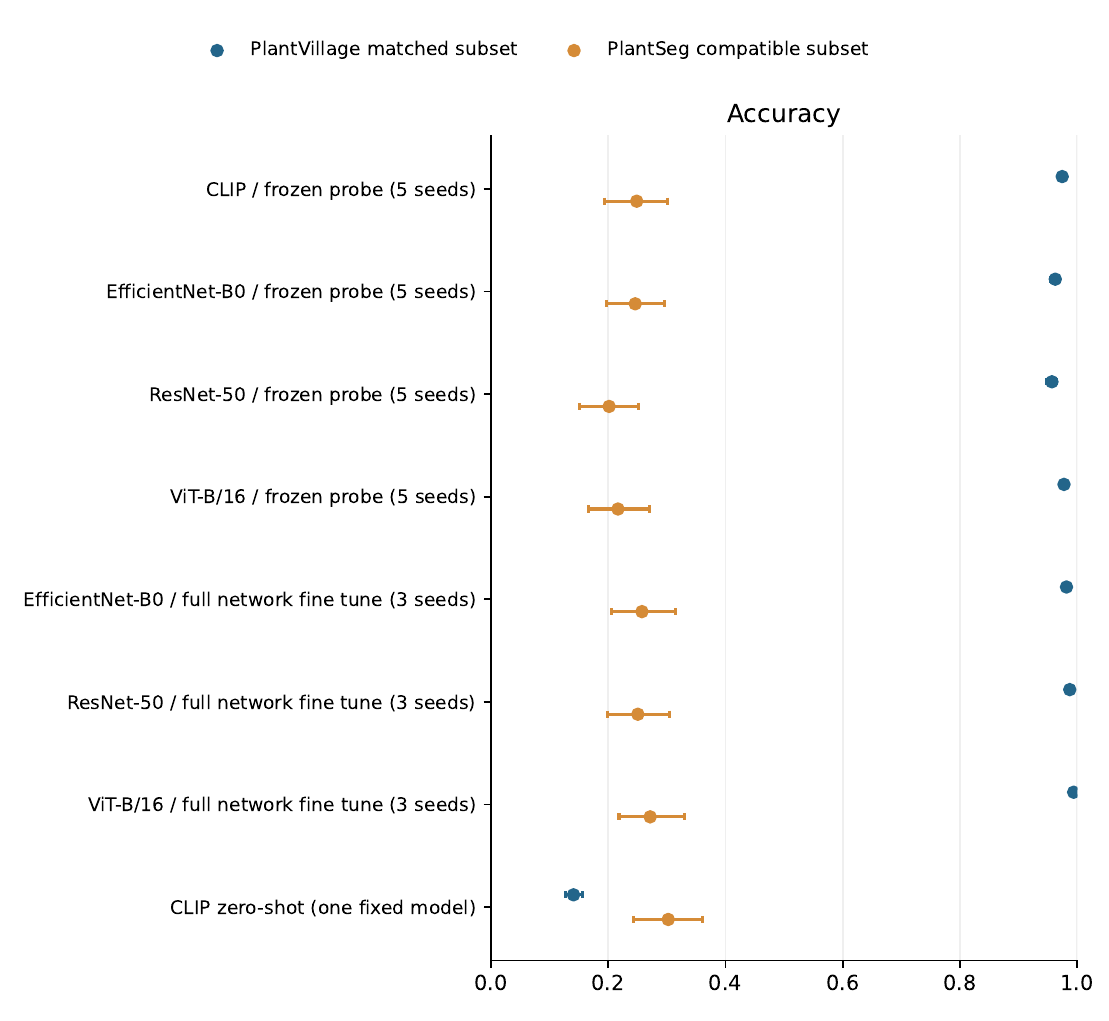}
\caption{\rev{\revsecond{Shared-true-class accuracy contrast. A} post hoc filter retains the same fifteen true class names in 4,106 PlantVillage test images and 238 external images, without restricting the 38-way prediction space. Intervals are approximate crossed seed/group percentile bounds, not a pure causal domain effect. \revsecond{The complete accuracy/F1/calibration transfer panels, the full-network PlantVillage comparison, and the underlying tables remain in \Supplement{2}.}}}
\label{fig:image_comparison}
\end{figure*}

\rev{Figure~\ref{fig:image_comparison} visualizes the shared-class transfer comparison; the large external performance loss limits claims of visual generalization.}

\subsection{\rev{Localization and navigation in the virtual greenhouse}}
\label{sec:lio_results}

\begin{figure*}[!tb]
\centering\rmfamily
\begin{subfigure}[t]{0.48\textwidth}\centering\rmfamily
\includegraphics[width=\linewidth]{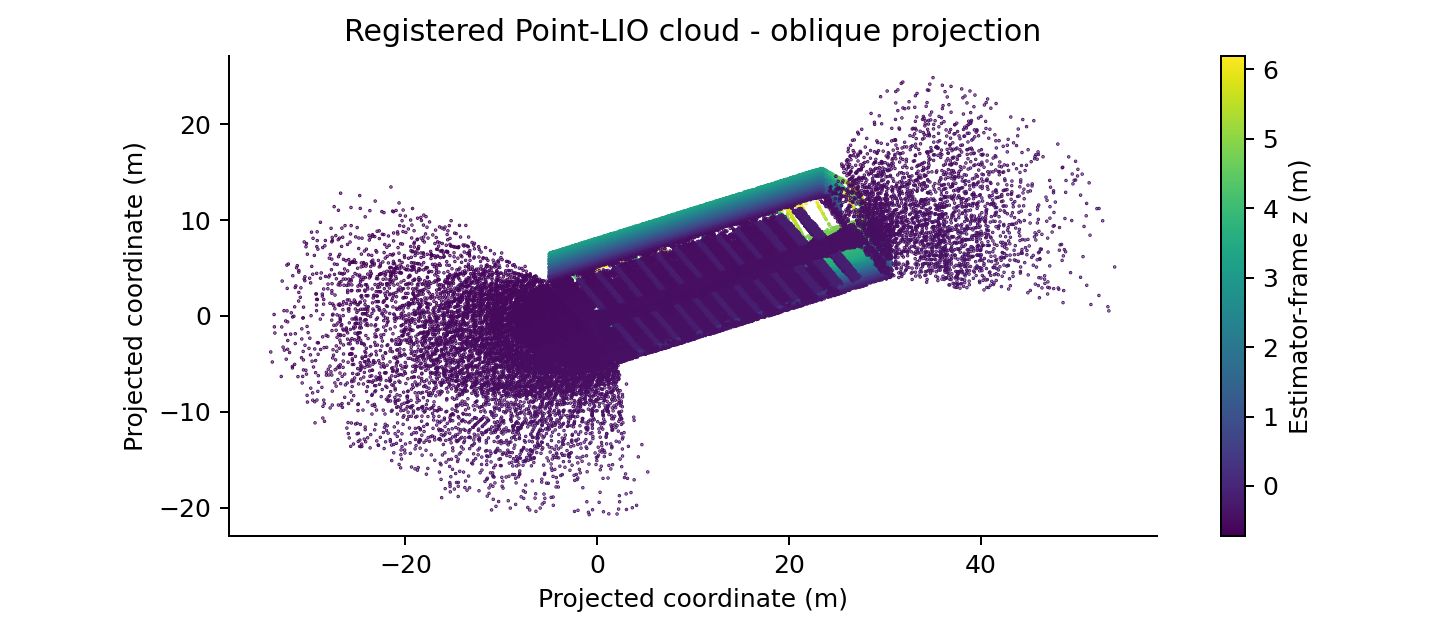}
\caption{\rev{Registered Point-LIO map from route 00.}}
\end{subfigure}\hfill
\begin{subfigure}[t]{0.48\textwidth}\centering\rmfamily
\includegraphics[width=\linewidth]{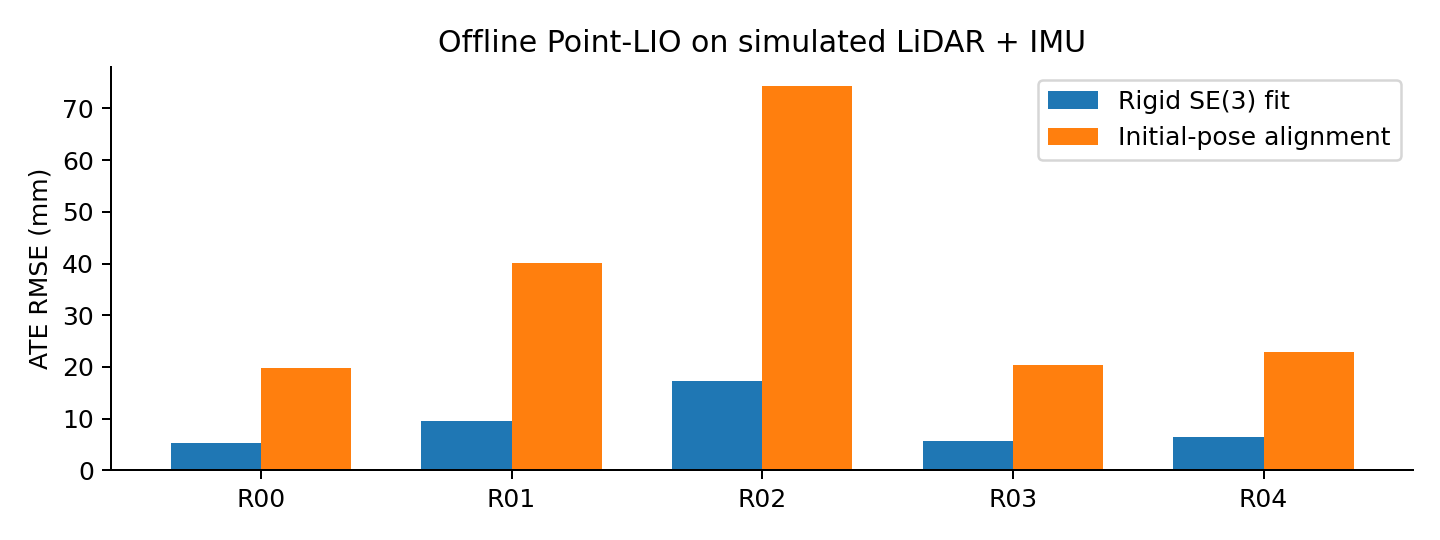}
\caption{\rev{Initial-pose and rigid-fit trajectory errors.}}
\end{subfigure}
\caption{\rev{Actual offline LiDAR--IMU estimation in the virtual greenhouse. Both alignment conventions are shown; navigation still uses simulator pose. The reconstruction therefore provides estimator evidence, not a demonstration of closed-loop Point-LIO-driven navigation or measured physical greenhouse accuracy.}}
\label{fig:pointlio}
\end{figure*}

\rev{The new localization evaluation contains five actual Point-LIO reconstructions with route lengths of 26.83--27.12~m (Table~\ref{tab:pointlio}). The registered cloud in Fig.~\ref{fig:pointlio} is the route-00 PCD output, with 513,833 voxel-deduplicated points at 0.05~m spacing; the displayed view subsamples every sixth stored point. It is not greenhouse geometry transformed by ground-truth pose and relabelled as an estimator map.}

\begin{table*}[!tb]
\centering\rmfamily\footnotesize\color{revisionblue}
\caption{\rev{Five offline Point-LIO routes. ATE columns use different alignment conventions and must not be interchanged.}}
\label{tab:pointlio}
\setlength{\tabcolsep}{3.6pt}
\renewcommand{\arraystretch}{1.14}
\begin{tabularx}{\textwidth}{YZZZZ}
\toprule
Route & Length (m) & Initial ATE (mm) & Rigid-fit ATE (mm) & RPE 1 s (mm) \\
\midrule
00 & 26.89 & 19.80 & 5.22 & 3.80 \\
01 & 26.83 & 40.12 & 9.63 & 4.77 \\
02 & 26.97 & 74.26 & 17.32 & 6.26 \\
03 & 27.05 & 20.37 & 5.71 & 4.16 \\
04 & 27.12 & 22.89 & 6.35 & 4.36 \\
\bottomrule
\end{tabularx}
\par\smallskip\begin{minipage}{\textwidth}\footnotesize \rev{Initial ATE uses the first overlapping full pose; rigid-fit ATE uses least-squares SE(3) alignment over all overlapping positions with scale fixed to one. RPE measures one-second relative pose differences. The five routes evaluate offline Point-LIO on synthetic LiDAR--IMU streams; the separate navigation experiments use simulator pose. The historical three-mission cohort is reported separately in the supplement.}\end{minipage}
\end{table*}

\rev{The difference between initial and fitted ATE is substantial on route~02 (74.26 versus 17.32~mm), illustrating why the smaller fitted value alone would \rev{obscure the influence of frame alignment and accumulated drift.} RPE of 3.80--6.26~mm characterizes short-horizon relative motion under the supplied synthetic streams. Low aligned errors cannot establish real-sensor calibration, greenhouse survey accuracy, or plant-target registration. The previous three long-route traces remain in \SuppFig{fig:trajectory_error_results}, with full historical records in \Supplement{6}, as diagnostics rather than silently pooled with the five new routes or retained as verified LIO-driven navigation evidence.}

\rev{The separate arm-proxy navigation/control suite reports 10/10 nominal route completions, 6/6 clear robustness cases, two intended obstacle stops, and 20/20 valid controls. Nominal routes use four scheduled stops and a 300~s cap; start pose and policy are fixed for those repeated seeds, and estimator-noise changes do not drive navigation. The descriptive Wilson interval for 10/10 is 72.2--100.0\%. A 0.50~m/s aisle-transit command is not a minimum instantaneous speed: docking and stops are slower or stationary. This suite characterizes the executed policy/controller under its specified conditions, not autonomous performance under estimated localization or field reliability.}

\rev{Figure~\ref{fig:pointlio} shows the newly reconstructed Point-LIO map and both trajectory-alignment conventions.}

\subsection{\rev{Environmental dynamics and operating limits}}
\label{sec:climate}

The greenhouse-state logs contain temperature, humidity, and soil-moisture measurements for all four management zones. \rev{The} \rev{original} end-to-start \rev{changes and raw/rolling-median plots are retained} in \rev{\Supplement{6};} \rev{they} \rev{characterize} \rev{state variation rather than controller transients}. \rev{The additional six-case evaluation provides long-horizon regulation evidence under specified forcing and saturation conditions.}

\begin{table*}[!tb]
\centering\rmfamily\footnotesize\color{revisionblue}
\caption{\rev{Environmental-control outcomes and reported limitations. Recovery and sustained settling are different criteria.}}
\label{tab:climate}
\setlength{\tabcolsep}{3.6pt}
\renewcommand{\arraystretch}{1.14}
\begin{tabularx}{\textwidth}{p{.24\textwidth}Y Y}
\toprule
Case/quantity & Reported outcome & Interpretation \\
\midrule
High temperature & Within 0.5~$^\circ$C after 2,461~s & Sustained settling reported for this case \\
High humidity & Within 2 percentage points after 3,448~s & Sustained settling reported for this case \\
Dry-substrate VWC & 90\% recovery after 7,457~s; final 0.33266 & Does not meet sustained $\pm0.01$ VWC settling \\
Hot/dry & Final 42.97~$^\circ$C and 10.64\% RH & Temperature, RH, and VWC fail settling \\
Combined disturbance & VWC fails sustained settling & Multi-disturbance operating limit \\
Actuator saturation & Temperature fails sustained settling & Actuator-limited operating envelope \\
\bottomrule
\end{tabularx}
\par\smallskip\begin{minipage}{\textwidth}\footnotesize Six of eighteen variable/case rows fail sustained settling. Water totals are model bookkeeping, not measured physical consumption. VWC denotes modeled volumetric water content; it is not interchangeable with historical percentage-scaled soil channels without a calibration mapping.\end{minipage}
\end{table*}

\rev{\rev{These results provide quantitative evidence of environmental regulation within the tested conditions, while also identifying operating limits.} The hot/dry failure shows that the modeled actuator/environment combination cannot reach every reference under every disturbance. \rev{A supervisory controller should therefore distinguish proposed references from references that the actuators can achieve under the current conditions.} The experiments quantify this limitation; they do not yet identify a calibrated physical greenhouse envelope because the complete model coefficients, weather calibration, and cross-zone dynamics are not supplied.}

\begin{figure*}[!tb]
\centering\rmfamily
\includegraphics[width=.97\textwidth,height=5.6cm,keepaspectratio]{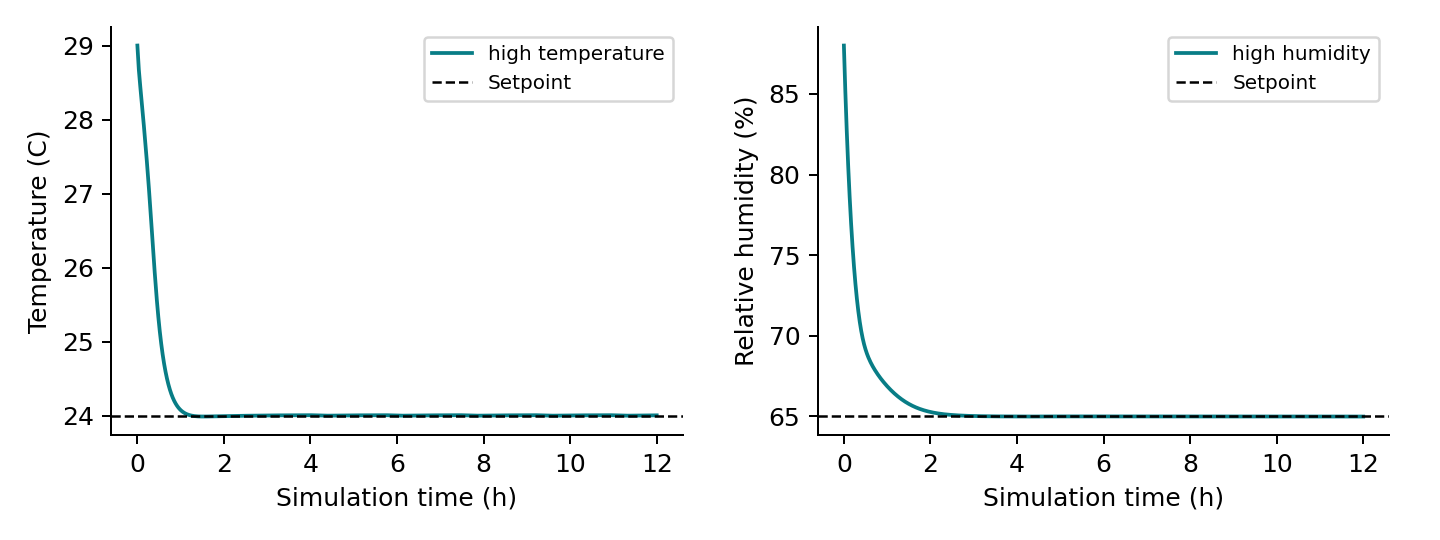}
\caption{\rev{Temperature and humidity regulation examples from the six-case environmental-model suite. The displayed curves are sampled every 60~s from 12-hour, 1~Hz records. These two successful examples do not represent every case: six of eighteen variable/case combinations fail sustained settling, as reported in the main environmental-response table. Weather inputs are synthetic rather than measured UAE weather.}}
\label{fig:climate}
\end{figure*}

\rev{Figure~\ref{fig:climate} shows representative successful temperature and humidity responses. Table~\ref{tab:climate} retains all reported operating limits and failed settling cases.}

\subsection{\rev{Manipulator geometry and integrated mission execution}}
\label{sec:arm_results}

After reaching \rev{an} \rev{admitted} inspection pose, the robot acquires a close robot-view RGB observation. \rev{In} \rev{the integrated evaluation, diagnosis} and \rev{target} \rev{coordinates} are \rev{taken} \rev{from} the \rev{scenario} \rev{state} \rev{so} \rev{that} the \rev{navigation,} \rev{alignment,} \rev{and intervention sequence can be tested separately from uncertain learned perception}.

\rev{The isolated arm benchmark comprises forty pre-specified requests: 32 crop-edge geometric targets and eight far-boundary challenges. All 32 accepted targets meet pose and virtual-disk criteria; the eight unreachable requests are rejected before servo control and are excluded from the accepted-target denominator. Mean position error is $0.449\pm0.154$~mm (maximum 0.853~mm); ray/target-plane error is $0.829\pm0.440$~mm. These are different quantities. The modeled nozzle is the finger-link origin, and nozzle roll is unspecified. Full-body/self-collision contacts remain uninstrumented.}

\begin{table*}[!tb]
\centering\rmfamily\footnotesize\color{revisionblue}
\caption{\rev{Integrated precision-arm missions, separate from the isolated geometric benchmark and arm-proxy controls.}}
\label{tab:integrated}
\setlength{\tabcolsep}{3.6pt}
\renewcommand{\arraystretch}{1.14}
\begin{tabularx}{\textwidth}{YZZ}
\toprule
Case & Completion time (s) & Maximum treated-target ray error (mm) \\
\midrule
Four-stop full route & 130.302 & 12.94 \\
Left berry & 30.802 & 10.99 \\
Right berry & 33.802 & 6.40 \\
\bottomrule
\end{tabularx}
\par\smallskip\begin{minipage}{\textwidth}\footnotesize All three cases complete. The full route visits beds 1, 7, 13, and 19 and performs two modeled interventions. These are selected virtual cases, not three independent field-reliability trials.\end{minipage}
\end{table*}

\rev{The full mission records 4,900 treatment-phase physics-step checks with zero motion violations and 100 admitted instantaneous dose-command samples. These counts describe within-mission checks, not 5,000 independent interventions. The terminal dose is an instantaneous model update on transition to stow; it belongs to dose-command checks but not to the subsequent treatment-phase step count. Treatment units are abstract rather than calibrated liquid volume. The larger integrated target-ray errors compared with the isolated benchmark show why submillimetre pose errors should not be promoted to system-level spray accuracy.}

\rev{Table~\ref{tab:integrated} does not establish reduced chemical use.} Such a
claim requires direct comparison of spray volume, deposition coverage,
off-target application, and treatment effectiveness.

\begin{figure*}[!tb]
\centering\rmfamily
\begin{subfigure}[t]{0.48\textwidth}\centering\rmfamily
\includegraphics[width=\linewidth]{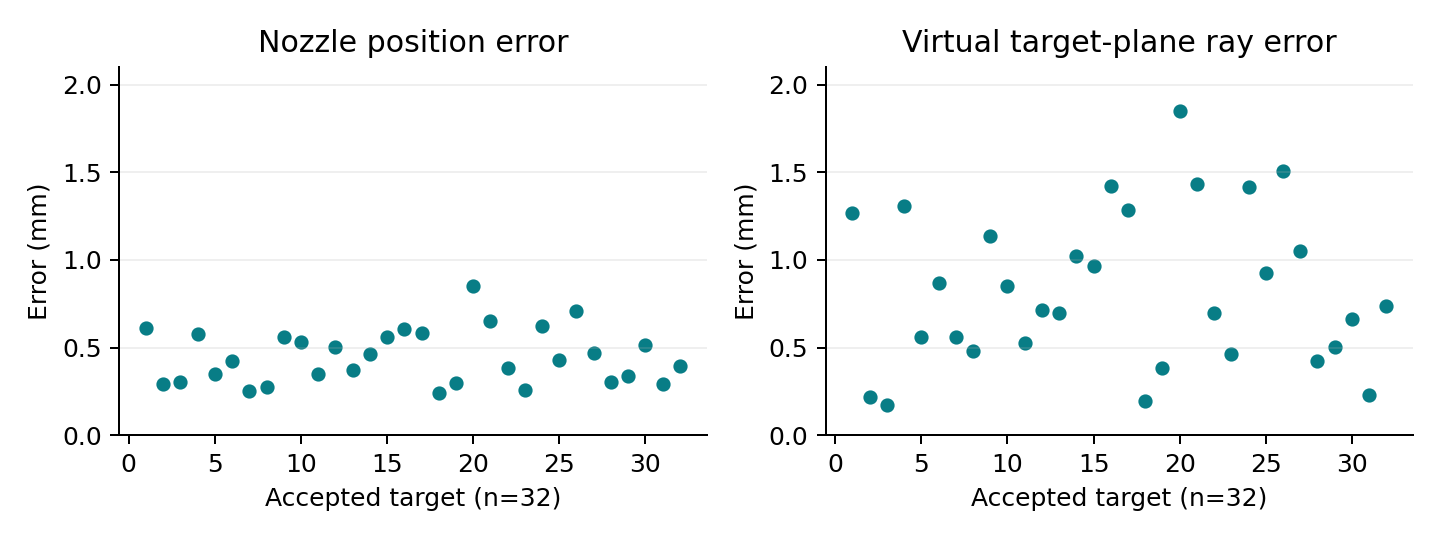}
\caption{\rev{Thirty-two accepted virtual geometric targets.}}
\end{subfigure}\hfill
\begin{subfigure}[t]{0.48\textwidth}\centering\rmfamily
\includegraphics[width=\linewidth]{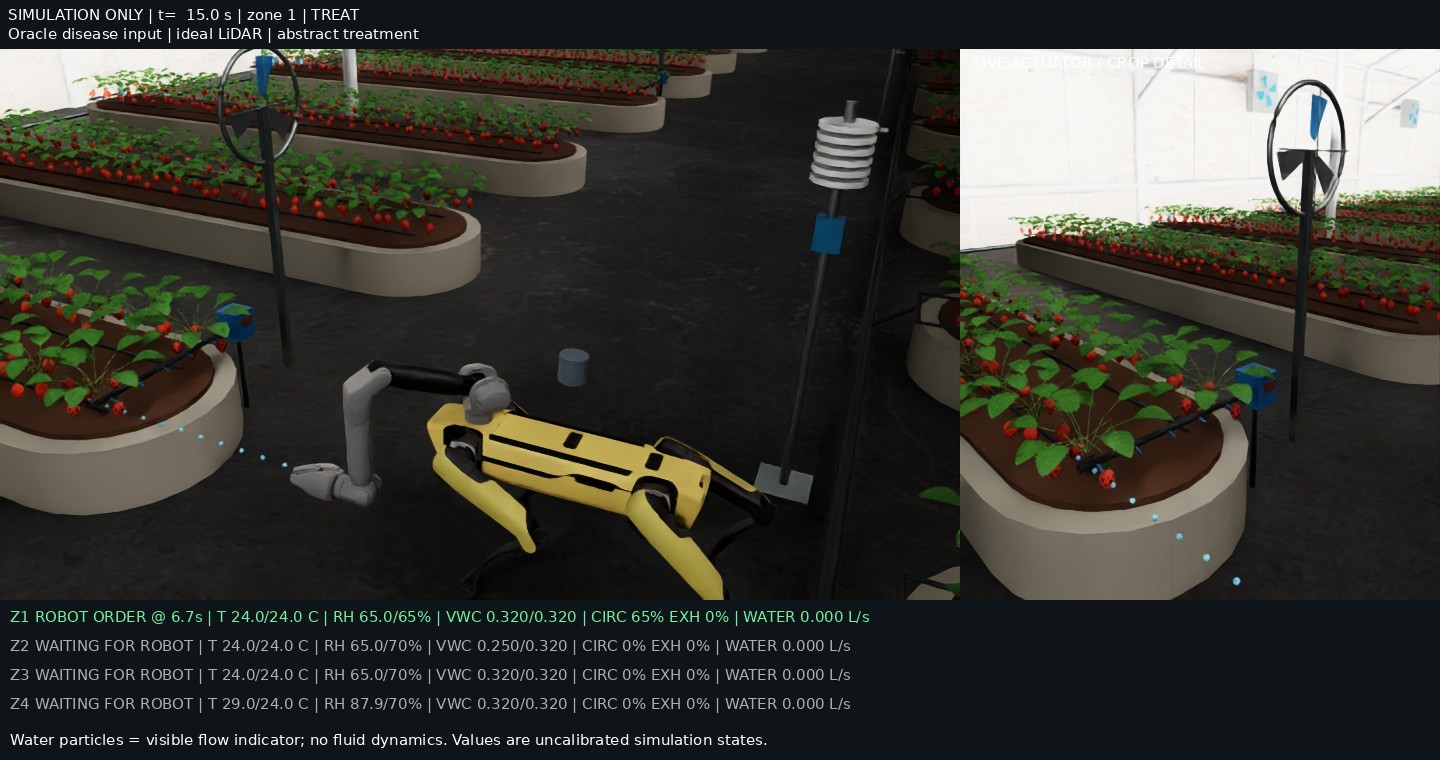}
\caption{\rev{Recorded integrated mission with authored targets.}}
\end{subfigure}
\caption{\rev{Manipulator geometry and mission integration are evaluated on distinct target sets. The isolated benchmark uses targets at $z=0.65$--0.85~m and a 0.95~m standoff; the mission uses lower authored berries and simulator-state diagnosis/positions. Geometric ray error is not pesticide mass deposition, off-target exposure, or biological treatment efficacy.}}
\label{fig:arm}
\end{figure*}

\rev{Figure~\ref{fig:arm} pairs the geometric-targeting results with a recorded authored-target mission; these are separate protocols.}

\subsection{\rev{Input readiness, fault handling, and evidence summary}}
\label{sec:safety_results}

\rev{The source-faithful neural-interface readiness test and the robot-controller fault tests assess different software. The exact readiness method rejects 52/52 engineered missing/stale/desynchronized fixtures and accepts one valid fixture. It also accepts 64/64 non-finite/extreme numerical probes and one synchronized future timestamp. Therefore its demonstrated function is input presence/age/synchrony checking, not complete numerical-range or non-finite protection. \rev{Accepting the single valid fixture does not establish a population false-block rate.} These tests used dummy state and fixed time, without invoking ROS or physical actuation.}

\rev{The separate virtual-controller harness rejects 90/90 isolated faults with zero counted invalid dose and maximum safe-dwell time 0.864~s. Nine faults injected after actual mission treatment admission all stop with safe dwell and no post-fault dose; maximum safe-dwell time is 0.500~s. Fault types include stale sensor, stale diagnosis, non-finite command, out-of-range command, unreachable target, unsafe clearance, disconnect, invalid state, and abort. This shows that the controller's command guard is not the same function as the original neural-interface readiness check. Legitimate dose before injection is retained, and subsequent output and cumulative-dose changes are checked separately.}

\rev{Treatment-phase injections begin near rest. The two moving obstacle cases have stop-to-dwell times of 3.412 and 3.282~s; one exhausts the logged conservative clearance margin to zero, and contacts are not instrumented. Neither these cases nor the near-rest fault tests establish collision-free avoidance, emergency braking from maximum speed, or recovery/restart. Same-supervisor-tick invalid-command detection does not measure perception or network latency. The practical contribution is explicit rejection/stop evidence under stated protocols, rather than an undifferentiated claim of a ``safe'' end-to-end system.}

\begin{table*}[!tb]
\centering\rmfamily\footnotesize\color{revisionblue}
\caption{\rev{Summary of evidence across the original framework and added experiments. Denominators are not pooled.}}
\label{tab:evidence}
\setlength{\tabcolsep}{3.6pt}
\renewcommand{\arraystretch}{1.14}
\begin{tabularx}{\textwidth}{p{.20\textwidth}Y Y}
\toprule
Component & Quantitative evidence & Boundary \\
\midrule
VLDA & Original 41/43; grouped full/no-image/no-text/sensor-only 43/44 per seed; language-only 68.64\%; per-variable MAE & One episode and generated labels; no demonstrated zone synergy \\
Visual baselines & 20 PV probes; nine full-network runs; external and corruption results & Different target space from VLDA; poor compatible transfer \\
Point-LIO & Five actual offline reconstructions; ATE and 1~s RPE & Synthetic streams; not navigation feedback \\
Navigation/control & 10 nominal, 6 clear, 2 obstacle, 20 valid controls & Arm-proxy protocol; simulator pose \\
Manipulator & 32 accepted targets; eight boundary rejects; three missions & Geometry/oracle targets, not deposition or crop response \\
Environmental model & Six 12-hour cases; 6/18 settling failures & Synthetic weather; incomplete process parameterization \\
Fault handling & 90 isolated and nine mission faults; stop/dose checks & Different guards; near-rest tests; no recovery or certification \\
Physical greenhouse & Not performed & Field accuracy, synchronization, agronomic efficacy remain unestablished \\
\bottomrule
\end{tabularx}
\end{table*}

\subsection{\rev{Interpretation, limitations and future work}}
\label{sec:limitations}
\label{sec:discussion}
\label{subsec:discussion_results}
\rev{AgriSense provides a fixed-format observation-to-proposal interface and a common virtual environment for studying its supporting robotic operations. The experimental contribution is the joint characterization of prediction, modality dependence, spatial reconstruction, targeting, environmental response and command admission. The architecture uses established encoders, feature concatenation, Point-LIO and deterministic controllers; novelty is not attributed to those individual components.}

\rev{The modality results distinguish predictive agreement from evidence that each input is necessary. On one grouped episode, image and language removal separately preserve zone accuracy, image removal improves reference MAE, and prompt keywords reproduce the generated intervention labels. These observations expose a supervision problem: aggregate accuracy can reward solutions that need little visual evidence. They do not show that the network uses a particular shortcut, nor establish a general failure of multimodal agricultural learning. The sensor/current-zone-only comparison matches full zone accuracy and lowers mean reference errors; a matched-input alternative multimodal model remains a future comparison. The public-image experiments evaluate crop--condition recognition under distribution shift; they do not supply the missing greenhouse decision labels or a robotics sim-to-real test.}

\label{subsec:system_originality}
\label{subsec:discussion_digital_twin}
\rev{The virtual greenhouse makes sensor placement, robot viewpoints, zone context and geometric intervention constraints explicit. Its useful role is controlled component evaluation before physical deployment. The present integrated missions use simulator pose, authored diagnosis and known target positions, while Point-LIO is tested offline. Consequently, the experiments do not establish an estimated-pose, learned-diagnosis autonomous treatment loop.}

\label{subsec:limitations}
\rev{Four limitations bound the conclusions. First, no physical-greenhouse experiments or biological treatment trials were performed. Sensor calibration, foliage motion, illumination, contact, deposition, crop response and long-duration operation require physical measurements. Second, the decision dataset contains only 283 generated-label records from one episode. Grouping removes identified source links but cannot create independent plants or sites; independent expert decisions and varied prompts are required to break the observed prompt/intervention correspondence. Third, the climate campaign reports logged responses, including six settling failures among 18 variable/case outcomes, but complete governing equations, coefficients, integration settings, controller gains and cross-zone coupling are unavailable in the retained experimental materials. The climate protocol is therefore incompletely reproducible and cannot support calibrated climate-prediction claims. Fourth, geometric targeting and specified rejection/stop tests do not measure chemical deposition, agronomic efficacy, complete collision coverage or recovery/restart reliability. The neural readiness check and virtual-controller command guard remain distinct implementations.}

\label{subsec:future_work}
\rev{Future evaluation should begin with an independently annotated decision dataset in which visual condition, prompt wording, current zone and environmental state vary separately. The completed joint-branch ablations should be extended to independent decision data and a matched-input vision--language comparator using a common output schema and evaluation budget. Physical greenhouse trials should then calibrate sensing and robot geometry, compare estimated-pose navigation with reference measurements, and evaluate perception-derived targets, deposition and crop outcomes. Environmental studies require a fully disclosed process model and controller settings before testing reference changes, disturbances and cross-zone coupling. The extended activity-state, confidence, freshness and dispatch requirements in Supplementary Section S9 define future integration work, to be implemented and tested with fault-injection and recovery protocols.}

\section{\rev{Conclusions}}
\label{sec:conclusions}

This paper presented \AgriSense{}, a framework for \rev{multimodal decision support and robotic intervention in a virtual} greenhouse. The framework integrates an Isaac Sim environment, a quadruped mobile manipulator, RGB, LiDAR, and inertial sensing, \rev{a} \rev{supervised} \rev{vision--language--environmental} \rev{predictor}, environmental regulation, and plant-level \rev{intervention} \rev{interfaces}. \rev{It} \rev{separates} \rev{fixed-format} \rev{learned} \rev{proposals from navigation}, \rev{control}, and \rev{manipulator} \rev{execution}.

\rev{The original held-out target-zone result, 41/43 correct decisions and macro-F1 0.9311, is reproducible. Five-seed retraining on a grouped split gives 43/44 correct target-zone predictions for} the \rev{full}, \rev{no-image}, and \rev{no-language} \rev{variants;} \rev{the} \rev{data} \rev{therefore do not establish an incremental multimodal zone-accuracy advantage}. \rev{The added joint-removal experiments confirm that sensor/current-zone-only inputs match full zone accuracy, with lower mean reference errors, whereas language-only retains perfect generated intervention-label agreement despite weaker zone predictions. These results sharpen the dataset-specific account of modality dependence.}

\rev{Added component experiments provide five offline Point-LIO reconstructions, 32 accepted geometric targets with eight boundary rejections,} three \rev{completed} \rev{integrated} missions, \rev{six long-horizon climate cases}, and \rev{separate fault-handling protocols}. \rev{These results make} the \rev{implementation} \rev{more} \rev{reproducible} and \rev{expose} \rev{operating} \rev{limits} \rev{that} \rev{the} \rev{original} qualitative \rev{demonstrations} \rev{could} \rev{not} \rev{quantify}.

The results \rev{support} the feasibility of \rev{the} \rev{specified} \rev{decision-support} \rev{interface} and \rev{selected} robotic\rev{/control} \rev{operations} within a \rev{common virtual} greenhouse\rev{. Their scientific value lies in making modality contributions, data shortcuts, geometry, and rejection behaviour explicit rather than inferring full} autonomy \rev{from} \rev{component} \rev{accuracy}. Physical greenhouse trials, \rev{independent} \rev{expert-labelled} \rev{decisions}, \rev{estimated-pose} \rev{closed-loop} \rev{navigation}, \rev{calibrated} \rev{climate} \rev{dynamics}, and \rev{deposition/biological} \rev{tests} \rev{remain} \rev{necessary} before \rev{field-readiness} or \rev{treatment-effectiveness} \rev{claims} can be made.

\section*{Code and Data Availability}

The \AgriSense{} project webpage is available at \url{https://lyessaadsaoud.github.io/AgriSense/}. The \rev{accompanying revision package contains the reported result tables, split/provenance descriptions, original retained figure assets, and separate supplementary material. Source} code, experimental configurations, and \rev{permitted} reproducibility materials will be released through the project repository upon publication\rev{.} \rev{Public benchmark images remain subject to their respective licenses;} the \rev{present statement does not imply that every raw checkpoint, image, or simulation log is included in} the \rev{manuscript} \rev{ZIP}.

\section*{Declaration of the Use of Artificial Intelligence}

During the preparation of this manuscript, the authors used OpenAI's ChatGPT and Grammarly to assist with language editing, grammar correction, and \rev{improvements in clarity and readability.} \rev{AI-assisted image augmentation and offline teacher labeling are described separately in the dataset methodology and are not treated as independent biological ground truth.}

\renewcommand{\bibfont}{\fontsize{8pt}{8.5pt}\selectfont}

\clearpage
\onecolumn
\appendix
\section*{Supplementary Material}
\addcontentsline{toc}{section}{Supplementary Material}
\noindent The following appendix reproduces the clean supplementary material in the same LaTeX document. Section and figure numbering use the S-prefix for clarity.
\setcounter{section}{0}
\setcounter{figure}{0}
\setcounter{table}{0}
\setcounter{equation}{0}
\setcounter{algorithm}{0}
\renewcommand{\thesection}{S\arabic{section}}
\renewcommand{\thefigure}{S\arabic{figure}}
\renewcommand{\thetable}{S\arabic{table}}
\renewcommand{\theequation}{S\arabic{equation}}
\renewcommand{\thealgorithm}{S\arabic{algorithm}}
\providecommand{\supppanel}[2]{\begin{minipage}[t]{0.47\textwidth}\centering\includegraphics[width=\linewidth]{Figures/#1}\par\small #2\end{minipage}}
\providecommand{\SuppFigure}[3]{\begin{figure}[p]\centering\includegraphics[width=\linewidth,height=0.70\textheight,keepaspectratio]{Figures/#1}\caption{#2}\label{#3}\end{figure}}
\newif\ifnewmaterial
\newmaterialtrue

\noindent\rev{Sections S1--S5 provide datasets, statistical protocols, diagnostic results and the reproducibility catalogue. Section S6 retains the original experimental records, Section S7 the standard LiDAR--inertial derivation, and Section S8 additional visual records. Section S9 separates the extended conceptual architecture from evaluated methods. This is the clean, consolidated manuscript and supplementary appendix.}

\begin{revblock}

\clearpage
\FloatBarrier
\section{\rev{Source, datasets, grouping and evaluation protocols}}
\label{sec:s-source}

\subsection{\rev{Implemented model and task contract}}
\revsecond{The implementation, \texttt{GreenhouseVLA}, is the three-output supervised predictor described in the main article.} Its checkpoint SHA-256 is
\path{5cb6396ed9eab64384b0e53053694b143efaad7729dc3dc5ff6e9683ff2691eb}.
Strict loading matches every parameter key. Frozen DINOv2-small
\citep{oquab2023dinov2} and BERT-base-uncased \citep{devlin2019bert} supply
384- and 768-dimensional final CLS representations. Their projections produce
256 and 128 features. A separate 19-input branch combines four current-zone
one-hot coordinates and 15 continuous measurements. Concatenated features enter
a 512-to-256-to-256 MLP. Three heads predict 12 environmental setpoints, four
destination-zone logits and one binary pesticide-label logit. There is no
disease-class, lesion-mask, trajectory, uncertainty or neural safety head.

\providecommand{\ValidationRevisionColor}{\color{blue}}
\begingroup\ValidationRevisionColor
\begin{table*}[t]
\centering\scriptsize
\caption{Recovered implemented VLDA, not a replacement disease classifier.}
\label{tab:validation-v1-architecture}
\setlength{\tabcolsep}{3pt}
\renewcommand{\arraystretch}{1.15}
\begin{tabularx}{\textwidth}{p{0.19\textwidth}X p{0.29\textwidth}}
\toprule
Component & Implemented operation & State \\
\midrule
Image encoder & DINOv2-small CLS 384; projection 384 to 256 & Backbone frozen; projection trainable \\
Text encoder & BERT-base CLS 768; max 128 tokens; projection 768 to 128 & Backbone frozen; projection trainable \\
Sensor/context branch & 19 inputs: current-zone one-hot (4) + continuous context (15); MLP 19 to 128 to 128 & Trainable; training-fitted standardization \\
Fusion & Concatenate 256+128+128=512; MLP 512 to 256 to 256 & Trainable; concatenation, not cross-modal attention \\
Three heads & Each 256 to 128, then 12 setpoints / 4 zone logits / 1 pesticide logit & Trainable; no pixel-mask or trajectory output \\
Objective & MSE + class-weighted CE + positive-weighted BCE-with-logits & Outer loss coefficients all 1 \\
\bottomrule
\end{tabularx}
\par\smallskip
\begin{minipage}{\textwidth}\footnotesize Strict checkpoint loading verifies 132,052,881 total parameters: 131,538,816 frozen and 514,065 trainable. The saved run used AdamW, learning rate \(10^{-4}\), weight decay \(10^{-4}\), batch 16, seed 42, maximum 300 epochs, patience 10; history records 50 epochs, with minimum validation loss at epoch 40. The later alternative trainer's mean pooling/Smooth-L1 defaults do not describe this checkpoint. Fresh FP32 batch-one forward latency was median 7.1006 ms, p95 7.4684 ms on an RTX 5090 Laptop GPU, excluding preprocessing/I/O; this is not an end-to-end control latency.
\end{minipage}
\end{table*}
\endgroup


The image path converts to RGB, rescales by $1/255$, resizes the shortest side
to 256 with bicubic interpolation, centre-crops to $224\times224$, and uses
channel means $(0.485,0.456,0.406)$ and standard deviations
$(0.229,0.224,0.225)$. BERT tokenization lowercases using the uncased tokenizer,
truncates and pads to 128 tokens, and passes input IDs and the attention mask.
Both backbones use CLS pooling, not mean pooling. Image/text projections each
use ReLU and dropout 0.1. The sensor MLP is 19-to-128/ReLU/dropout0.1-to-128/ReLU.
Fusion uses ReLU, dropout 0.2, then ReLU. Each head is 256-to-128/ReLU/dropout0.1
followed by its output layer, with no final regression activation.

The sensor order is current-zone one-hot coordinates 1--4, then
global temperature, humidity and soil moisture, followed by the same three
measurements for zones 1, 2, 3 and 4 in order. All 19 coordinates are standardized
using training-only statistics. The 12 regression coordinates are temperature,
humidity and soil-moisture targets for zones 1--4 in that order, also standardized
using training-only statistics and inverse-transformed for physical-unit errors.
The original code replaces a fitted standard deviation below $10^{-6}$ by one.

\providecommand{\ValidationRevisionColor}{\color{blue}}
\begingroup\ValidationRevisionColor
For an RGB image $x_i$, tokenized prompt $t_i$ (with its tokenizer attention
mask), and a training-standardized sensor/context vector $s_i\in\mathbb{R}^{19}$,
the recovered predictor computes
\begin{align}
v_i &= P_v\!\left(D_{\phi}(x_i)_{\mathrm{CLS}}\right)\in\mathbb{R}^{256}, &
\ell_i &= P_t\!\left(B_{\psi}(t_i)_{\mathrm{CLS}}\right)\in\mathbb{R}^{128},\\
e_i &= S_{\eta}(s_i)\in\mathbb{R}^{128}, &
h_i &= F_{\theta}([v_i;\ell_i;e_i])\in\mathbb{R}^{256}.
\end{align}
Here $D_{\phi}$ is frozen DINOv2-small (384-dimensional CLS output),
$B_{\psi}$ is frozen BERT-base (768-dimensional CLS output), and the
projections, sensor MLP, concatenation-fusion MLP and heads are trainable.
The sensor vector includes four current-zone one-hot coordinates and 15
continuous global/zonal temperature, humidity and soil-moisture values.
The heads output
\begin{align}
\widehat{\widetilde r}_i &= H_r(h_i)\in\mathbb{R}^{12}, &
z_i &= H_z(h_i)\in\mathbb{R}^{4}, & a_i &= H_a(h_i)\in\mathbb{R}.
\end{align}
Setpoints in original physical units are obtained by the training-fitted
inverse transformation
$\widehat r_i=\mu_r+\sigma_r\odot\widehat{\widetilde r}_i$.
The zone prediction is $1+\arg\max_{k\in\{0,1,2,3\}}z_{ik}$; the binary
pesticide prediction is $\mathbf{1}\{\operatorname{sigmoid}(a_i)>0.5\}$.
For a minibatch $\mathcal B$, the implemented original objective is
\begin{equation}
\mathcal L_{\mathcal B}=
\operatorname{MSE}(\widehat{\widetilde R},\widetilde R)
+\operatorname{CE}_{w}(Z,y^{\mathrm{zone}})
+\operatorname{BCEWithLogits}_{q}(a,y^{\mathrm{pesticide}}),
\end{equation}
where uppercase $\widehat{\widetilde R}$, $\widetilde R$, and $Z$ collect
minibatch outputs/targets. The code uses default mean reductions: MSE averages
all $12|\mathcal B|$ continuous values, weighted cross-entropy is
\begin{equation}
\operatorname{CE}_{w}(Z,y)=
\frac{\sum_{i\in\mathcal B}w_{y_i}
[-\log\operatorname{softmax}(z_i)_{y_i}]}
{\sum_{i\in\mathcal B}w_{y_i}},
\end{equation}
and positive-weighted BCE-with-logits averages the binary terms over
$|\mathcal B|$; $q$ multiplies the positive-label term. Zone targets use the
zero-based training-class indices $0,1,2,3$. Outer task coefficients are all one,
$w_c=N/(4n_c)$ from training-zone counts and
$q=\max(1,N_{\mathrm{negative}}/N_{\mathrm{positive}})$ from training binary-label
counts. This is MSE with CLS pooling; mean pooling, Smooth-L1, focal loss or a
scheduler from the different historical trainer must not be attributed to
this checkpoint.

These equations describe a supervised finite-parameter predictor. They do
not supply a trajectory optimizer, certified admissible-action set,
calibrated posterior, lesion-mask output or safety theorem. Offline
confidence/OOD utilities were tested separately and are not integrated into
the recovered VLDA actuation path. A validity predicate or constrained
intervention policy remains a proposed extension until implemented and
validated against an explicit expert/rule contract.
\endgroup


The original saved run used one training seed, 42, AdamW with learning rate
$10^{-4}$ and weight decay $10^{-4}$, batch 16, a 300-epoch limit and patience 10.
Its history records 50 epochs and the lowest validation loss at epoch 40.
The historical alternative trainer's mean-pooling, Smooth-L1/focal-loss and
scheduling defaults do not describe this checkpoint. Original training-time
wall-clock cost and peak VRAM are unavailable; present replay timing does not
reconstruct them. Software configurations are associated with their respective training protocols.

\subsection{\rev{Original episode, label provenance and dependence}}
The supervised CSV contains 283 rows: 150 base JPEG frames of
$1280\times720$ pixels and 133 disease-named augmented PNG images of
$1536\times1024$ pixels. All indexed images decode. The raw directory contains
2,917 images, but only 150 enter this labelled CSV; unindexed frames are not
additional labelled samples. The augmentations have 35 botrytis, 52 anthracnose
and 46 powdery-mildew filenames. Source metadata and historical
scripts support offline teacher labelling and image-edit augmentation; the
trimmed final CSV does not preserve a complete per-row teacher/API history.
No paid labelling or image-generation API was run in the validation campaign.
An exact final-CSV filename-stem reconciliation identifies six base-to-augmented
pairs. All six preserve current zone, all 15 continuous context values and all
12 setpoints, but all six change the final prompt, destination-zone target and
pesticide label. \revsecond{The source parents of the remaining 127 augmented records are not represented among the 150 supervised base rows;} this does not imply that their original raw images are
absent. Thus historical initial row-copy code does not support an equation of
complete final-label inheritance. API model, quality, fidelity, seed and weighting
script defaults do not establish per-image API settings, and scene
preservation instructions are not verified geometric preservation.

\begingroup\color{blue}\footnotesize
\setlength{\tabcolsep}{3pt}
\renewcommand{\arraystretch}{1.15}
\begin{longtable}{@{}rrlllll@{}}
\caption{Exact filename-stem matches in the final supervised CSV, with zero-based row indices. \revsecond{All six preserve the source current-zone value, all 15 context values, and all 12 setpoint values, but differ in prompt, target-zone label, and intervention flag.} \revsecond{The source parents of the remaining 127 augmented records are not represented among the 150 supervised base rows;} no claim is made that their raw images are absent. Historical initial row-copy code does not establish complete final-label inheritance.}\label{tab:s-augmentation}\\
\toprule
Base row & Aug. row & Context & \revsecond{12 targets} & Prompt & Zone & Pesticide \\
\midrule\endfirsthead
\multicolumn{7}{l}{\textit{Continued from previous page}}\\
\toprule
Base row & Aug. row & Context & \revsecond{12 targets} & Prompt & Zone & Pesticide \\
\midrule\endhead
\midrule\multicolumn{7}{r}{\textit{Continued on next page}}\\\endfoot
\bottomrule\endlastfoot
102 & 152 & Identical & Identical & Different & 3 to 2 & False to True \\
130 & 163 & Identical & Identical & Different & 3 to 2 & False to True \\
122 & 164 & Identical & Identical & Different & 3 to 2 & False to True \\
108 & 170 & Identical & Identical & Different & 3 to 2 & False to True \\
26 & 181 & Identical & Identical & Different & 3 to 1 & False to True \\
131 & 252 & Identical & Identical & Different & 3 to 2 & False to True \\
\end{longtable}
\endgroup


All base rows have a false pesticide label, and all augmented rows have a true
label. A prompt rule matching the three disease names agrees with all 283 labels,
including all 43 historical test labels. This is a measured shortcut and confound,
not an expert intervention baseline or proof of real disease recognition. The
generated binary labels must not be treated as expert treatment decisions.

The historical split is 198/42/43 training/validation/test rows, stratified jointly
by zone and pesticide label with seed 42. Indices are disjoint, but five
source-lineage pairs cross subsets; three test rows have training relatives.
Twenty cross-split dHash pairs at Hamming distance at most four provide additional
similarity-screening evidence. Source inheritance and image similarity need not
be disjoint categories. Exact encoded/decoded equality alone misses this lineage.
All samples are from one acquisition/simulation episode.

A separate split was frozen before \revsecond{retraining}: 265 connected components
link common source-frame keys or dHash64 distance at most four, producing
197/42/44 rows with components kept together. This addresses discovered relations,
but does not establish independent physical plants, episodes or greenhouses.
For the 20 historical image-only probes, a documented post-training/pre-final
uncertainty amendment replaces per-filename bootstrap units with the previously
audited components. It changes neither predictions nor point estimates and must
not be called an entirely prespecified uncertainty protocol.

\subsection{\rev{Public disease-image data and annotation compatibility}}

\providecommand{\ValidationRevisionColor}{\color{blue}}
\begingroup\ValidationRevisionColor
\begin{table*}[t]
\centering\scriptsize
\caption{Audited sample counts and evaluation units. Derived test subsets are not additional acquired datasets.}
\label{tab:validation-v1-data}
\setlength{\tabcolsep}{3pt}
\renewcommand{\arraystretch}{1.15}
\begin{tabularx}{\textwidth}{p{0.25\textwidth}r p{0.23\textwidth}X}
\toprule
Data or derived evaluation & Samples & Train / validation / test & Interpretation \\
\midrule
Original episode, historical split & 283 & 198 / 42 / 43 & 150 base + 133 disease-named augmentations; teacher/simulation labels \\
Same original episode, grouped split & 283 & 197 / 42 / 44 & 265 screened source/similarity components; one episode \\
PlantVillage retained v2 & 49,098 & 34,360 / 4,910 / 9,828 & 54,305 decode; 5,207 quarantined \\
PlantSeg eligible & 7,551 & 5,228 / 821 / 1,502 & 7,774 image/mask pairs decode; 223 unique exclusions \\
Compatible external test & 238 & -- / -- / 238 & 15 true classes; 238 groups; all 38 outputs retained \\
Novel-host OOD test & 671 & -- / -- / 671 & 19 literal host labels; 666 groups \\
Known-host condition OOD test & 131 & -- / -- / 131 & 15 annotated conditions; 131 groups; exploratory \\
Plant Pathology 2021 & 0 & -- & HTTP 401 access blocker; no images downloaded \\
\bottomrule
\end{tabularx}
\par\smallskip
\begin{minipage}{\textwidth}\footnotesize Historical indices are disjoint but five source-lineage pairs cross subsets; three test rows have training relatives. The prompt-keyword rule matches all 283 pesticide labels. Group/hash screens address discovered relations, not physical-plant, episode or pretraining independence. PlantVillage quarantine includes false-positive pHash chains; 13,194 original image records lack resolved physical-leaf IDs. PlantSeg exclusions are 20 annotation issues plus 205 conflicting exact-duplicate rows, with two rows overlapping: 223 unique.
\end{minipage}
\end{table*}
\endgroup


PlantVillage \citep{mohanty2016plantvillage} was obtained from the specified
author repository. The colour archive has 54,305 valid decoded $256\times256$
images in 38 crop--condition classes. The v2 policy links class-qualified leaf-ID
suggestions, exact encoded and decoded hashes, and DCT pHash64 distance at most
four. Forty-five cross-label components containing 5,207 images are quarantined,
including conservative false-positive similarity chains. The retained 49,098
images are frozen as 34,360/4,910/9,828 train/validation/test images. Their observed
group counts are 13,535/1,932/3,864. A legacy leaf-name alias is handled without
changing image class labels. Across the original image inventory, 13,194 image
records lack resolved physical-leaf IDs. Thus group-disjointness is a documented
screen, not complete plant/site or pretraining independence. The recorded author
declaration is CC BY-SA 3.0; attribution and the archived licence text are retained.

PlantSeg v7 \citep{plantseg2025} contains 7,774 decoded image/mask pairs, 34 host
labels and 115 crop--condition categories. Twenty annotation inconsistencies and
205 rows in exact-image groups with conflicting labels overlap in two rows:
223 unique rows are excluded, leaving 7,551 eligible pairs, including 1,502 official
test images. Masks use background zero and disease index plus one. The dataset has
no healthy class. Archive metadata record 7,741 per-image CC-BY-NC abbreviations
and 33 CC0 entries; the dataset-level declaration is CC BY-NC 4.0. No version is
inferred for an abbreviated per-image licence. Use remains noncommercial research.

The compatible classification subset has 238 official-test images in 238 screened
source/similarity groups with 15 supported true classes. Source URLs, encoded and
decoded equality and pHash are checked against all retained PlantVillage images;
ambiguous taxonomies and conflicting components are excluded. All 38 output
classes and unrestricted argmax remain. A post hoc comparison restricts existing
PV test true classes to those same 15 labels, yielding 4,106 images/1,948 groups;
it is not a new independent split or a causal domain intervention.

Plant Pathology 2021 FGVC8 remains unexecuted: official file-list access returned
HTTP 401 and no exposed authorized session was available. Zero images were
downloaded; no new terms were accepted. Its multilabel task was not replaced by
synthetic labels or an incompatible single-label merge. The public disease/mask
datasets do not supply the original model's 19 measured context inputs, 12
setpoints, destination-zone labels or expert pesticide decisions. They cannot
validate the unchanged VLDA control-label task by inventing missing annotations.

\rev{The checkpoint architecture is shown in main Figure 5; its input and parameter details are provided in the architecture table above.}

\begin{figure}[!h]
\centering
\includegraphics[width=\linewidth,height=0.7\textheight,keepaspectratio]{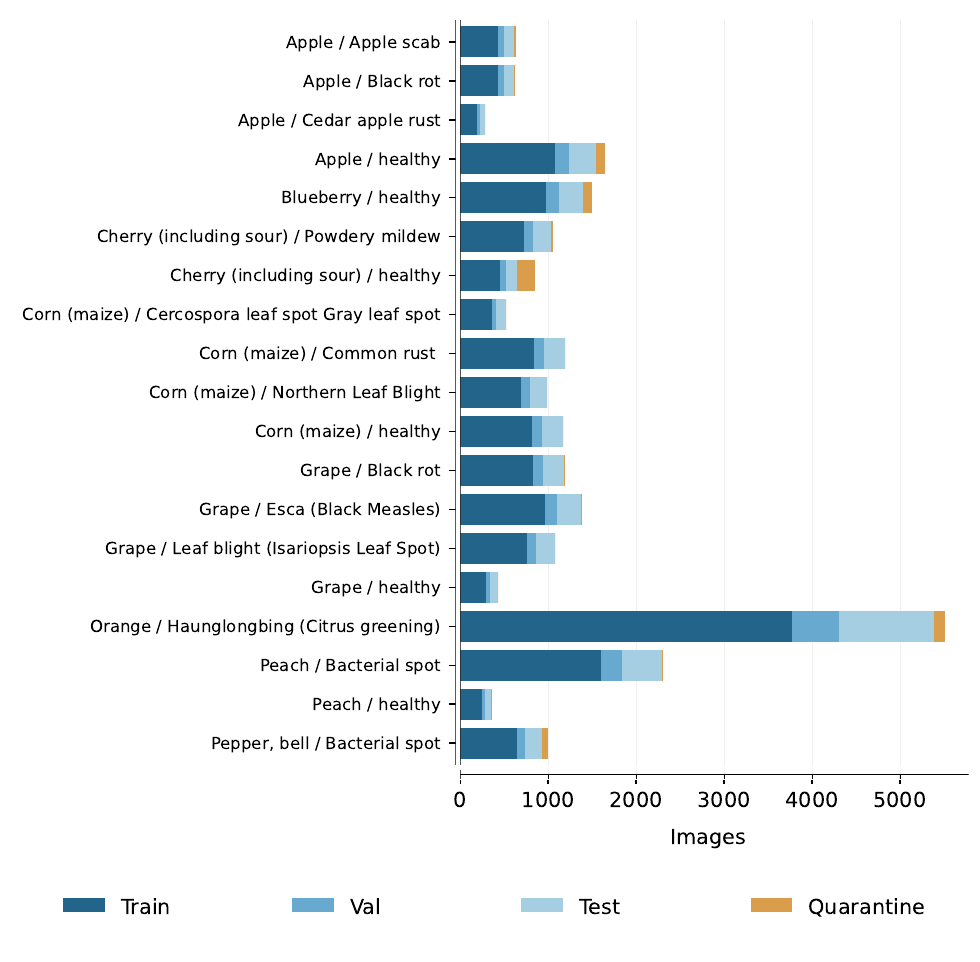}
\caption{\rev{Complete PV class distribution and conservative quarantine. The split is frozen before training; leaf suggestions and hash grouping do not prove biological/site independence. The \revsecond{two panels include} all 38 classes, rather than selected favourable categories. \revsecond{Panel (a) shows the first 19 classes; panel (b) shows the remaining 19, with the same linear horizontal scale. The dataset contains 54,305 valid images and 49,098 eligible images. Quarantine includes cross-label pHash similarity components; this screening has false positives.}}}
\label{fig:s-pv-splits}
\end{figure}

\begin{figure}[!ht]
\ContinuedFloat
\centering
\includegraphics[width=\linewidth,height=0.7\textheight,keepaspectratio]{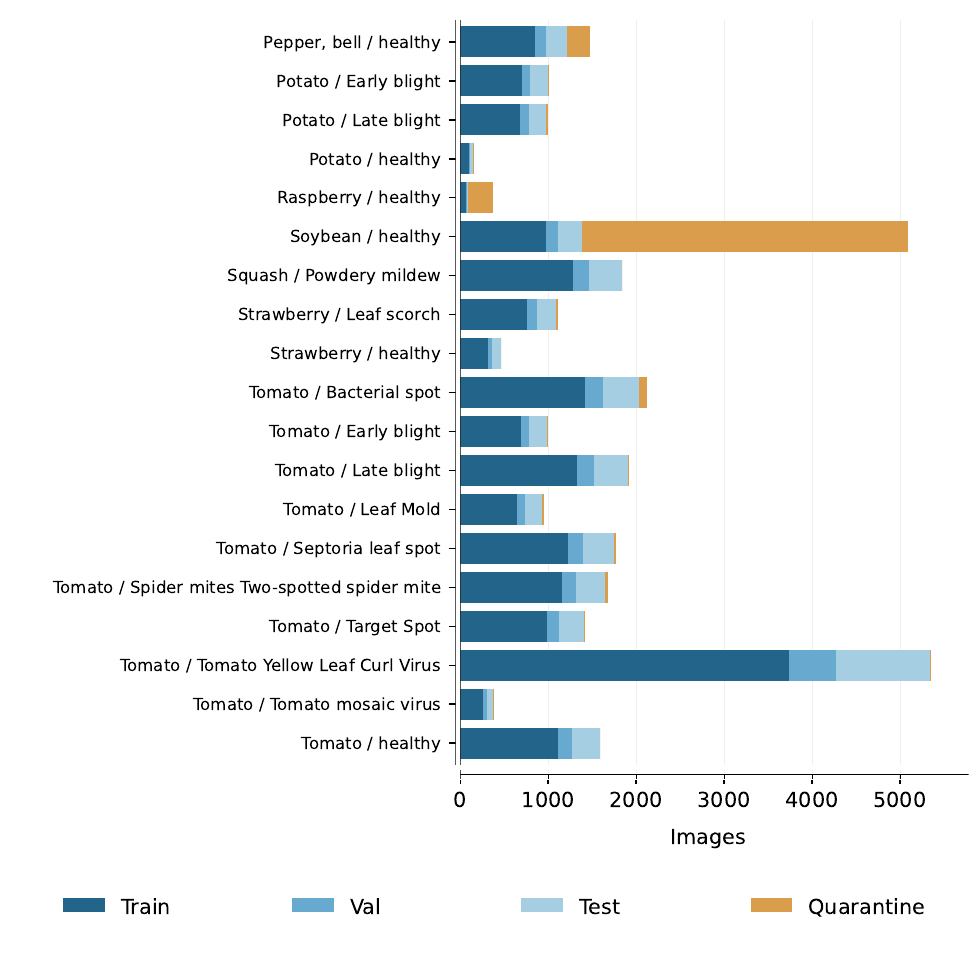}
\caption{\revsecond{Continued: (b) the remaining 19 PV classes, using the same horizontal scale and split legend as panel (a).}}
\end{figure}

\clearpage
\FloatBarrier
\section{\rev{Training, full classification results and uncertainty}}
\label{sec:s-results}

\subsection{\rev{Completed run census and validation selection}}
The historical campaign completed 109 trainings: 40 \revsecond{VLDA retrainings}
(two splits, four conditions, five seeds), 40 image-only zone probes
(two splits, four encoders, five seeds), 20 PV disease probes and nine full-network
PV fine-tunes. The additional grouped-split campaign contributes fifteen runs (five matched full controls, five sensor-only and five language-only), bringing the combined total to 124 training executions. Historical/current full seeds are not pooled. Fixed CLIP zero-shot introduces no new training. The 230
classification exports include clean, corruption, compatible external and original
zone-task evaluations, not 230 independently trained models. Twelve original
checkpoint inference conditions and the two separate OOD diagnostic cohorts
have their own output registries and are not added as new biological samples.

The official pretrained comparison models are ResNet-50 \citep{he2016resnet},
EfficientNet-B0 \citep{tan2019efficientnet}, ViT-B/16
\citep{dosovitskiy2021vit} and CLIP ViT-B/32 \citep{radford2021clip}.
Frozen encoders run their official deterministic preprocessing in BF16, storing
FP32 embeddings. Each linear probe standardizes embeddings using training means
and standard deviations, lower-bounded at $10^{-6}$, and uses FP32 AdamW,
learning rate $10^{-3}$, weight decay $10^{-4}$, batch 512, maximum 100 epochs,
patience 10 and unweighted cross-entropy. The earliest validation-cross-entropy
minimum, with improvement tolerance $10^{-6}$, determines the immutable checkpoint.
No training augmentation is used for these cached representations.
Five seeds are 17, 29, 43, 71 and 101. CLIP zero-shot uses one deterministic
class-name prompt per class and its original logit scale, without prompt
ensembling or retuning after test inspection.

Full-network runs update every parameter for each of the three non-CLIP
architectures, using seeds 17, 29 and 43. They use AdamW backbone/head rates
$10^{-5}/10^{-4}$, weight decay $10^{-4}$, batch 64, maximum ten epochs,
patience three, fixed learning rates, BF16, and validation-cross-entropy selection.
Training-only random-resized $224\times224$ crops use scale $[0.7,1]$ and
horizontal flips with probability 0.5; official evaluation transforms remain
unchanged. Every run completed ten epochs. On-the-fly augmented presentations
total 3,092,400 across nine runs, rather than that many independent images;
no augmented image files were saved. Training plus selection budgets, updated
capacity and augmentation differ from frozen probes, so their contrast is not
an isolated causal ablation of backbone unfreezing.

Genuine VLDA branch removals retain the original 514,065 stored trainable
parameters and fusion width, with removed features zeroed throughout training
and evaluation. Their common objective, AdamW $10^{-4}$, weight decay $10^{-4}$,
batch 16, maximum 100 epochs and patience 10 are matched. Frozen CLS embeddings
are cached in evaluation mode. This disables the frozen BERT dropout enabled by
the original trainer's recursive \texttt{model.train()}, so these runs are
explicitly \revsecond{retrainings with cached frozen-backbone features} rather than an exact rerun of
the original stochastic training path. Inactive zeroed branches remain in the
stored parameter count.

\begingroup\color{blue}\footnotesize
\setlength{\tabcolsep}{3pt}
\renewcommand{\arraystretch}{1.15}
\begin{longtable}{@{}p{0.26\textwidth}rrrrr@{}}
\caption{Measured resources on the local RTX 5090 Laptop GPU. Frozen extraction is a shared one-time 49,098-image cost; training seconds are mean per seed for heads or full networks, respectively. Peak memory is extraction peak for frozen rows and maximum training allocation for full rows, so those columns are different operations. Device batch-one latency excludes decoding, transforms and host transfer; it is not system control latency.}\label{tab:s-resources}\\
\toprule
Model regime & Trainable params & Extract s & Train s & Peak MiB & Latency ms \\
\midrule\endfirsthead
\multicolumn{6}{l}{\textit{Continued from previous page}}\\
\toprule
Model regime & Trainable params & Extract s & Train s & Peak MiB & Latency ms \\
\midrule\endhead
\midrule\multicolumn{6}{r}{\textit{Continued on next page}}\\\endfoot
\bottomrule\endlastfoot
ResNet-50 frozen & 77862 & 25.59 & 1.60 & 471.2 & 2.800 \\
EffNet-B0 frozen & 48678 & 13.72 & 1.52 & 437.9 & 3.863 \\
ViT-B/16 frozen & 29222 & 49.96 & 2.26 & 799.5 & 5.130 \\
CLIP probe frozen & 19494 & 44.49 & 3.52 & 773.4 & 9.763 \\
ResNet-50 full & 23585894 & -- & 427.95 & 3322.5 & 2.815 \\
EffNet-B0 full & 4056226 & -- & 272.86 & 3268.3 & 3.688 \\
ViT-B/16 full & 85827878 & -- & 1141.77 & 6173.1 & 4.546 \\
\end{longtable}
\endgroup


\providecommand{\ValidationRevisionColor}{\color{blue}}
\begingroup\ValidationRevisionColor
\begin{table*}[t]
\centering\scriptsize
\caption{One recovered checkpoint on the historical 43-row test: descriptive reproduction, not independent field validation.}
\label{tab:validation-v1-original-replay}
\setlength{\tabcolsep}{3pt}
\renewcommand{\arraystretch}{1.15}
\begin{tabularx}{\textwidth}{X>{\centering\arraybackslash}X>{\centering\arraybackslash}X>{\centering\arraybackslash}X}
\toprule
Metric & Point & Observed-group interval & Valid draws \\
\midrule
Zone accuracy (41/43), \% & 95.349 & \([87.805,100.000]\) & 1000/1000 \\
Zone macro-F1, fixed four classes & 0.9311 & \([0.7185,1.0000]\) & 1000/1000 \\
Zone balanced accuracy & 0.9494 & \([0.8500,1.0000]\) & 971/1000 \\
Pesticide generated-label agreement (43/43) & 1.0000 & \([1,1]\) & 1000/1000 \\
\bottomrule
\end{tabularx}
\par\smallskip
\begin{minipage}{\textwidth}\footnotesize Intervals are approximate 95\% percentile bounds from 1,000 whole-group draws over 41 screened components, conditional on one fixed model. Balanced-accuracy bounds condition on retaining all four classes (971/1,000 draws); fixed-four-class F1 assigns zero to absent sampled classes. The degenerate pesticide interval reflects zero observed generated-label errors and does not bound unseen intervention risk. Saved-split contamination and one-episode dependence remain; these intervals are not field guarantees.
\end{minipage}
\end{table*}
\endgroup


\begin{table*}[t]
\centering\scriptsize
\caption{VLDA cached-feature ablations on the frozen grouped split: five seeds per condition.}
\label{tab:validation-v1-ablations}
\resizebox{\textwidth}{!}{%
\begin{tabular}{lrrrrr}
\toprule
Condition & Zone accuracy \% & Four-class macro-F1 & Temperature MAE $^\circ$C & RH MAE pp & Soil MAE pp \\
\midrule
Full & \(97.727\pm0.000\) & \(0.9762\pm0.0000\) & \(0.249\pm0.011\) & \(2.030\pm0.061\) & \(0.997\pm0.034\) \\
Without language & \(97.727\pm0.000\) & \(0.9762\pm0.0000\) & \(0.283\pm0.020\) & \(2.172\pm0.104\) & \(1.158\pm0.036\) \\
Without sensor/current-zone & \(78.636\pm4.129\) & \(0.6685\pm0.0751\) & \(0.329\pm0.007\) & \(2.369\pm0.040\) & \(1.188\pm0.037\) \\
Without image & \(97.727\pm0.000\) & \(0.9762\pm0.0000\) & \(0.168\pm0.030\) & \(1.302\pm0.257\) & \(0.664\pm0.074\) \\
Sensor/current-zone only & \(97.727\pm0.000\) & \(0.9762\pm0.0000\) & \(0.222\pm0.021\) & \(1.657\pm0.262\) & \(0.911\pm0.115\) \\
Language only & \(68.636\pm2.963\) & \(0.3988\pm0.0673\) & \(0.314\pm0.017\) & \(2.375\pm0.034\) & \(1.123\pm0.054\) \\
\bottomrule
\end{tabular}}
\par\smallskip
\begin{minipage}{\textwidth}\footnotesize Values are means $\pm$ sample SD across seeds 17, 29, 43, 71 and 101 on the unchanged 197/42/44 grouped, single-episode split. Zone macro-F1 explicitly includes all four classes, with undefined class contributions set to zero. RH and soil errors are percentage points. Current full controls reproduce all historical full prediction arrays, trainable checkpoint tensors, normalization tensors, training histories and comparable metrics exactly. The three earlier single-branch removals are retained from the verified historical campaign; full and the two new single-branch conditions use five current runs each. All branches remain in the topology, are evaluated before masking, and removed projected vectors are zeroed during training, validation and testing. Sensor/current-zone contains four one-hot coordinates and 15 continuous values. Cached DINOv2/BERT CLS features were extracted in evaluation mode; this differs from earliest online frozen-BERT dropout training. Seed SD is not a sampling confidence interval. All intervention metrics describe simulation-label agreement, not field treatment safety.\end{minipage}
\end{table*}

\providecommand{\ValidationRevisionColor}{\color{blue}}
\begingroup\ValidationRevisionColor
\begin{table*}[t]
\centering\scriptsize
\caption{Disease-image classification on the frozen PlantVillage v2 test (9,828 images, 3,864 screened groups, 38 output classes).}
\label{tab:validation-v1-pv}
\setlength{\tabcolsep}{3pt}
\renewcommand{\arraystretch}{1.15}
\begin{tabularx}{\textwidth}{X>{\centering\arraybackslash}X>{\centering\arraybackslash}X>{\centering\arraybackslash}X>{\centering\arraybackslash}X}
\toprule
Cohort / model & Runs & Accuracy \% \(\pm\) SD pp & Accuracy interval \% & Macro-F1 \(\pm\) SD \\
\midrule
Frozen probe / ResNet-50 & 5 & \(96.803\pm0.059\) & \([96.351,97.224]\) & \(0.9623\pm0.0010\) \\
Frozen probe / EfficientNet-B0 & 5 & \(97.271\pm0.064\) & \([96.885,97.673]\) & \(0.9682\pm0.0009\) \\
Frozen probe / ViT-B/16 & 5 & \(98.209\pm0.075\) & \([97.892,98.521]\) & \(0.9773\pm0.0007\) \\
Frozen probe / CLIP image probe & 5 & \(97.900\pm0.042\) & \([97.520,98.231]\) & \(0.9746\pm0.0007\) \\
CLIP image--text zero-shot & 1 & \(12.566\) & \([11.516,13.564]\) & \(0.1195\) \\
Full network / ResNet-50 & 3 & \(98.911\pm0.097\) & \([98.610,99.194]\) & \(0.9864\pm0.0012\) \\
Full network / EfficientNet-B0 & 3 & \(98.589\pm0.117\) & \([98.279,98.882]\) & \(0.9821\pm0.0012\) \\
Full network / ViT-B/16 & 3 & \(99.342\pm0.149\) & \([99.110,99.555]\) & \(0.9906\pm0.0019\) \\
\bottomrule
\end{tabularx}
\par\smallskip
\begin{minipage}{\textwidth}\footnotesize Intervals are approximate 95\% crossed seed/group percentile bounds (1,000 draws), conditional on observed seeds, groups and the frozen subset. Frozen heads use five seeds; full-network runs use three seeds. Fixed CLIP zero-shot has no new training or seed SD. Full-network training changes optimizer, augmentation, capacity and epoch budget and is not an isolated unfreezing ablation. These disease-label results are not the recovered VLDA zone/setpoint/action task. All 230 per-run, 56 aggregate and eight paired balanced-accuracy intervals are finite-only conditional on required class support; aggregate valid counts are 867--981/1,000, including 872/1,000 for each full-network clean family. Fixed-label F1 assigns zero to absent sampled classes; its wider lower bootstrap tails are retained.
\end{minipage}
\end{table*}
\endgroup


\subsection{\rev{Metrics and distinct uncertainty quantities}}
Accuracy uses each declared sample denominator. Macro-F1 averages the fixed
declared class universe with zero contribution at a zero denominator; balanced
accuracy uses the fixed originally supported true classes. ROC/AP use applicable
one-versus-rest class support; unsupported AUCs are not filled with invented zero
values. Confusion matrices and class precision, recall, specificity and support
are exported. Calibration includes multiclass Brier sum, negative log-likelihood
and 15-bin top-label ECE. Raw softmax confidence is not assumed calibrated.
Offline risk--coverage curves retain confidence ties together; their test-score
sweep is descriptive and is not an implemented treatment abstention policy.

Optimization-seed sample SD measures dispersion among fitted runs. Per-run
source-group percentile intervals instead resample complete observed groups;
crossed aggregate intervals additionally resample the observed fitted seeds.
They condition on these models, groups and chosen subsets and exclude new-site,
calibration, pretraining and broader model-selection uncertainty. The fixed
zero-shot model has no optimization-seed SD. Paired contrasts retain shared
within-domain group draws, while external domains are resampled separately.
Intervals use 1,000 draws with seed 20261015 unless a preserved historical
analysis is explicitly identified otherwise.

Balanced accuracy is undefined in a replicate that omits a required true class.
Finite-only intervals condition on retaining that support and report valid versus
requested counts; they do not recode the undefined value to zero. Across the
56 final aggregate families, balanced-accuracy valid counts range from
867 to 981 of 1,000; the three clean full-network families each have 872.
Fixed-class F1 follows a different convention: an absent sampled class can lower
the macro average even when all observed predictions are correct.

For the historical checkpoint, the 43 saved test rows map to 41 screened groups:
source IDs 1, 6 and 7 form one component and the remaining 40 are singletons.
The full-model zone accuracy is $41/43$, with group-percentile bounds
$[0.878049,1]$; macro-F1 is 0.931061 with $[0.718478,1]$.
Balanced accuracy is 0.949405 with conditional bounds $[0.85,1]$ from
971/1,000 class-complete draws. All 12 inference conditions retain the same
source alignment and grouping. Generated pesticide-label agreement is one with
degenerate empirical bounds $[1,1]$, which do not estimate unseen treatment risk.
Historical independent-row Wilson/row-bootstrap files remain descriptive;
group resampling cannot repair the saved split's leakage or one-episode origin.

Regression errors are reported separately in degrees Celsius and humidity/soil
percentage points. Mixed-unit MAE is retained only to reproduce the historical
file. Full, no-language and no-image grouped models each obtain $43/44$ zone
accuracy in all five seeds; removing images \revsecond{reduces} all three recorded
physical-unit MAEs. Removing the combined sensor/current-zone branch lowers
accuracy, but cannot isolate an abiotic causal effect because current-zone
one-hot coordinates are removed with the 15 continuous values. These are retained
negative findings against a claim of demonstrated image/language synergy.

\rev{Table S8 catalogs the 56 historical campaign families. The additional sensor-only and language-only grouped conditions and matched current full controls are reported separately in Table S6; they are not pooled into those historical aggregates.}

\begingroup\color{blue}\footnotesize
\setlength{\tabcolsep}{3pt}
\renewcommand{\arraystretch}{1.15}
\begin{longtable}{@{}p{0.35\textwidth}rrrrr@{}}
\caption{All 56 classification families. Each row retains its own task, split and sample count; these rows are not a unified VLDA/disease leaderboard. Values are means with sample SD across completed seeds; accuracy is percent and SD is percentage points. Other metrics are on $[0,1]$. Fixed zero-shot has no seed SD. The complete sampling intervals and valid counts are supplied separately in CSV.}\label{tab:s-all-families}\\
\toprule
Cohort / model / condition & Runs / $N$ & Acc. & Macro-F1 & Bal. acc. & ECE \\
\midrule\endfirsthead
\multicolumn{6}{l}{\textit{Continued from previous page}}\\
\toprule
Cohort / model / condition & Runs / $N$ & Acc. & Macro-F1 & Bal. acc. & ECE \\
\midrule\endhead
\midrule\multicolumn{6}{r}{\textit{Continued on next page}}\\\endfoot
\bottomrule\endlastfoot
External frozen / CLIP probe / external clean & 5 / 238 & 24.87$\pm$0.69 & 0.087$\pm$0.003 & 0.250$\pm$0.005 & 0.532$\pm$0.014 \\
External frozen / CLIP zero-shot / external clean & 1 / 238 & 30.25 & 0.100 & 0.240 & 0.091 \\
External frozen / EffNet-B0 / external clean & 5 / 238 & 24.62$\pm$0.70 & 0.091$\pm$0.004 & 0.262$\pm$0.016 & 0.416$\pm$0.013 \\
External frozen / ResNet-50 / external clean & 5 / 238 & 20.17$\pm$1.19 & 0.070$\pm$0.003 & 0.203$\pm$0.010 & 0.567$\pm$0.016 \\
External frozen / ViT-B/16 / external clean & 5 / 238 & 21.68$\pm$0.97 & 0.064$\pm$0.005 & 0.193$\pm$0.009 & 0.500$\pm$0.007 \\
External full / EffNet-B0 / external clean & 3 / 238 & 25.77$\pm$0.87 & 0.085$\pm$0.005 & 0.252$\pm$0.013 & 0.355$\pm$0.007 \\
External full / ResNet-50 / external clean & 3 / 238 & 25.07$\pm$0.49 & 0.082$\pm$0.007 & 0.247$\pm$0.015 & 0.414$\pm$0.017 \\
External full / ViT-B/16 / external clean & 3 / 238 & 27.17$\pm$1.70 & 0.091$\pm$0.007 & 0.261$\pm$0.023 & 0.280$\pm$0.009 \\
PV corruption full / EffNet-B0 / blur radius2 & 3 / 9828 & 36.91$\pm$1.97 & 0.271$\pm$0.011 & 0.265$\pm$0.012 & 0.312$\pm$0.026 \\
PV corruption full / ResNet-50 / blur radius2 & 3 / 9828 & 59.14$\pm$1.00 & 0.528$\pm$0.017 & 0.512$\pm$0.012 & 0.231$\pm$0.003 \\
PV corruption full / ViT-B/16 / blur radius2 & 3 / 9828 & 85.54$\pm$2.62 & 0.855$\pm$0.027 & 0.846$\pm$0.018 & 0.049$\pm$0.014 \\
PV corruption full / EffNet-B0 / illumination 0.5 & 3 / 9828 & 91.12$\pm$0.31 & 0.893$\pm$0.003 & 0.895$\pm$0.002 & 0.015$\pm$0.001 \\
PV corruption full / ResNet-50 / illumination 0.5 & 3 / 9828 & 93.07$\pm$1.34 & 0.911$\pm$0.017 & 0.900$\pm$0.019 & 0.012$\pm$0.002 \\
PV corruption full / ViT-B/16 / illumination 0.5 & 3 / 9828 & 97.81$\pm$1.07 & 0.974$\pm$0.013 & 0.977$\pm$0.010 & 0.005$\pm$0.004 \\
PV corruption full / EffNet-B0 / noise std0.05 & 3 / 9828 & 67.11$\pm$2.41 & 0.594$\pm$0.031 & 0.592$\pm$0.033 & 0.060$\pm$0.017 \\
PV corruption full / ResNet-50 / noise std0.05 & 3 / 9828 & 95.24$\pm$1.00 & 0.944$\pm$0.013 & 0.941$\pm$0.011 & 0.006$\pm$0.002 \\
PV corruption full / ViT-B/16 / noise std0.05 & 3 / 9828 & 98.63$\pm$0.32 & 0.982$\pm$0.004 & 0.983$\pm$0.004 & 0.004$\pm$0.001 \\
Original saved / VLDA full / clean & 5 / 43 & 96.74$\pm$2.08 & 0.956$\pm$0.026 & 0.960$\pm$0.023 & 0.055$\pm$0.024 \\
Original saved / VLDA no sensor branch / clean & 5 / 43 & 68.84$\pm$2.08 & 0.521$\pm$0.041 & 0.540$\pm$0.044 & 0.162$\pm$0.037 \\
Original saved / VLDA no image / clean & 5 / 43 & 100.00$\pm$0.00 & 1.000$\pm$0.000 & 1.000$\pm$0.000 & 0.018$\pm$0.005 \\
Original saved / VLDA no language / clean & 5 / 43 & 92.09$\pm$3.89 & 0.914$\pm$0.043 & 0.930$\pm$0.039 & 0.067$\pm$0.026 \\
Original saved / CLIP probe / clean & 5 / 43 & 67.91$\pm$1.95 & 0.478$\pm$0.045 & 0.505$\pm$0.043 & 0.189$\pm$0.033 \\
Original saved / EffNet-B0 / clean & 5 / 43 & 64.19$\pm$4.22 & 0.417$\pm$0.093 & 0.448$\pm$0.097 & 0.225$\pm$0.050 \\
Original saved / ResNet-50 / clean & 5 / 43 & 57.67$\pm$4.16 & 0.420$\pm$0.033 & 0.458$\pm$0.046 & 0.177$\pm$0.031 \\
Original saved / ViT-B/16 / clean & 5 / 43 & 67.91$\pm$2.55 & 0.573$\pm$0.035 & 0.615$\pm$0.041 & 0.143$\pm$0.038 \\
PV full / EffNet-B0 / clean & 3 / 9828 & 98.59$\pm$0.12 & 0.982$\pm$0.001 & 0.982$\pm$0.002 & 0.004$\pm$0.000 \\
PV full / ResNet-50 / clean & 3 / 9828 & 98.91$\pm$0.10 & 0.986$\pm$0.001 & 0.987$\pm$0.001 & 0.002$\pm$0.001 \\
PV full / ViT-B/16 / clean & 3 / 9828 & 99.34$\pm$0.15 & 0.991$\pm$0.002 & 0.992$\pm$0.001 & 0.002$\pm$0.001 \\
PV frozen / CLIP probe / clean & 5 / 9828 & 97.90$\pm$0.04 & 0.975$\pm$0.001 & 0.976$\pm$0.000 & 0.003$\pm$0.001 \\
PV frozen / CLIP zero-shot / clean & 1 / 9828 & 12.57 & 0.120 & 0.153 & 0.139 \\
PV frozen / EffNet-B0 / clean & 5 / 9828 & 97.27$\pm$0.06 & 0.968$\pm$0.001 & 0.969$\pm$0.001 & 0.003$\pm$0.000 \\
PV frozen / ResNet-50 / clean & 5 / 9828 & 96.80$\pm$0.06 & 0.962$\pm$0.001 & 0.961$\pm$0.001 & 0.005$\pm$0.001 \\
PV frozen / ViT-B/16 / clean & 5 / 9828 & 98.21$\pm$0.08 & 0.977$\pm$0.001 & 0.978$\pm$0.001 & 0.004$\pm$0.001 \\
PV corruption frozen / CLIP probe / blur radius2 & 5 / 9828 & 66.86$\pm$0.40 & 0.603$\pm$0.005 & 0.582$\pm$0.005 & 0.196$\pm$0.006 \\
PV corruption frozen / CLIP zero-shot / blur radius2 & 1 / 9828 & 10.03 & 0.080 & 0.138 & 0.152 \\
PV corruption frozen / EffNet-B0 / blur radius2 & 5 / 9828 & 72.71$\pm$0.63 & 0.716$\pm$0.007 & 0.692$\pm$0.007 & 0.125$\pm$0.007 \\
PV corruption frozen / ResNet-50 / blur radius2 & 5 / 9828 & 76.52$\pm$0.33 & 0.741$\pm$0.004 & 0.721$\pm$0.002 & 0.124$\pm$0.004 \\
PV corruption frozen / ViT-B/16 / blur radius2 & 5 / 9828 & 70.63$\pm$0.44 & 0.722$\pm$0.003 & 0.688$\pm$0.003 & 0.177$\pm$0.006 \\
PV corruption frozen / CLIP probe / illumination 0.5 & 5 / 9828 & 86.90$\pm$0.24 & 0.844$\pm$0.005 & 0.836$\pm$0.004 & 0.058$\pm$0.004 \\
PV corruption frozen / CLIP zero-shot / illumination 0.5 & 1 / 9828 & 13.05 & 0.099 & 0.142 & 0.093 \\
PV corruption frozen / EffNet-B0 / illumination 0.5 & 5 / 9828 & 93.71$\pm$0.16 & 0.933$\pm$0.002 & 0.928$\pm$0.001 & 0.014$\pm$0.002 \\
PV corruption frozen / ResNet-50 / illumination 0.5 & 5 / 9828 & 95.14$\pm$0.08 & 0.940$\pm$0.001 & 0.944$\pm$0.001 & 0.010$\pm$0.002 \\
PV corruption frozen / ViT-B/16 / illumination 0.5 & 5 / 9828 & 97.29$\pm$0.08 & 0.967$\pm$0.001 & 0.969$\pm$0.001 & 0.004$\pm$0.001 \\
PV corruption frozen / CLIP probe / noise std0.05 & 5 / 9828 & 82.79$\pm$0.26 & 0.813$\pm$0.004 & 0.812$\pm$0.003 & 0.097$\pm$0.003 \\
PV corruption frozen / CLIP zero-shot / noise std0.05 & 1 / 9828 & 12.19 & 0.088 & 0.119 & 0.119 \\
PV corruption frozen / EffNet-B0 / noise std0.05 & 5 / 9828 & 94.03$\pm$0.09 & 0.929$\pm$0.001 & 0.928$\pm$0.001 & 0.012$\pm$0.002 \\
PV corruption frozen / ResNet-50 / noise std0.05 & 5 / 9828 & 79.08$\pm$2.34 & 0.773$\pm$0.014 & 0.788$\pm$0.021 & 0.122$\pm$0.022 \\
PV corruption frozen / ViT-B/16 / noise std0.05 & 5 / 9828 & 94.27$\pm$0.06 & 0.940$\pm$0.001 & 0.940$\pm$0.001 & 0.014$\pm$0.002 \\
Original grouped / VLDA full / clean & 5 / 44 & 97.73$\pm$0.00 & 0.976$\pm$0.000 & 0.964$\pm$0.000 & 0.029$\pm$0.016 \\
Original grouped / VLDA no sensor branch / clean & 5 / 44 & 78.64$\pm$4.13 & 0.669$\pm$0.075 & 0.700$\pm$0.077 & 0.126$\pm$0.023 \\
Original grouped / VLDA no image / clean & 5 / 44 & 97.73$\pm$0.00 & 0.976$\pm$0.000 & 0.964$\pm$0.000 & 0.021$\pm$0.000 \\
Original grouped / VLDA no language / clean & 5 / 44 & 97.73$\pm$0.00 & 0.976$\pm$0.000 & 0.964$\pm$0.000 & 0.029$\pm$0.012 \\
Original grouped / CLIP probe / clean & 5 / 44 & 73.64$\pm$2.59 & 0.554$\pm$0.073 & 0.569$\pm$0.078 & 0.147$\pm$0.008 \\
Original grouped / EffNet-B0 / clean & 5 / 44 & 64.55$\pm$6.55 & 0.444$\pm$0.105 & 0.457$\pm$0.119 & 0.158$\pm$0.054 \\
Original grouped / ResNet-50 / clean & 5 / 44 & 65.00$\pm$4.13 & 0.447$\pm$0.075 & 0.456$\pm$0.081 & 0.158$\pm$0.046 \\
Original grouped / ViT-B/16 / clean & 5 / 44 & 70.00$\pm$6.31 & 0.550$\pm$0.082 & 0.571$\pm$0.080 & 0.189$\pm$0.060 \\
\end{longtable}
\endgroup


\begingroup\color{blue}\footnotesize
\setlength{\tabcolsep}{3pt}
\renewcommand{\arraystretch}{1.15}
\begin{longtable}{@{}p{0.37\textwidth}rrrrrr@{}}
\caption{Per-class F1 averaged over completed optimization seeds on the 9,828-image PV test. Full-network families use three seeds; frozen ViT uses five; fixed CLIP zero-shot uses one model. Complete precision, recall, specificity, AUC/AP and all family/condition rows are delivered in the per-class CSV.}\label{tab:s-perclass}\\
\toprule
True class & $N$ & Full R50 & Full EB0 & Full ViT & Frozen ViT & CLIP zero \\
\midrule\endfirsthead
\multicolumn{7}{l}{\textit{Continued from previous page}}\\
\toprule
True class & $N$ & Full R50 & Full EB0 & Full ViT & Frozen ViT & CLIP zero \\
\midrule\endhead
\midrule\multicolumn{7}{r}{\textit{Continued on next page}}\\\endfoot
\bottomrule\endlastfoot
Apple: Apple scab & 123 & 0.997 & 0.999 & 0.997 & 0.985 & 0.000 \\
Apple: Black rot & 124 & 1.000 & 1.000 & 0.996 & 0.996 & 0.000 \\
Apple: Cedar apple rust & 56 & 0.994 & 1.000 & 0.994 & 1.000 & 0.000 \\
Apple: healthy & 308 & 0.993 & 0.997 & 0.999 & 0.993 & 0.000 \\
Blueberry: healthy & 279 & 0.997 & 0.999 & 1.000 & 0.998 & 0.000 \\
Cherry (including sour): Powdery mildew & 206 & 0.998 & 0.997 & 1.000 & 0.995 & 0.000 \\
Cherry (including sour): healthy & 131 & 0.984 & 0.986 & 0.982 & 0.984 & 0.000 \\
Corn (maize): Cercospora leaf spot Gray leaf spot & 103 & 0.896 & 0.859 & 0.900 & 0.876 & 0.339 \\
Corn (maize): Common rust  & 238 & 0.994 & 0.997 & 0.998 & 1.000 & 0.722 \\
Corn (maize): Northern Leaf Blight & 197 & 0.944 & 0.929 & 0.936 & 0.932 & 0.093 \\
Corn (maize): healthy & 232 & 1.000 & 0.999 & 0.999 & 0.998 & 0.288 \\
Grape: Black rot & 236 & 0.987 & 0.982 & 0.994 & 0.980 & 0.030 \\
Grape: Esca (Black Measles) & 276 & 0.990 & 0.990 & 0.996 & 0.984 & 0.000 \\
Grape: Leaf blight (Isariopsis Leaf Spot) & 216 & 0.999 & 0.995 & 0.999 & 0.999 & 0.105 \\
Grape: healthy & 85 & 1.000 & 0.996 & 1.000 & 0.995 & 0.437 \\
Orange: Haunglongbing (Citrus greening) & 1079 & 0.999 & 0.999 & 1.000 & 0.999 & 0.000 \\
Peach: Bacterial spot & 458 & 0.999 & 0.997 & 0.999 & 0.997 & 0.103 \\
Peach: healthy & 71 & 0.988 & 0.979 & 0.993 & 0.986 & 0.014 \\
Pepper, bell: Bacterial spot & 187 & 0.973 & 0.980 & 0.991 & 0.957 & 0.000 \\
Pepper, bell: healthy & 243 & 0.990 & 0.993 & 0.994 & 0.987 & 0.000 \\
Potato: Early blight & 200 & 0.996 & 0.990 & 0.998 & 0.978 & 0.036 \\
Potato: Late blight & 192 & 0.990 & 0.978 & 0.992 & 0.968 & 0.226 \\
Potato: healthy & 28 & 0.994 & 0.964 & 0.983 & 0.900 & 0.000 \\
Raspberry: healthy & 19 & 1.000 & 1.000 & 1.000 & 1.000 & 0.647 \\
Soybean: healthy & 278 & 0.993 & 0.993 & 0.997 & 0.996 & 0.145 \\
Squash: Powdery mildew & 367 & 1.000 & 1.000 & 1.000 & 1.000 & 0.667 \\
Strawberry: Leaf scorch & 218 & 0.996 & 0.994 & 0.999 & 0.995 & 0.050 \\
Strawberry: healthy & 92 & 1.000 & 1.000 & 1.000 & 1.000 & 0.000 \\
Tomato: Bacterial spot & 407 & 0.987 & 0.974 & 0.993 & 0.966 & 0.000 \\
Tomato: Early blight & 196 & 0.930 & 0.925 & 0.972 & 0.917 & 0.087 \\
Tomato: Late blight & 380 & 0.976 & 0.968 & 0.986 & 0.961 & 0.196 \\
Tomato: Leaf Mold & 184 & 0.988 & 0.974 & 0.986 & 0.980 & 0.085 \\
Tomato: Septoria leaf spot & 350 & 0.976 & 0.962 & 0.995 & 0.948 & 0.078 \\
Tomato: Spider mites Two-spotted spider mite & 328 & 0.983 & 0.981 & 0.993 & 0.968 & 0.000 \\
Tomato: Target Spot & 280 & 0.973 & 0.966 & 0.988 & 0.949 & 0.000 \\
Tomato: Tomato Yellow Leaf Curl Virus & 1069 & 0.998 & 0.996 & 0.998 & 0.994 & 0.174 \\
Tomato: Tomato mosaic virus & 74 & 0.982 & 0.982 & 0.998 & 0.984 & 0.022 \\
Tomato: healthy & 318 & 0.999 & 1.000 & 0.999 & 0.989 & 0.000 \\
\end{longtable}
\endgroup


\SuppFigure{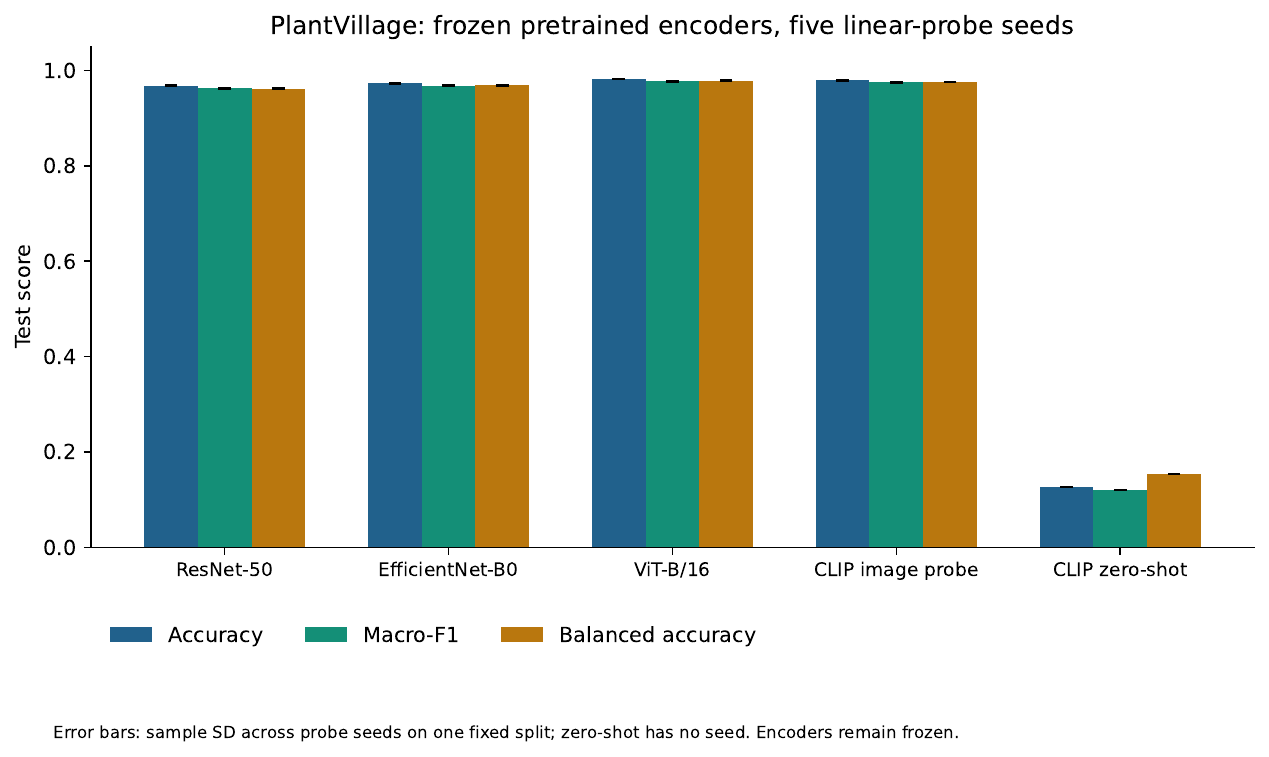}{Complete frozen
representation and fixed CLIP comparison under the same PV split. Seed SD and
conditional sampling intervals are distinct quantities; this disease-image task
is separate from original VLDA zone/setpoint prediction.}{fig:s-frozen-results}
\SuppFigure{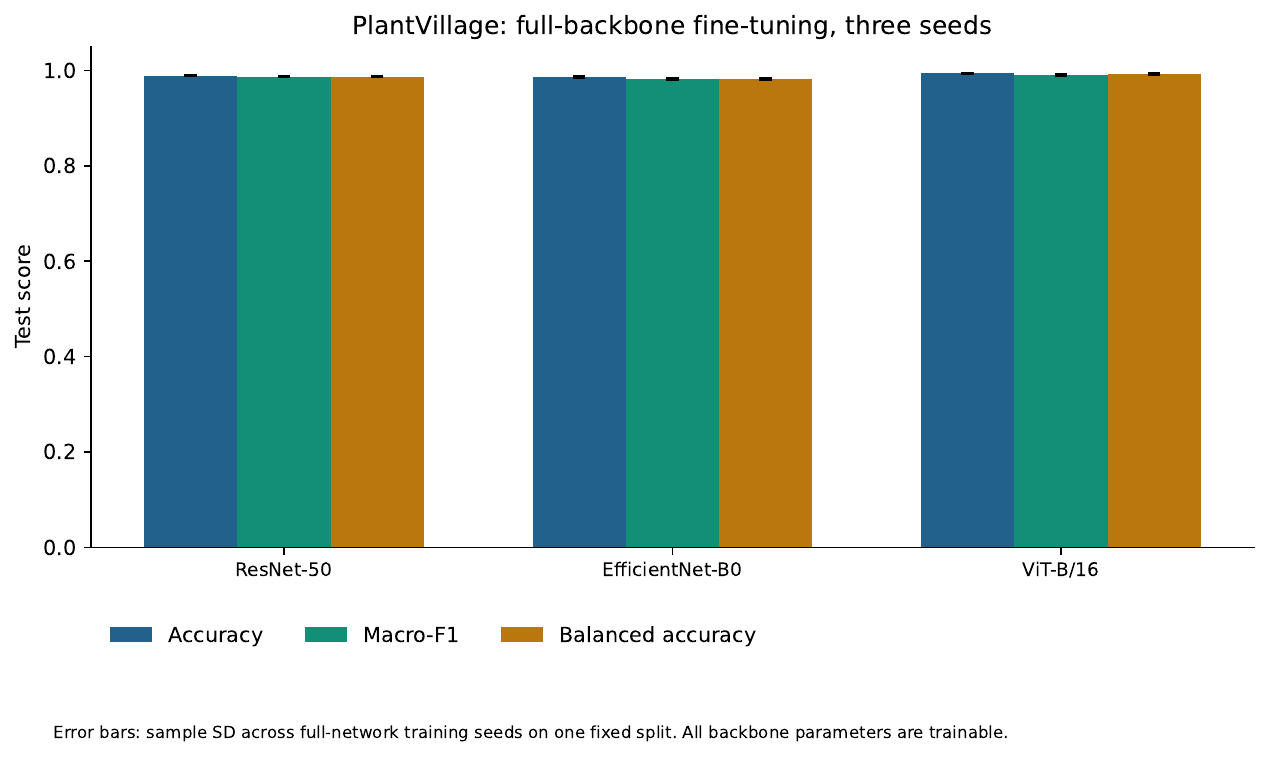}{All nine
full-network trainings, three seeds per architecture. The observed clean mean
performance does not imply compatible external or field accuracy.}{fig:s-full-results}
\SuppFigure{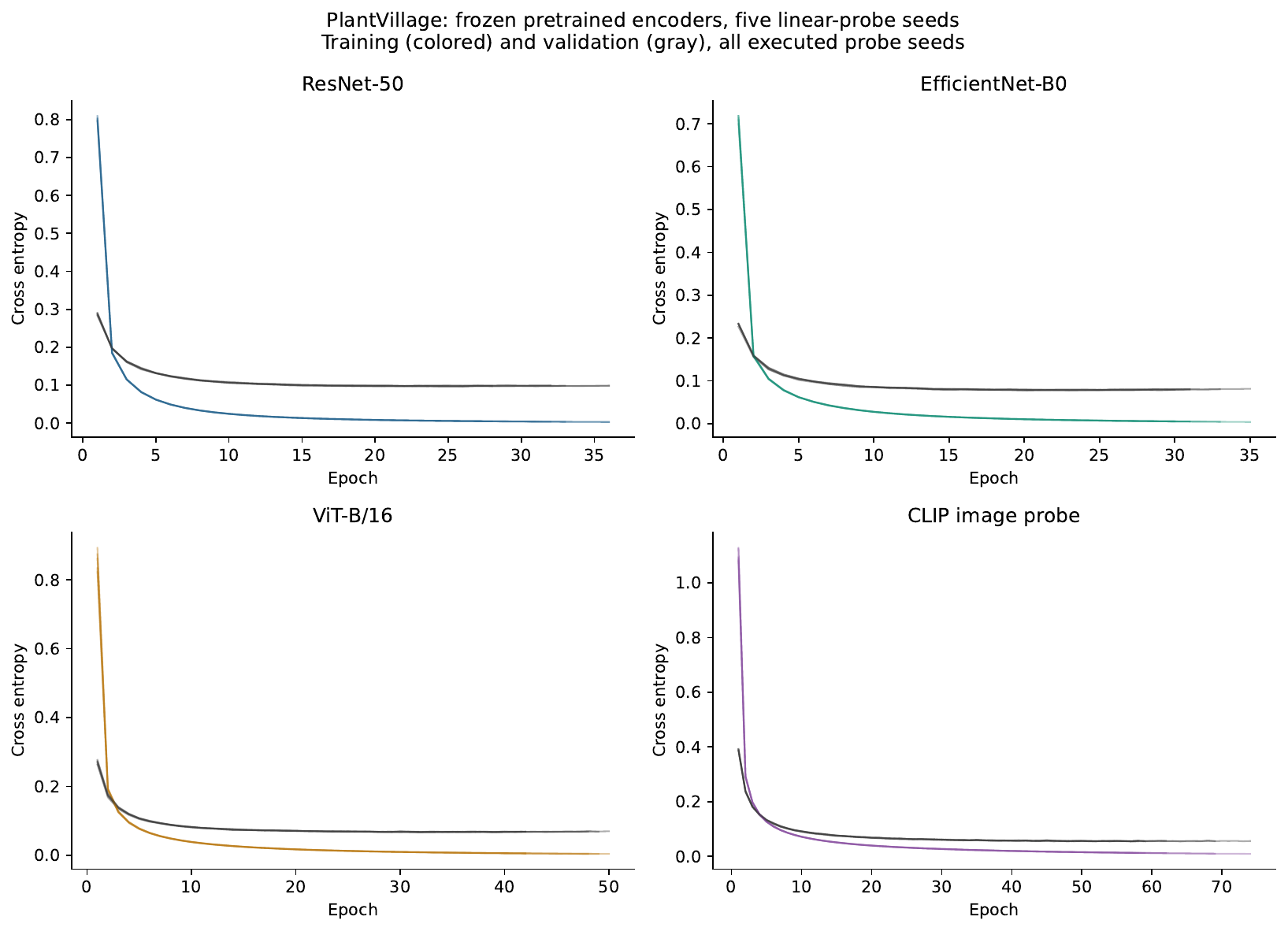}{Completed frozen-head
learning histories. Checkpoints are selected by validation loss, not these test
outcomes. Shared encoder extraction is separate from head-training time.}{fig:s-frozen-learning}
\SuppFigure{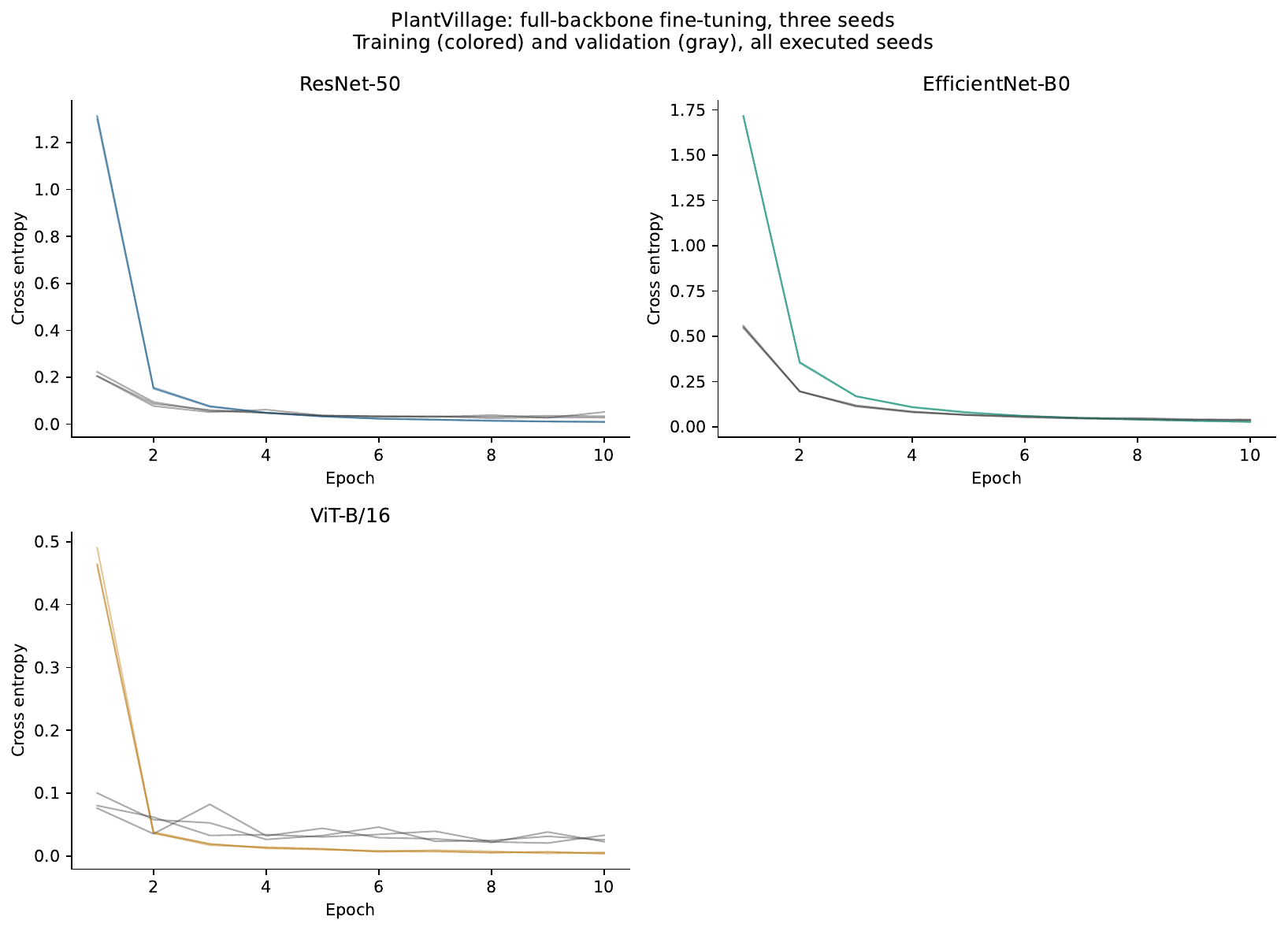}{All full-network
learning histories under the ten-epoch protocol. Epochs and repeated augmented
views are not independent sample counts.}{fig:s-full-learning}
\SuppFigure{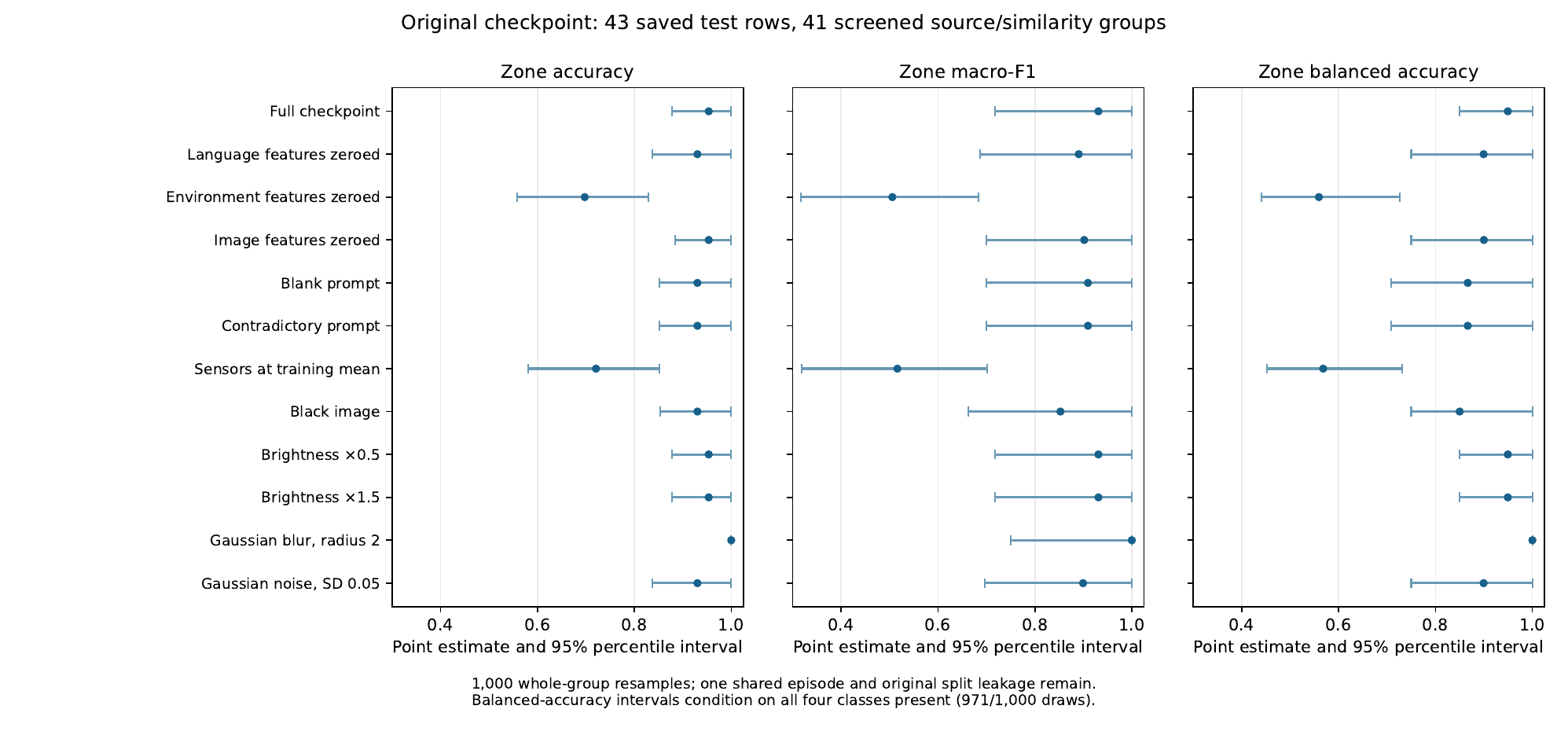}{One
historical checkpoint, 12 inference conditions, 43 saved rows and 41 screened
groups. Balanced-accuracy intervals condition on 971 class-complete draws;
fixed-four-class F1 uses zero contribution for an absent class. Original leakage
and one-episode dependence remain.}{fig:s-original-intervals}

\rev{The added sensor/current-zone-only condition matches the full model at 43/44 correct zones and macro-F1 0.9762 in every seed. Its temperature/RH/soil MAEs are $0.222\pm0.021$~$^\circ$C, $1.657\pm0.262$ and $0.911\pm0.115$ percentage points, lower in mean than the full model. Language-only reaches $68.636\pm2.963$\% zone accuracy and macro-F1 $0.3988\pm0.0673$, with respective reference MAEs $0.314\pm0.017$, $2.375\pm0.034$ and $1.123\pm0.054$. Sensor-only intervention-label agreement is 43/44 with F1 0.9744 in every seed; language-only retains 44/44 agreement. Thus the structured inputs suffice for the observed zone accuracy, while generated intervention labels remain predictable from language alone. These findings characterize this supervision and split, without identifying a unique causal feature pathway or establishing treatment validity.}

\rev{An additional campaign trains five sensor/current-zone-only models, five language-only models and five matched full controls on the unchanged grouped split. All branches execute before post-projection masking, preserving dropout random-number consumption and the 514,065-parameter topology; masks apply during training, validation and test. Sensor-only retains all 19 structured inputs; language-only zeros the complete structured branch. The AdamW, normalization, loss, epoch limit, patience and validation-selection settings are unchanged. The current full controls exactly reproduce historical full predictions, tensors, normalization and histories. The new runs use CPU execution; timings are not compared with prior GPU latency. No historical/current seed pooling is performed.}

\clearpage
\FloatBarrier
\section{\rev{Corruptions, compatible transfer and OOD diagnostics}}
\label{sec:s-robust}

\providecommand{\ValidationRevisionColor}{\color{blue}}
\begingroup\ValidationRevisionColor
\begin{table*}[t]
\centering\scriptsize
\caption{Fixed image-corruption evaluations on the same 9,828 PlantVillage test images; accuracy in percent, mean \(\pm\) optimization-seed sample SD in percentage points.}
\label{tab:validation-v1-corruption}
\setlength{\tabcolsep}{3pt}
\renewcommand{\arraystretch}{1.15}
\begin{tabularx}{\textwidth}{X>{\centering\arraybackslash}X>{\centering\arraybackslash}X>{\centering\arraybackslash}X}
\toprule
Cohort / model & Brightness \(\times0.5\) & Blur radius 2 & Noise \(\sigma=0.05\) \\
\midrule
Frozen probe / ResNet-50 & \(95.14\pm0.08\) & \(76.52\pm0.33\) & \(79.08\pm2.34\) \\
Frozen probe / EfficientNet-B0 & \(93.71\pm0.16\) & \(72.71\pm0.63\) & \(94.03\pm0.09\) \\
Frozen probe / ViT-B/16 & \(97.29\pm0.08\) & \(70.63\pm0.44\) & \(94.27\pm0.06\) \\
Frozen probe / CLIP image probe & \(86.90\pm0.24\) & \(66.86\pm0.40\) & \(82.79\pm0.26\) \\
CLIP image--text zero-shot & \(13.05\) & \(10.03\) & \(12.19\) \\
Full network / ResNet-50 & \(93.07\pm1.34\) & \(59.14\pm1.00\) & \(95.24\pm1.00\) \\
Full network / EfficientNet-B0 & \(91.12\pm0.31\) & \(36.91\pm1.97\) & \(67.11\pm2.41\) \\
Full network / ViT-B/16 & \(97.81\pm1.07\) & \(85.54\pm2.62\) & \(98.63\pm0.32\) \\
\bottomrule
\end{tabularx}
\par\smallskip
\begin{minipage}{\textwidth}\footnotesize Corruptions precede the unchanged official model transforms. There are five seeds per frozen probe, three per full network and one fixed zero-shot model. Negative outcomes, including full EfficientNet-B0 blur accuracy 36.91\%, remain unchanged. An executed clean replay of the first 64 test images for seed 17 of each full-network architecture matches saved probabilities bitwise (maximum absolute difference zero). This checked-subset parity does not prove every corruption path or a causal blur mechanism. No checkpoint/threshold/test-driven correction was made. Regime differences prevent an isolated causal interpretation of unfreezing.
\end{minipage}
\end{table*}
\endgroup


\providecommand{\ValidationRevisionColor}{\color{blue}}
\begingroup\ValidationRevisionColor
\begin{table*}[t]
\centering\scriptsize
\caption{Post hoc shared-true-class transfer comparison: PlantVillage 4,106 images/1,948 groups versus PlantSeg 238 images/238 groups, both with the same 15 true classes.}
\label{tab:validation-v1-transfer}
\setlength{\tabcolsep}{3pt}
\renewcommand{\arraystretch}{1.15}
\begin{tabularx}{\textwidth}{X>{\centering\arraybackslash}X>{\centering\arraybackslash}X>{\centering\arraybackslash}X>{\centering\arraybackslash}X>{\centering\arraybackslash}X}
\toprule
Cohort / model & Runs & PV accuracy \% & External accuracy \(\pm\) SD pp & External F1\(_{15}\) \(\pm\) SD & External--PV gap pp [interval] \\
\midrule
Frozen probe / CLIP image probe & 5 & \(97.428\) & \(24.874\pm0.690\) & \(0.2198\pm0.0075\) & \shortstack{\(-72.554\)\\\([-77.921,-67.295]\)} \\
Frozen probe / EfficientNet-B0 & 5 & \(96.249\) & \(24.622\pm0.703\) & \(0.2305\pm0.0094\) & \shortstack{\(-71.628\)\\\([-76.511,-66.624]\)} \\
Frozen probe / ResNet-50 & 5 & \(95.675\) & \(20.168\pm1.188\) & \(0.1764\pm0.0088\) & \shortstack{\(-75.507\)\\\([-80.608,-70.492]\)} \\
Frozen probe / ViT-B/16 & 5 & \(97.735\) & \(21.681\pm0.967\) & \(0.1611\pm0.0128\) & \shortstack{\(-76.054\)\\\([-81.077,-70.759]\)} \\
Full network / EfficientNet-B0 & 3 & \(98.157\) & \(25.770\pm0.875\) & \(0.2149\pm0.0126\) & \shortstack{\(-72.387\)\\\([-77.560,-66.782]\)} \\
Full network / ResNet-50 & 3 & \(98.717\) & \(25.070\pm0.485\) & \(0.2076\pm0.0172\) & \shortstack{\(-73.647\)\\\([-78.962,-68.307]\)} \\
Full network / ViT-B/16 & 3 & \(99.383\) & \(27.171\pm1.698\) & \(0.2293\pm0.0177\) & \shortstack{\(-72.212\)\\\([-77.662,-66.355]\)} \\
Fixed zero-shot / CLIP image--text zero-shot & 1 & \(14.101\) & \(30.252\) & \(0.2546\) & \shortstack{\(16.151\)\\\([10.150,21.814]\)} \\
\bottomrule
\end{tabularx}
\par\smallskip
\begin{minipage}{\textwidth}\footnotesize All 38 output probabilities and unrestricted argmax predictions remain; there is no 15-class softmax restriction or external fine-tuning. The gap uses approximate 95\% crossed seed/group percentile bounds (1,000 draws) with common within-domain group draws across seeds, independent domain draws and paired seed weights. The derived subset is post hoc; frequencies, organs and acquisition differ. This is a conditional transfer contrast, not a pure causal domain effect or a field-validation estimate. F1\(_{15}\) averages supported true classes; full-38 F1 is a separate metric. SD describes optimization seeds; a fixed model has no seed SD.
\end{minipage}
\end{table*}
\endgroup


\subsection{\rev{Adverse image and context conditions}}
Brightness multiplier 0.5, Gaussian blur radius 2 and Gaussian noise standard
deviation 0.05 precede unchanged official model preprocessing on the same 9,828
PV test images. All requested model/seed outcomes are retained. Full
EfficientNet-B0 blur accuracy of 36.91\% is a measured negative result. An
independent clean replay of the first 64 test images for seed 17 of each
full-network architecture matches saved probabilities bitwise, with maximum
absolute difference zero. That bounded parity check does not prove every
corruption implementation or a causal explanation for blur sensitivity.

The historical checkpoint's separate inference conditions include zeroed image,
language or sensor features; blank and contradictory prompts; sensors at the
training mean; black images; brightness 0.5 and 1.5; blur radius 2; and noise
standard deviation 0.05. Those 12 fixed-checkpoint evaluations are perturbations,
not retrained ablations. Generated pesticide-label agreement stays perfect,
consistent with its severe label shortcuts rather than verified intervention
robustness. Missing-feature representations may themselves lie outside training.

Compatible external evaluation keeps all 38 outputs and unrestricted argmax on
238 PlantSeg images with 15 supported true classes. Supported-15 macro-F1 and
fixed-38 macro-F1 are different metrics; the latter includes unsupported classes
under the declared zero-division convention. The shared-true-class contrast uses
4,106 existing PV test images versus the 238 external images. It retains the
poor transfer outcomes and does not remove frequency, acquisition, organ or
annotation differences. Neither side supplies the original VLDA action/context
supervision. No external prompt tuning, training or restricted-class softmax is
performed.

\subsection{\rev{Frozen threshold, score direction and distinct OOD cohorts}}
For every model, the OOD score is $1-\max_{1\leq c\leq38}p_c$.
The original global PV-validation confidence fifth percentile is frozen using
the lower order statistic; a sample is accepted as ID when confidence is greater
than or equal to that threshold, with ties accepted. No OOD image selects a
threshold. Later condition diagnostics reference those exact prior calibration
files by SHA, without recomputation. Validation coverage near 95\% does not
guarantee 95\% coverage for new ID data.

ROC and AP use OOD as positive. ROC false positives are therefore ID rejections;
operating OOD acceptance is a different quantity: OOD images accepted as ID at
the unchanged validation-derived threshold. It is not test-tuned FPR at 95\% OOD
recall, nor a false-intervention rate. Each diagnostic preserves predictions,
scores, probabilities, calibrated threshold identities and 1,000 separate-domain
whole-source-group bootstrap draws. Per-run intervals exclude calibration
variation; seed SD is reported separately.

\providecommand{\ValidationRevisionColor}{\color{blue}}
\begingroup\ValidationRevisionColor
\begin{table*}[t]
\centering\scriptsize
\caption{Separate novel-host OOD diagnostics: 671 official PlantSeg test images, 19 literal host labels and 666 groups; ID reference is the 9,828-image PlantVillage test.}
\label{tab:validation-v1-novel-host-ood}
\setlength{\tabcolsep}{3pt}
\renewcommand{\arraystretch}{1.15}
\begin{tabularx}{\textwidth}{X>{\centering\arraybackslash}X>{\centering\arraybackslash}X>{\centering\arraybackslash}X>{\centering\arraybackslash}X>{\centering\arraybackslash}X}
\toprule
Cohort / model & Runs & OOD AUROC \(\pm\) SD & OOD AP \(\pm\) SD & ID rejected \% \(\pm\) SD pp & OOD accepted \% \(\pm\) SD pp \\
\midrule
Frozen probe / ResNet-50 & 5 & \(0.8411\pm0.0062\) & \(0.3205\pm0.0086\) & \(5.31\pm0.12\) & \(61.16\pm1.24\) \\
Frozen probe / EfficientNet-B0 & 5 & \(0.9270\pm0.0043\) & \(0.5118\pm0.0079\) & \(5.17\pm0.05\) & \(37.73\pm0.78\) \\
Frozen probe / ViT-B/16 & 5 & \(0.9539\pm0.0028\) & \(0.6231\pm0.0157\) & \(5.48\pm0.12\) & \(25.16\pm1.17\) \\
Frozen probe / CLIP image probe & 5 & \(0.8903\pm0.0051\) & \(0.3821\pm0.0130\) & \(6.07\pm0.17\) & \(47.78\pm1.81\) \\
Full network / ResNet-50 & 3 & \(0.9836\pm0.0030\) & \(0.8171\pm0.0205\) & \(5.45\pm0.20\) & \(6.16\pm1.86\) \\
Full network / EfficientNet-B0 & 3 & \(0.9825\pm0.0002\) & \(0.8169\pm0.0062\) & \(5.36\pm0.23\) & \(7.35\pm0.23\) \\
Full network / ViT-B/16 & 3 & \(0.9972\pm0.0004\) & \(0.9638\pm0.0037\) & \(5.04\pm0.41\) & \(0.45\pm0.15\) \\
CLIP fixed image--text zero-shot & 1 & 0.2454 & 0.0389 & 5.91 & 99.25 \\
\bottomrule
\end{tabularx}
\par\smallskip
\begin{minipage}{\textwidth}\footnotesize OOD score is \(1-\max_k p_k\) over all 38 outputs. Each acceptance threshold uses only 4,910 global PV validation images (lower fifth confidence percentile, with ties accepted), then is fixed. It does not guarantee 95\% unseen-ID coverage. SD across seeds is separate from each run's 1,000 conditional source-group bootstrap intervals; calibration/model-selection uncertainty is excluded. Constant-score AP equals the pooled OOD prevalence 0.063911. Host names, organs and acquisition are confounded; these are not intervention-safety or globally novel-pathogen claims.
\end{minipage}
\end{table*}
\endgroup


\providecommand{\ValidationRevisionColor}{\color{blue}}
\begingroup\ValidationRevisionColor
\begin{table*}[t]
\centering\scriptsize
\caption{Exploratory known-host condition OOD: 131 images/131 groups across 15 annotated conditions on Blueberry, Raspberry and Soybean host names represented only as healthy in PV38.}
\label{tab:validation-v1-known-condition-ood}
\setlength{\tabcolsep}{3pt}
\renewcommand{\arraystretch}{1.15}
\begin{tabularx}{\textwidth}{X>{\centering\arraybackslash}X>{\centering\arraybackslash}X>{\centering\arraybackslash}X>{\centering\arraybackslash}X}
\toprule
Cohort / model & Runs & Global-ID AUROC \(\pm\) SD & Restricted-ID AUROC \(\pm\) SD & OOD accepted \% \(\pm\) SD pp \\
\midrule
CLIP fixed image--text zero-shot & 1 & \(0.37403\) & \(0.22450\) & \(98.47\) \\
Frozen probe / CLIP image probe & 5 & \(0.89432\pm0.00811\) & \(0.89452\pm0.00959\) & \(37.86\pm3.44\) \\
Frozen probe / EfficientNet-B0 & 5 & \(0.92992\pm0.00294\) & \(0.95075\pm0.00260\) & \(37.86\pm1.58\) \\
Frozen probe / ResNet-50 & 5 & \(0.86550\pm0.00941\) & \(0.89455\pm0.00997\) & \(53.28\pm4.02\) \\
Frozen probe / ViT-B/16 & 5 & \(0.96672\pm0.00046\) & \(0.98421\pm0.00077\) & \(18.17\pm2.12\) \\
Full network / EfficientNet-B0 & 3 & \(0.98229\pm0.00067\) & \(0.99157\pm0.00127\) & \(7.89\pm1.17\) \\
Full network / ResNet-50 & 3 & \(0.97900\pm0.00509\) & \(0.98651\pm0.00520\) & \(8.14\pm1.17\) \\
Full network / ViT-B/16 & 3 & \(0.99617\pm0.00105\) & \(0.99912\pm0.00015\) & \(0.76\pm0.76\) \\
\bottomrule
\end{tabularx}
\par\smallskip
\begin{minipage}{\textwidth}\footnotesize Values with $\pm$ are optimization-seed means and sample SD; fixed zero-shot has no seed SD. Conditional source-group sampling intervals are separate. Thirty model diagnostics yield 60 reference analyses, not 60 independent models. All prior global validation thresholds and 38 outputs remain unchanged. Global ID comprises 9,828 images/3,864 groups; the host-set restricted healthy ID reference contains 576 images/58 groups (Blueberry 279, Raspberry 19, Soybean 278). This restricts host-label support; it is not paired or equal-proportion matching. AP prevalence differs: 1.315393\% globally versus 18.528996\% restricted. The same OOD scores/threshold imply identical OOD acceptance under both references. Inputs/protocol were frozen before extension inference, after the main campaign was defined; selection is post hoc/exploratory. Sparse conditions, host proportions, organs and web acquisition remain confounded. No globally novel pathogen, causal disease isolation, healthy-class correctness or intervention safety follows.
\end{minipage}
\end{table*}
\endgroup


\begingroup\color{blue}\footnotesize
\setlength{\tabcolsep}{3pt}
\renewcommand{\arraystretch}{1.15}
\begin{longtable}{@{}p{0.25\textwidth}p{0.18\textwidth}rrrr@{}}
\caption{Additional known-host condition OOD statistics with unchanged prior global validation thresholds and all 38 outputs. Global ID has 9,828 images/3,864 groups; host-restricted healthy ID has 576/58. AP prevalence is 0.01315393 versus 0.18528996; differences in AP across references do not alone imply better detection. Values with SD describe seeds, not group-bootstrap sampling intervals.}\label{tab:s-known-ap}\\
\toprule
Model regime & ID reference & Runs & AP & ID reject \% & OOD accept \% \\
\midrule\endfirsthead
\multicolumn{6}{l}{\textit{Continued from previous page}}\\
\toprule
Model regime & ID reference & Runs & AP & ID reject \% & OOD accept \% \\
\midrule\endhead
\midrule\multicolumn{6}{r}{\textit{Continued on next page}}\\\endfoot
\bottomrule\endlastfoot
Fixed zero / CLIP probe & Global & 1 & 0.0096 & 5.91 & 98.47 \\
Fixed zero / CLIP probe & Host-restricted & 1 & 0.1155 & 10.59 & 98.47 \\
Frozen / CLIP probe & Global & 5 & 0.2226$\pm$0.0138 & 6.07$\pm$0.17 & 37.86$\pm$3.44 \\
Frozen / CLIP probe & Host-restricted & 5 & 0.7693$\pm$0.0121 & 3.92$\pm$0.26 & 37.86$\pm$3.44 \\
Frozen / EffNet-B0 & Global & 5 & 0.1982$\pm$0.0254 & 5.17$\pm$0.05 & 37.86$\pm$1.58 \\
Frozen / EffNet-B0 & Host-restricted & 5 & 0.7876$\pm$0.0083 & 3.37$\pm$0.26 & 37.86$\pm$1.58 \\
Frozen / ResNet-50 & Global & 5 & 0.1642$\pm$0.0072 & 5.31$\pm$0.12 & 53.28$\pm$4.02 \\
Frozen / ResNet-50 & Host-restricted & 5 & 0.7497$\pm$0.0090 & 2.15$\pm$0.10 & 53.28$\pm$4.02 \\
Frozen / ViT-B/16 & Global & 5 & 0.3706$\pm$0.0112 & 5.48$\pm$0.12 & 18.17$\pm$2.12 \\
Frozen / ViT-B/16 & Host-restricted & 5 & 0.9380$\pm$0.0069 & 1.60$\pm$0.33 & 18.17$\pm$2.12 \\
Full / EffNet-B0 & Global & 3 & 0.5755$\pm$0.0150 & 5.36$\pm$0.23 & 7.89$\pm$1.17 \\
Full / EffNet-B0 & Host-restricted & 3 & 0.9616$\pm$0.0067 & 2.60$\pm$0.17 & 7.89$\pm$1.17 \\
Full / ResNet-50 & Global & 3 & 0.5971$\pm$0.0120 & 5.45$\pm$0.20 & 8.14$\pm$1.17 \\
Full / ResNet-50 & Host-restricted & 3 & 0.9529$\pm$0.0131 & 2.03$\pm$0.66 & 8.14$\pm$1.17 \\
Full / ViT-B/16 & Global & 3 & 0.8613$\pm$0.0384 & 5.04$\pm$0.41 & 0.76$\pm$0.76 \\
Full / ViT-B/16 & Host-restricted & 3 & 0.9964$\pm$0.0004 & 2.66$\pm$0.56 & 0.76$\pm$0.76 \\
\end{longtable}
\endgroup


Novel-host OOD contains 671 official PlantSeg test images, 666 screened groups
and 19 predeclared literal host labels absent from the downstream PV crop labels.
Citrus and Zucchini are conservatively excluded because their relation to PV
Orange and generic Squash is ambiguous. Crop/domain/organ differences remain
confounded and global pathogen novelty is not established. The fixed CLIP
zero-shot negative outcome is retained: AUROC 0.2454 and OOD acceptance 99.25\%,
with no score reversal or prompt retuning.

The separate known-host condition cohort was selected post hoc/exploratorily
and frozen before its own inference: 131 images/groups and 15 source annotations
on Blueberry (27), Raspberry (12) and Soybean (92), crop labels represented only
as healthy in the downstream PV taxonomy. This defines annotated host--condition
categories outside that closed label space, not globally novel pathogens or
absence from pretraining. Thirty models produce 60 reference analyses: global
ID has 9,828 images/3,864 groups; the host-set restricted healthy ID reference has
576 images/58 groups (279 Blueberry, 19 Raspberry, 278 Soybean). True-class
membership and source positions were frozen before OOD scores. It is not
pair-matching, equal-proportion matching or organ matching. All 38 outputs and
the original global validation thresholds remain.

The identical OOD scores/thresholds imply identical OOD acceptance in both ID
references, whereas AUROC, AP and ID rejection can differ. AP prevalence is
1.315393\% globally and 18.528996\% in the restricted reference, so a direct AP
change cannot alone imply improved detection. Full ViT has mean global/restricted
AUROC 0.996169/0.999121 with seed SD 0.001050/0.000146, while the fixed CLIP
zero-shot model has 0.374028/0.224502 and 98.47\% OOD acceptance. These contrastive
outcomes remain exploratory and acquisition-confounded. In the full-ViT seed-17
global analysis, group-percentile AUROC bounds are $[0.992958,0.997642]$ and OOD
acceptance bounds $[0,0.038168]$; the observed zero-acceptance seed-43 bootstrap is
degenerate and cannot prove zero unseen acceptance.

\SuppFigure{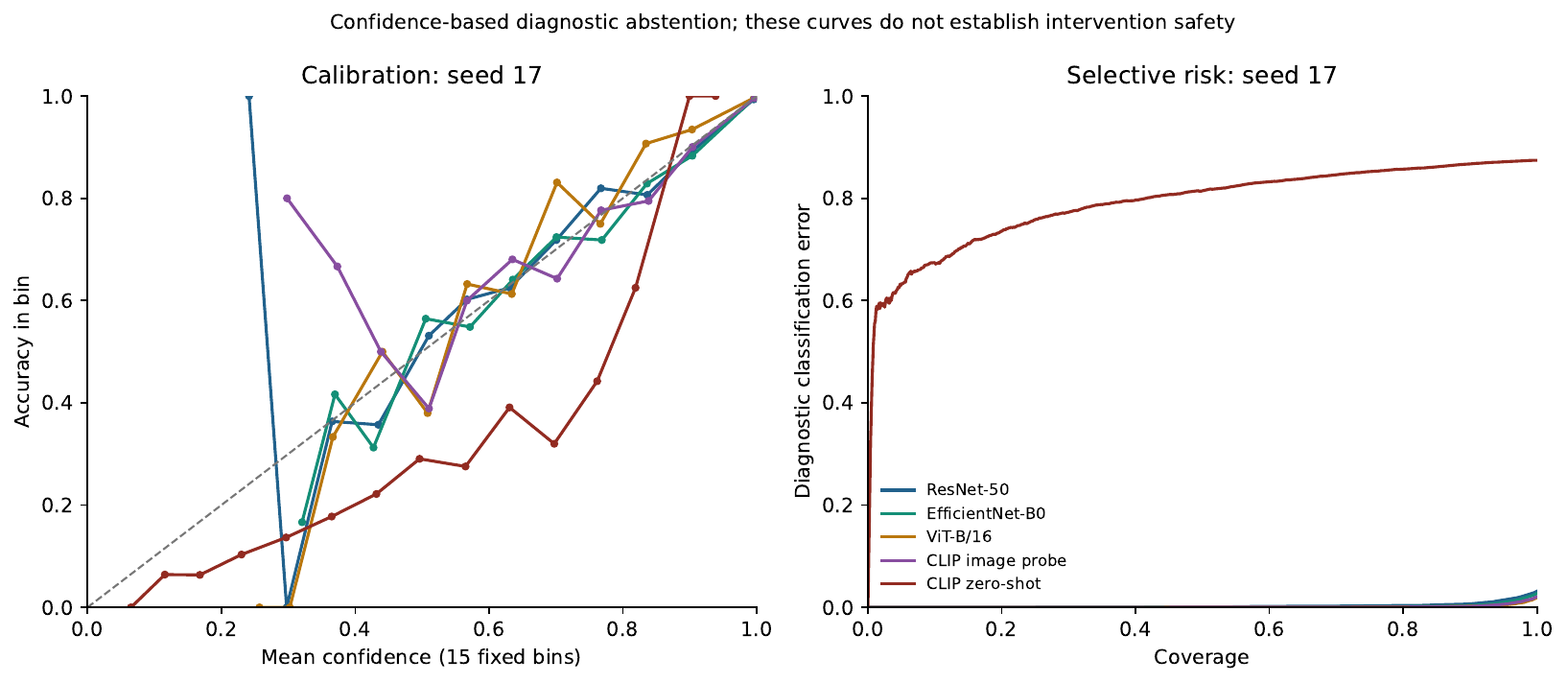}{Raw-confidence
calibration and descriptive selective risk--coverage evidence for declared
classification tasks. Test-confidence sweeps do not implement or certify
\revsecond{an agricultural treatment abstention rule.}}{fig:s-calibration}

\begin{figure}[ht]
\centering
\includegraphics[width=\linewidth,height=0.78\textheight,keepaspectratio]{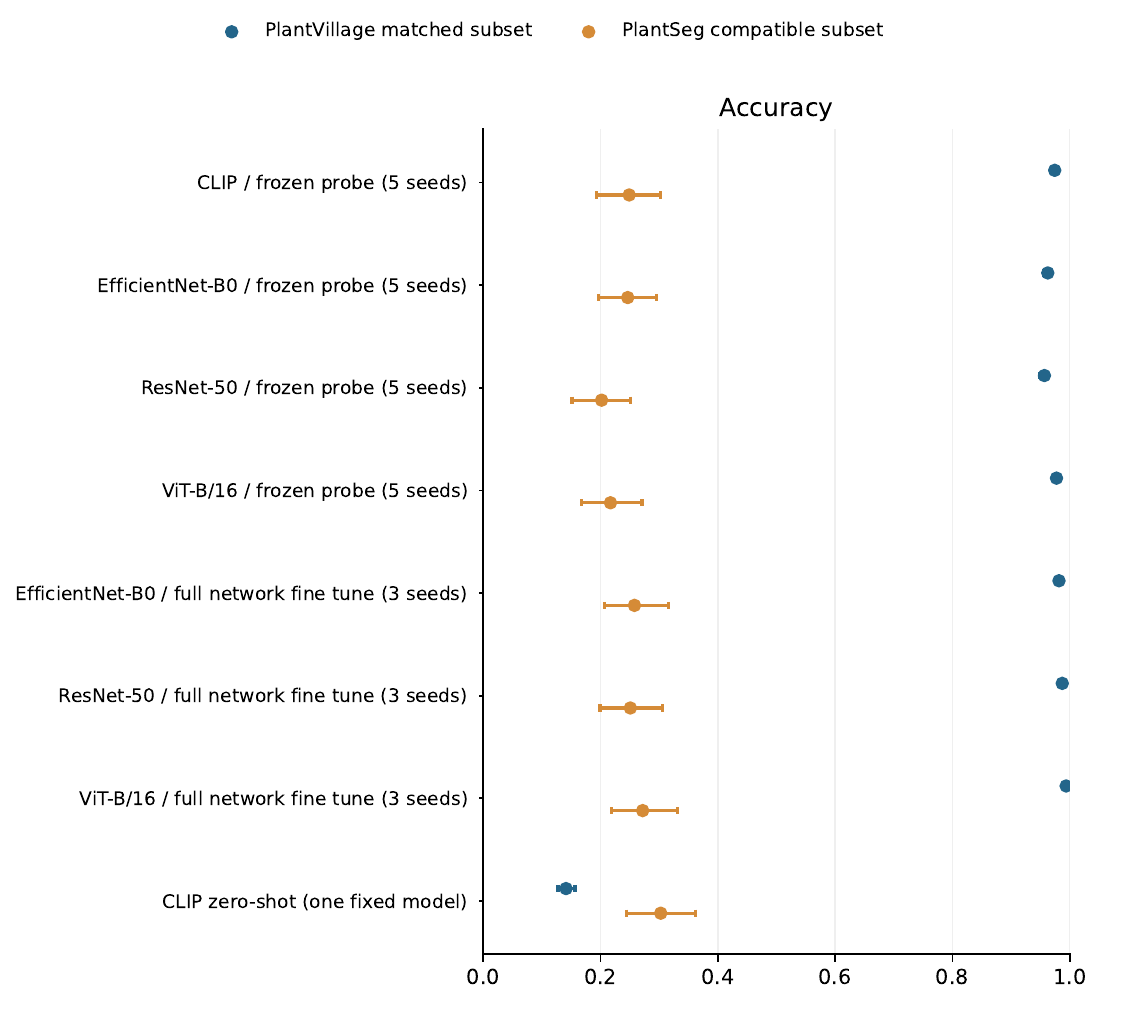}
\caption{\rev{Post hoc shared true-class support with unrestricted 38-output predictions. Approximate crossed seed/group intervals condition on the selected subsets and fitted seeds; frequency, organ and acquisition differences remain. \revsecond{(a) Accuracy. The derived saved-prediction analysis uses 4,106 PV images and 238 external images. Lines show 95\% observed-group and paired-seed percentile intervals.}}}
\label{fig:s-transfer}
\end{figure}

\begin{figure}[ht]
\ContinuedFloat
\centering
\includegraphics[width=\linewidth,height=0.78\textheight,keepaspectratio]{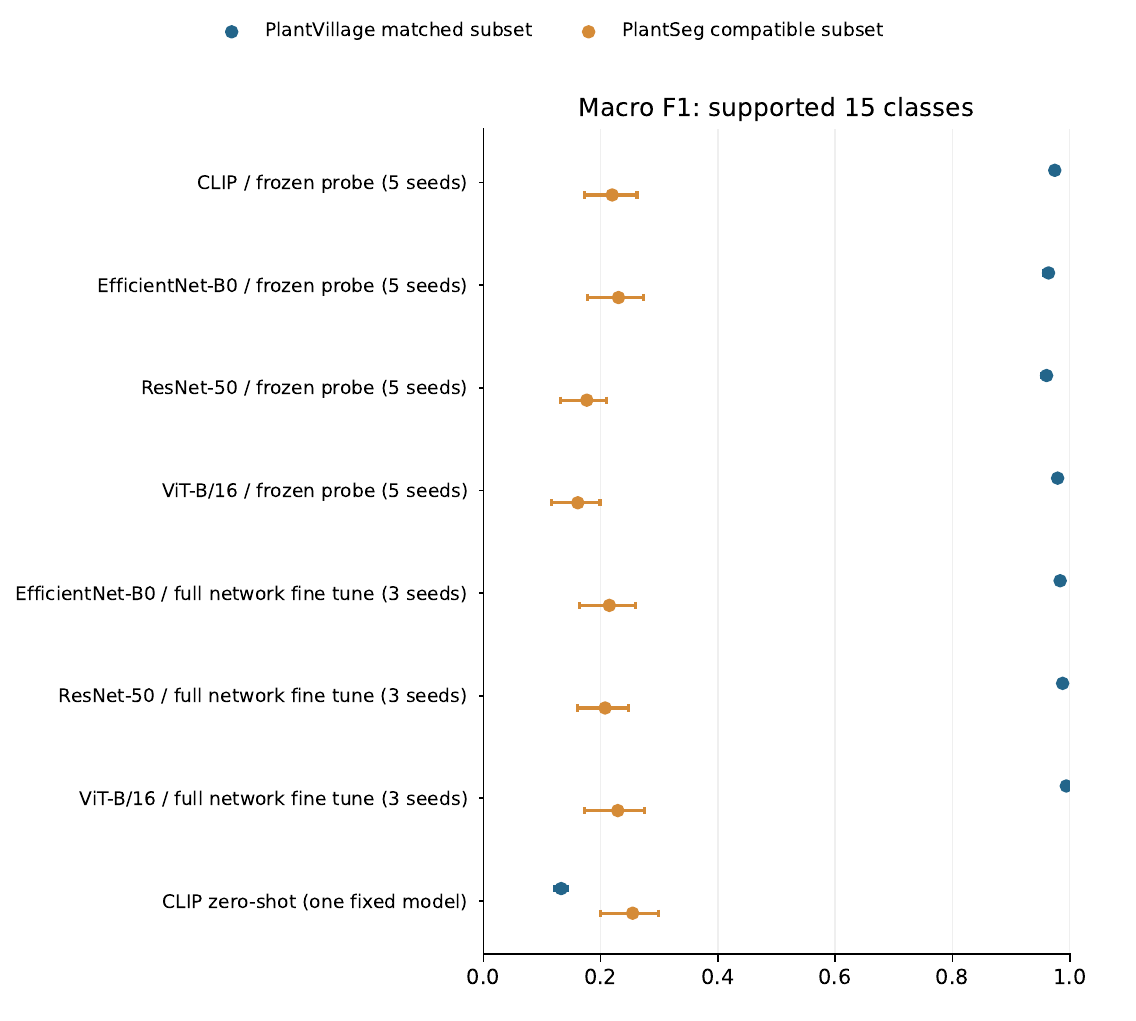}
\caption{\revsecond{Continued: (b) macro F1 over the 15 shared true classes. The cohorts, unrestricted 38-output predictions and interval definition are unchanged from panel (a).}}
\end{figure}

\begin{figure}[ht]
\ContinuedFloat
\centering
\includegraphics[width=\linewidth,height=0.78\textheight,keepaspectratio]{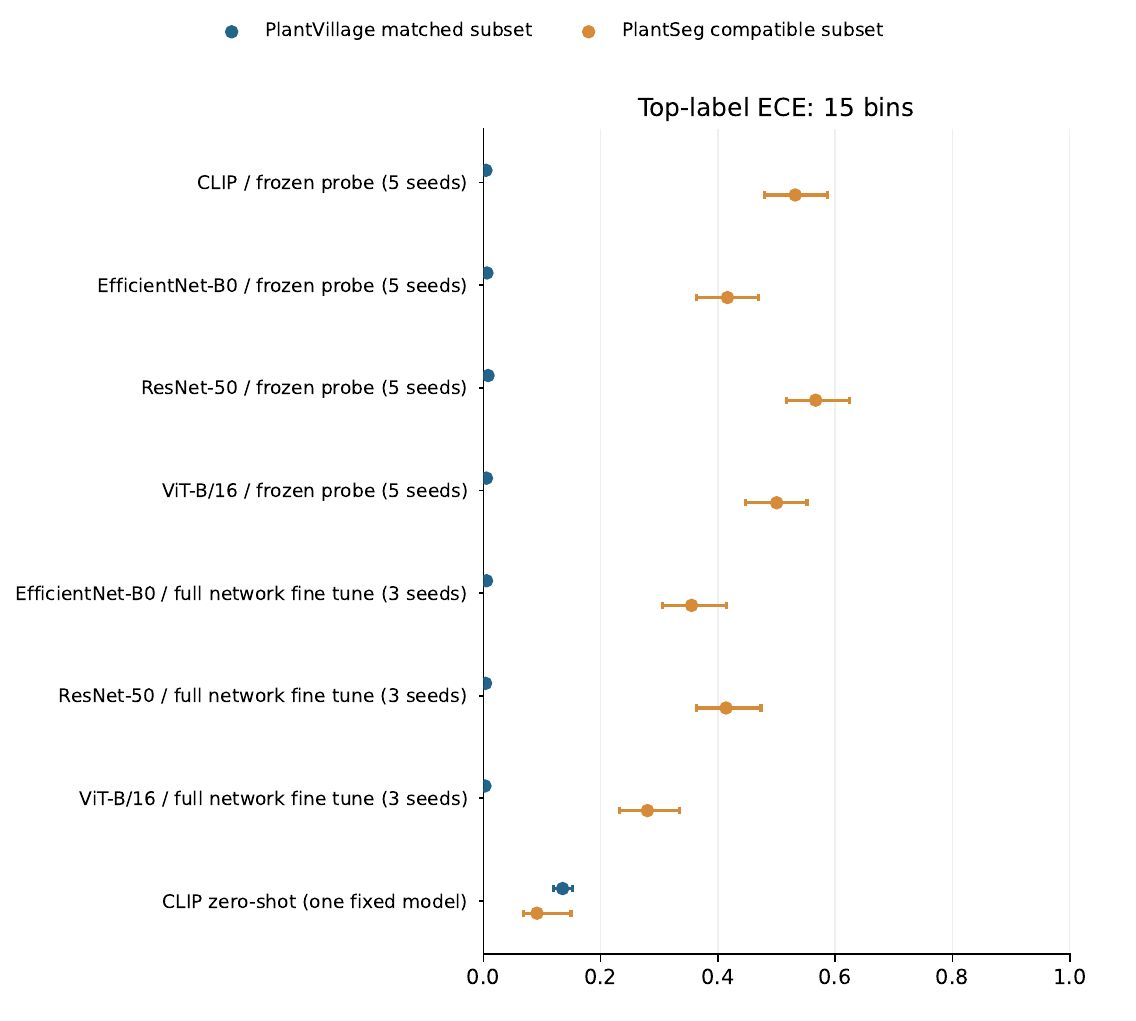}
\caption{\revsecond{Continued: (c) top-label ECE with 15 bins. The cohorts, unrestricted 38-output predictions and interval definition are unchanged from panel (a).}}
\end{figure}

\SuppFigure{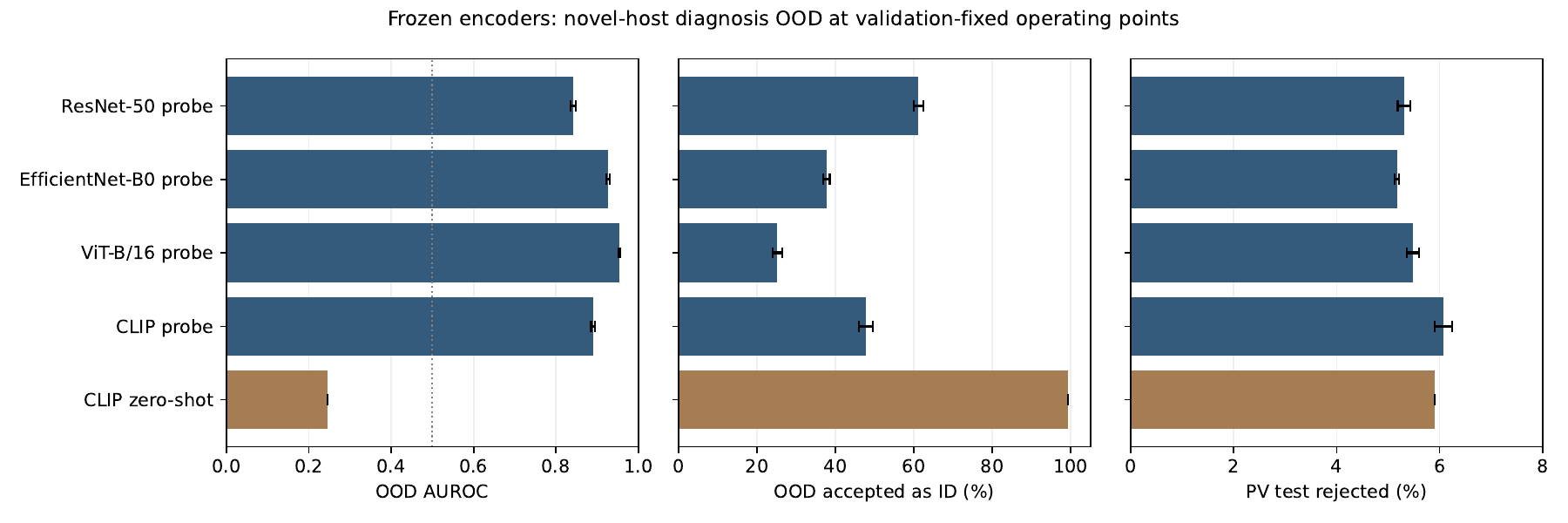}{Novel-host confidence
diagnostics for the frozen encoders and fixed CLIP. Thresholds are validation-only;
whiskers are optimization-seed SD. Per-run sampling intervals are separate.
OOD acceptance is not unsafe treatment. Full-network rows are given in the
complete numerical table and CSV.}{fig:s-novel-ood}

\begin{figure}[ht]
\centering
\includegraphics[width=\linewidth,height=0.78\textheight,keepaspectratio]{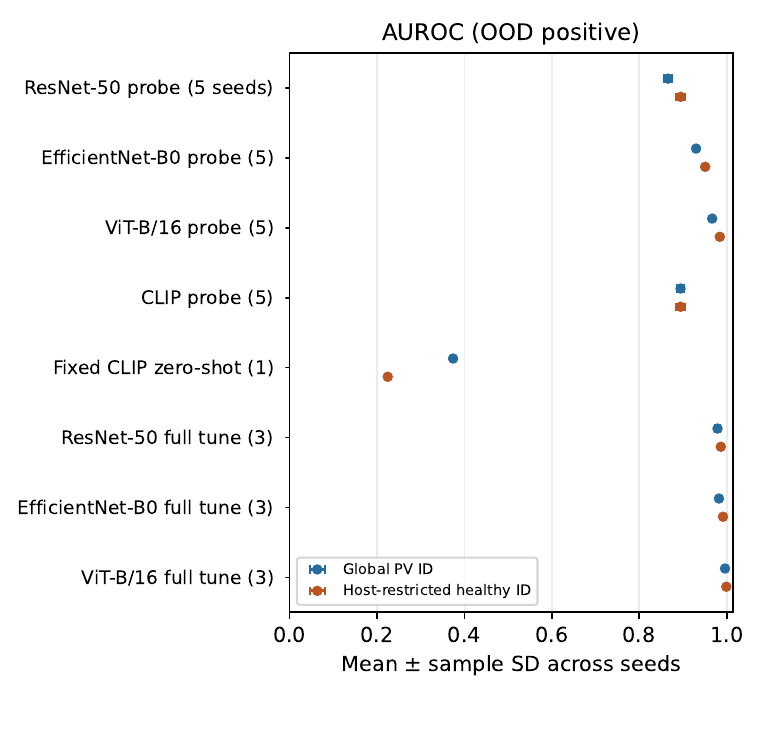}
\caption{\rev{Exploratory known-host condition OOD against global and host-restricted healthy ID references. Identical OOD scores and unchanged thresholds give identical OOD acceptance; reference composition and AP prevalence differ. Bars show seed SD, not group confidence intervals or causal disease effects. \revsecond{(a) AUROC, with OOD as the positive class. The cohort contains 131 PlantSeg images. All 38 outputs and the original global PV validation thresholds are retained. Host restriction is not pair matching; domain, organ and condition differences remain confounded. Seed SD applies to repeatedly trained configurations; the fixed CLIP model has no optimization-seed SD.}}}
\label{fig:s-known-ood}
\end{figure}

\begin{figure}[ht]
\ContinuedFloat
\centering
\includegraphics[width=\linewidth,height=0.78\textheight,keepaspectratio]{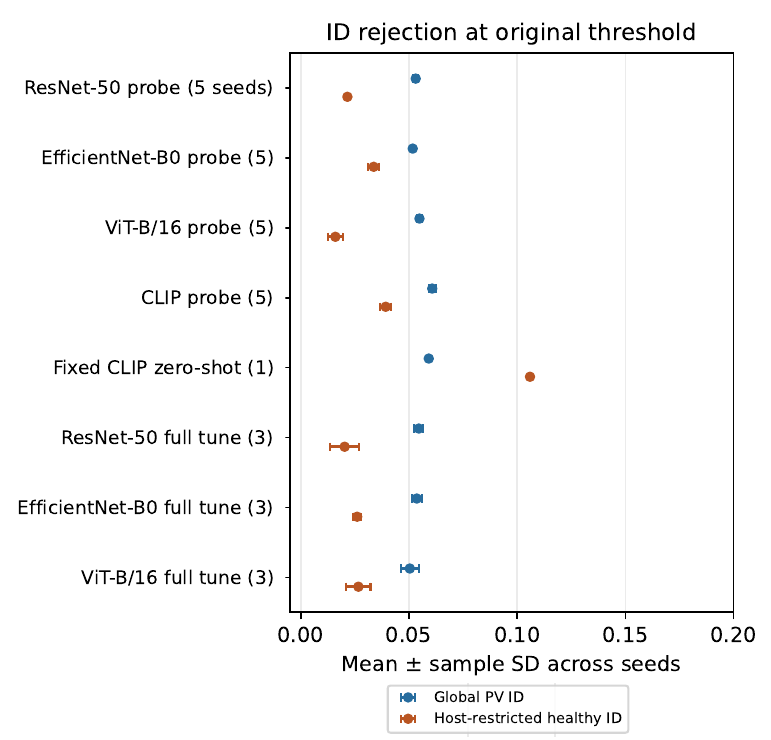}
\caption{\revsecond{Continued: (b) ID rejection at the original validation threshold, for the same global and host-restricted healthy ID references as panel (a). Bars show sample SD across seeds.}}
\end{figure}

\begin{figure}[ht]
\ContinuedFloat
\centering
\includegraphics[width=\linewidth,height=0.78\textheight,keepaspectratio]{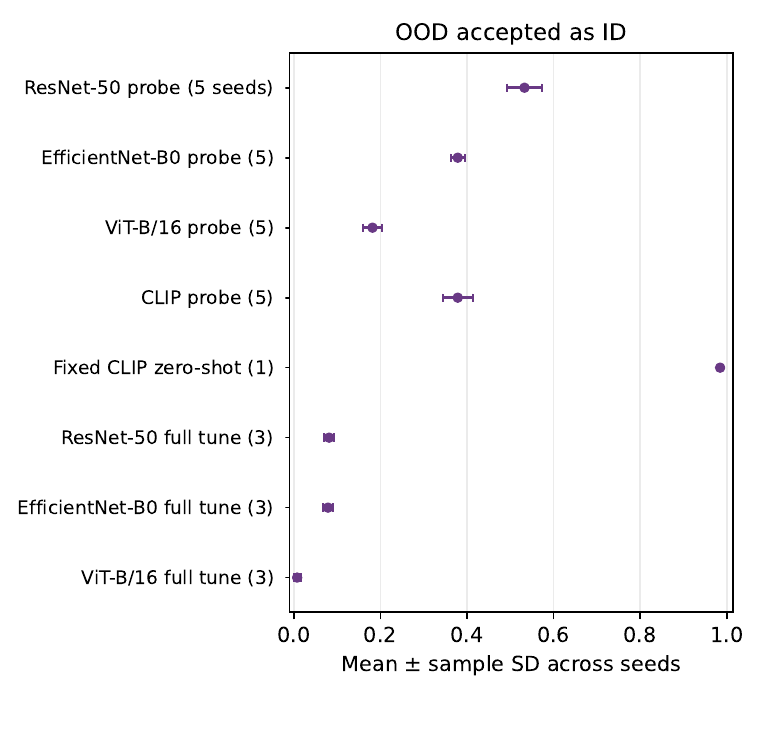}
\caption{\revsecond{Continued: (c) OOD accepted as ID at the original validation threshold. The OOD scores and threshold are identical for the two ID-reference analyses. Bars show sample SD across seeds.}}
\end{figure}

\clearpage
\FloatBarrier
\section{\rev{Virtual-greenhouse simulation protocols and outcomes}}
\label{sec:s-simulation}
\subsection{\rev{Protocol cohorts and reporting scope}}
\rev{This section reports the team's virtual-greenhouse component experiments. Navigation/control, arm geometry, integrated missions, environmental responses and command faults use separate protocols and denominators. Numerical summaries are transcribed from the retained experiment reports; the present revision does not recompute them from raw simulation traces. Complete climate parameterization and raw-trace reproducibility remain limitations, as stated in the main discussion.}

The 38 navigation/control cases comprise ten nominal repeats, six clear-layout
robustness cases, two intended obstacle stops and 20 valid controls under an
arm-proxy protocol. They are distinct from three precision-arm integrated
missions, nine integrated faults, 40 arm requests and 90 forced-state isolated
faults. These denominators must not be pooled into an overall reliability rate.
The aisle speed 0.50 m/s is a command, not a minimum measured speed. Nominal
starting pose and learned policy stay fixed; estimator noise does not drive
navigation. The first five nominal runs start the mission at 5 s and later runs
at 3 s, and the total-time chart preserves that initialization difference.

\begingroup\color{blue}\footnotesize
\setlength{\tabcolsep}{3pt}
\renewcommand{\arraystretch}{1.15}
\begin{longtable}{@{}p{0.28\textwidth}p{0.36\textwidth}p{0.29\textwidth}@{}}
\caption{Navigation/control and fault-test cohorts. Counts retain their protocol-specific denominators. Near-rest treatment dwell and moving obstacle stopping are separate measurements.}\label{tab:s-simulation-safety}\\
\toprule
Protocol & Metric & Reported value \\
\midrule\endfirsthead
\multicolumn{3}{l}{\textit{Continued from previous page}}\\
\toprule
Protocol & Metric & Reported value \\
\midrule\endhead
\midrule\multicolumn{3}{r}{\textit{Continued on next page}}\\\endfoot
\bottomrule\endlastfoot
arm-proxy nominal & completion & 10/10 cases \\
arm-proxy nominal & reported Wilson95 interval & 72.2--100.0 percent \\
arm-proxy clear robustness & completion & 6/6 cases \\
arm-proxy physical obstacles & intended safe-stop outcomes & 2/2 cases \\
arm-proxy valid controls & completion & 20/20 cases \\
navigation command & aisle-transit speed command & 0.50 m/s \\
precision-arm integrated full route & route completion duration & 130.302 s \\
isolated forced-TREAT fallback harness & invalid-input rejection & 90/90 trials \\
isolated forced-TREAT fallback harness & maximum safe-dwell time & 0.864 s \\
integrated faults after treatment admission & stopped with safe dwell & 9/9 faults \\
integrated faults after treatment admission & no post-fault dose & 9/9 faults \\
integrated faults after treatment admission & maximum safe-dwell time & 0.500 s \\
physical safe-dwell definition & planar speed and yaw-rate criteria & <0.035;<0.06;0.5 m/s;rad/s;s \\
moving-route obstacle stop case1 & stop-to-dwell time & 3.412 s \\
moving-route obstacle stop case2 & stop-to-dwell time & 3.282 s \\
one moving-route obstacle case & minimum conservative logged clearance margin & 0 m \\
\end{longtable}
\endgroup

\begingroup\color{blue}\footnotesize
\setlength{\tabcolsep}{3pt}
\renewcommand{\arraystretch}{1.15}
\begin{longtable}{@{}p{0.28\textwidth}p{0.36\textwidth}p{0.29\textwidth}@{}}
\caption{Modeled scene components and authored placement coordinates. Additional visual demonstration assemblies are separate from these primary counts.}\label{tab:s-simulation-scene-components}\\
\toprule
Protocol & Metric & Reported value \\
\midrule\endfirsthead
\multicolumn{3}{l}{\textit{Continued from previous page}}\\
\toprule
Protocol & Metric & Reported value \\
\midrule\endhead
\midrule\multicolumn{3}{r}{\textit{Continued on next page}}\\\endfoot
\bottomrule\endlastfoot
Capacitive soil VWC probes & authored component count & 24 components \\
Shielded temperature/RH stations & authored component count & 4 components \\
Zone irrigation valves & authored component count & 4 components \\
Exhaust fans & authored component count & 4 components \\
Suspended circulation fans & authored component count & 3 components \\
\end{longtable}
\endgroup

The authored placement table specifies soil-probe prong centres at approximately
$z=0.270$ m; four shielded temperature/RH stations at $x=1.05$ m,
$y=4,11,18,25$ m and shield $z=1.3375$ m; four irrigation-valve anchors at
$x=-6.7$ m, the same four $y$ positions and $z=0.18$ m; four exhaust fans at
$x=-7.4$ m, those four $y$ positions and $z=2.7$ m; and three suspended
circulation fans at $x=0$, $y=6,17,27$ m and $z=3.1$ m. These are modeled
placements, not surveyed/calibrated measurements. Additional near-aisle visual
assemblies are not silently added to the primary counts.

\subsection{\rev{Targeting geometry and oracle-driven missions}}
\begingroup\color{blue}\footnotesize
\setlength{\tabcolsep}{3pt}
\renewcommand{\arraystretch}{1.15}
\begin{longtable}{@{}p{0.28\textwidth}p{0.36\textwidth}p{0.29\textwidth}@{}}
\caption{Virtual-target geometric benchmark: 32 accepted requests and eight boundary rejections. Pose-success and disk-hit rates use accepted targets only.}\label{tab:s-simulation-targeting}\\
\toprule
Protocol & Metric & Reported value \\
\midrule\endfirsthead
\multicolumn{3}{l}{\textit{Continued from previous page}}\\
\toprule
Protocol & Metric & Reported value \\
\midrule\endhead
\midrule\multicolumn{3}{r}{\textit{Continued on next page}}\\\endfoot
\bottomrule\endlastfoot
virtual crop-edge targets & pre-specified total requests & 40 requests \\
accepted virtual targets & position error mean \ensuremath{\pm} sampleSD & 0.449 \ensuremath{\pm} 0.154 mm \\
accepted virtual targets & maximum position error & 0.853 mm \\
accepted virtual targets & mean nozzle-axis error & 0.000502 rad \\
accepted virtual targets & pose acceptance & 32/32 requests \\
accepted virtual targets & virtual disk hits & 32/32 requests \\
accepted virtual targets & ray error mean \ensuremath{\pm} sampleSD & 0.829 \ensuremath{\pm} 0.440 mm \\
far-boundary requests & rejected before servo control & 8/8 requests \\
virtual benchmark & observed upright failures & 0 failures \\
accepted virtual targets & settling and control windows & 3;6 s \\
\end{longtable}
\endgroup

\begingroup\color{blue}\footnotesize
\setlength{\tabcolsep}{3pt}
\renewcommand{\arraystretch}{1.15}
\begin{longtable}{lrr}
\caption{Integrated authored-target missions, each completed. These three cases use simulator-state diagnosis and targets and are separate from the 32-target geometric benchmark.}\label{tab:s-sim-berries}\\
\toprule
Case / seed & Duration s & Max treated-target ray error mm \\
\midrule\endfirsthead
\multicolumn{3}{l}{\textit{Continued from previous page}}\\
\toprule
Case / seed & Duration s & Max treated-target ray error mm \\
\midrule\endhead
\midrule\multicolumn{3}{r}{\textit{Continued on next page}}\\\endfoot
\bottomrule\endlastfoot
full route / 1200 & 130.302 & 12.94 \\
left berry / 1201 & 30.802 & 10.99 \\
right berry / 1202 & 33.802 & 6.40 \\
\end{longtable}
\endgroup

The virtual geometric benchmark comprises 32 accepted crop-edge targets and eight
far-boundary requests. The \revsecond{mean position error across accepted targets} is $0.449\pm0.154$ mm
(sample SD), maximum 0.853 mm; virtual target-plane ray error is
$0.829\pm0.440$ mm. The \revsecond{32/32 pose-success and 32/32 disk-hit counts for accepted targets} exclude all
eight pre-servo boundary rejections. They use 3 s settling, 6 s control, 0.95 m
standoff and virtual target heights 0.65--0.85 m. Pose criteria are position below
30 mm and nozzle-axis error below 0.1 rad; virtual disk radius is 30 mm. The
finger-link origin is the modeled nozzle and roll is unspecified.

These selected submillimetre virtual statistics are not the integrated authored
berry results. Integrated full-route, left-berry and right-berry maximum treated
target ray errors are 12.94, 10.99 and 6.40 mm, respectively. Diagnosis and target
positions in those missions use simulator-state oracles, not demonstrated camera
localization or the \revsecond{VLDA inference pathway.} Modelled treatment
units, fan fractions and water particles are not calibrated pesticide volume,
deposition, CFD or biological efficacy. Physical joint/contact control is reported,
but full-body/self/environment contact coverage is not instrumented.

The full-route mission is reported COMPLETE at 130.302 simulated seconds,
visiting beds 1, 7, 13 and 19 and issuing two modeled interventions. Its 4,900
TREAT-phase physics-step checks and 100 instantaneous dose-command checks are
different denominators. A terminal dose on transition to STOW belongs to the
command checks, not subsequent TREAT-step checks. A separate recorded repeat
contains 255 frames at 2 fps, 127.5 s of playback sampling simulated 3.0--130.0 s;
the terminal COMPLETE event at 130.302 s follows the final sampled frame.
Paused-physics rendering preserves the recorded capture pose; playback frame count and simulated elapsed time have distinct meanings.

\subsection{\rev{Offline localization and modeled climate}}
\begingroup\color{blue}\footnotesize
\setlength{\tabcolsep}{3pt}
\renewcommand{\arraystretch}{1.15}
\begin{longtable}{lrrrr}
\caption{Offline Point-LIO on simulated LiDAR/IMU streams. Initial ATE uses first-overlap full-pose alignment; fit ATE uses all-overlap rigid SE(3) alignment at unit scale. Navigation uses simulator pose.}\label{tab:s-sim-lio}\\
\toprule
Route & Length m & Initial ATE mm & Fit ATE mm & 1 s RPE mm \\
\midrule\endfirsthead
\multicolumn{5}{l}{\textit{Continued from previous page}}\\
\toprule
Route & Length m & Initial ATE mm & Fit ATE mm & 1 s RPE mm \\
\midrule\endhead
\midrule\multicolumn{5}{r}{\textit{Continued on next page}}\\\endfoot
\bottomrule\endlastfoot
route\_00 & 26.89 & 19.80 & 5.22 & 3.80 \\
route\_01 & 26.83 & 40.12 & 9.63 & 4.77 \\
route\_02 & 26.97 & 74.26 & 17.32 & 6.26 \\
route\_03 & 27.05 & 20.37 & 5.71 & 4.16 \\
route\_04 & 27.12 & 22.89 & 6.35 & 4.36 \\
\end{longtable}
\endgroup

Point-LIO \citep{he2023pointlio} consumes simulated LiDAR and IMU together in
offline replay. The documented inputs are 16 rings, 200 azimuth samples/ring,
rolling acquisition at 500 Hz over 0.1 s, 10 Hz scans, a body-origin IMU at 100 Hz,
35 m range and body-aligned LiDAR translation $(0,0,0.33)$ m. Range/dropout/inertial
perturbations vary across five routes, without real optical returns, foliage
motion or calibration errors. RGB/environment do not enter that estimator.
Navigation continues to use simulator pose; offline reconstruction is not
Point-LIO-driven closed-loop navigation.

Fit ATE uses rigid SE(3) Kabsch alignment across all overlapping trajectory
positions with scale fixed at one. Initial ATE uses the first overlapping full
pose. Ground-truth positions are linearly interpolated and orientation uses SLERP
without extrapolation; one-second RPE has a 0.12 s pairing tolerance. Raw
unregistered frame RMSE is not physical localization error when origins/yaw
differ. The figure uses the registered route-00 PCD, with 513,833
points after 0.05 m voxel deduplication and every sixth stored point displayed.
Neither low aligned error nor simulated input consistency establishes surveyed
greenhouse accuracy.

\begingroup\color{blue}\footnotesize
\setlength{\tabcolsep}{3pt}
\renewcommand{\arraystretch}{1.15}
\begin{longtable}{@{}p{0.28\textwidth}p{0.36\textwidth}p{0.29\textwidth}@{}}
\caption{Six-case virtual climate suite under synthetic weather. Outcomes use the stated recovery and settling criteria; process parameterization remains incomplete.}\label{tab:s-simulation-climate}\\
\toprule
Protocol & Metric & Reported value \\
\midrule\endfirsthead
\multicolumn{3}{l}{\textit{Continued from previous page}}\\
\toprule
Protocol & Metric & Reported value \\
\midrule\endhead
\midrule\multicolumn{3}{r}{\textit{Continued on next page}}\\\endfoot
\bottomrule\endlastfoot
environmental suite & Cases, duration and logging rate & 6 cases; 12 h each; 1 Hz logging \\
high temperature & Sustained settling within 0.5 degrees C & 2461 s \\
high humidity & Sustained settling within 2 RH percentage points & 3448 s \\
dry substrate & 90\% recovery time & 7457 s \\
dry substrate & final VWC & 0.33266 fraction \\
all variable/case rows & sustained settling failures & 6/18 rows \\
hot/dry & final temperature & 42.97 degrees C \\
hot/dry & final relative humidity & 10.64\% RH \\
\end{longtable}
\endgroup

The six 12-hour, 1 Hz climate cases are dry substrate, high humidity, high
temperature, hot/dry, combined disturbance and actuator saturation. Six of 18
variable/case rows fail sustained settling: dry-substrate VWC; hot/dry
temperature, RH and VWC; combined-disturbance VWC; saturation temperature.
High temperature settles within 0.5 degrees Celsius after 2,461 s and high RH
within two percentage points after 3,448 s. Dry substrate reaches 90\% recovery
after 7,457 s but fails sustained $\pm0.01$ VWC settling, ending at 0.33266 VWC.
Recovery and settling are different criteria. Hot/dry finishes at 42.97 degrees
Celsius and 10.64\% RH. Synthetic weather and model bookkeeping do not validate
measured UAE weather or a physical greenhouse. \revgreen{Complete process equations, coefficients, numerical integration settings, controller gains and cross-zone coupling remain unavailable; this limits reproducibility as discussed in the main article. The 1 Hz value denotes logging frequency.}

\subsection{\rev{Original learned-model readiness versus simulation command guards}}
\providecommand{\ValidationRevisionColor}{\color{blue}}
\begingroup\ValidationRevisionColor
\begin{table*}[!htbp]
\centering\scriptsize
\caption{Exact source-method readiness fixtures and limits; no ROS/actuator invocation or physical treatment.}
\label{tab:validation-v1-readiness}
\setlength{\tabcolsep}{3pt}
\renewcommand{\arraystretch}{1.15}
\begin{tabularx}{\textwidth}{p{0.27\textwidth}p{0.28\textwidth}X}
\toprule
Fixture or evidence & Observed outcome & Permitted interpretation \\
\midrule
Engineered missing/stale/desynchronized inputs & 52 rejected / 52 invalid fixtures & Existing presence/age/synchrony branches match specification \\
Fresh valid control & 1 accepted / 1 valid control; false blocks 0/1 & Only one valid control; no population false-block rate \\
Nonfinite/extreme numeric inputs & 64 accepted / 64 probes & This method does not enforce finite/range bounds \\
Synchronized future timestamp & 1 accepted / 1 probe & Future-time admissibility not enforced by this method \\
Expert intervention outcomes & Unavailable; no unsafe-action/false-intervention rate & Generated-label agreement is not field safety \\
\bottomrule
\end{tabularx}
\par\smallskip
\begin{minipage}{\textwidth}\footnotesize The 53 presence/age/synchrony cases comprise 52 invalid fixtures and one valid control; they are not 53 invalid rejection trials. The exact method was tested with dummy state and fixed time; the control interface was not run. Confidence/OOD utilities are offline and are not integrated into the VLDA actuation path. Numerical command guards and alternative fusion architectures are not part of this readiness test. Physical sensor/LiDAR/manipulator evidence and expert labels remain separate requirements. The corresponding controller tests use the separate protocols in this section.
\end{minipage}
\end{table*}
\endgroup

The original source readiness function checks presence, receipt age at most 2 s
and timestamp spread at most 1 s. Exact isolated tests reject 52/52 engineered
invalid presence/age/synchrony fixtures and accept one valid control, giving
0/1 false blocks in that sole valid fixture. All 64 nonfinite/extreme numerical
probes and one synchronized future timestamp pass this readiness method.
No ROS module, control interface or physical actuator was run in these tests.

The separate virtual-controller command guard reports
nonfinite/out-of-range command rejection, bookkeeping restoration, treatment
closure and physical dwell. The latter's 90/90 forced isolated faults have a
maximum safe-dwell time of 0.864 s; nine faults injected after genuine mission
treatment admission stop with no post-fault dose and maximum dwell time 0.500 s.
Legitimate prefault dose is retained. Safe dwell is upright posture, planar speed
below 0.035 m/s and yaw rate below 0.06 rad/s for 0.5 s. The retained protocol does not specify a numerical upright-angle
bound. Faults include stale sensor/diagnosis, nonfinite
or out-of-range command, unreachable target, unsafe clearance, disconnect,
invalid state and abort. Treatment-phase starts near rest do not measure
travel-speed emergency braking. Separate moving obstacle stops take 3.412 and
3.282 s; one logged conservative clearance margin reaches 0 m and contacts are
uninstrumented, so collision-free avoidance is not established.

Neither readiness nor the virtual-controller guard is a neural safety
head in the predictor. Same-tick rejection does not measure perception
or network latency, and restoring unexecuted dose bookkeeping is not physical
spray reversibility. Recovery/restart, complete collision instrumentation,
expert false-intervention labels and field certification remain absent.

\begin{figure}[!ht]\centering
\includegraphics[width=\linewidth]{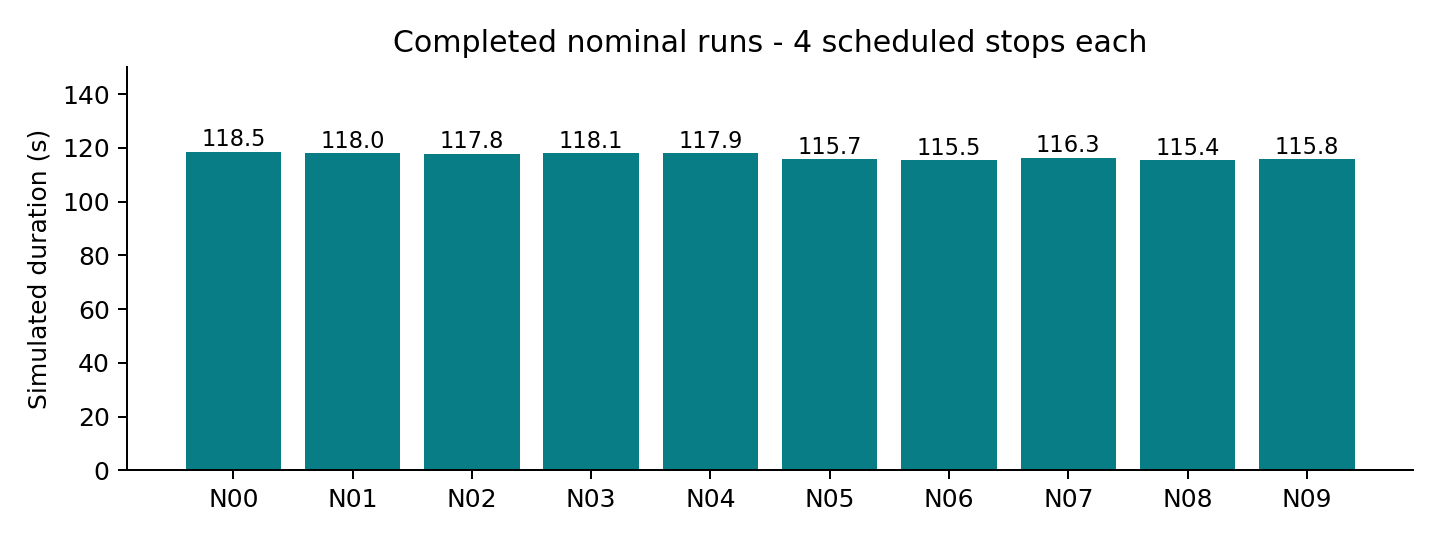}
\caption{\rev{Rounded total simulated durations for ten nominal
arm-proxy runs, with four scheduled stops. The first five mission starts are at
5 s and the later five at 3 s; this difference remains in the plotted elapsed
time. The nominal runs use one fixed mission configuration and are not independent field-reliability trials.}}\label{fig:s-simulation-navigation}\end{figure}
%

\clearpage
\FloatBarrier
\section{\rev{Reproducibility catalogue and remaining experiments}}
\label{sec:s-reproduce}
\subsection{\rev{Portable tables and full evidence package}}
The accompanying \path{supplement_data/CSV_CATALOG.csv} and
\path{supplement_data/PROVENANCE.json} identify copied versus derived files,
row counts, source paths, byte sizes and SHA-256 hashes. The small tabular exports
are included beside this source. \revsecond{The restoration manifest, \path{RESTORED_EVIDENCE_PROVENANCE.json}, records their original source identities and hashes. The restoration preserves historical evidence; the additional 24 October campaign contributes fifteen separately identified runs.} Full per-run probabilities, predictions,
checkpoints, configuration/history records and large curve/bootstrap files remain
in the complete validation-project deliverable. They are indexed, not falsely
claimed embedded as thousands of miniature supplement plots.
The two source reports are included in the evidence\_reports directory. Larger campaign artifacts are indexed in the supplied catalogues; they are not claimed to be embedded in this manuscript package.

\begin{longtable}{@{}p{0.48\textwidth}p{0.46\textwidth}@{}}
\caption{\rev{Supplementary CSV catalogue; exact hashes and source identities are in the machine-readable catalogue.}}\label{tab:s-catalog}\\
\toprule File under \path{supplement_data/} & Contents and interpretation \\
\midrule\endfirsthead
\toprule File & Contents and interpretation \\\midrule\endhead
\bottomrule\endlastfoot
\path{classification_per_run_230.csv} & All 230 portable classification evaluations, metrics and conditional CI valid counts; not 230 trainings.\\
\path{classification_model_means_56.csv} & 56 task/condition/model families, seed means and sample SD.\\
\path{classification_interval_rows_224.csv} & Four aggregate interval metrics per family, with finite/requested counts and scope.\\
\path{classification_all_per_class_230.csv} & All class precision/recall/F1, confusion counts, applicable ROC/AP and calibration rows with run identity.\\
\path{classification_curve_and_draw_catalog_230.csv} & Relative paths, sizes and hashes for full per-run curves, confusion, calibration, risk--coverage and draws in the validation package.\\
\path{matched_taxonomy_transfer.csv} & Per-seed and aggregate post hoc shared-15-class transfer contrasts; all 38 outputs retained.\\
\path{frozen_model_resources.csv}; \path{full_network_resources.csv} & Shared feature extraction versus head/full training costs, memory and device-only latency scopes.\\
\path{genuine_vlda_40_runs.csv} & Historical ledger: two splits, four branch conditions and five seeds; the fifteen additional runs are reported separately in Table S6.\\
\path{original_checkpoint_12_group_intervals.csv} & Fixed original checkpoint, 12 conditions, 43 rows/41 screened groups.\\
\path{novel_host_frozen_21_diagnostics.csv}; \path{novel_host_full_9_diagnostics.csv} & Separate novel-host diagnostic cohorts; no new training.\\
\path{known_host_condition_60_references.csv}; \path{known_host_condition_seed_aggregates.csv} & Thirty models, both global/restricted ID references, fixed prior thresholds and prevalence-sensitive AP.\\
\path{plantvillage_38_class_splits.csv} & Original/retained/quarantined class support and frozen assignments.\\
\path{virtual_greenhouse_simulation_74_facts.csv} & Simulation protocol definitions, reported values, units, denominators and source locations; raw-trace recomputation status is recorded per entry.\\
\end{longtable}

The reproducibility package indexes per-run metrics, group assignments, model configurations, curve data and figure assets. Historical and repeated experiments retain their own software and hardware records; PyTorch 2.9.1+cu128, torchvision 0.24.1 and transformers 4.37.2 describe the added campaign. Historical checkpoints and original assets are preserved. Numerical summaries in Section S4 come from retained experiment reports; raw-trace recomputation and complete climate parameterization remain outstanding.

\subsection{\rev{Specific experiments that remain unexecuted}}
\begin{enumerate}
\item Authorized Plant Pathology 2021 acquisition and a correct multilabel
training/evaluation protocol remain blocked by access, rather than being
substituted with artificial labels. PlantSeg masks are audited, but no segmentation
model or IoU/Dice result is claimed.
\item Independent greenhouse episodes, physical plants and field sites with
expert disease and intervention labels are required to resolve the single-episode
teacher-data shortcuts. No generated prompt/context annotation is measured
ground truth for that missing external control task.
\item Camera-based target localization, surveyed LiDAR/IMU calibration,
sensor-to-controller estimator integration and Point-LIO-driven closed-loop
navigation need physical, time-synchronized evaluation; offline aligned maps and
simulator pose do not close those requirements.
\item Actual learned VLDA output driving integrated missions remains to be
demonstrated with an explicit contract. Simulator diagnosis/target oracles and
independent disease-image classifiers cannot substitute for this evidence.
\item Expert-verified pesticide applicability, dose, timing, contraindications,
false interventions, deposition and biological efficacy remain unmeasured.
Confidence ranking, generated binary labels and simulated dose bookkeeping are
not treatment-safety outcomes.
\item Complete physical contact instrumentation, moving-stop clearance,
recovery/restart and timing under real perception/network delays remain open.
The one zero-clearance logged obstacle case and uninstrumented contacts are
retained as hazards, not described as collision-free avoidance.
\item The climate model requires verified governing equations/parameters,
measured weather provenance, physical calibration and validation across its
failed envelope. Proposed balance equations must be labelled proposed until
matched to executed source and traces; no UAE-weather validation is supplied.
\item Alternative fusion, a genuine deployed numerical intervention guard,
calibrated model uncertainty and a matched-budget causal unfreezing study remain
unexecuted. No missing component, theorem, convergence, novelty or optimality
claim is invented to fill these gaps.
\end{enumerate}

The supported contribution is a reproducible implementation/evaluation with
explicit negative results and evidence boundaries. High clean image scores,
generated-label agreement on a small test and bounded oracle-driven simulation outcomes
answer different questions; they are not pooled into a universal system accuracy
or a physical digital-twin/safety validation.

\end{revblock}

\newmaterialfalse
\color{black}

\clearpage
\FloatBarrier
\section{\rev{Original experimental records and supplementary results}}
\label{sec:s-original}
\label{sec:s-historical-results}
\rev{This section preserves the earlier manuscript's reported results and detailed narrative as historical records, not as newly validated experiments. The original 43-row model test, three historical missions and their LIO-labelled channels are distinct from the later grouped model tests and five offline Point-LIO routes in Sections S1--S5. The archived records do not independently establish which estimator drove each historical mission or that its controller excluded simulator pose. Design-level validation conditions are not evidence that the learned-model readiness gate passed. The historical and new cohorts retain their separate interpretations.}
{
This section reports the simulation-centric evaluation of \AgriSense{} across prediction, localization, mapping, environmental monitoring, intervention replay, and supervision outputs. The results are organized to separate quantitative evidence from qualitative demonstrations. Quantitative evaluation includes held-out \VLDA{} prediction metrics, GNSS-denied localization accuracy, trajectory analysis, point-cloud reconstruction, and variation in four-zone environmental state. Qualitative evaluation includes digital-twin views, plant-level inspection and intervention replay, the remote-supervision interface, and mission-replay evidence. The section explicitly distinguishes system-integration evidence from biological treatment or field-deployment validation.
}
\subsection{Experimental Setup and Evaluation Scope}
\label{subsec:experimental_setup}

The proposed \AgriSense{} framework was implemented under
Ubuntu~22.04.5~LTS using Python~3.10.12 and evaluated on a workstation
equipped with a 13th Gen Intel Core i9-13980HX processor, an NVIDIA
GeForce RTX~4060 Laptop GPU with 8~GB of dedicated memory, 15.2~GB of
RAM, and NVMe solid-state storage.

\rev{The original evaluation covers held-out supervised \VLDA{} predictions, historical LIO-labelled localization and mapping records, four-zone environmental-state monitoring, spatially grounded digital-twin visualization, non-real-time supervision, and qualitative plant-level inspection/intervention replay. Their evidentiary status is separated below rather than treating all components as one validated autonomous experiment.}

Three inspection missions are considered: an all-zone mission, a
Zone~2-focused mission, and a Zone~4-focused mission. Each mission
contains synchronized ground-truth and LiDAR--inertial odometry
trajectories, greenhouse-state measurements for four management zones,
localization-error records, reconstructed LiDAR maps, digital-twin
outputs, and robot-view visual logs.

\rev{The historical archive contains LIO-labelled and simulator-reference channels, but does not independently establish the online estimator identity or the pose stream consumed by navigation. The later five-route Point-LIO replay excludes reference pose from estimation; its robot-control results still use simulator pose, as distinguished in Sections S4--S5.} Localization
errors are computed from temporally synchronized ground-truth--LIO pairs. This avoids direct
row-by-row comparison of streams recorded at different sampling rates.

The results are simulation-centric. Plant-level spray demonstrations
evaluate command sequencing, geometric reachability, and integration
with the mobile platform; they do not constitute measurements of spray
deposition, chemical-use reduction, crop response, or biological
treatment efficacy.

\subsection{\VLDA{} Decision-Output Evaluation}
\label{subsec:vlda_results}

The \VLDA{} model maps prompt text, robot-view RGB observations,
greenhouse sensor states, and current-zone context into three types of
structured proposal: four-zone environmental references, target-zone
selection, and a binary pesticide-intervention decision. Table~
\ref{tab:vlda_metrics} summarizes performance on the held-out test
split.

\begin{table}[!t]
\centering
\caption{\VLDA{} performance on the \rev{historical 43-row} held-out test split. \rev{These are the original reported values, not the later strict replay or grouped-test results. The overall setpoint MAE mixes units and is retained for historical comparison only; unit-specific errors are reported separately.}}
\label{tab:vlda_metrics}
\footnotesize
\renewcommand{\arraystretch}{1.12}
\begin{tabularx}{\columnwidth}{
>{\raggedright\arraybackslash}X
>{\centering\arraybackslash}p{1.65cm}}
\toprule
\textbf{Metric} & \textbf{Value} \\
\midrule
Number of test samples             & 43 \\
Overall setpoint MAE                & 1.266 \\
Target-zone accuracy                & 95.35\% \\
Target-zone macro-F1                & 93.11\% \\
{Spray-label accuracy (exploratory)}                  & 100.00\% \\
{Spray-label F1 (exploratory)}                        & 100.00\% \\
Overall loss                        & 1.172 \\
Regression loss                     & 1.046 \\
Zone-classification loss            & 0.126 \\
Pesticide-decision loss             & 0.0002 \\
\bottomrule
\end{tabularx}
\end{table}

The target-zone metrics indicate that the model generally distinguishes
the requested or inferred management region while retaining balanced
performance across classes. The {spray-label result} should be
interpreted cautiously because it was obtained from a test split of only
43 samples. It establishes consistency on the present dataset but does
not establish field-level disease-diagnosis or treatment performance.

\begin{figure}[!tb]
\centering
\captionsetup[subfigure]{font=footnotesize,justification=centering}

\begin{subfigure}[t]{0.49\textwidth}
    \centering
    \includegraphics[width=\linewidth]{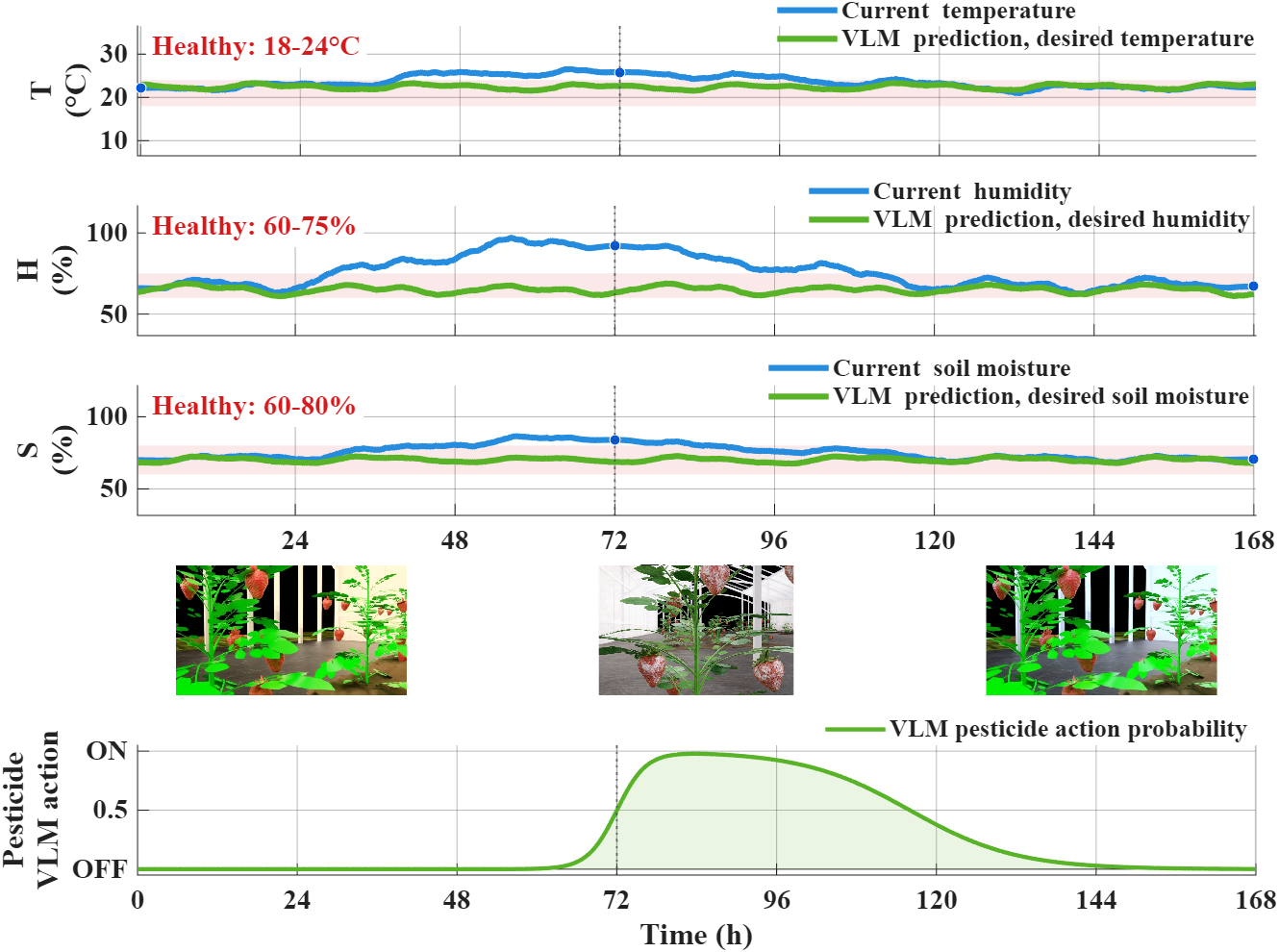}
    \caption{Representative Case~1.}
    \label{fig:vlda_zone_response_z1}
\end{subfigure}
\hfill
\begin{subfigure}[t]{0.49\textwidth}
    \centering
    \includegraphics[width=\linewidth,height=6.5cm]{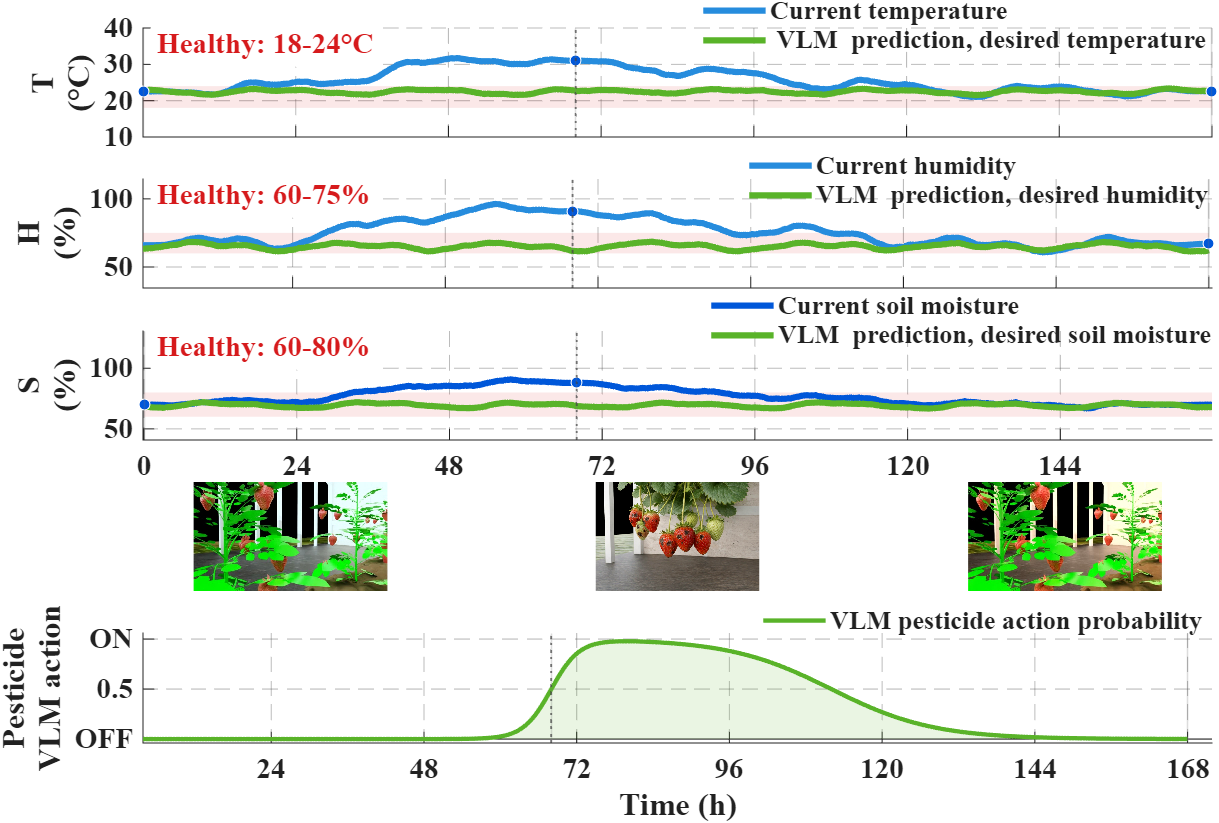}
    \caption{Representative Case~2.}
    \label{fig:vlda_zone_response_z2}
\end{subfigure}

\vspace{0.5em}

\begin{subfigure}[t]{0.49\textwidth}
    \centering
    \includegraphics[width=\linewidth,height=6.5cm]{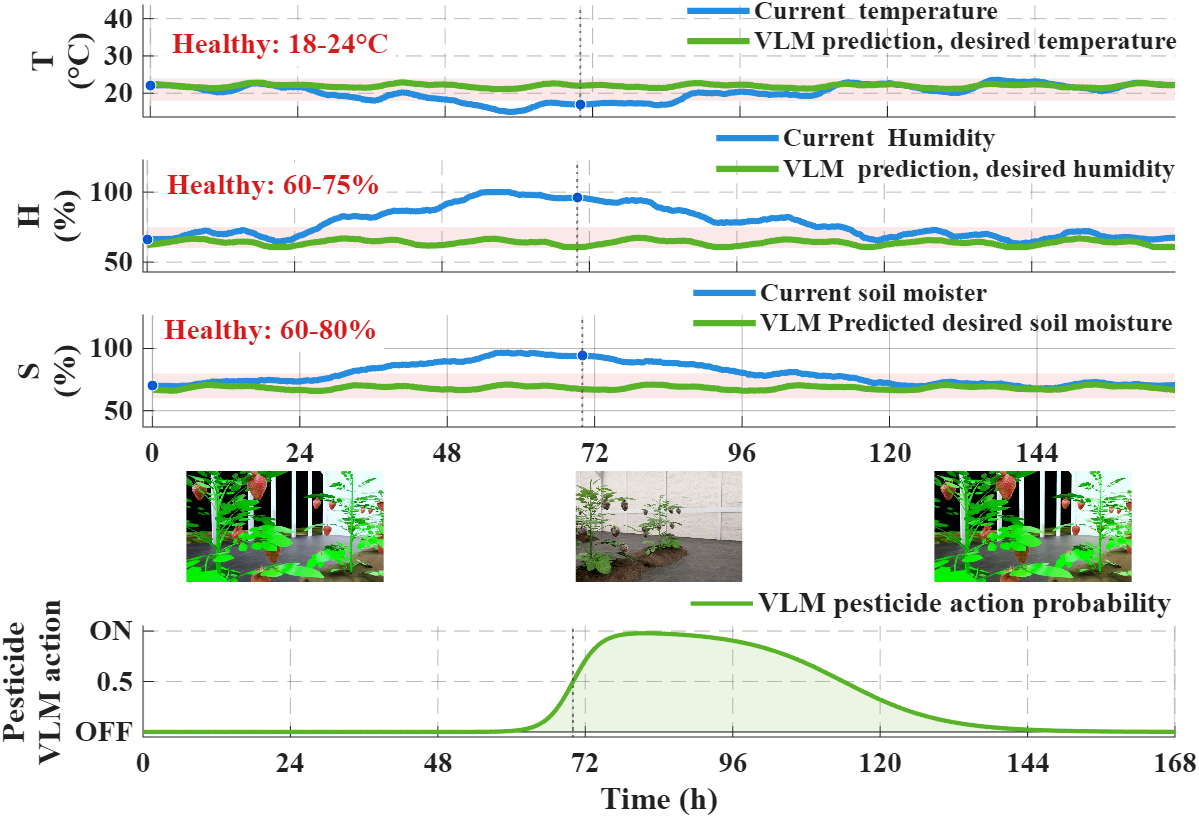}
    \caption{Representative Case~3.}
    \label{fig:vlda_zone_response_z3}
\end{subfigure}
\hfill
\begin{subfigure}[t]{0.49\textwidth}
    \centering
    \includegraphics[width=\linewidth]{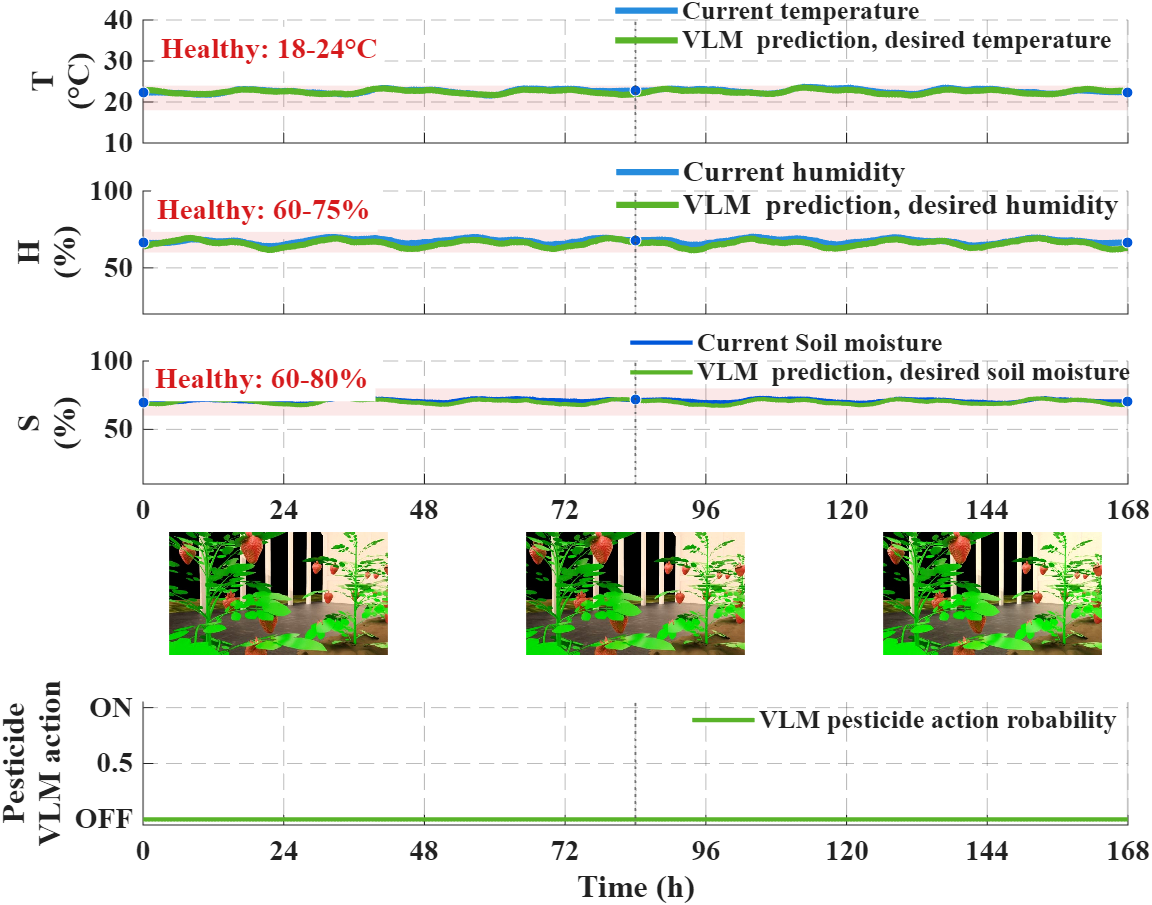}
    \caption{Representative Case~4.}
    \label{fig:vlda_zone_response_z4}
\end{subfigure}

\caption{Representative \VLDA{} decision-output examples under different
greenhouse conditions. Each panel shows current and predicted or desired
temperature, humidity, and soil-moisture values together with
pesticide-action probability and robot-view RGB context. The panels
illustrate how multimodal observations are transformed into structured
environmental and intervention proposals before validation and
deterministic execution. \rev{These historical prediction panels do not establish independent generalization; the checkpoint replay and grouped evaluation below provide the revised quantitative interpretation.}}
\label{fig:vlda_zone_response}
\end{figure}

Figure~\rev{\ref{fig:vlda_zone_response}} presents four representative
decision-output examples. Each case combines environmental-reference
prediction, pesticide-action probability, and robot-view RGB context,
thereby showing the relationship between sensor state, visual evidence,
and the resulting high-level proposal.

\rev{The historical outputs illustrate a joint supervised interface for environmental references, zone labels and intervention labels. They do not establish causal benefit from every modality or learned end-to-end control. Confidence, freshness, range, localization consistency, feasibility and reachability are retained as the original design-level validation contract; the completed readiness and simulator-command-guard tests are reported separately in Section S4.}

\subsection{GNSS-Denied Localization and Mapping}
\label{subsec:gnss_results}

\rev{The original manuscript reports three simulated greenhouse inspection missions with LiDAR, inertial and reference records. Their archived metrics are retained without treating the historical estimator or controller provenance as independently verified.} Table~\ref{tab:agrisense_navigation_metrics} reports the
localization and mission-level results computed from synchronized
ground-truth--LIO pairs.

\begin{table}[!tb]
\centering
\caption{\rev{Historical} localization and mission metrics computed from
synchronized ground-truth and \rev{LIO-labelled records. These original numerical summaries do not independently establish estimator identity or controller-pose provenance.}}
\label{tab:agrisense_navigation_metrics}
\scriptsize
\renewcommand{\arraystretch}{1.16}
\setlength{\tabcolsep}{3pt}

\begin{tabularx}{\textwidth}{
>{\raggedright\arraybackslash}p{2.4cm}
*{8}{>{\centering\arraybackslash}X}}
\toprule
\textbf{Mission} &
\textbf{Pairs} &
\textbf{Time [s]} &
\textbf{Path [m]} &
\textbf{XY RMSE [cm]} &
\textbf{XY P95 [cm]} &
\textbf{XY Max [m]} &
\textbf{3D RMSE [cm]} &
\textbf{Sync [ms]} \\
\midrule
Inspect All Zones
& 7643 & 607.95 & 151.23 & 5.34 & 9.94 & 0.18 & 5.64 & 69.93 \\

Inspect Zone~2
& 2893 & 235.54 & 50.21 & 3.43 & 7.60 & 0.10 & 3.69 & 80.78 \\

Inspect Zone~4
& 3181 & 222.94 & 81.34 & 5.16 & 8.98 & 0.11 & 5.39 & 80.38 \\
\bottomrule
\end{tabularx}
\end{table}

The all-zone mission covered 151.23~m over 607.95~s, while the Zone~2
and Zone~4 missions covered 50.21~m and 81.34~m, respectively. Across
the three missions, XY RMSE ranged from 3.43~cm to 5.34~cm and 3D RMSE
ranged from 3.69~cm to 5.64~cm. The XY 95th-percentile error remained
below 10~cm for all missions.

The instantaneous error was mostly below approximately 15~cm, although
the all-zone mission reached a maximum XY error of 0.18~m. This
distinction avoids describing 15~cm as a strict upper bound.

Figure~\rev{\ref{fig:trajectory_error_results}} compares the ground-truth and
LIO trajectories and shows the evolution of localization error.

\rev{The archived path alignment documents agreement between the recorded LIO-labelled and reference channels. It does not by itself verify the estimator identity, independence of the control pose source, or adequacy for deployment.} Plant-level target
localization, however, requires an additional close-range perception
and manipulator-registration stage and should not be inferred directly
from the global trajectory RMSE.

\begin{figure}[!tb]
\centering
\captionsetup[subfigure]{font=footnotesize,justification=centering}

\begin{subfigure}[t]{0.33\textwidth}
    \centering
    \includegraphics[width=\linewidth,height=5.8cm]
    {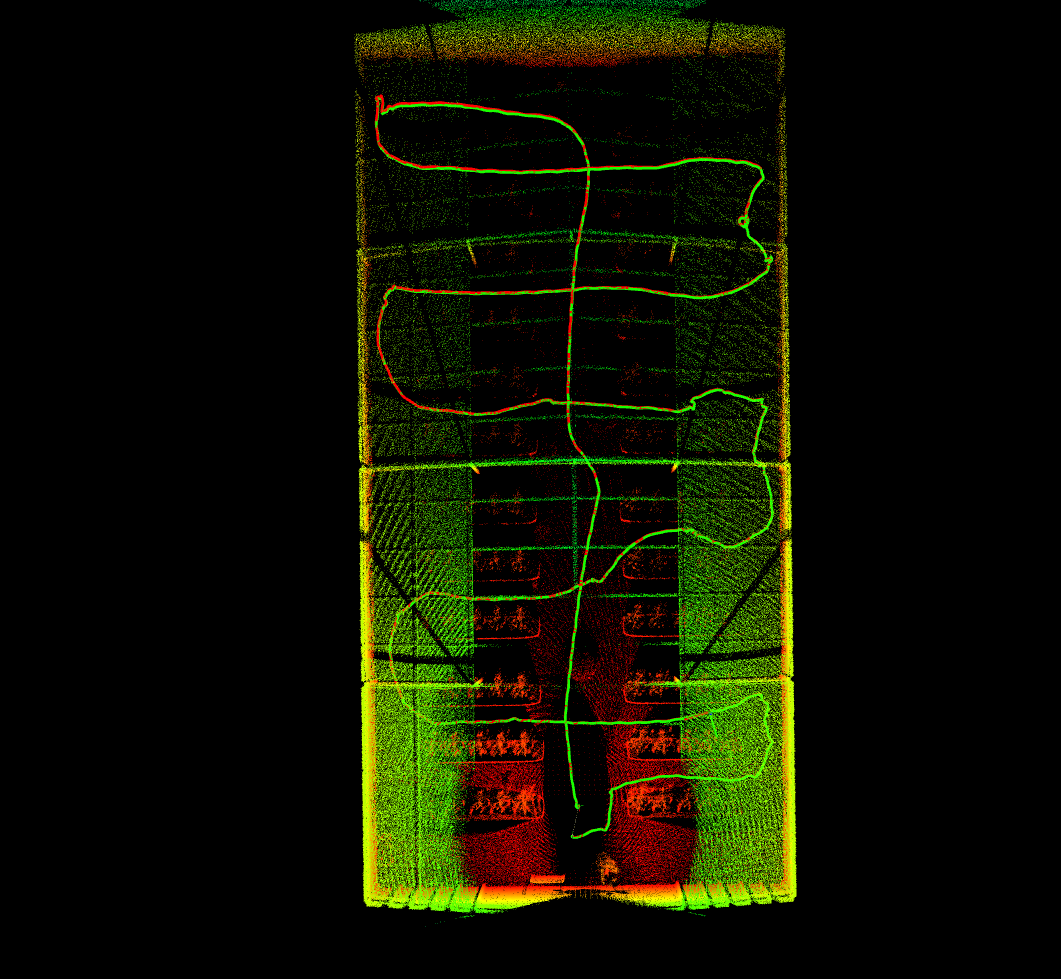}
    \caption{All-zone mission.}
    \label{fig:pc_allzones}
\end{subfigure}
\hfill
\begin{subfigure}[t]{0.32\textwidth}
    \centering
    \includegraphics[width=\linewidth]
    {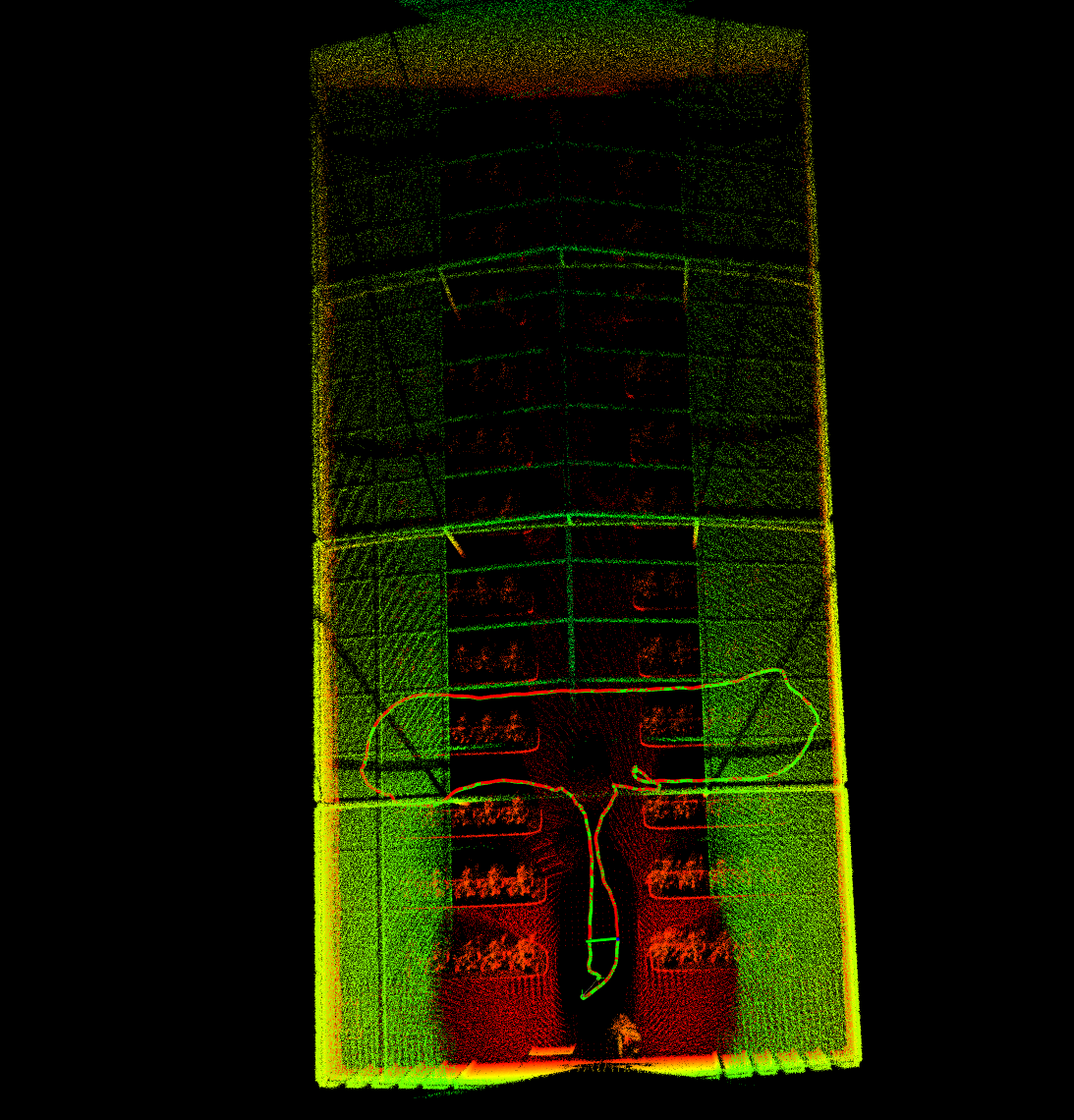}
    \caption{Zone~2 mission.}
    \label{fig:pc_zone2}
\end{subfigure}
\hfill
\begin{subfigure}[t]{0.32\textwidth}
    \centering
    \includegraphics[width=\linewidth]
    {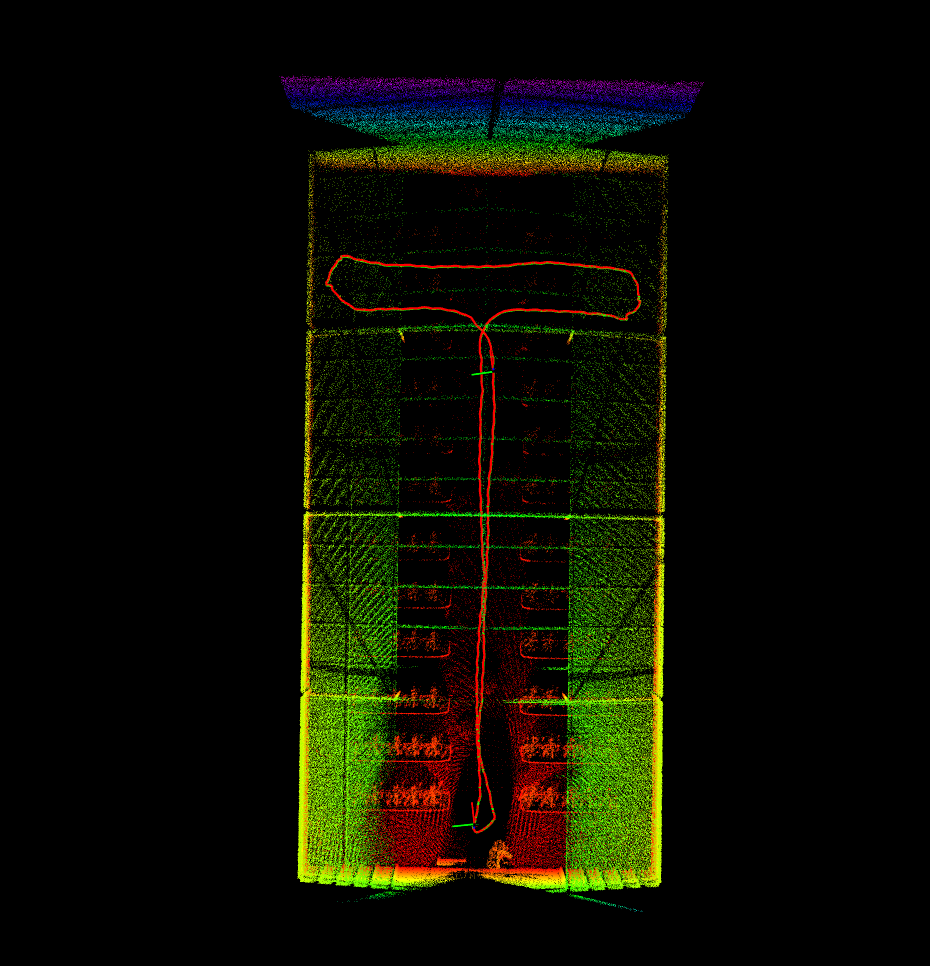}
    \caption{Zone~4 mission.}
    \label{fig:pc_zone4}
\end{subfigure}

\caption{LiDAR point-cloud reconstructions \rev{retained from the three historical greenhouse mission records.} The maps preserve the main
aisle, cultivation-bed, and greenhouse structural geometry over
missions with different spatial extents.}
\label{fig:pointcloud_comparison}
\end{figure}

\begin{figure}[!tb]
\centering

\begin{subfigure}[t]{0.49\textwidth}
    \centering
    \includegraphics[width=\linewidth]
    {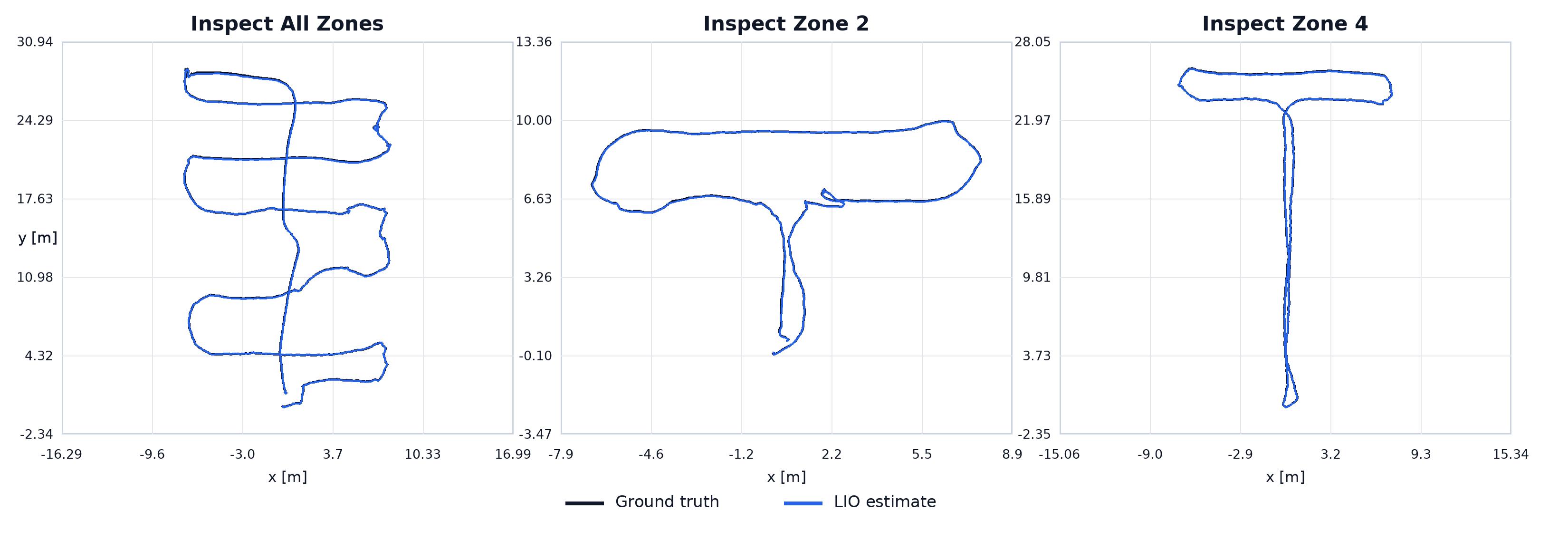}
    \caption{Ground-truth and LIO path alignment.}
    \label{fig:gt_lio_path_tracking}
\end{subfigure}
\hfill
\begin{subfigure}[t]{0.49\textwidth}
    \centering
    \includegraphics[width=\linewidth]
    {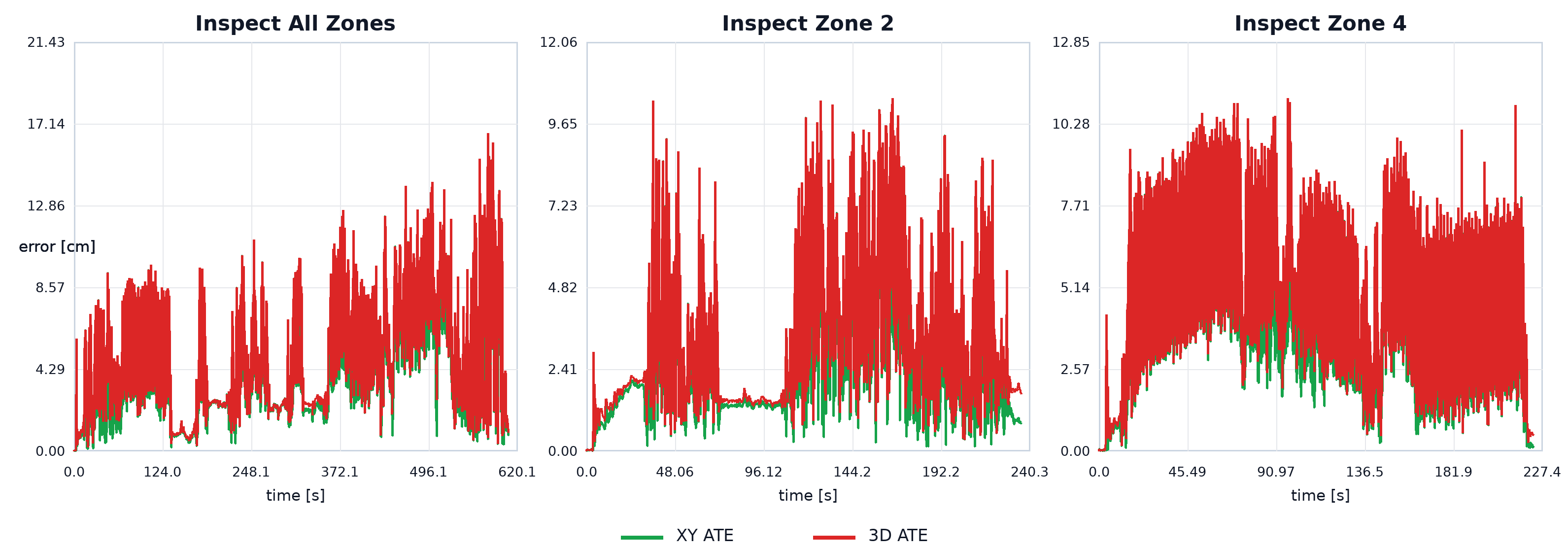}
    \caption{Localization-error evolution.}
    \label{fig:localization_error_evolution}
\end{subfigure}

\caption{\rev{Historical logged trajectories and error traces for the all-zone, Zone~2 and Zone~4 missions. Original plot labels are retained, but the estimator provenance of these records is not independently established. They are not the new offline Point-LIO experiment or proof of LIO-driven robot navigation.}}
\label{fig:trajectory_error_results}
\end{figure}

Figure~\ref{fig:pointcloud_comparison} presents the reconstructed
point-cloud maps. Despite differences in mission length and spatial
coverage, the maps preserve the principal greenhouse structure,
including aisle geometry, cultivation-bed boundaries, and structural
elements.

\subsection{Four-Zone Environmental-State Variation}
\label{subsec:zone_environment_results}

\begin{figure}[!tb]
\centering
\includegraphics[width=0.98\textwidth]
{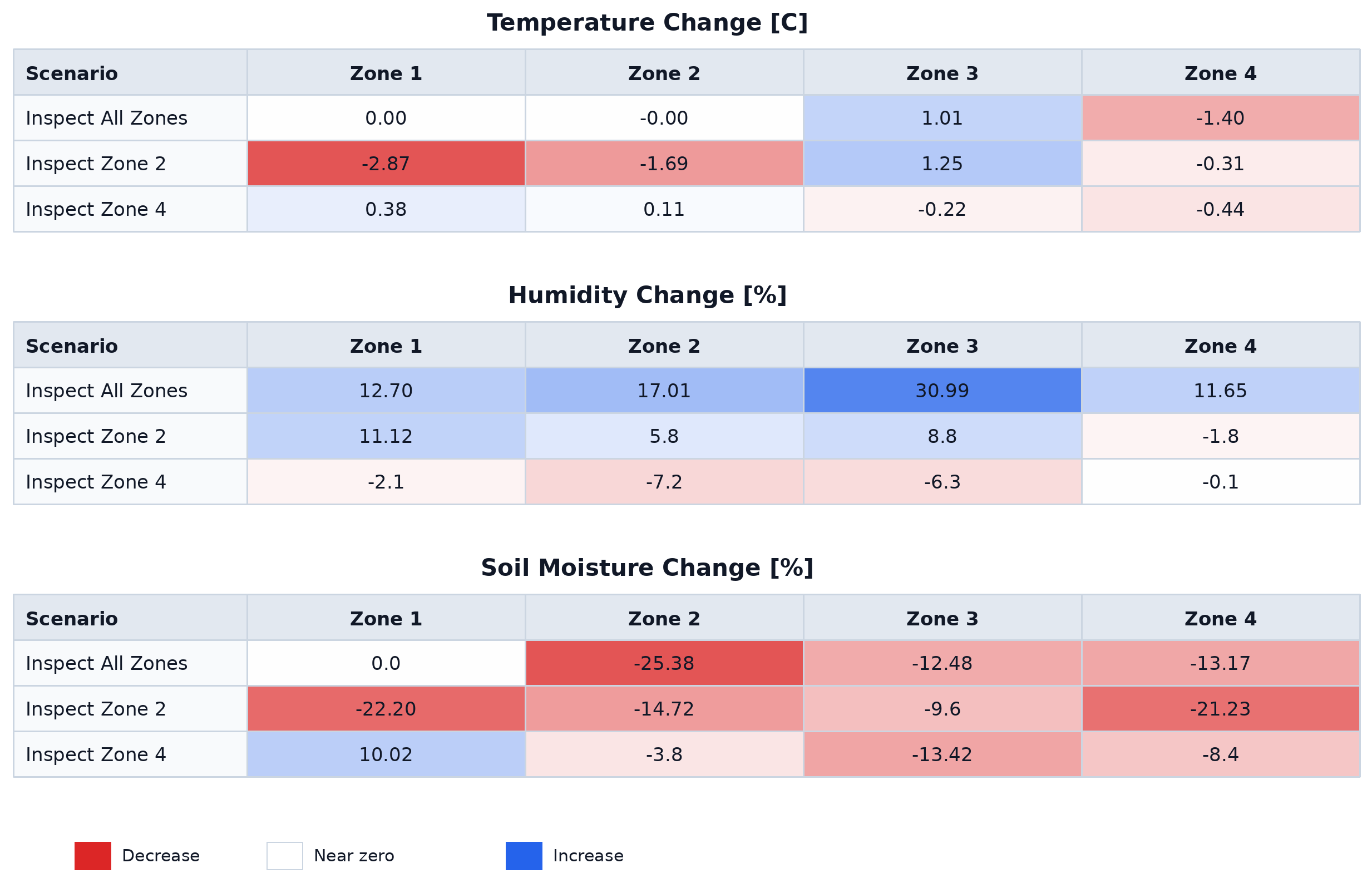}
\caption{End-to-start environmental-state variation across the
all-zone, Zone~2, and Zone~4 missions. Positive and negative values
indicate increases and decreases, respectively, in temperature,
humidity, and soil moisture for each management zone. \rev{Original historical environmental-state visualization recovered from the reference project history. These state changes do not alone measure closed-loop controller quality.}}
\label{fig:environment_delta_map}
\end{figure}

The greenhouse-state logs contain temperature, humidity, and
soil-moisture measurements for all four management zones. Figure~
\rev{\ref{fig:environment_delta_map}} summarizes the end-to-start change in
each variable during each mission.

\begin{table}[!tb]
\centering
\caption{Four-zone greenhouse-state statistics across the three
inspection missions. Here, \(\Delta\) denotes the final value minus the
initial value.}
\label{tab:agrisense_greenhouse_state_statistics}
\scriptsize
\renewcommand{\arraystretch}{1.15}
\setlength{\tabcolsep}{3pt}

\begin{tabularx}{\textwidth}{
>{\raggedright\arraybackslash}p{2.15cm}
>{\raggedright\arraybackslash}p{2.15cm}
*{6}{>{\centering\arraybackslash}X}}
\toprule
\textbf{Mission} &
\textbf{Variable} &
\textbf{Zone~1 \(\Delta\)} &
\textbf{Zone~2 \(\Delta\)} &
\textbf{Zone~3 \(\Delta\)} &
\textbf{Zone~4 \(\Delta\)} &
\textbf{Min} &
\textbf{Max} \\
\midrule

Inspect All Zones
& Temperature [$^\circ$C]
& 0.00 & -0.00 & 1.01 & -1.40 & 20.27 & 25.38 \\

Inspect All Zones
& Humidity [\%]
& 12.70 & 17.01 & 30.99 & 11.65 & 40.68 & 81.00 \\

Inspect All Zones
& Soil moisture [\%]
& 0.02 & -25.38 & -12.48 & -13.17 & 62.00 & 92.18 \\

\midrule

Inspect Zone~2
& Temperature [$^\circ$C]
& -2.87 & -1.69 & 1.25 & -0.31 & 20.28 & 25.37 \\

Inspect Zone~2
& Humidity [\%]
& 11.12 & 5.79 & 8.79 & -1.84 & 58.00 & 81.00 \\

Inspect Zone~2
& Soil moisture [\%]
& -22.20 & -14.72 & -9.58 & -21.23 & 62.00 & 92.18 \\

\midrule

Inspect Zone~4
& Temperature [$^\circ$C]
& 0.38 & 0.11 & -0.22 & -0.44 & 20.27 & 25.38 \\

Inspect Zone~4
& Humidity [\%]
& -2.09 & -7.22 & -6.30 & -0.11 & 31.15 & 81.00 \\

Inspect Zone~4
& Soil moisture [\%]
& 10.02 & -3.75 & -13.42 & -8.42 & 62.00 & 91.69 \\

\bottomrule
\end{tabularx}
\end{table}

Temperature variation is comparatively limited, whereas humidity and
soil moisture exhibit stronger zone-dependent changes. This supports
the use of spatially associated measurements rather than relying only
on greenhouse-wide averages.

These statistics characterize the environmental conditions recorded
during the missions. They should not be interpreted as complete PI/PID
performance metrics because settling time, rise time, overshoot,
steady-state error, and disturbance-rejection experiments are not
reported here. The four representative response plots in
Fig.~\rev{\ref{fig:vlda_zone_response}} instead illustrate the
decision-to-control interface qualitatively.

\subsection{Spatially Grounded Digital-Twin Views}
\label{subsec:local_digital_twin_results}

Figure~\rev{3 (main manuscript)} presents complementary views of the
greenhouse digital twin. The top-down view shows greenhouse geometry,
cultivation beds, management regions, and robot position, while the
interior views show the crop rows, navigation corridors, sensing
infrastructure, and local operating context.

The views demonstrate how robot observations and environmental
measurements can be associated with a common local map. They are
qualitative digital-twin visualizations and do not independently
constitute localization-accuracy evidence.

\subsection{Plant-Level Inspection and Intervention Demonstration}
\label{subsec:plant_level_results}

The plant-level workflow begins with zone navigation and close-range
inspection. Once the robot reaches a validated inspection pose, a
scenario-defined or separately annotated target is checked for
localization consistency, base stability, and manipulator reachability.
\rev{In the original design, a bounded spray primitive is dispatched only after the intervention proposal and validation conditions pass. The retained replay does not establish that the learned model passed this gate; the later simulator tests use separately supplied proposals.}

\begin{figure}[!tb]
\centering
\includegraphics[width=0.98\textwidth]{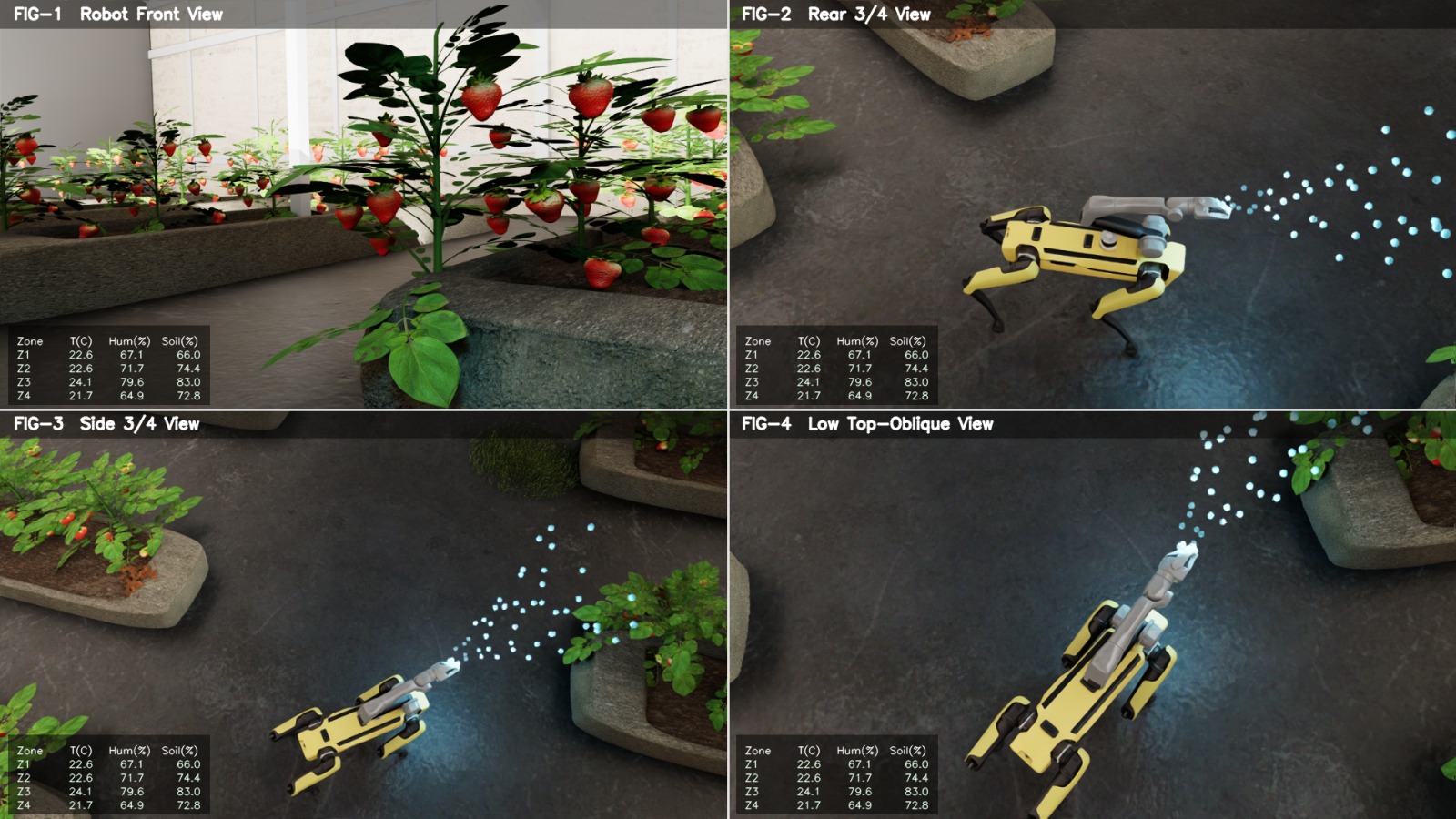}
\caption{Qualitative plant-level inspection and intervention
demonstration in \rev{the \AgriSense{} virtual greenhouse}. The four views show crop
inspection, robot approach, manipulator alignment, and execution of a
bounded directed-spray primitive. The environmental overlay links the
action with its management-zone context. \rev{Historical simulated intervention illustration; scenario-defined targeting does not establish learned diagnosis or biological treatment efficacy.}}
\label{fig:plant_level_intervention}
\end{figure}

Figure~\rev{\ref{fig:plant_level_intervention}} presents four digital-twin
views of the intervention sequence. The figure illustrates inspection
context, robot approach, manipulator alignment, and directed spray
execution. It should be interpreted as a qualitative demonstration of
system integration rather than a quantitative evaluation of spraying
accuracy or biological effectiveness.

\begin{table}[!t]
\centering
\caption{Comparison of zone-level regulation and plant-level
intervention.}
\label{tab:zone_vs_plant}
\footnotesize
\renewcommand{\arraystretch}{1.12}

\begin{tabularx}{\columnwidth}{
>{\raggedright\arraybackslash}p{1.85cm}
>{\raggedright\arraybackslash}X
>{\raggedright\arraybackslash}X}
\toprule
\textbf{Criterion} &
\textbf{Zone-level regulation} &
\textbf{Plant-level intervention} \\
\midrule

Spatial scope
& Entire management zone
& Selected plant or local region \\

Primary actuator
& Fixed greenhouse actuator
& Mobile manipulator and spray end-effector \\

Spatial requirement
& Zone association
& Zone, inspection pose, and local target \\

Decision context
& Environmental measurements and references
& Image, state, pose, and reachability \\

Available evidence
& Recorded zone states and proposed references
& Qualitative simulated command execution \\

Main uncertainty
& Environmental dynamics and controller tuning
& Target localization, reachability, and spray deposition \\

\bottomrule
\end{tabularx}
\end{table}

The table intentionally avoids claiming reduced chemical use. Such a
claim requires direct comparison of spray volume, deposition coverage,
off-target application, and treatment effectiveness.

\subsection{\AgriSense{} Application and Remote Supervision}
\label{subsec:remote_supervision_results}

The \AgriSense{} application functions as a non-real-time supervision
layer. It visualizes edge-generated telemetry, greenhouse measurements,
perception outputs, local maps, mission coverage, diagnostics, and
validated action states. \rev{Operator inputs are intended as high-level requests within the original validation architecture. The screenshots alone do not verify enforcement against every bypass or failure mode.}

\begin{figure}[p]
\centering
\captionsetup[subfigure]{font=footnotesize,justification=centering}

\begin{subfigure}[t]{0.32\textwidth}
\centering
\includegraphics[width=\linewidth]{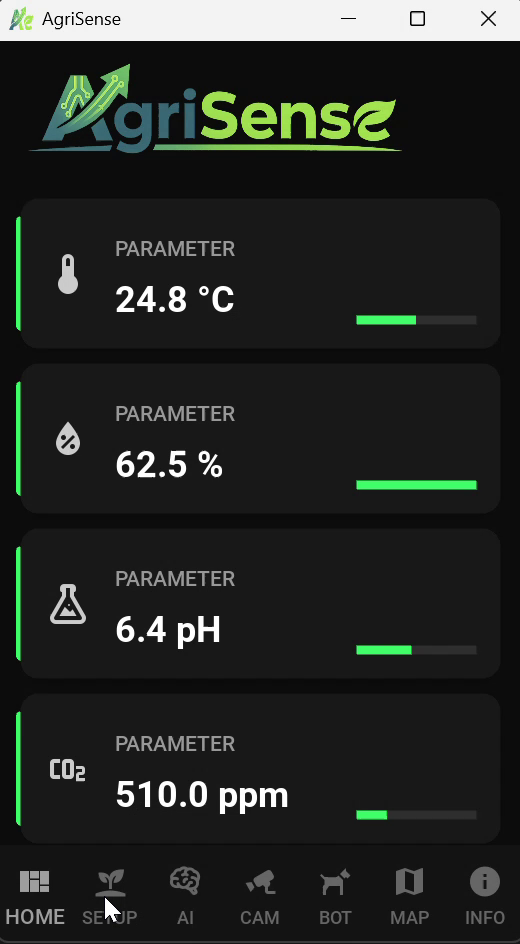}
\caption{Mission dashboard.}
\end{subfigure}
\hfill
\begin{subfigure}[t]{0.32\textwidth}
\centering
\includegraphics[width=\linewidth]{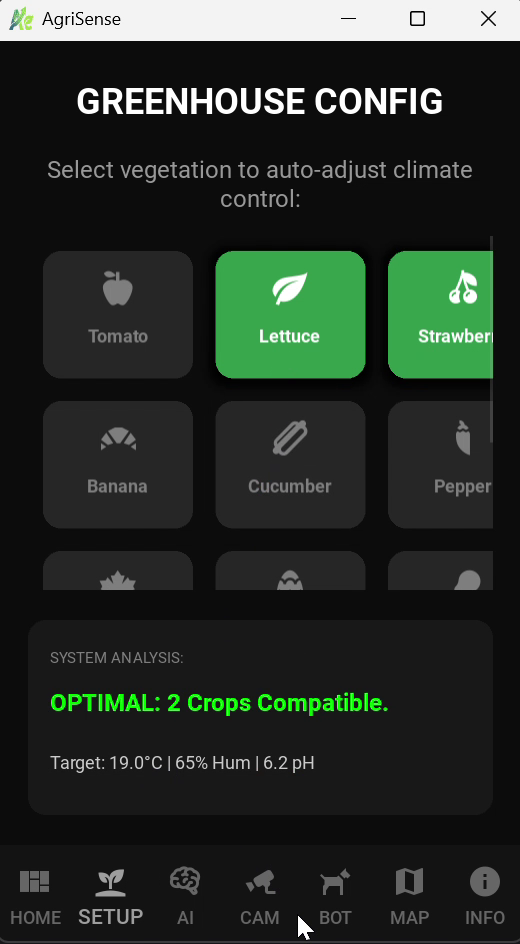}
\caption{Climate regulation.}
\end{subfigure}
\hfill
\begin{subfigure}[t]{0.32\textwidth}
\centering
\includegraphics[width=\linewidth]{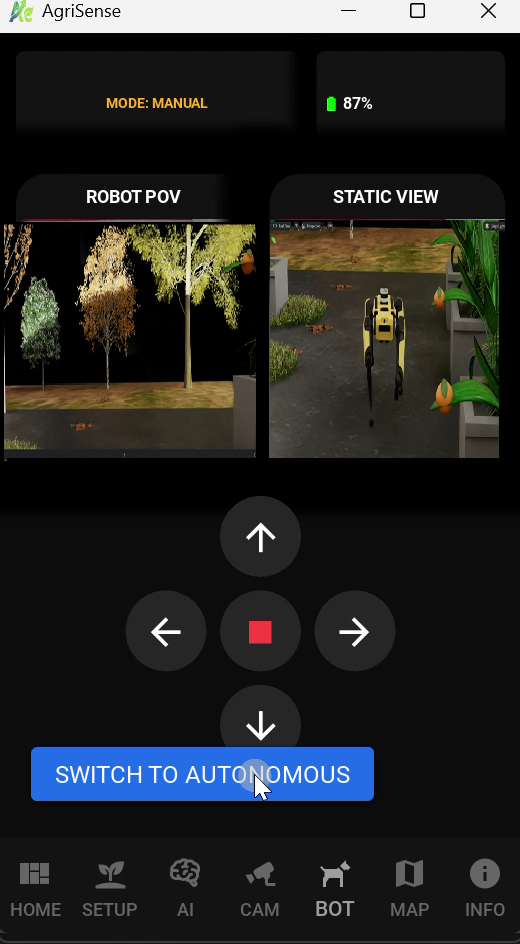}
\caption{Perception output.}
\end{subfigure}

\vspace{0.5em}

\begin{subfigure}[t]{0.32\textwidth}
\centering
\includegraphics[width=\linewidth]{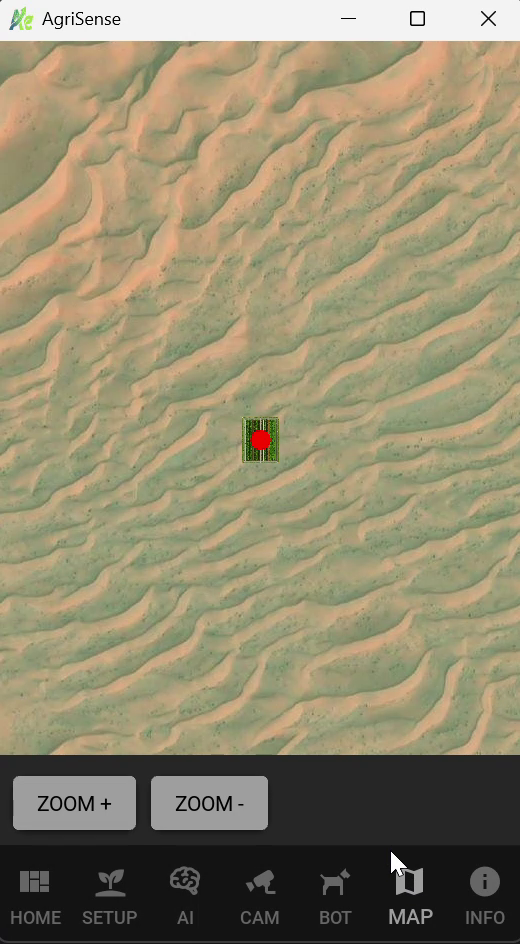}
\caption{Local navigation map.}
\end{subfigure}
\hfill
\begin{subfigure}[t]{0.32\textwidth}
\centering
\includegraphics[width=\linewidth]{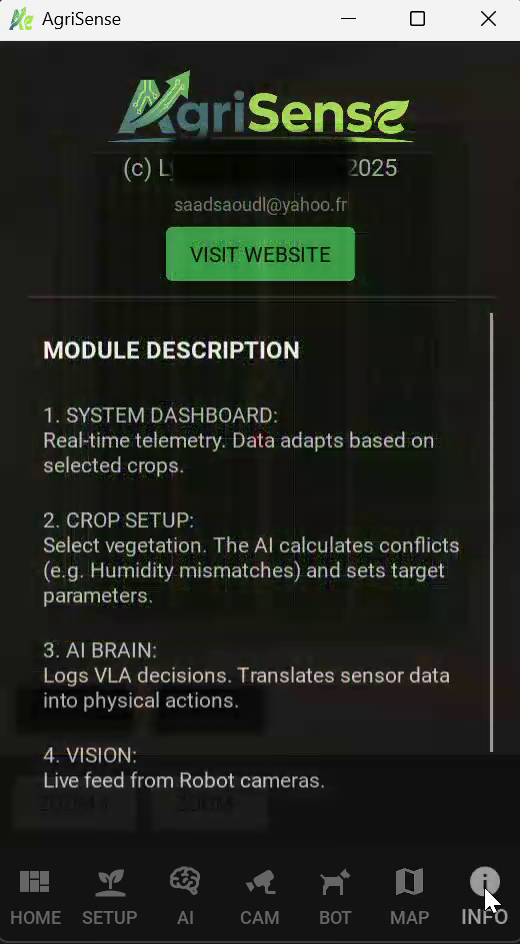}
\caption{\rev{Application information.}}
\end{subfigure}
\hfill
\begin{subfigure}[t]{0.32\textwidth}
\centering
\includegraphics[width=\linewidth]{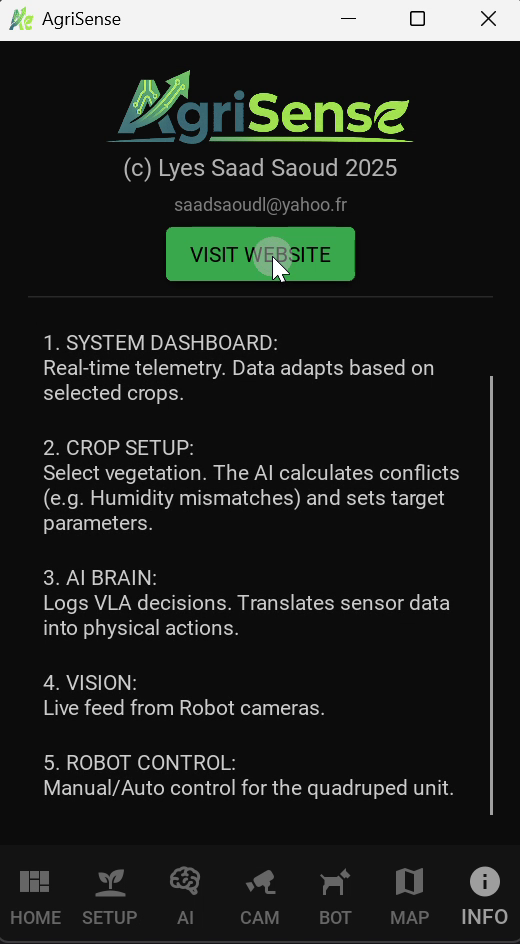}
\caption{\rev{Module descriptions.}}
\end{subfigure}

\caption{\AgriSense{} application and remote-supervision interface. The
selected views show mission telemetry, environmental-control status,
perception output, GNSS-denied mapping, and \rev{application and module information displayed by the supervision interface.} \rev{The original six screenshots are retained unchanged as qualitative supervision records. Their mapping labels do not independently establish estimator provenance.}}
\label{fig:app_ui}
\end{figure}

For readability, Fig.~\rev{\ref{fig:app_ui}} includes
six representative interface views.

\subsection{Qualitative Mission Replay}
\label{subsec:qualitative_demo_replay}

\begin{figure}[!tb]
\centering
\captionsetup[subfigure]{font=footnotesize,justification=centering}

\begin{subfigure}[t]{0.24\textwidth}
\centering
\includegraphics[width=\linewidth]{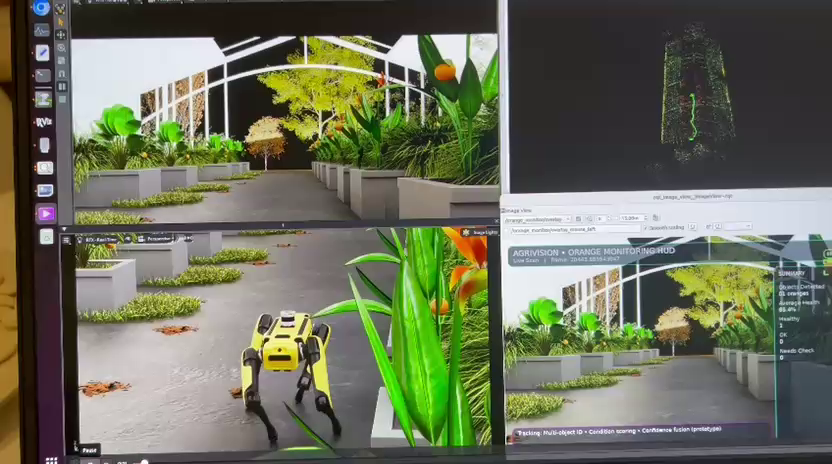}
\caption{Mission start.}
\end{subfigure}
\hfill
\begin{subfigure}[t]{0.24\textwidth}
\centering
\includegraphics[width=\linewidth]{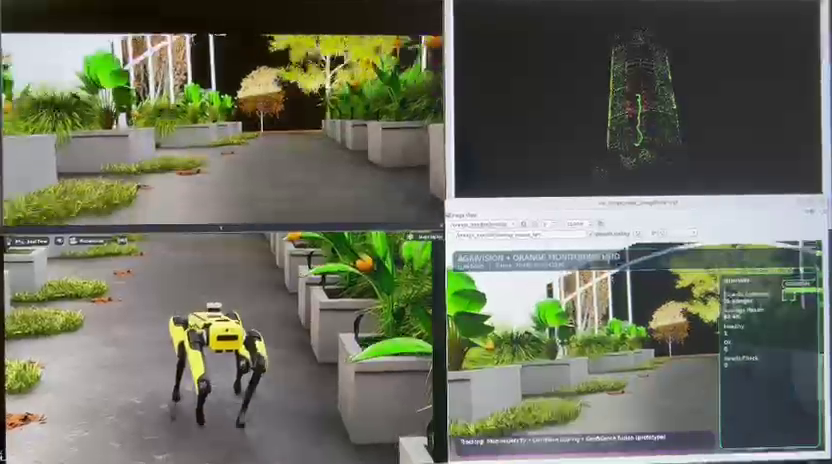}
\caption{Aisle traversal.}
\end{subfigure}
\hfill
\begin{subfigure}[t]{0.24\textwidth}
\centering
\includegraphics[width=\linewidth]{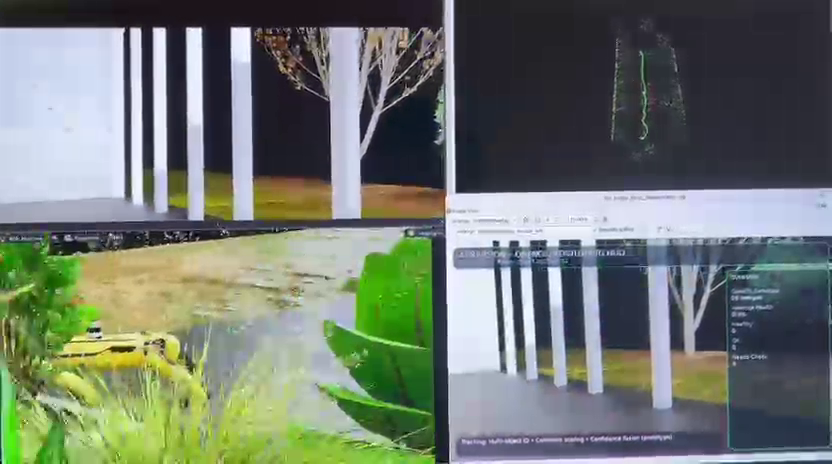}
\caption{Turning maneuver.}
\end{subfigure}
\hfill
\begin{subfigure}[t]{0.24\textwidth}
\centering
\includegraphics[width=\linewidth]{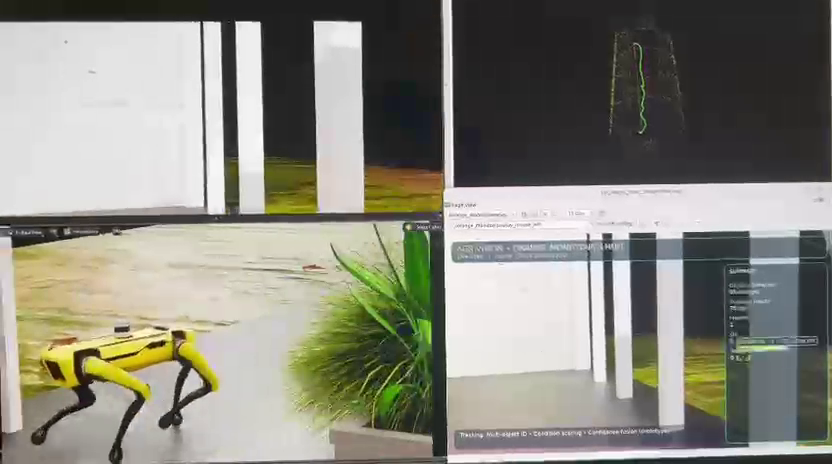}
\caption{Row alignment.}
\end{subfigure}

\vspace{0.5em}

\begin{subfigure}[t]{0.24\textwidth}
\centering
\includegraphics[width=\linewidth]{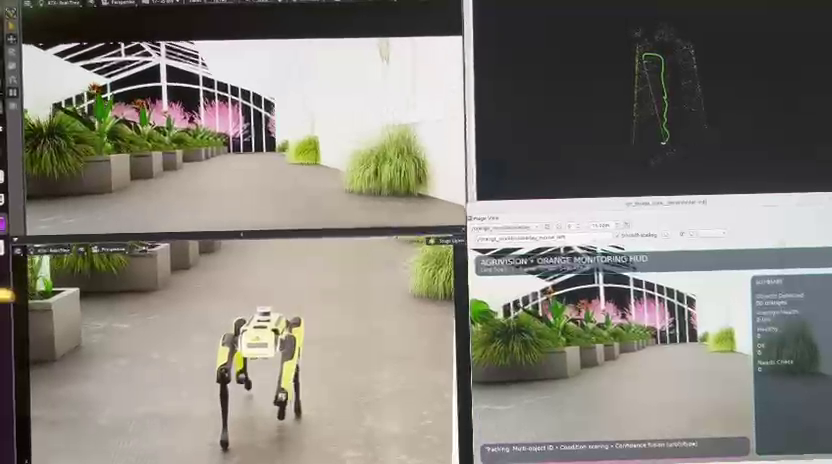}
\caption{Crop inspection.}
\end{subfigure}
\hfill
\begin{subfigure}[t]{0.24\textwidth}
\centering
\includegraphics[width=\linewidth]{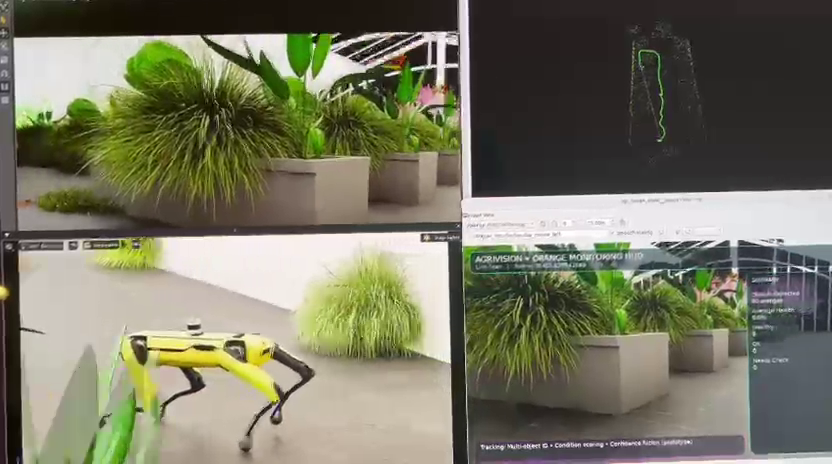}
\caption{Obstacle-aware motion.}
\end{subfigure}
\hfill
\begin{subfigure}[t]{0.24\textwidth}
\centering
\includegraphics[width=\linewidth]{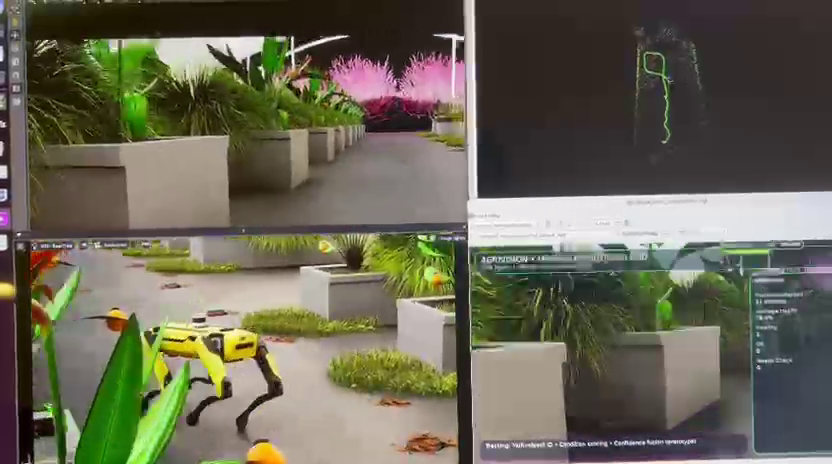}
\caption{Return traversal.}
\end{subfigure}
\hfill
\begin{subfigure}[t]{0.24\textwidth}
\centering
\includegraphics[width=\linewidth]{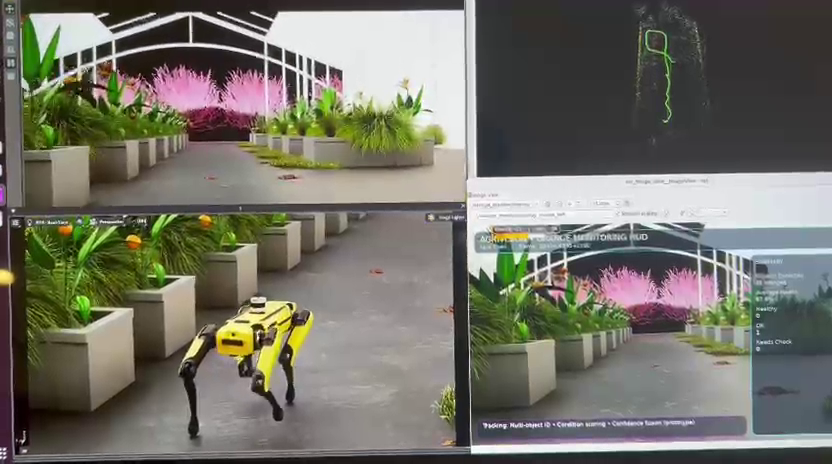}
\caption{Mission completion.}
\end{subfigure}

\caption{Representative frames from the \AgriSense{} greenhouse mission
replay. The sequence illustrates aisle navigation, turning,
row alignment, close-range inspection, obstacle-aware motion, and
mission completion. \rev{The frames do not independently verify which estimator or pose stream drove the historical navigation controller.}}
\label{fig:demo_collage}
\end{figure}

A continuous greenhouse inspection mission was recorded and replayed to
complement the numerical localization results. Figure~
\ref{fig:demo_collage} shows eight representative frames covering
mission initialization, aisle traversal, turning, row alignment,
inspection, obstacle-aware motion, return, and completion.

The replay provides qualitative evidence of the robot behavior associated
with the recorded trajectories. It does not replace the synchronized
trajectory metrics reported in Table~
\ref{tab:agrisense_navigation_metrics}.

\subsection{Summary of Experimental Evidence}
\label{subsec:results_summary}

\rev{The historical evaluation records three kinds of output: joint supervised environmental-reference, zone and intervention predictions; centimeter-scale agreement in three archived LIO-labelled/reference trajectory records; and qualitative digital-twin views of monitoring, navigation, supervision, inspection and manipulation. These are component or documentary records, not proof of a validated learned end-to-end autonomous pipeline. Later verified component evaluations are retained separately in Sections S1--S5.}

The current evidence remains simulation-centric. Physical-greenhouse
evaluation, biological disease validation, closed-loop PI/PID transient
metrics, quantitative target-localization accuracy, manipulator
reachability success rate, and spray-deposition measurements remain
necessary before claims of field readiness or treatment effectiveness
can be made.

\subsection{\rev{Original supplementary traces and metric definitions}}
\label{sec:s-original-traces}

The supplementary material provides the mission-level audit trail supporting
the quantitative and qualitative results reported in
the main Results and Discussion section. It retains the synchronized localization-error
histories, cumulative localization metrics, and raw and rolling-median
greenhouse-state traces for the all-zone, Zone~2, and Zone~4 inspection
missions. The supplementary files are intended to expose the underlying
recorded signals rather than only the mission-level summary statistics
reported in the main manuscript.

\rev{The original design requires LiDAR--inertial spatial estimation without privileged reference pose. The retained historical files do not independently verify that this requirement held for every estimator and controller component. Their ground-truth channels are therefore treated as evaluator records, without inferring controller independence. The later offline Point-LIO and simulator-pose robot tests have separate protocols in Section S4.}

For each temporally synchronized ground-truth--LIO pair \(n\), the planar
localization error is

\begin{equation}
e_{xy,n}
=
\sqrt{
\left(x^{\mathrm{LIO}}_n-x^{\mathrm{GT}}_n\right)^2
+
\left(y^{\mathrm{LIO}}_n-y^{\mathrm{GT}}_n\right)^2
},
\label{eq:appendix_xy_error}
\end{equation}

and the cumulative planar RMSE through sample \(N\) is

\begin{equation}
\operatorname{RMSE}_{xy}(N)
=
\sqrt{
\frac{1}{N}
\sum_{n=1}^{N}e_{xy,n}^{2}
}.
\label{eq:appendix_cumulative_rmse}
\end{equation}

The same synchronization and metric definitions are used for all three
missions. When estimated and reference trajectories already share the same
simulation coordinate frame, no additional rigid alignment is introduced
before computing these errors. \rev{These historical errors use their recorded simulation frame and are distinct from the rigid-fit ATE in the later offline Point-LIO routes.}

The environmental plots retain the original recorded temperature, relative
humidity, and soil-moisture signals for all four management zones. The
rolling-median traces are provided only as visualization aids for identifying
slow trends in the presence of short-term fluctuations. All numerical
environmental statistics reported in the main manuscript are derived from the
recorded raw signals, not from the smoothed traces. The supplementary
environmental plots should therefore be interpreted as evidence of
zone-dependent state variation and data-recording consistency rather than as
a complete PI/PID transient-response evaluation.


\renewcommand{\supppanel}[2]{%
\begin{minipage}[t]{0.485\textwidth}
    \centering
    \includegraphics[width=\linewidth]{Figures/#1}
    \vspace{0.15em}
    
    {\footnotesize #2\par}
\end{minipage}%
}

\subsubsection{Supplementary Methodological Traceability}
\label{app:methodological_traceability}

The supplementary results should be interpreted together with the
methodological definitions provided in the main paper. In the localization
experiments, high-rate IMU measurements propagate robot motion between LiDAR
updates, while LiDAR geometric constraints correct accumulated inertial drift
through scan-to-map registration. \rev{In the design specification, the corrected state updates the local pose and map for management-zone association and navigation; the historical logs alone do not establish that complete causal data path.} \AgriSense{} does not claim a new LiDAR--inertial odometry
algorithm; the localization stack provides the spatial estimate required by
the integrated greenhouse decision-to-action framework.

\rev{The synchronized historical records pair LIO-labelled channels with simulator reference channels. They preserve the numerical error calculations, but do not independently establish the estimator implementation or the navigation controller input. These plots must not be presented as verification that privileged simulator pose was excluded from robot control.}

The multimodal dataset used for \VLDA{} development is also distinct from the
mission-level localization logs presented below. The base robot-view RGB
observations were generated in the Isaac Sim greenhouse together with their
management-zone and environmental context. Disease-like strawberry
appearances were introduced afterward through controlled image-to-image
augmentation and were not physically modeled as plant pathology inside Isaac
Sim. \rev{The later record-level label reconciliation in Section S1 supersedes the assumption that all prompt and decision labels were inherited unchanged. The controlled image edits are a separate model-development process and do not themselves modify the recorded historical localization or environmental-state histories.}

These distinctions are included to separate three different forms of
evidence used in the paper: localization evaluation against simulator ground
truth, recorded greenhouse-state variation, and controlled visual
augmentation for multimodal model development.

\subsubsection{Dashboard Outputs and Reproducibility Files}
\label{app:dashboard_reproducibility}

The supplementary package contains the numerical logs and interactive
dashboard artifacts used to generate and review the reported results. The
HTML dashboards are supplied separately because their interactive behavior
cannot be preserved in the static manuscript PDF. They provide
non-real-time mission replay and diagnostic review and are not part of the
safety-critical localization, navigation, environmental-control, or
manipulator-execution loops.

\begin{table}[!htbp]
\centering
\caption{Supplementary files provided for mission-level result reproduction.}
\label{tab:appendix_reproducibility_files}
\scriptsize
\renewcommand{\arraystretch}{1.10}
\setlength{\tabcolsep}{3pt}

\begin{tabularx}{\columnwidth}{
>{\raggedright\arraybackslash}p{1.35cm}
>{\raggedright\arraybackslash}X}
\toprule
\textbf{Mission} &
\textbf{Directory and supplied artifacts} \\
\midrule

All zones
&
\path{inspect_allzones/}: \path{synced_errors.csv},
\path{greenhouse_state.csv}, \path{metrics_summary.csv}, and
\path{live_dashboard.html} \\

Zone~2
&
\path{inspect_zone2/}: \path{synced_errors.csv},
\path{greenhouse_state.csv}, \path{metrics_summary.csv}, and
\path{live_dashboard.html} \\

Zone~4
&
\path{inspect_zone4/}: \path{synced_errors.csv},
\path{greenhouse_state.csv}, \path{metrics_summary.csv}, and
\path{live_dashboard.html} \\

\bottomrule
\end{tabularx}
\end{table}

For each mission, \path{synced_errors.csv} contains the temporally
synchronized ground-truth and LIO positions, instantaneous position errors,
cumulative RMSE traces, and synchronization information.
\path{greenhouse_state.csv} contains the recorded four-zone temperature,
relative-humidity, and soil-moisture histories.
\path{metrics_summary.csv} stores the corresponding mission-level evaluator
outputs and configuration metadata. The interactive
\path{live_dashboard.html} file provides non-real-time inspection of the same
recorded telemetry.

The synchronization record is important because the simulator ground-truth
and localization streams can operate at different sampling rates. Errors are
therefore evaluated from matched timestamp pairs rather than from direct
row-by-row comparison of independently sampled files.

\subsubsection{All-Zone Inspection Results}
\label{app:allzones_results}

The all-zone mission provides the longest and most spatially extensive
\rev{historical trajectory record.} The robot traversed all four management regions
over a path length of \(151.23~\mathrm{m}\), producing 7,643 synchronized
ground-truth--LIO pairs over \(607.95~\mathrm{s}\). The mission-level
XY RMSE was \(5.34~\mathrm{cm}\), the 95th-percentile planar error was
\(9.94~\mathrm{cm}\), and the 3D position RMSE was \(5.64~\mathrm{cm}\).

\begin{figure}[p]
\centering
\begin{subfigure}[t]{0.485\textwidth}
    \centering
    \includegraphics[width=\linewidth,height=6.0cm,keepaspectratio]
    {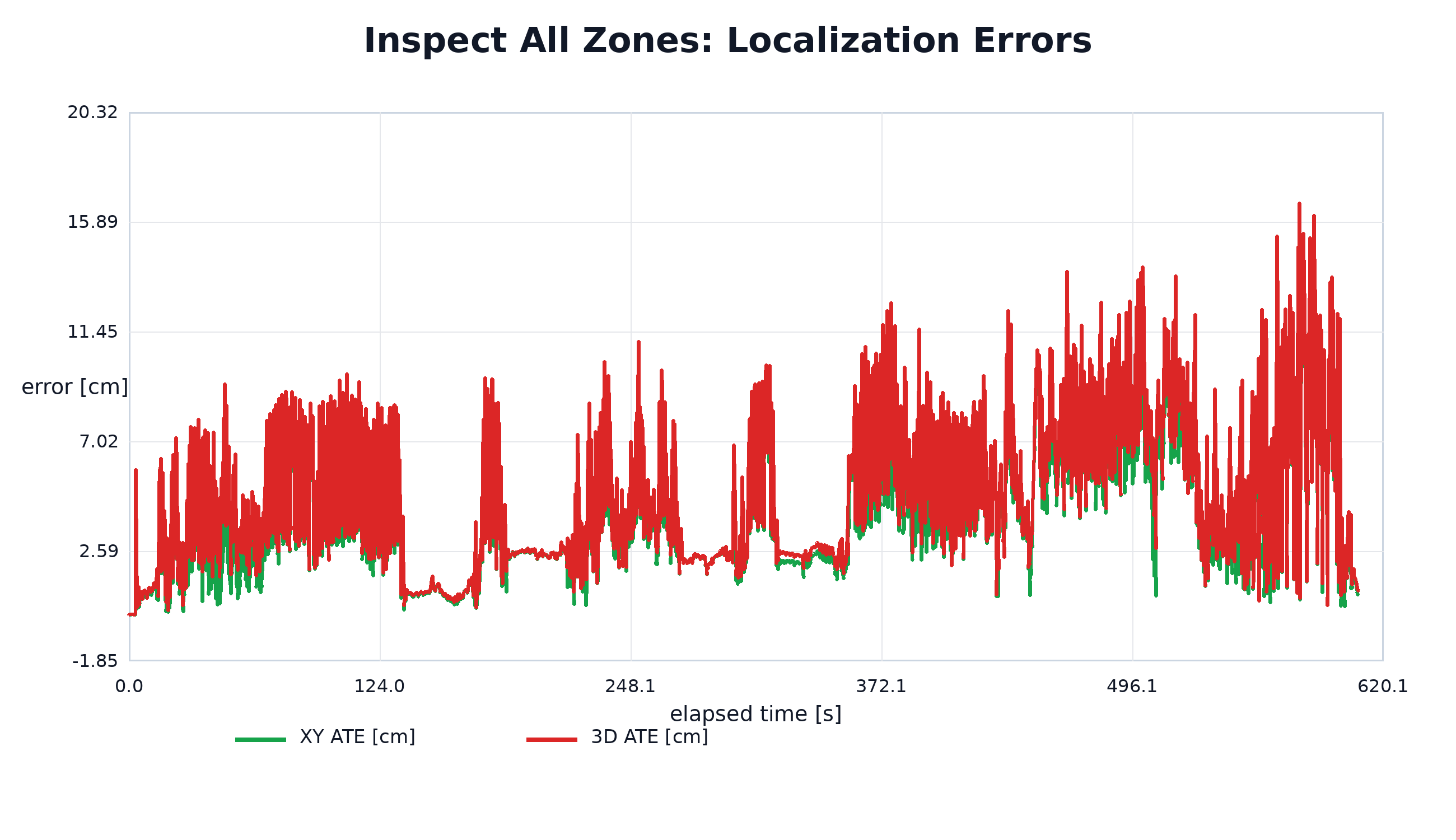}
    \caption{Instantaneous planar localization error.}
\end{subfigure}\hfill
\begin{subfigure}[t]{0.485\textwidth}
    \centering
    \includegraphics[width=\linewidth,height=6.0cm,keepaspectratio]
    {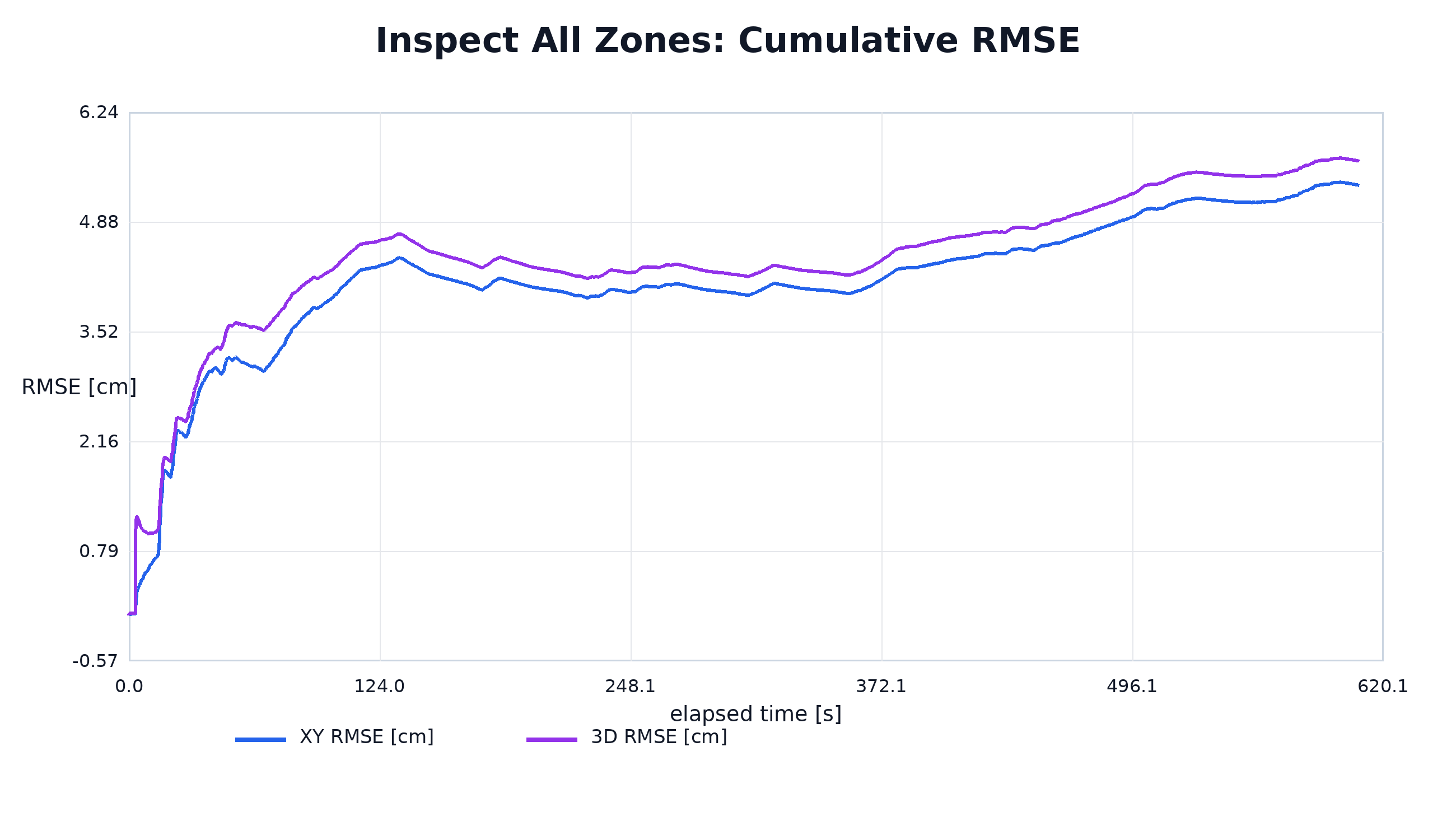}
    \caption{Cumulative planar localization RMSE.}
\end{subfigure}\par\medskip
\begin{subfigure}[t]{0.485\textwidth}
    \centering
    \includegraphics[width=\linewidth,height=6.0cm,keepaspectratio]
    {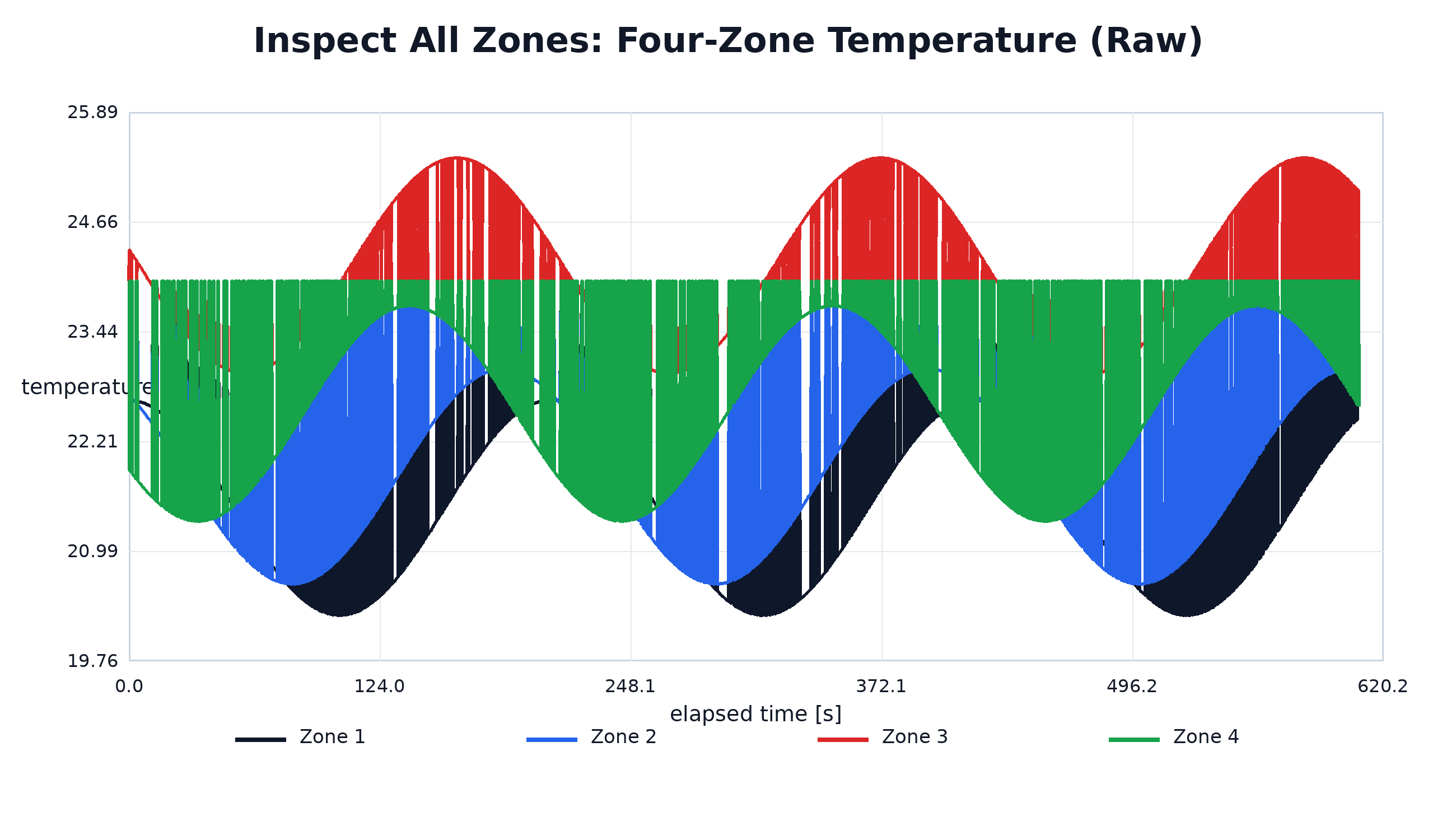}
    \caption{Raw four-zone temperature.}
\end{subfigure}\hfill
\begin{subfigure}[t]{0.485\textwidth}
    \centering
    \includegraphics[width=\linewidth,height=6.0cm,keepaspectratio]
    {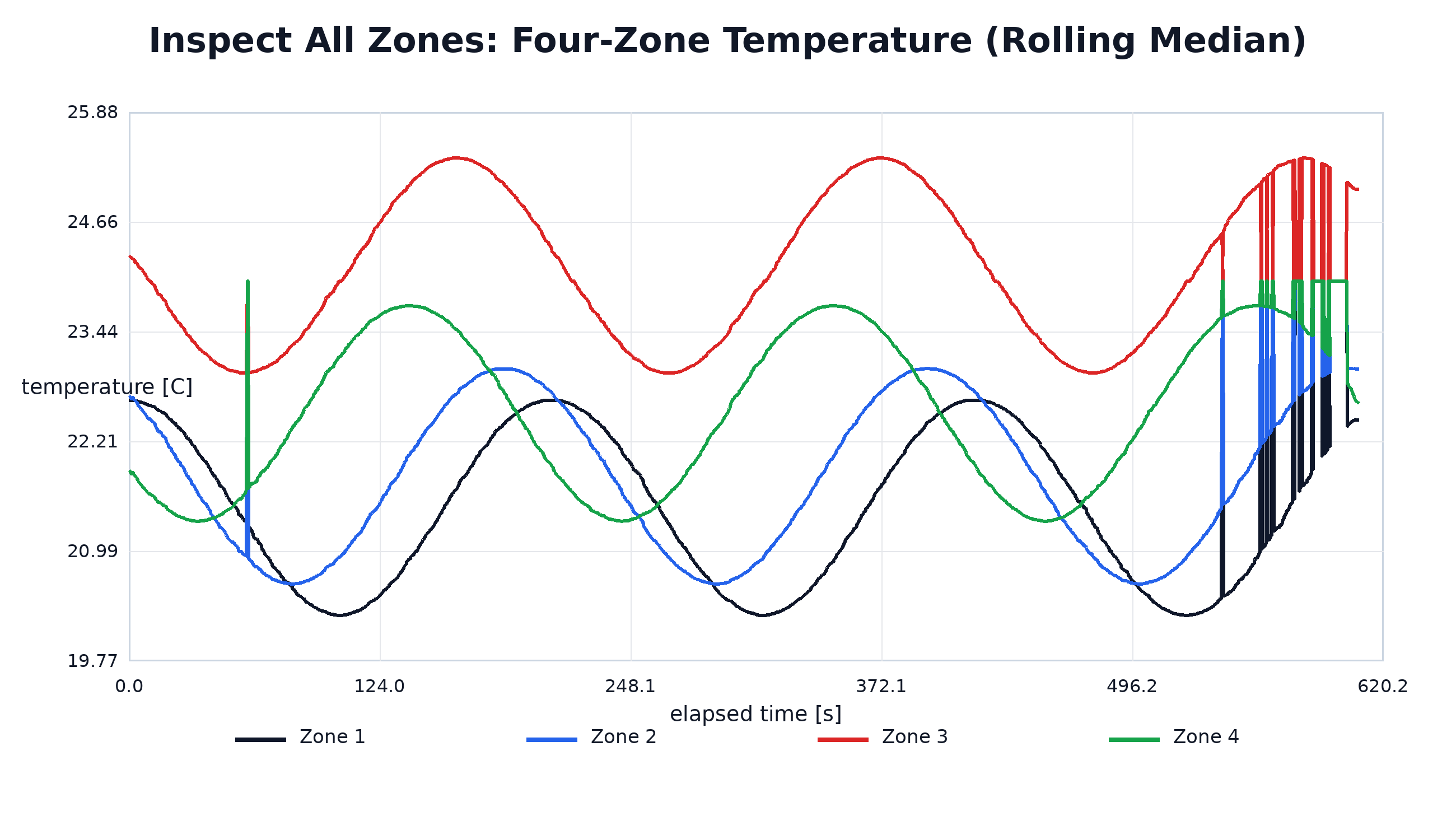}
    \caption{Rolling-median temperature.}
\end{subfigure}\par\medskip
\caption{\rev{All-zone original supplementary traces, part 1. Raw measurements and visualization-only rolling medians are retained. The plotted localization stream is historical and is not the new offline Point-LIO result.}}
\label{fig:app_allzones_results}
\end{figure}
\begin{figure}[p]
\centering
\begin{subfigure}[t]{0.485\textwidth}
    \centering
    \includegraphics[width=\linewidth,height=6.0cm,keepaspectratio]
    {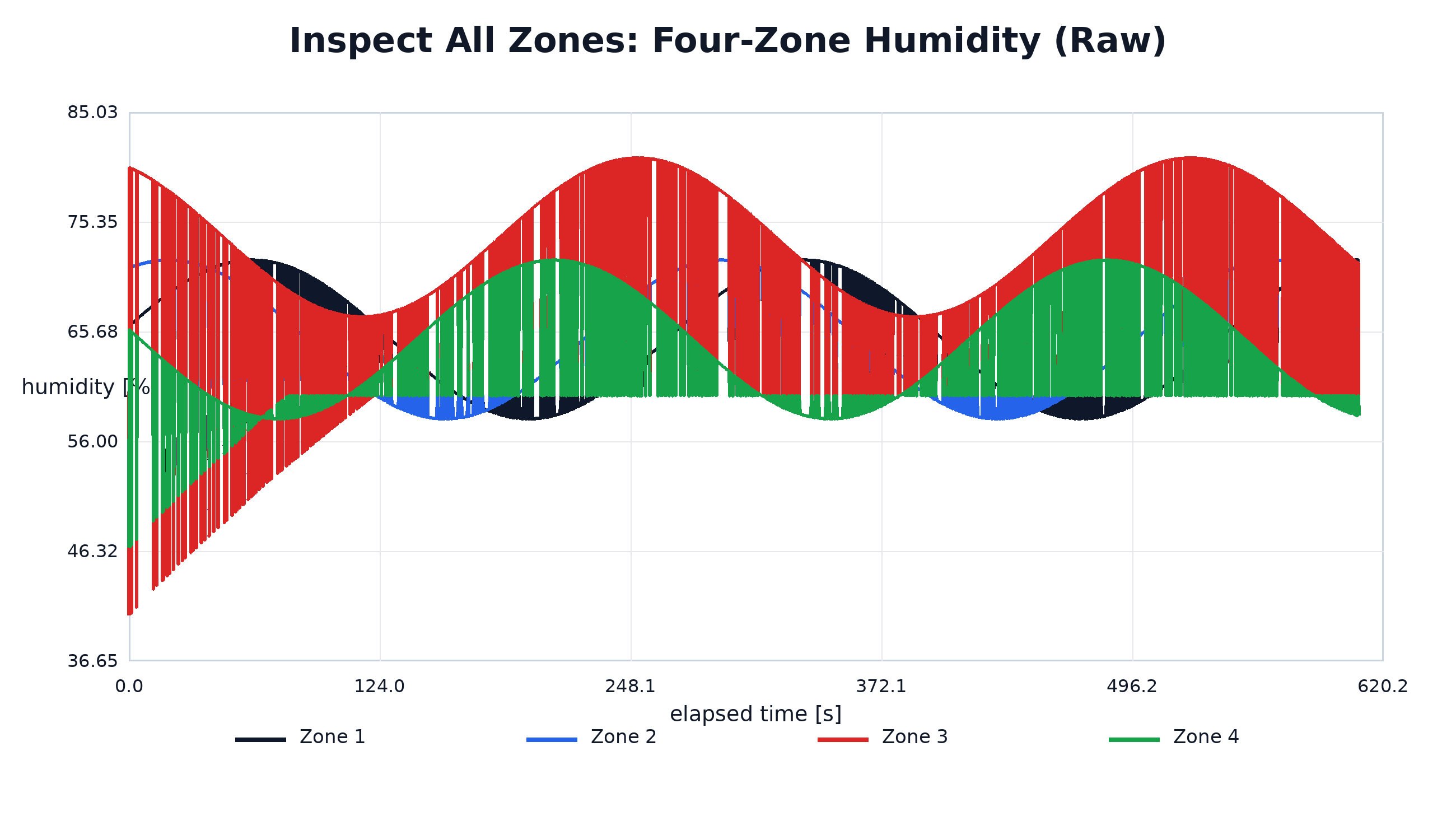}
    \caption{Raw four-zone relative humidity.}
\end{subfigure}\hfill
\begin{subfigure}[t]{0.485\textwidth}
    \centering
    \includegraphics[width=\linewidth,height=6.0cm,keepaspectratio]
    {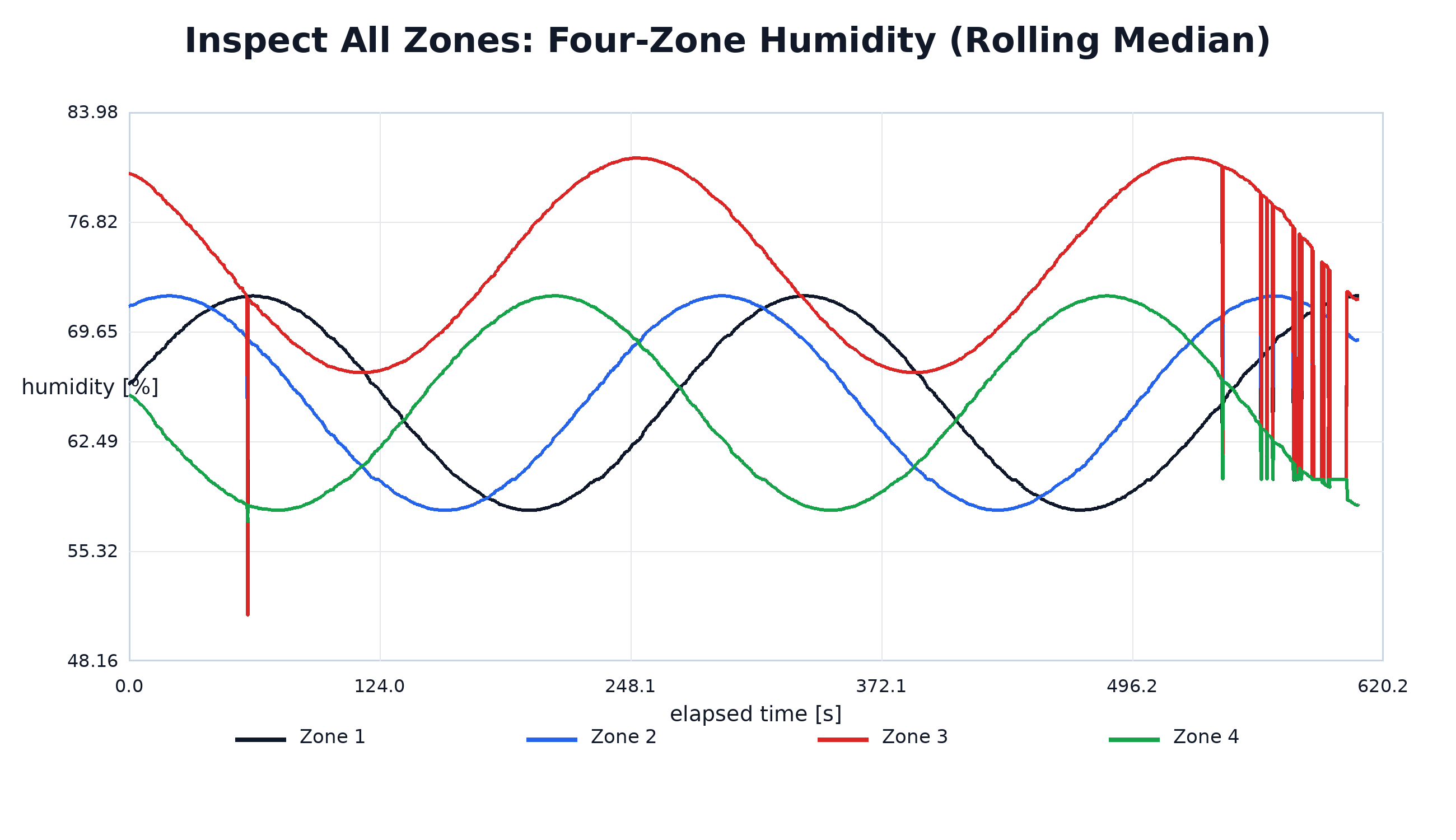}
    \caption{Rolling-median relative humidity.}
\end{subfigure}\par\medskip
\begin{subfigure}[t]{0.485\textwidth}
    \centering
    \includegraphics[width=\linewidth,height=6.0cm,keepaspectratio]
    {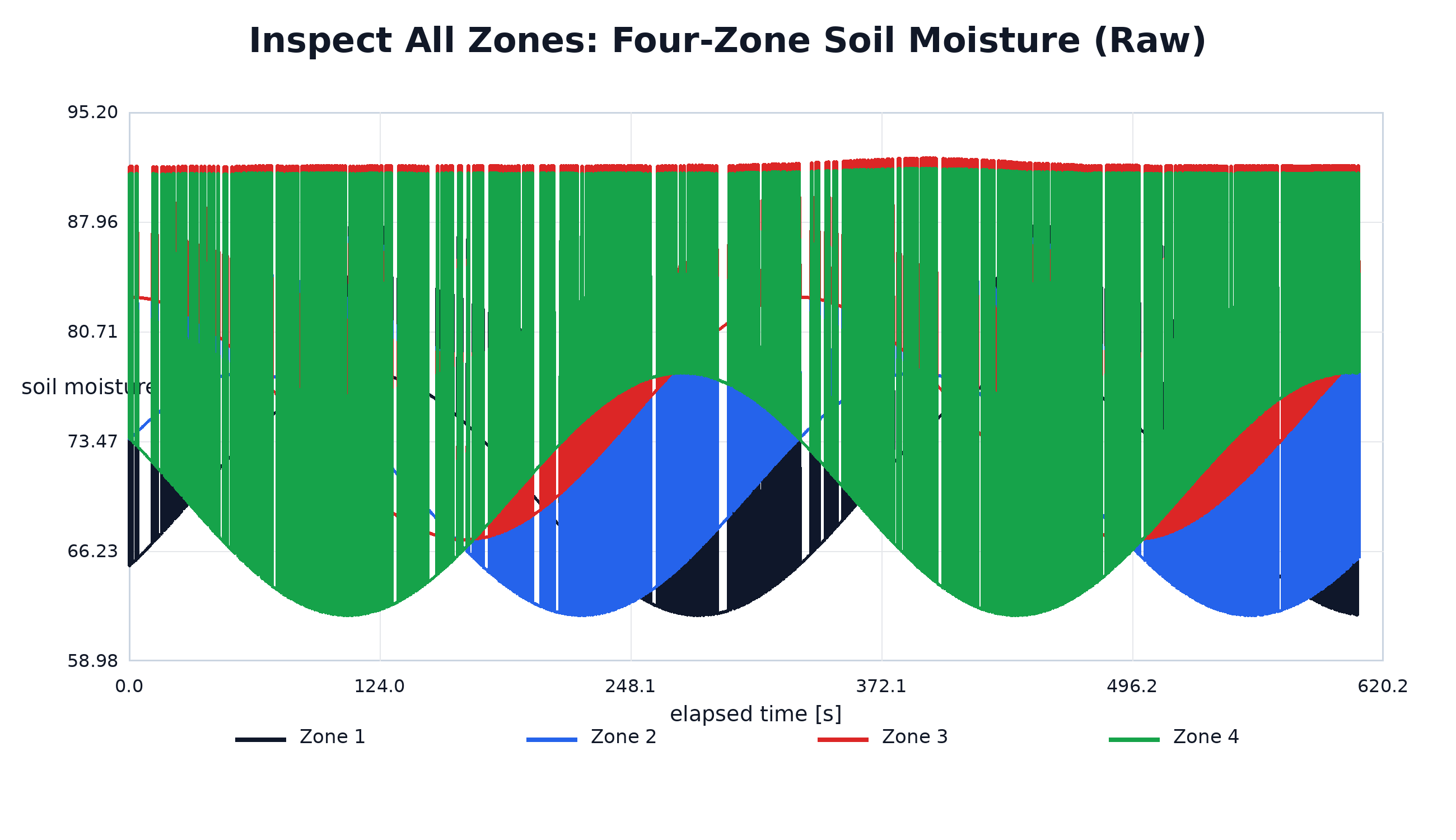}
    \caption{Raw four-zone soil moisture.}
\end{subfigure}\hfill
\begin{subfigure}[t]{0.485\textwidth}
    \centering
    \includegraphics[width=\linewidth,height=6.0cm,keepaspectratio]
    {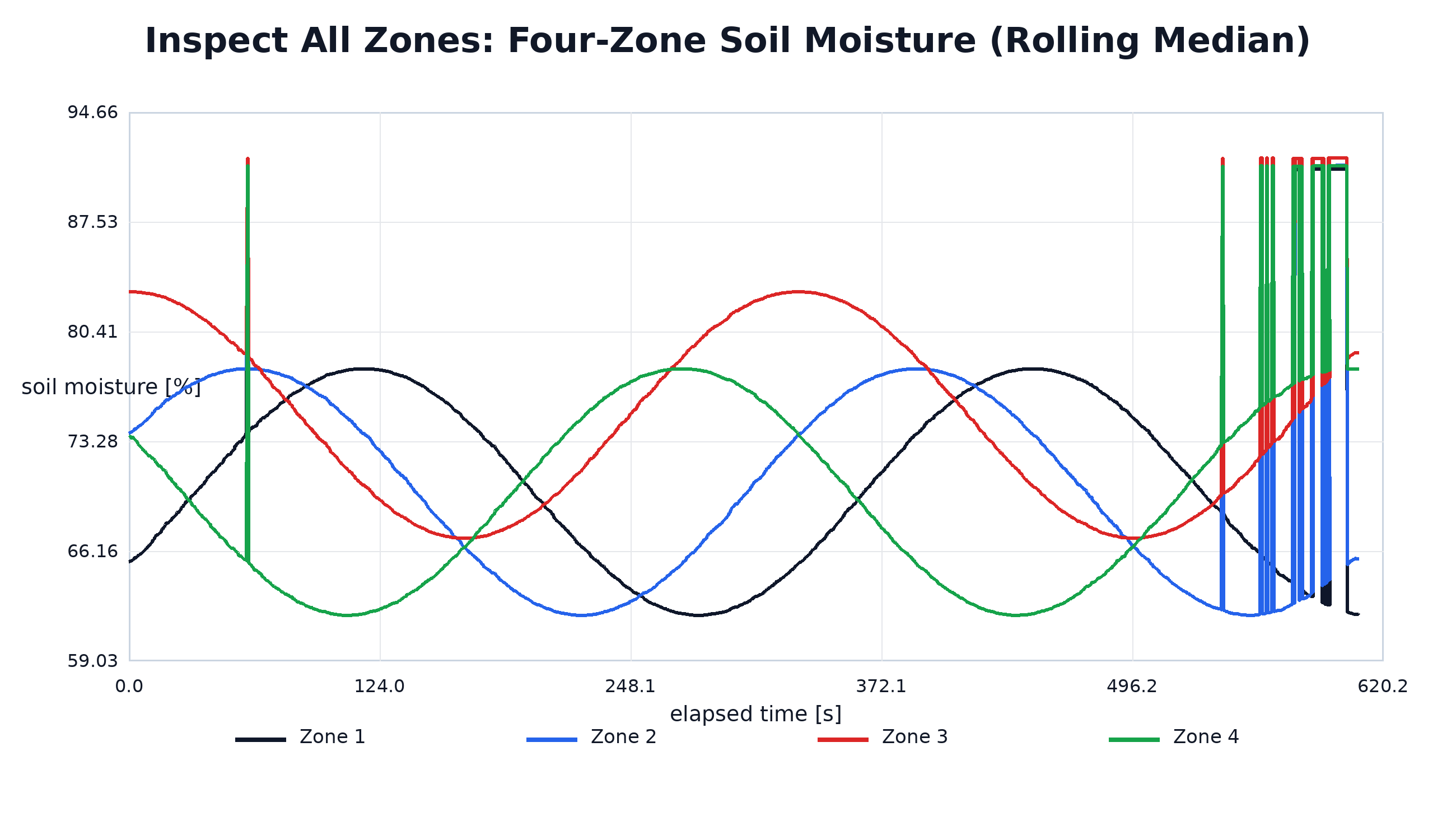}
    \caption{Rolling-median soil moisture.}
\end{subfigure}\par\medskip
\caption{\rev{All-zone original supplementary traces, part 2. Raw measurements and visualization-only rolling medians are retained. The plotted localization stream is historical and is not the new offline Point-LIO result.}}
\label{fig:app_allzones_results-continued}
\end{figure}

Figure~\ref{fig:app_allzones_results}(a) preserves the complete
instantaneous localization-error history, including short-duration
excursions that are not represented by a single mission-level statistic.
Figure~\ref{fig:app_allzones_results}(b) shows the evolution of cumulative
localization performance and converges to the reported \(5.34~\mathrm{cm}\)
XY RMSE.

The environmental panels provide the corresponding four-zone temporal
context. The left column retains the recorded measurements, whereas the
right column provides rolling-median representations to expose slower trends.
These traces demonstrate zone-dependent environmental variation but are not
used to claim PI/PID transient-performance characteristics such as settling
time, overshoot, or disturbance rejection.

\subsubsection{Zone~2 Inspection Results}
\label{app:zone2_results}

The Zone~2-focused mission covered \(50.21~\mathrm{m}\) over
\(235.54~\mathrm{s}\) and produced 2,893 synchronized ground-truth--LIO
pairs. The resulting XY RMSE was \(3.43~\mathrm{cm}\), the planar
95th-percentile error was \(7.60~\mathrm{cm}\), and the 3D RMSE was
\(3.69~\mathrm{cm}\).

\begin{figure}[p]
\centering
\begin{subfigure}[t]{0.485\textwidth}
    \centering
    \includegraphics[width=\linewidth,height=6.0cm,keepaspectratio]
    {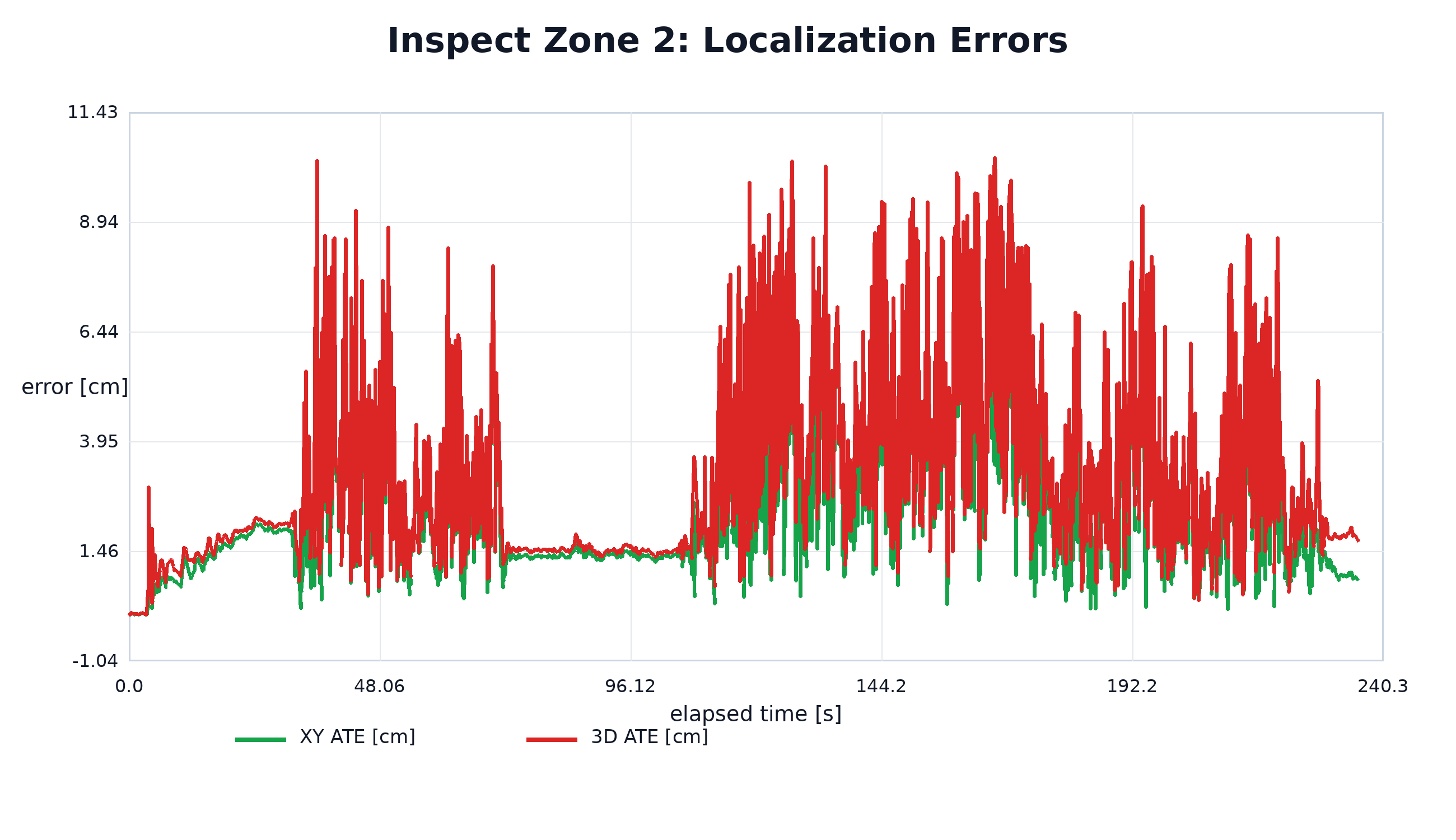}
    \caption{Instantaneous planar localization error.}
\end{subfigure}\hfill
\begin{subfigure}[t]{0.485\textwidth}
    \centering
    \includegraphics[width=\linewidth,height=6.0cm,keepaspectratio]
    {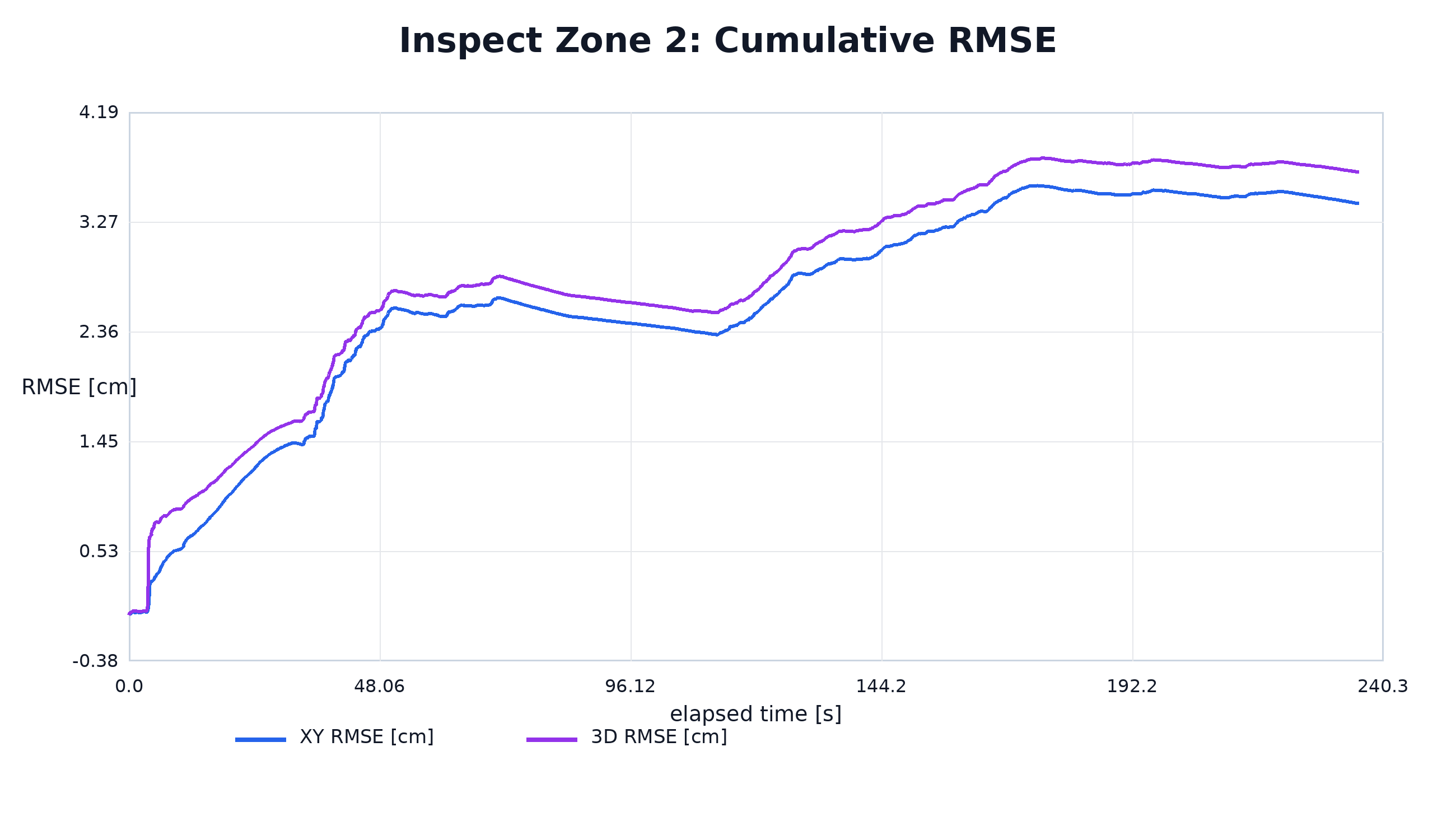}
    \caption{Cumulative planar localization RMSE.}
\end{subfigure}\par\medskip
\begin{subfigure}[t]{0.485\textwidth}
    \centering
    \includegraphics[width=\linewidth,height=6.0cm,keepaspectratio]
    {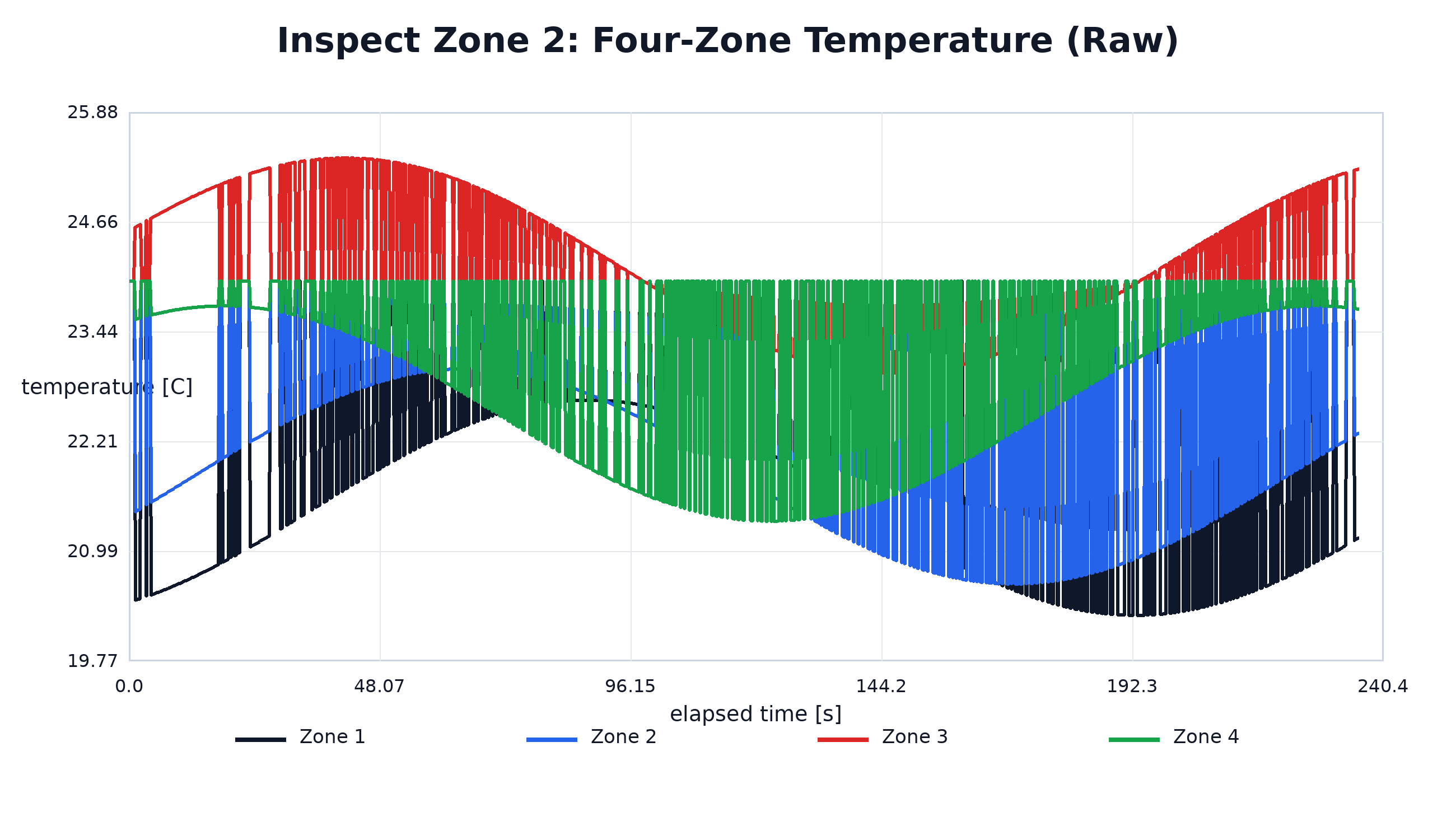}
    \caption{Raw four-zone temperature.}
\end{subfigure}\hfill
\begin{subfigure}[t]{0.485\textwidth}
    \centering
    \includegraphics[width=\linewidth,height=6.0cm,keepaspectratio]
    {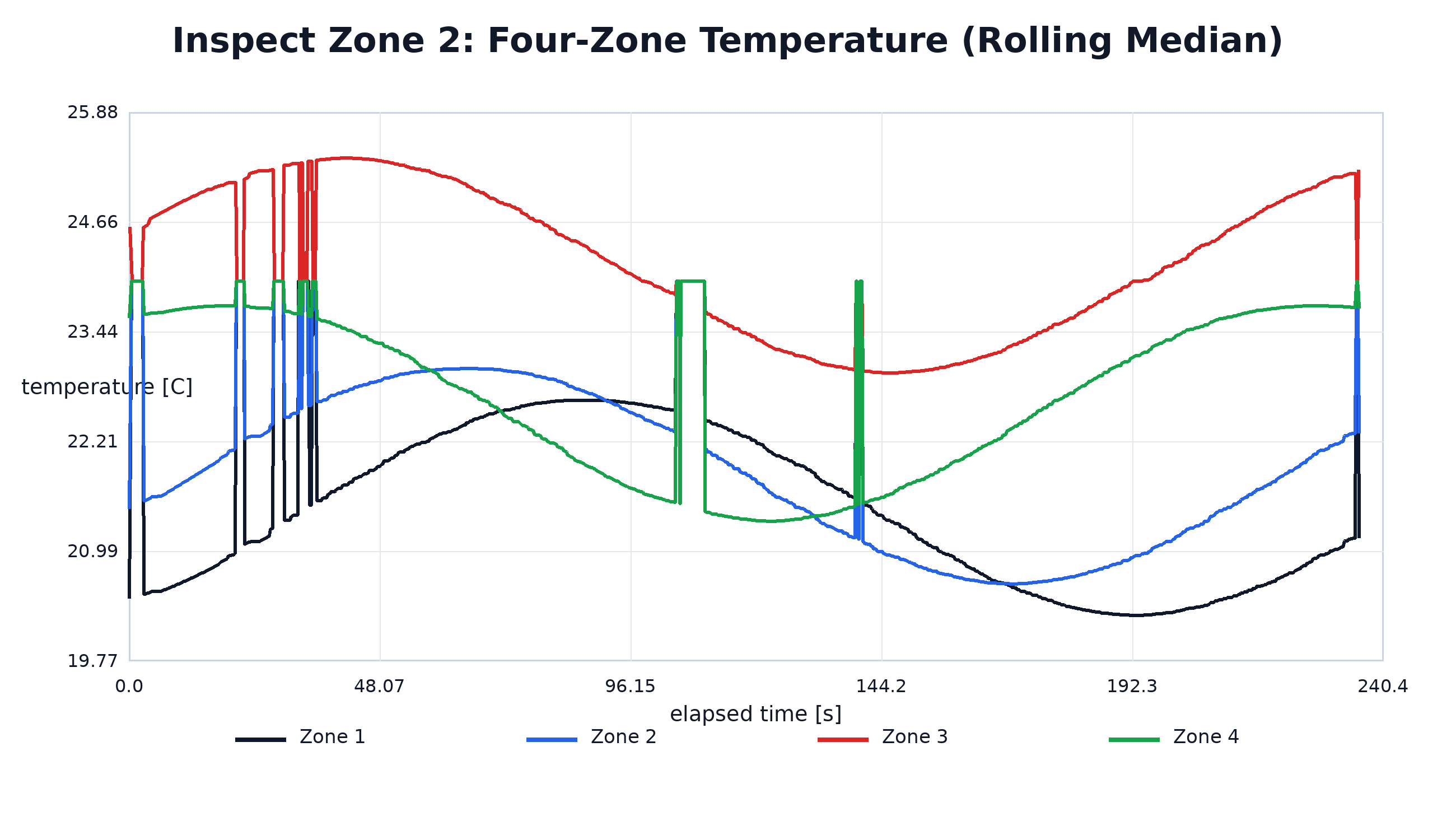}
    \caption{Rolling-median temperature.}
\end{subfigure}\par\medskip
\caption{\rev{Zone~2 original supplementary traces, part 1. Raw measurements and visualization-only rolling medians are retained. The plotted localization stream is historical and is not the new offline Point-LIO result.}}
\label{fig:app_zone2_results}
\end{figure}
\begin{figure}[p]
\centering
\begin{subfigure}[t]{0.485\textwidth}
    \centering
    \includegraphics[width=\linewidth,height=6.0cm,keepaspectratio]
    {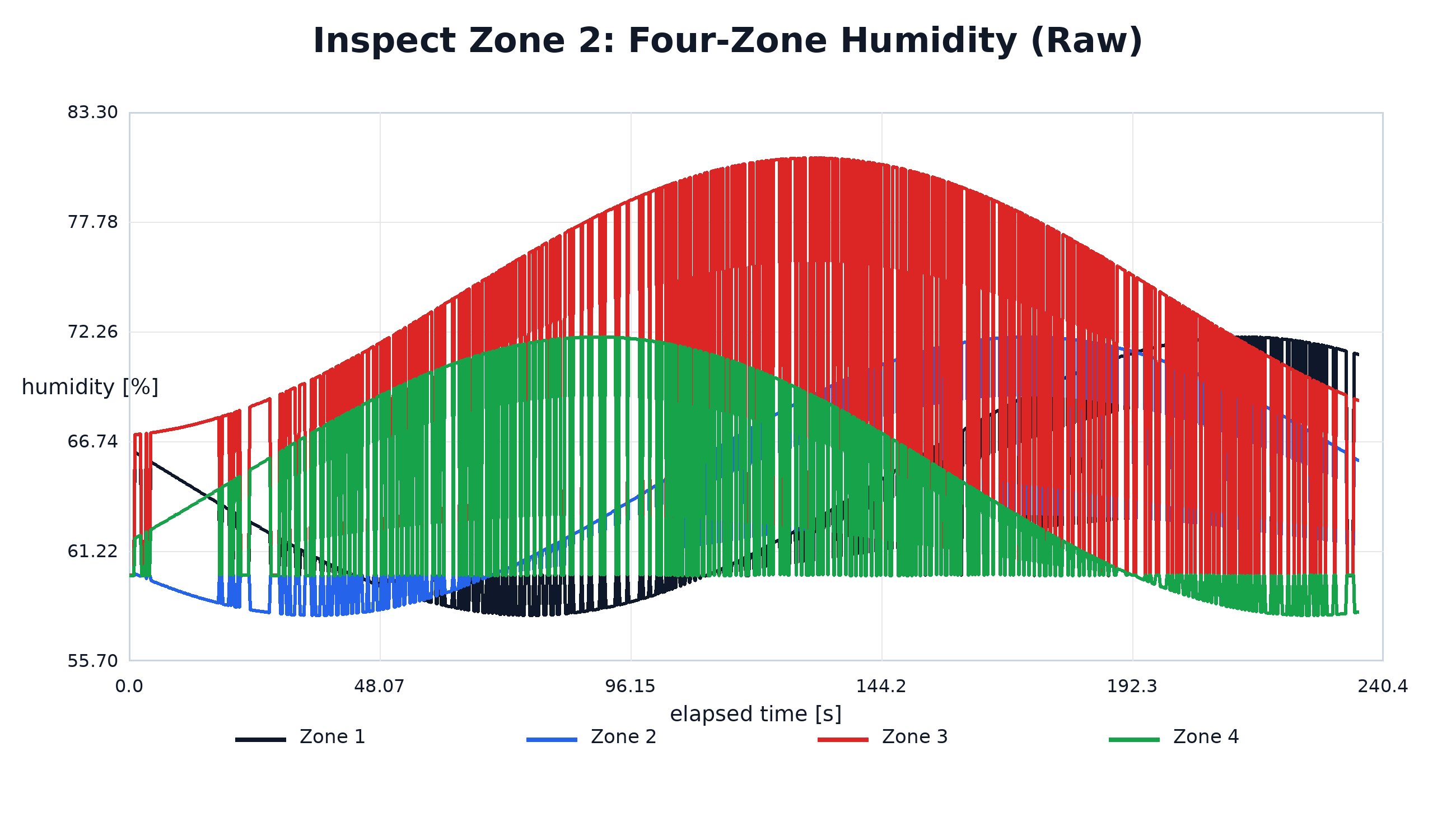}
    \caption{Raw four-zone relative humidity.}
\end{subfigure}\hfill
\begin{subfigure}[t]{0.485\textwidth}
    \centering
    \includegraphics[width=\linewidth,height=6.0cm,keepaspectratio]
    {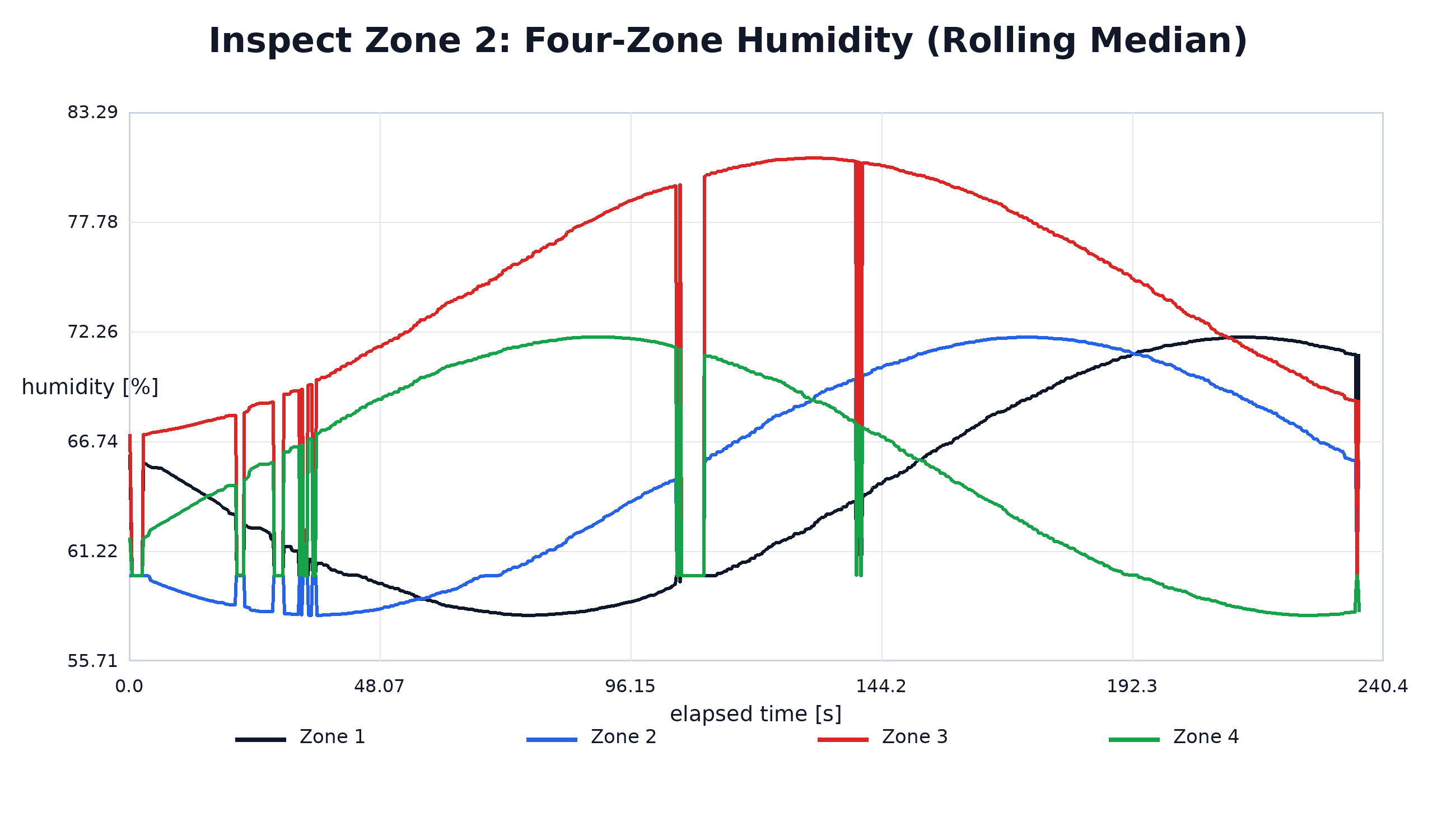}
    \caption{Rolling-median relative humidity.}
\end{subfigure}\par\medskip
\begin{subfigure}[t]{0.485\textwidth}
    \centering
    \includegraphics[width=\linewidth,height=6.0cm,keepaspectratio]
    {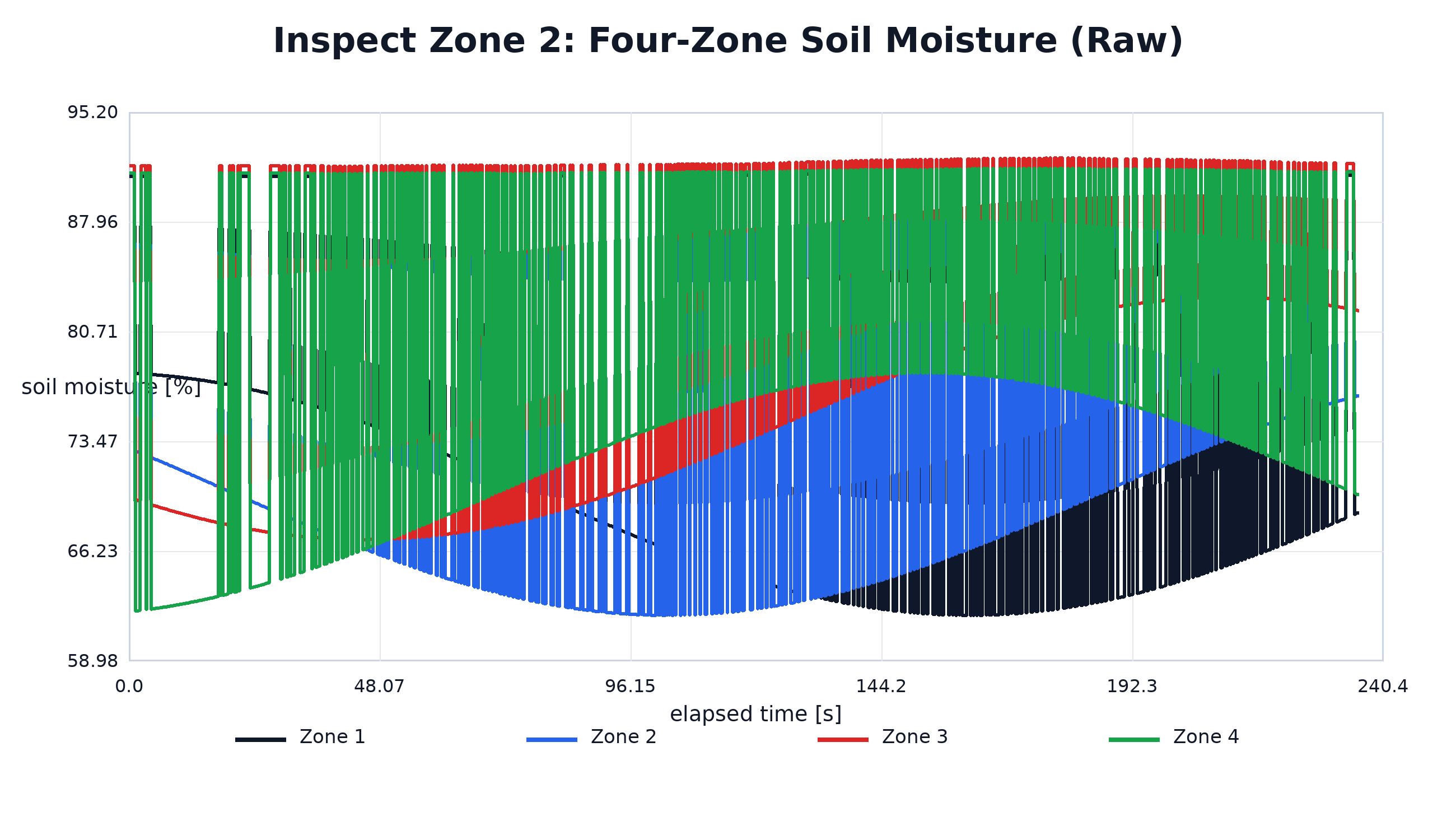}
    \caption{Raw four-zone soil moisture.}
\end{subfigure}\hfill
\begin{subfigure}[t]{0.485\textwidth}
    \centering
    \includegraphics[width=\linewidth,height=6.0cm,keepaspectratio]
    {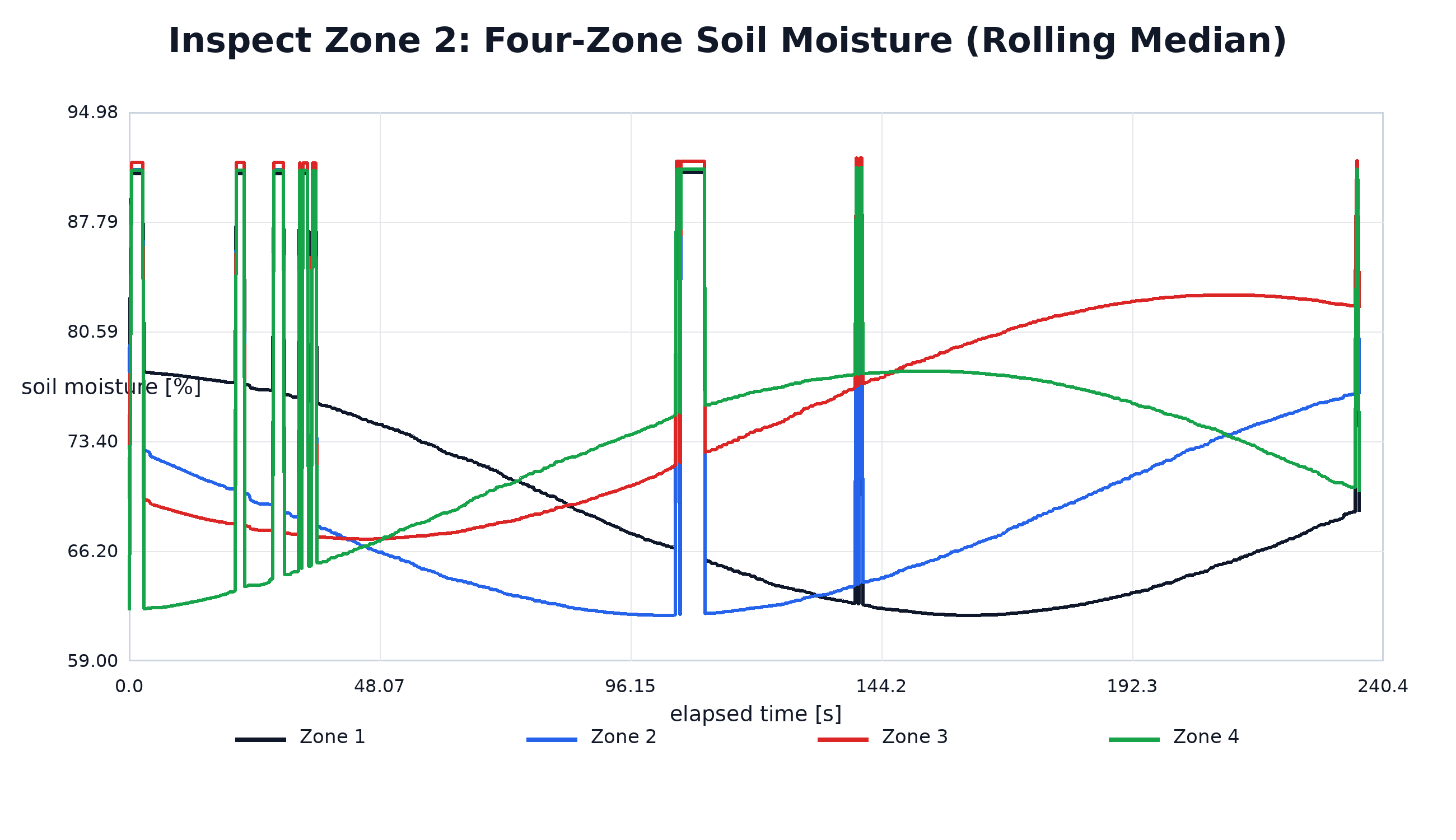}
    \caption{Rolling-median soil moisture.}
\end{subfigure}\par\medskip
\caption{\rev{Zone~2 original supplementary traces, part 2. Raw measurements and visualization-only rolling medians are retained. The plotted localization stream is historical and is not the new offline Point-LIO result.}}
\label{fig:app_zone2_results-continued}
\end{figure}

The Zone~2 mission produced the lowest localization error among the three
evaluated missions. The cumulative trace in
Fig.~\ref{fig:app_zone2_results}(b) converges to the reported
\(3.43~\mathrm{cm}\) XY RMSE. Although the navigation objective was
concentrated on Zone~2, the complete four-zone environmental state was
retained throughout the mission because the multimodal model uses both
current-zone and greenhouse-wide environmental context.

\subsubsection{Zone~4 Inspection Results}
\label{app:zone4_results}

The Zone~4-focused mission covered \(81.34~\mathrm{m}\) over
\(222.94~\mathrm{s}\) and produced 3,181 synchronized ground-truth--LIO
pairs. The resulting XY RMSE was \(5.16~\mathrm{cm}\), the planar
95th-percentile error was \(8.98~\mathrm{cm}\), and the 3D RMSE was
\(5.39~\mathrm{cm}\).

\begin{figure}[p]
\centering
\begin{subfigure}[t]{0.485\textwidth}
    \centering
    \includegraphics[width=\linewidth,height=6.0cm,keepaspectratio]
    {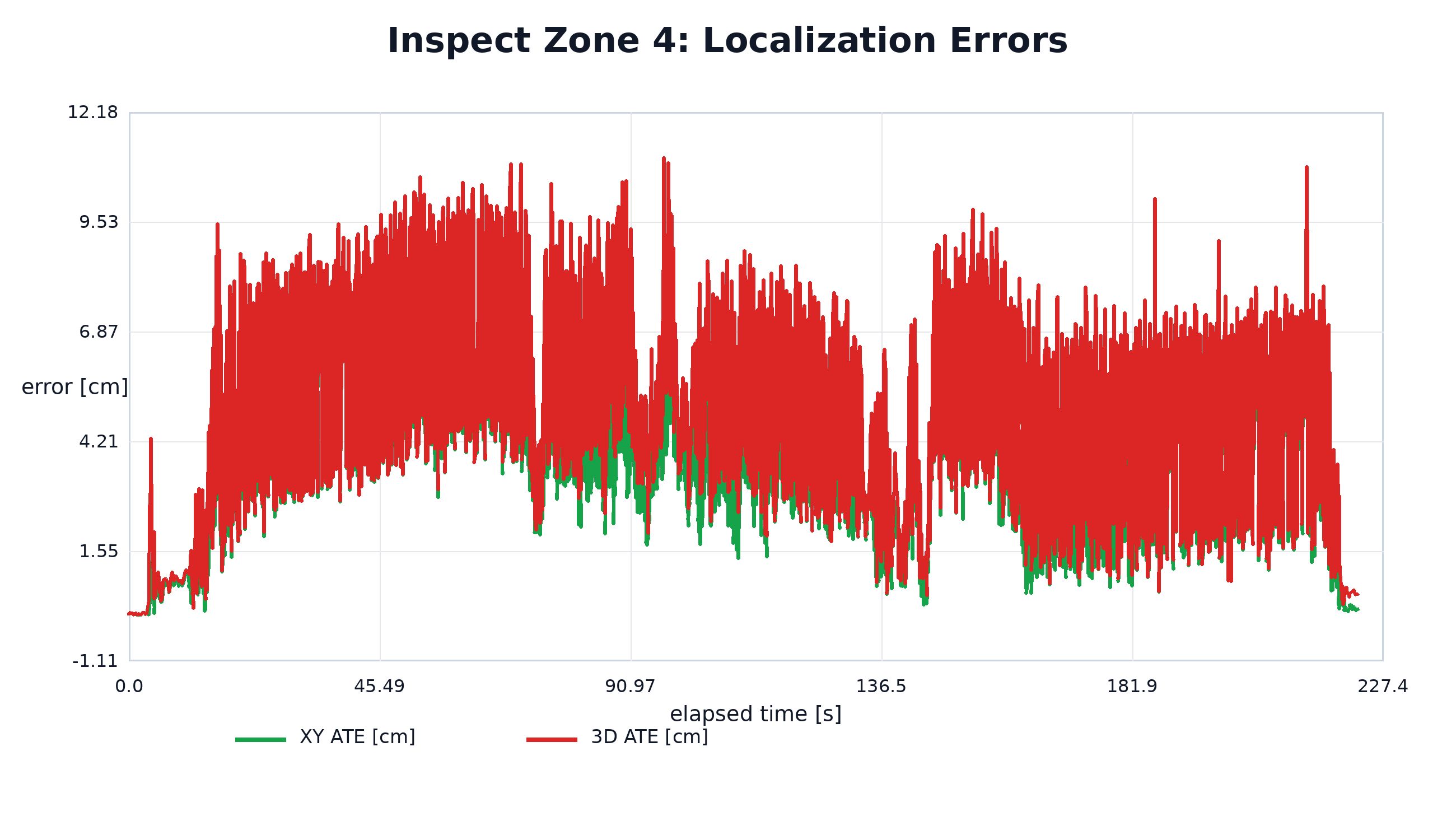}
    \caption{Instantaneous planar localization error.}
\end{subfigure}\hfill
\begin{subfigure}[t]{0.485\textwidth}
    \centering
    \includegraphics[width=\linewidth,height=6.0cm,keepaspectratio]
    {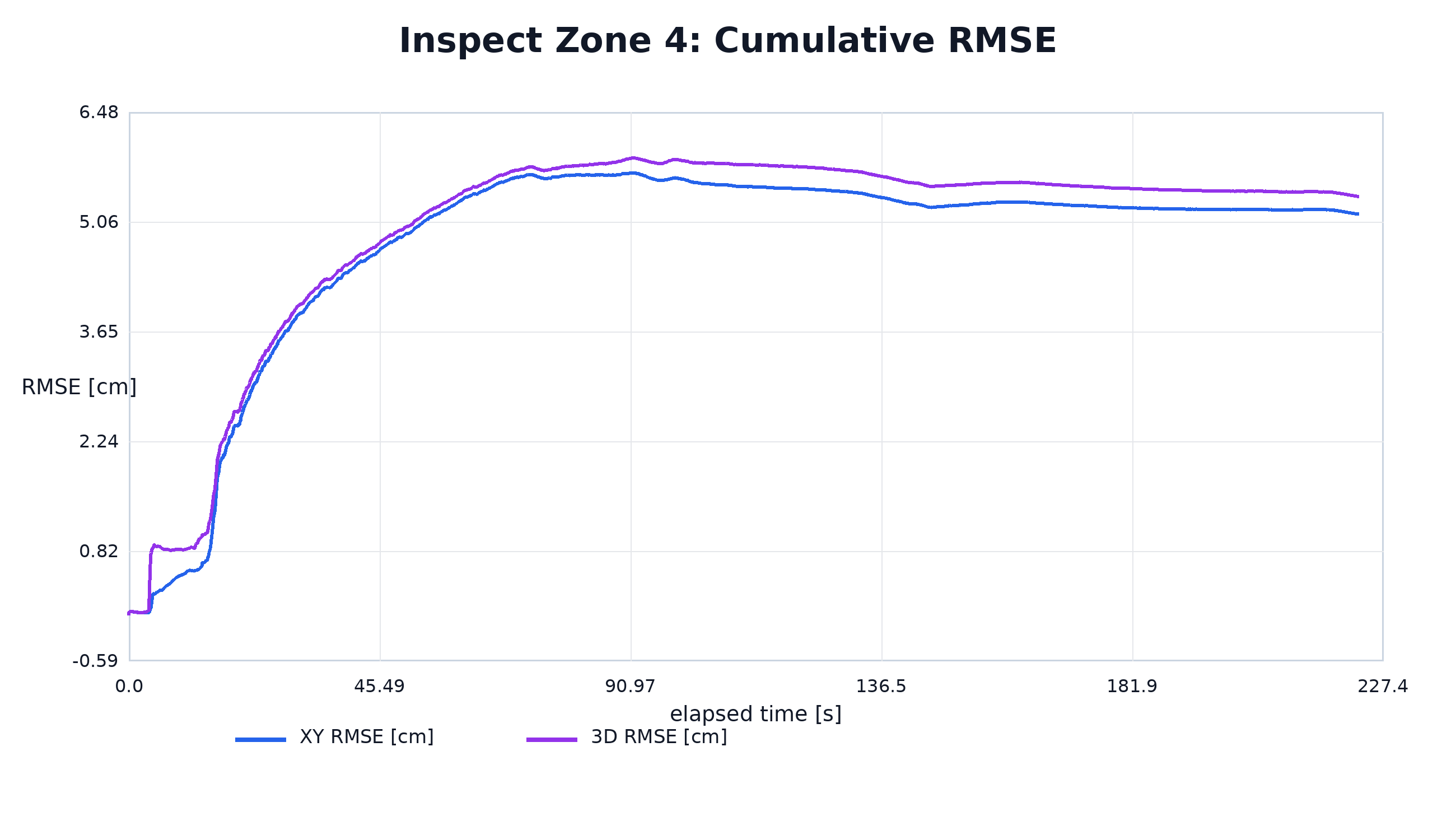}
    \caption{Cumulative planar localization RMSE.}
\end{subfigure}\par\medskip
\begin{subfigure}[t]{0.485\textwidth}
    \centering
    \includegraphics[width=\linewidth,height=6.0cm,keepaspectratio]
    {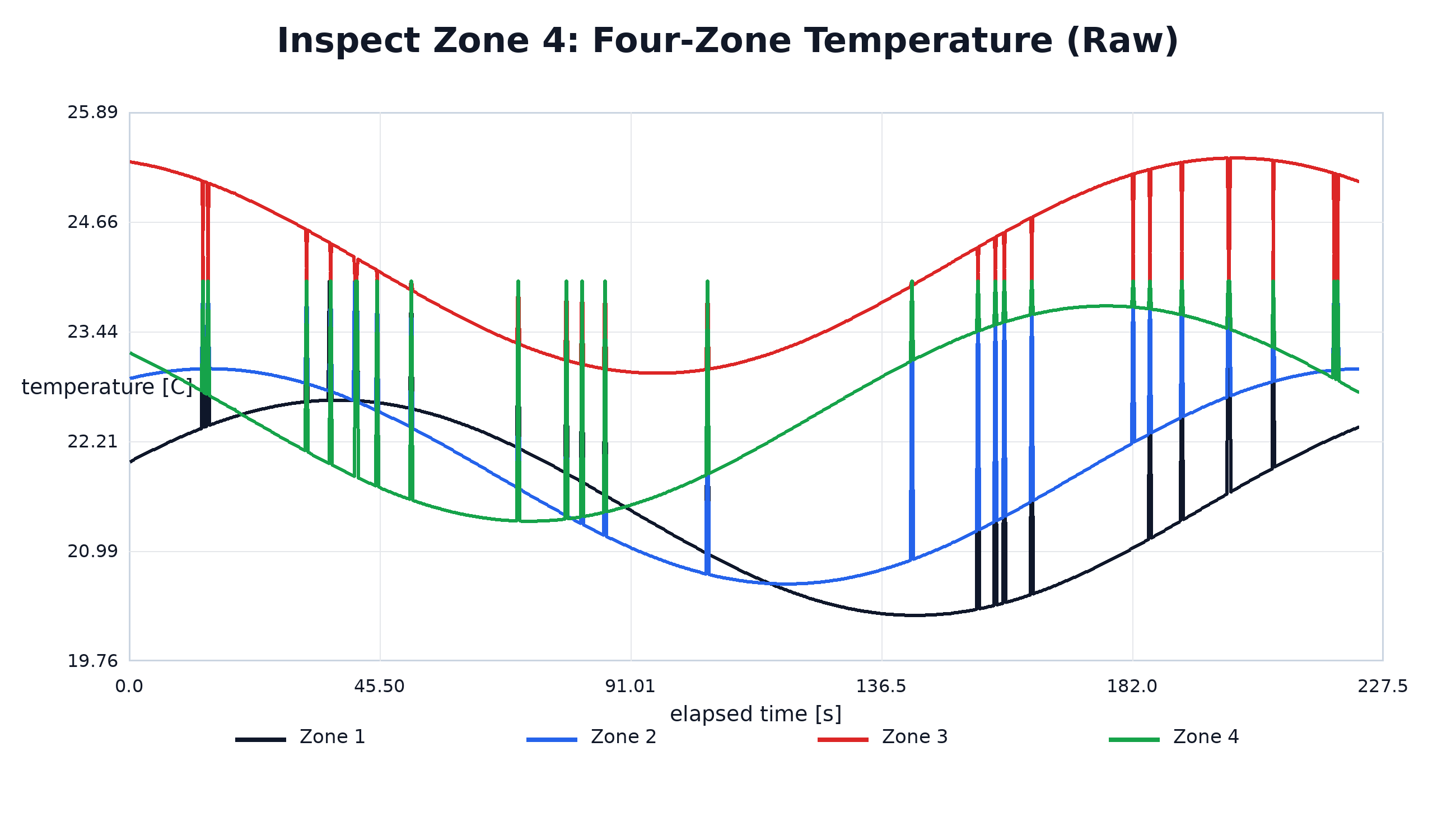}
    \caption{Raw four-zone temperature.}
\end{subfigure}\hfill
\begin{subfigure}[t]{0.485\textwidth}
    \centering
    \includegraphics[width=\linewidth,height=6.0cm,keepaspectratio]
    {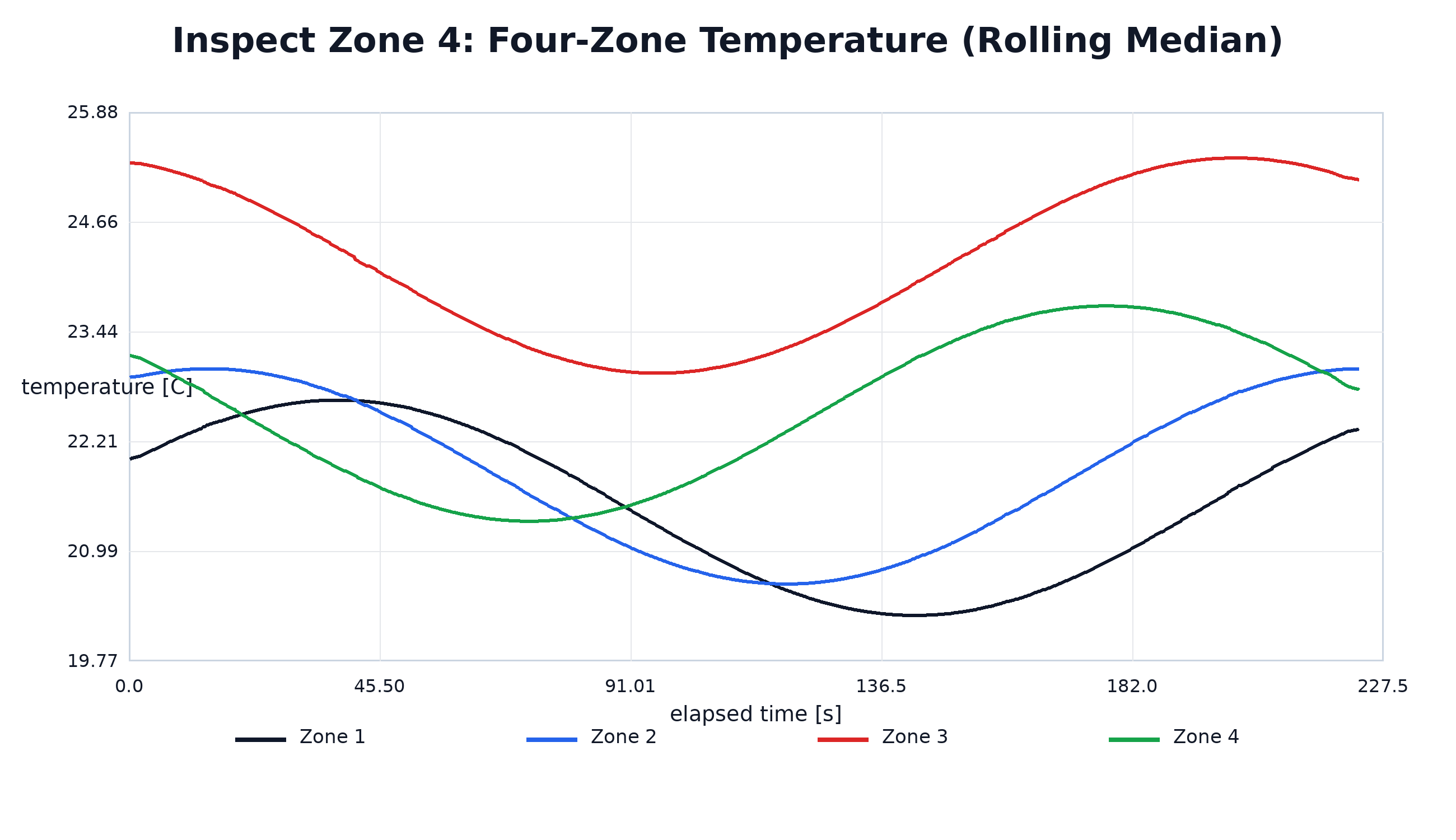}
    \caption{Rolling-median temperature.}
\end{subfigure}\par\medskip
\caption{\rev{Zone~4 original supplementary traces, part 1. Raw measurements and visualization-only rolling medians are retained. The plotted localization stream is historical and is not the new offline Point-LIO result.}}
\label{fig:app_zone4_results}
\end{figure}
\begin{figure}[p]
\centering
\begin{subfigure}[t]{0.485\textwidth}
    \centering
    \includegraphics[width=\linewidth,height=6.0cm,keepaspectratio]
    {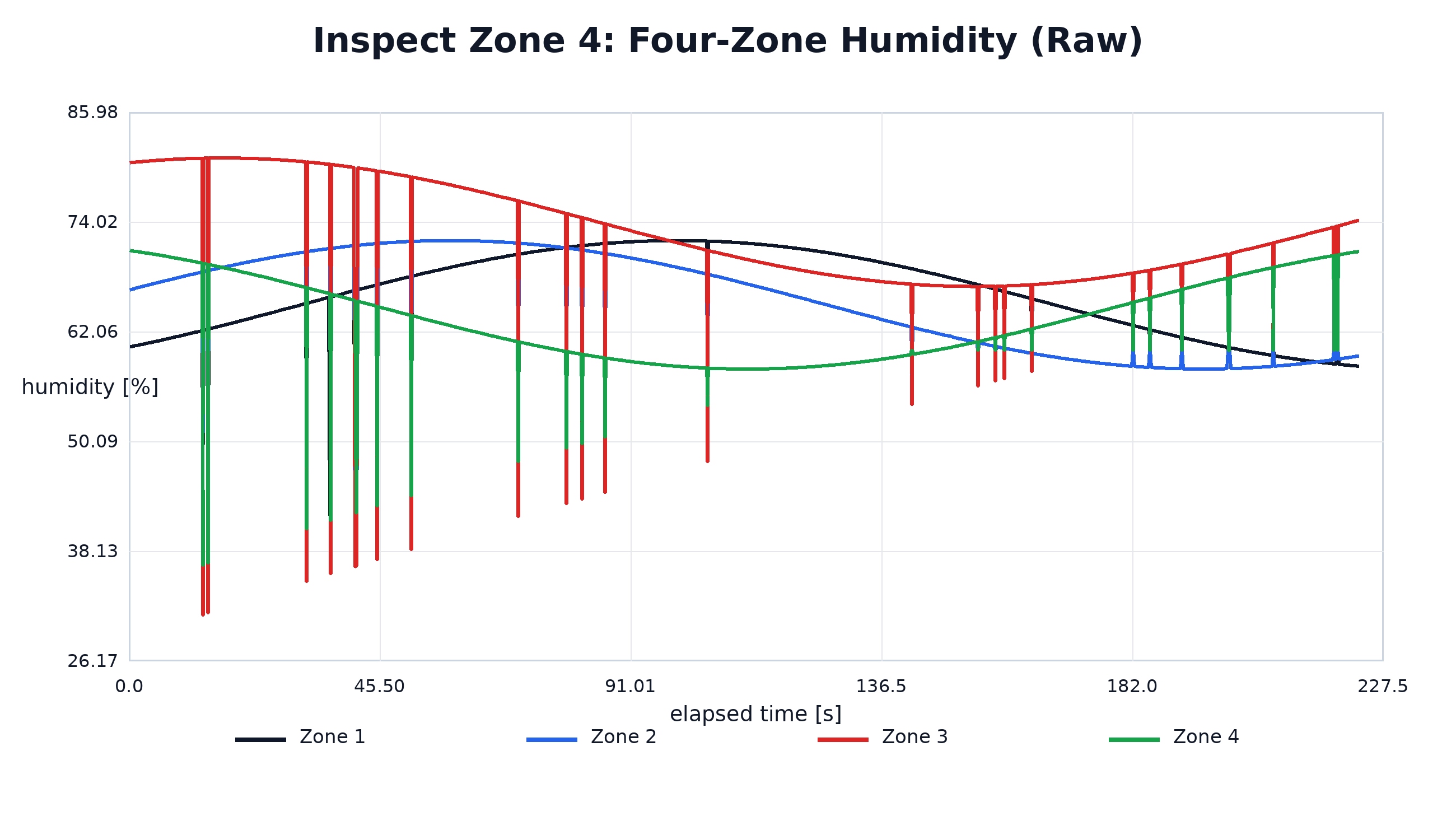}
    \caption{Raw four-zone relative humidity.}
\end{subfigure}\hfill
\begin{subfigure}[t]{0.485\textwidth}
    \centering
    \includegraphics[width=\linewidth,height=6.0cm,keepaspectratio]
    {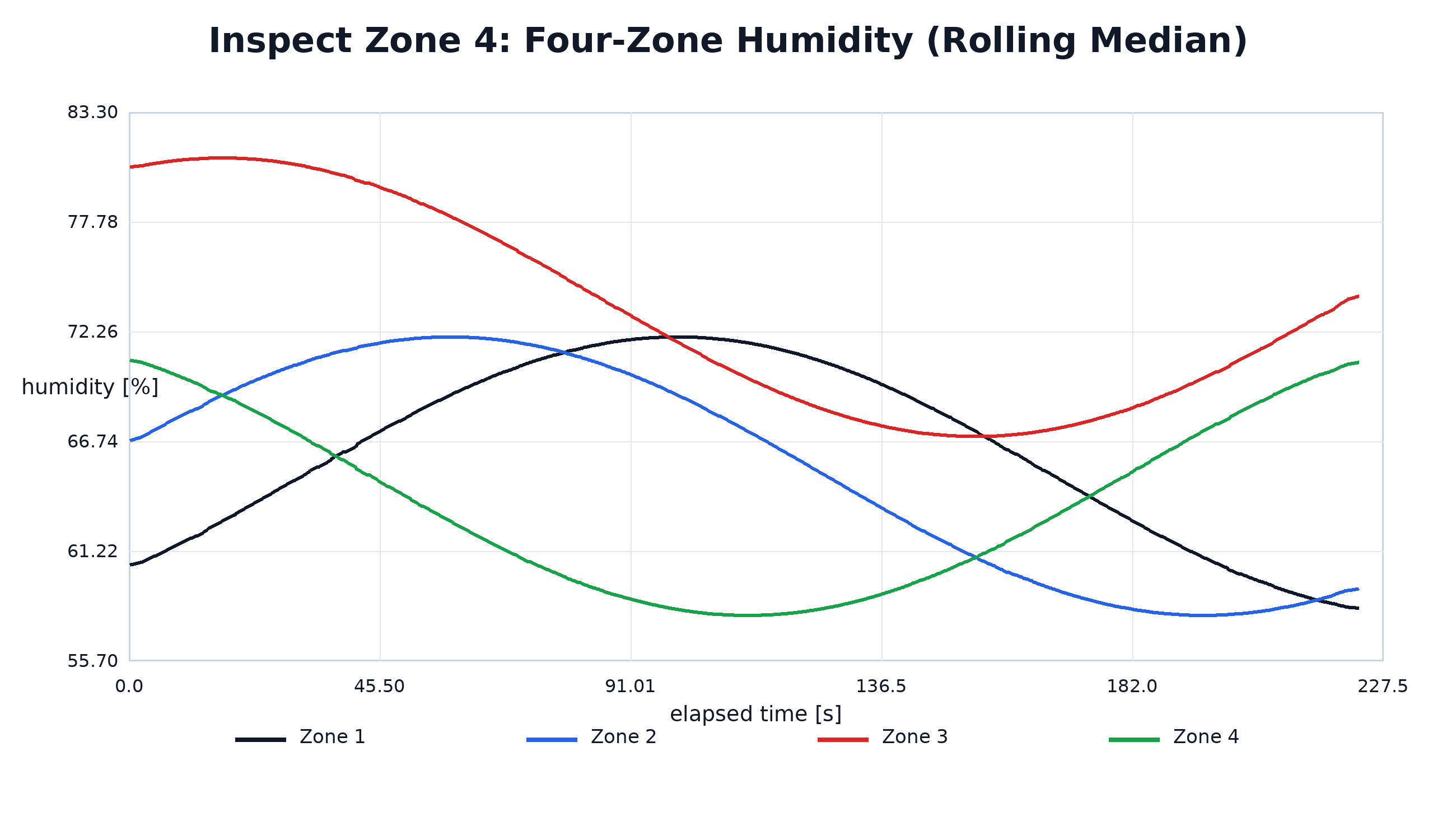}
    \caption{Rolling-median relative humidity.}
\end{subfigure}\par\medskip
\begin{subfigure}[t]{0.485\textwidth}
    \centering
    \includegraphics[width=\linewidth,height=6.0cm,keepaspectratio]
    {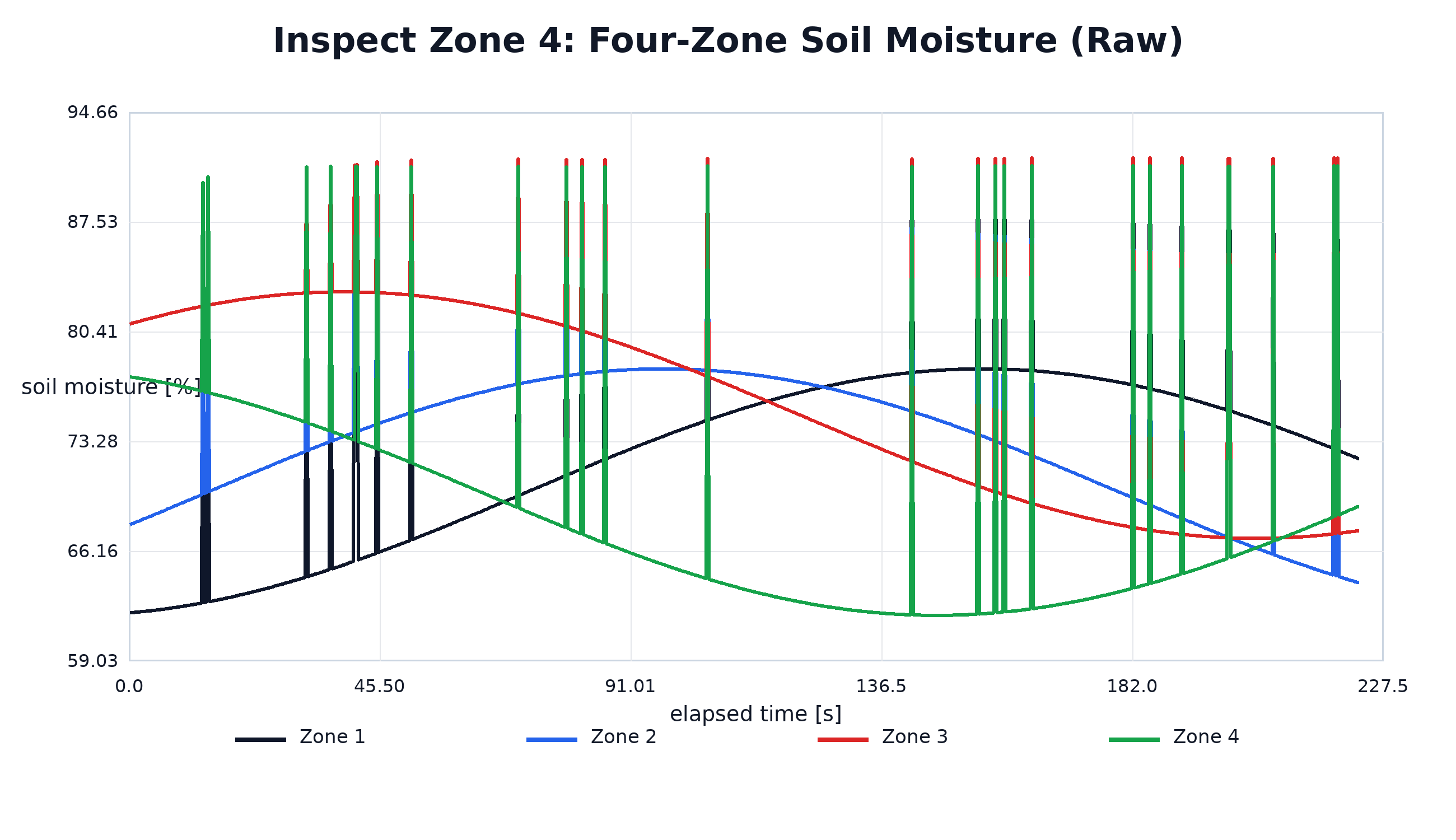}
    \caption{Raw four-zone soil moisture.}
\end{subfigure}\hfill
\begin{subfigure}[t]{0.485\textwidth}
    \centering
    \includegraphics[width=\linewidth,height=6.0cm,keepaspectratio]
    {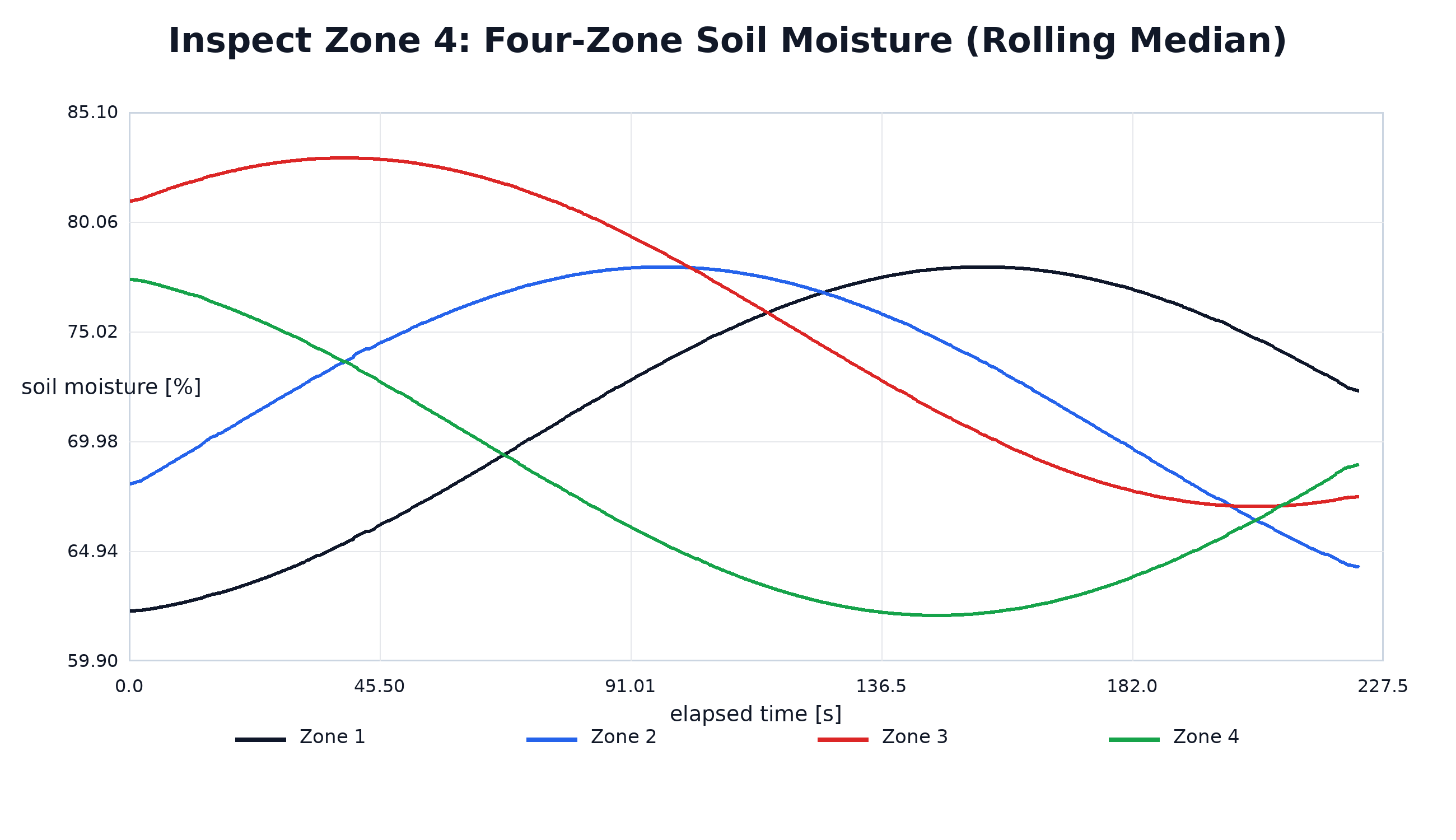}
    \caption{Rolling-median soil moisture.}
\end{subfigure}\par\medskip
\caption{\rev{Zone~4 original supplementary traces, part 2. Raw measurements and visualization-only rolling medians are retained. The plotted localization stream is historical and is not the new offline Point-LIO result.}}
\label{fig:app_zone4_results-continued}
\end{figure}

The Zone~4 localization traces show the complete sample-level behavior behind
the mission-level \(5.16~\mathrm{cm}\) XY RMSE. As in the other missions,
the environmental channels from all four management zones are retained even
though the navigation objective is zone-specific. This preserves the
greenhouse-wide state available to the \VLDA{} decision layer and supports
reproducible comparison across missions.
\subsubsection{Interpretation and Reporting Boundaries}
\label{app:reporting_boundaries}

\rev{The supplementary historical files preserve centimeter-scale agreement between recorded LIO-labelled and reference trajectories, corresponding spatial/environmental context, and raw numerical information for reproducing their summary metrics. They do not independently verify the estimator identity, controller pose consumption, or successful downstream learned-policy validation.}

Several conclusions are intentionally outside the scope of these
supplementary figures. The localization results do not establish absolute
global positioning or indefinitely drift-free operation without additional
global constraints or loop closure. They also do not establish plant-level
manipulator-targeting accuracy, which depends additionally on close-range
perception, camera--robot calibration, depth or target estimation,
base stability, and manipulator registration.

Likewise, the environmental traces demonstrate recorded zone-state evolution
but do not constitute a systematic PI/PID tuning study. Quantities such as
rise time, settling time, overshoot, steady-state error, disturbance
rejection, actuator effort, and saturation behavior require dedicated
closed-loop experiments.

Finally, the disease-like samples used by the multimodal model are controlled
visual proxies generated after simulation. They are not biologically
confirmed disease observations and do not establish field-level diagnosis,
disease severity, spray efficacy, crop response, or agronomic treatment
performance. These boundaries are consistent with the simulation-centric
scope of the present work.

\clearpage
\FloatBarrier
\section{\rev{Retained standard LiDAR--inertial mathematics}}
\label{sec:s-standard}

\rev{This section preserves the original standard LiDAR--inertial derivations outside the main narrative, following the reviewer request to shorten that non-novel derivation in the main text. Original activity-state and validation/dispatch specifications are retained in Section S9. They are future integration requirements rather than evaluated methods. The actual saved neural configuration and the two separately tested readiness/command guards are specified in Sections S1 and S4.}

\subsection{\rev{LiDAR--inertial estimation background}}
\label{sec:s-standard-lio}

Let \(\mathcal{F}_{M}\) denote the local greenhouse/map frame,
\(\mathcal{F}_{I}\) the IMU/body frame, and \(\mathcal{F}_{L}\) the LiDAR
sensor frame. A nominal LIO state can be written as

\begin{equation}
\mathcal{S}^{\mathrm{lio}}_t
=
\left(
R^{M}_{I,t},
\vect{p}^{M}_{I,t},
\vect{v}^{M}_{I,t},
\vect{b}_{g,t},
\vect{b}_{a,t},
\vect{g}
\right),
\label{eq:lio_state}
\end{equation}

where \(R^{M}_{I,t}\) is the IMU orientation in the local map frame,
\(\vect{p}^{M}_{I,t}\) and \(\vect{v}^{M}_{I,t}\) are its position and
velocity, \(\vect{b}_{g,t}\) and \(\vect{b}_{a,t}\) are gyroscope and
accelerometer biases, and \(\vect{g}\) is gravity.

For downstream navigation, the six-degree-of-freedom robot pose is summarized
as

\begin{equation}
\vect{x}^{M}_t
=
\begin{bmatrix}
\vect{p}^{M\top}_{I,t} &
\vect{\vartheta}^{\top}_t
\end{bmatrix}^{\top},
\qquad
\vect{\vartheta}_t
=
\begin{bmatrix}
\phi_t & \theta_t & \psi_t
\end{bmatrix}^{\top},
\label{eq:s-local_pose}
\end{equation}

where \(\phi_t\), \(\theta_t\), and \(\psi_t\) denote roll, pitch, and yaw.

\subsubsection{IMU propagation}

The IMU measures angular velocity and specific force. The measurements contain
slowly varying biases and measurement noise:

\begin{equation}
\begin{aligned}
\vect{\omega}_{m,t}
&=
\vect{\omega}_{t}
+
\vect{b}_{g,t}
+
\vect{n}_{g,t},
\\
\vect{a}_{m,t}
&=
R^{M\top}_{I,t}
\left(
\dot{\vect{v}}^{M}_{I,t}
-
\vect{g}
\right)
+
\vect{b}_{a,t}
+
\vect{n}_{a,t},
\end{aligned}
\label{eq:imu_measurement_model}
\end{equation}

where \(\vect{n}_{g,t}\) and \(\vect{n}_{a,t}\) denote gyroscope and
accelerometer noise.

Over an IMU sampling interval \(\Delta t\), the bias-corrected angular
velocity propagates orientation according to

\begin{equation}
\hat{R}^{M,-}_{I,k+1}
=
R^{M}_{I,k}
\operatorname{Exp}
\left[
\left(
\vect{\omega}_{m,k}
-
\vect{b}_{g,k}
\right)
\Delta t
\right],
\label{eq:imu_orientation_propagation}
\end{equation}

where \(\operatorname{Exp}(\cdot)\) denotes the exponential map from a
rotation vector to \(\mathrm{SO}(3)\).

The corresponding acceleration in the map frame is

\begin{equation}
\vect{a}^{M}_{k}
=
R^{M}_{I,k}
\left(
\vect{a}_{m,k}
-
\vect{b}_{a,k}
\right)
+
\vect{g},
\label{eq:imu_map_acceleration}
\end{equation}

and velocity and position are propagated as

\begin{equation}
\begin{aligned}
\hat{\vect{v}}^{M,-}_{I,k+1}
&=
\vect{v}^{M}_{I,k}
+
\vect{a}^{M}_{k}\Delta t,
\\
\hat{\vect{p}}^{M,-}_{I,k+1}
&=
\vect{p}^{M}_{I,k}
+
\vect{v}^{M}_{I,k}\Delta t
+
\frac{1}{2}
\vect{a}^{M}_{k}\Delta t^{2}.
\end{aligned}
\label{eq:imu_translational_propagation}
\end{equation}

The associated state covariance is propagated using the linearized inertial
dynamics and IMU-noise model. Biases are treated as slowly varying states.
IMU propagation alone is insufficient for sustained greenhouse navigation:
gyroscope bias produces increasing orientation error, while accelerometer
bias and gravity-projection errors accumulate through integration and produce
position drift.

The high-rate inertial estimate is also used to deskew the LiDAR scan.
Because individual LiDAR points are acquired at different instants during a
scan, each point is compensated to a common reference time before geometric
registration. This step reduces motion-induced distortion of greenhouse
walls, crop rows, beds, and other structural features.

\subsubsection{LiDAR geometric correction}

Let \(R_{IL}\) and \(\vect{t}_{IL}\) denote the calibrated rigid
LiDAR-to-IMU transformation. A deskewed LiDAR point
\(\vect{p}^{L}_{i}\) is transformed into the local map frame as

\begin{equation}
\vect{p}^{M}_{i}
=
R^{M}_{I,t}
\left(
R_{IL}\vect{p}^{L}_{i}
+
\vect{t}_{IL}
\right)
+
\vect{p}^{M}_{I,t}.
\label{eq:lidar_point_transform}
\end{equation}

For each transformed point, neighboring points in the current local map are
used to determine a local geometric surface. For a point-to-plane
formulation, let \(\vect{n}_{i}\) denote the estimated unit surface normal and
\(\vect{q}_{i}\) a point on the corresponding plane. The geometric residual
is

\begin{equation}
r^{L}_{i}(\vect{\xi}_t)
=
\vect{n}_{i}^{\top}
\left[
\vect{p}^{M}_{i}(\vect{\xi}_t)
-
\vect{q}_{i}
\right].
\label{eq:point_plane_residual}
\end{equation}

The LiDAR measurements then correct the IMU-propagated state. A compact
representation of the tightly coupled estimation problem is

\begin{equation}
\vect{\xi}^{\ast}_t
=
\arg\min_{\vect{\xi}_t}
\left[
\sum_{i\in\mathcal{C}_t}
\rho
\left(
\frac{
\left(r^{L}_{i}(\vect{\xi}_t)\right)^2
}{
\sigma^{2}_{L,i}
}
\right)
+
\left\|
\delta\vect{\xi}_t
\right\|^{2}_{P^{-1}_{I,t}}
\right],
\label{eq:lio_cost}
\end{equation}

where \(\mathcal{C}_t\) is the set of accepted LiDAR correspondences,
\(\rho(\cdot)\) is a robust loss, \(\sigma^{2}_{L,i}\) represents LiDAR
residual uncertainty, \(P_{I,t}\) is the covariance of the IMU-propagated
state, and \(\delta\vect{\xi}_t\) denotes the local error state relative to
the inertial prediction.

The first term in \eqref{eq:lio_cost} enforces agreement with observed
greenhouse geometry, whereas the second term incorporates the propagated
inertial prior. Weak or geometrically inconsistent correspondences are
rejected before the correction step. The corrected state updates robot pose,
velocity, IMU biases, and pose uncertainty.

This fusion follows the complementary sensing principle used by established
tightly coupled LIO approaches such as FAST-LIO2 and LIO-SAM
\citep{xu2022fastlio2,shan2020liosam}. \AgriSense{} does not claim a new
LiDAR--IMU estimator.

\subsubsection{Local-map update and LIO cycle}

After state correction, accepted LiDAR points are transformed with the
corrected pose and inserted into the local map:

\begin{equation}
\mathcal{M}_{t}
=
\operatorname{VoxelFilter}
\left(
\mathcal{M}_{t-1}
\cup
\left\{
\vect{p}^{M}_{i}(\vect{\xi}^{\ast}_t)
\right\}_{i\in\mathcal{C}_t}
\right).
\label{eq:local_map_update}
\end{equation}

Voxel filtering and bounded local-map maintenance limit redundant points and
keep the map suitable for repeated scan-to-map registration.

\rev{A} \rev{representative} \rev{LiDAR--IMU estimation} cycle is summarized in Algorithm~\ref{alg:lio_cycle}. \rev{This generic background algorithm is not an audited line-by-line description of the original three mission implementations.}

\begin{algorithm}[!t]
\caption{LiDAR--IMU Spatial Estimation Cycle}
\label{alg:lio_cycle}
\begin{algorithmic}[1]
\Require LiDAR scan \(P_t\), synchronized IMU measurements,
previous LIO state, covariance, and local map \(\mathcal{M}_{t-1}\)
\State Synchronize the LiDAR scan with the IMU stream and reject stale data
\State Deskew LiDAR points using high-rate IMU motion propagation
\State Propagate orientation, velocity, position, biases, and covariance
\State Transform the deskewed LiDAR points through the calibrated
LiDAR--IMU extrinsics
\State Search local-map neighbors and construct valid geometric
correspondences
\State Compute point-to-plane residuals and their Jacobians
\State Correct the propagated state using the LiDAR geometric constraints
\State Update pose, velocity, biases, and pose uncertainty
\State Insert accepted corrected points into the bounded local map
\State Publish pose, map, uncertainty, and localization timestamp
\end{algorithmic}
\end{algorithm}

The resulting localization packet is

\begin{equation}
\mathcal{O}^{\mathrm{lio}}_t
=
\left\{
\vect{x}^{M}_t,
\mathcal{M}_t,
\Sigma^{x}_t,
\tau^{x}_t
\right\},
\label{eq:lio_output}
\end{equation}

\rev{Here,} \rev{$\Sigma^x_t$} \rev{denotes} \rev{a} pose-uncertainty estimate and \rev{$\tau^x_t$} \rev{a} localization timestamp \rev{in the design-level packet}. \rev{Availability} and \rev{publication} \rev{of every packet field are not established by the offline trajectory metrics alone}.

\rev{A} localization-valid flag \rev{could} \rev{require} \rev{a} map, fresh timestamp, and \rev{acceptable} uncertainty. \rev{These} \rev{are} \rev{interface} \rev{requirements,} not \rev{measured} \rev{recovery} or indefinitely drift-free operation\rev{.} \rev{The} \rev{new} \rev{Point-LIO} \rev{experiment} \rev{remains} \rev{offline; navigation uses simulator pose}.

\subsection{\rev{Original bounded intervention-sequence specification}}
\rev{The earlier backup source additionally expresses the intended intervention primitive as four bounded phases:}

\begin{equation}
\begin{aligned}
\textsc{SafePosture}
&\rightarrow \textsc{TargetApproach}\\
&\rightarrow \textsc{DirectedSpray}
\rightarrow \textsc{SafeReturn}.
\end{aligned}
\label{eq:spray_sequence}
\end{equation}

\rev{This original sequence is retained as a design-level contract. The later documented targeting and integrated-mission experiments use separately supplied targets/proposals and do not establish successful execution driven by the learned interface.} It does not measure spray deposition, chemical-use reduction, crop response, or biological-treatment effectiveness.

\clearpage
\FloatBarrier
\section{\rev{Additional verified-result figures and original variants}}
\label{sec:s-added-figures}
\rev{The sensor-placement, external-transfer, offline Point-LIO, climate-response, and targeting/mission figures are presented with the corresponding main-text methods and results. The original qualitative and historical figures are retained in Section S6. This section provides additional historical representations and original variants; no new experiment was run for this reorganization.}
\newmaterialtrue
\begin{revblock}

\end{revblock}
\newmaterialfalse

\begin{figure}[p]\centering
\includegraphics[width=.94\textwidth,height=6.2cm,keepaspectratio]{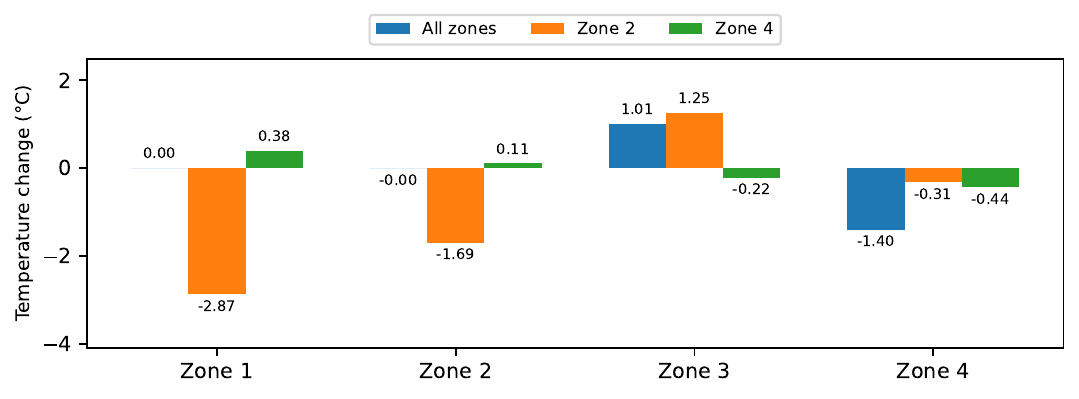}\par\smallskip
\includegraphics[width=.94\textwidth,height=6.2cm,keepaspectratio]{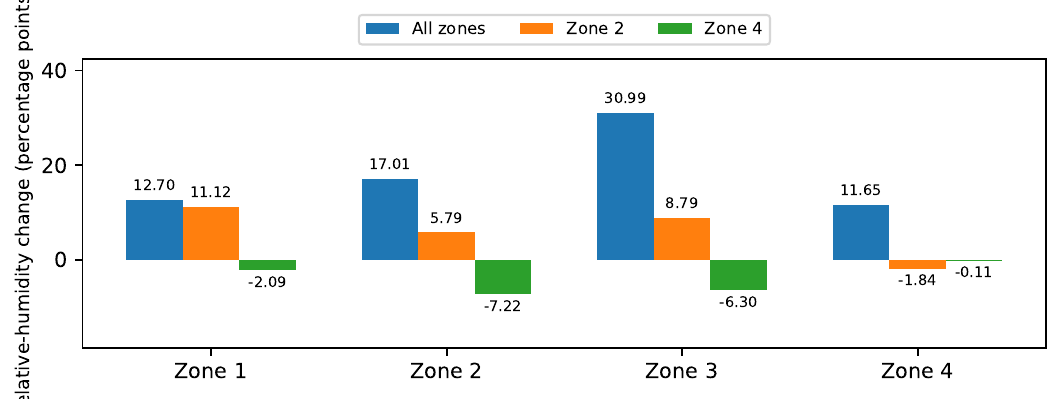}\par\smallskip
\includegraphics[width=.94\textwidth,height=6.2cm,keepaspectratio]{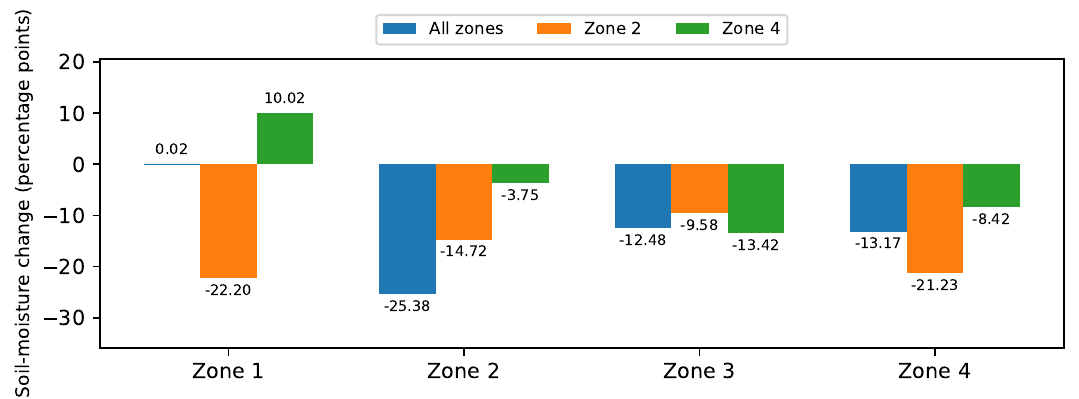}
\caption{\rev{Original three-mission environmental changes, reconstructed from the archived raw CSVs as an additional representation alongside the original delta-map asset, retained in Figure~\ref{fig:environment_delta_map} of this supplementary record. Each bar is the final recorded value minus the initial recorded value for the named zone. Humidity and soil-moisture differences are percentage points, not relative percentage changes. These original state histories remain separate from the added six-case control-response experiment.}}
\label{fig:s-restored-state-deltas}\end{figure}
\subsection{\rev{Original architecture and spatial-grounding variants}}
\rev{The designated reference project also contains a distinct earlier architecture rendition and references a spatial-grounding diagram. The spatial-grounding diagram is retained below; the repeated architecture rendition is omitted because the system overview appears in main Figure 1. The diagrams reflect the original design; they do not establish that every depicted gate or data path was implemented or validated in subsequent component tests.}


%

\clearpage
\FloatBarrier
\section{\rev{Conceptual activity-state and dispatch extensions}}
\label{sec:s-design}
\rev{This section retains the extended architecture specifications from the original manuscript. They are conceptual requirements for future integration, not methods used to obtain the experimental results. Main-text Methods now contains the evaluated neural architecture and component protocols. Parameters such as confidence bounds, reference slew limits and map-consistency thresholds in these formulations require implementation and prospective evaluation. These specifications are retained to preserve the original design contribution without attributing unexecuted methods to the experiments.}

\subsection{\rev{Conceptual design: System architecture, activity state and module interfaces}}
\label{design:sec:architecture}

{
This section describes the internal organization of \AgriSense{} and defines the state variables, interfaces, and execution logic used throughout the paper. The framework is presented as a hierarchical cyber--physical autonomy stack in which sensing, localization, multimodal proposal generation, validation, control, and supervision are separated into explicit layers. This separation is important because greenhouse autonomy requires semantic decision-making to remain coupled to spatial context while preventing direct access from learned models to safety-critical actuators. The section first introduces the physical, edge-autonomy, and supervisory layers; then defines the greenhouse-zone state representation, quadruped sensing and intervention model, activity-state packet, and module-level interfaces used for reproducible execution and post-mission analysis.
}
\paragraph{\rev{Hierarchical System Framework}}
\label{design:subsec:overall_framework}

\AgriSense{} is organized as a hierarchical autonomy stack comprising physical, edge-autonomy, and supervisory layers. \rev{In the virtual experiments, physical-layer elements are represented by simulator assets and state variables; no physical greenhouse installation is used.} The physical layer contains the greenhouse, crop rows, fixed environmental sensors, greenhouse actuators, and the quadruped mobile manipulator. The edge-autonomy layer performs sensor synchronization, LiDAR--IMU localization, zone association, \VLDA{} inference, decision validation, command buffering, navigation, {PI/PID environmental reference execution, and manipulator-action dispatch.} The supervisory layer provides non-real-time visualization, event review, alerts, diagnostics, and operator requests.

Safety-critical sensing, validation, and control remain on the local edge platform. {This local-execution design is consistent with recent edge-computing analyses in smart agriculture, which emphasize reduced latency, lower bandwidth dependence, and improved resilience relative to cloud-only pipelines \citep{sapna2026edgecomputing}.} The supervisory application can request a mission or review a proposed intervention, but it does not directly bypass the edge-side validation and control layers. \rev{This separation is intended to reduce dependence on supervisory connectivity; uninterrupted operation under arbitrary network loss is not established by the present experiments.}

The complete system state at time \(t\) is represented by
\begin{equation}
\mathcal{X}_t =
\left(
I_t,
P_t,
\vect{m}_t,
\vect{q}_t,
\vect{e}_t,
z_t,
\vect{x}_t,
\mathcal{M}_t,
\vect{u}_t
\right),
\label{eq:system_state}
\end{equation}
where \(I_t\) is the robot-view RGB observation, \(P_t\) is the LiDAR point cloud, \(\vect{m}_t\) is the IMU measurement, \(\vect{q}_t\) is the user prompt or mission request, \(\vect{e}_t\) is the greenhouse environmental state, \(z_t\) is the current management zone, \(\vect{x}_t\) is the robot pose in the local map frame, \(\mathcal{M}_t\) is the locally reconstructed greenhouse map, and \(\vect{u}_t\) denotes the active navigation, environmental-control, and intervention commands.

The system follows a proposal--validation--execution structure. The \VLDA{} module produces structured decision proposals, whereas {deterministic navigation, PI/PID reference-tracking control, and predefined manipulator primitives remain responsible for physical execution.} This separation prevents direct semantic-model access to safety-critical actuators.

\paragraph{\rev{Greenhouse Zones, Environmental States, and Control}}
\label{subsec:zones_control}

The greenhouse is partitioned into \(K\) management zones,
\(\mathcal{Z}_1,\ldots,\mathcal{Z}_K\). For zone \(k\), the environmental state is defined as
\begin{equation}
\vect{e}_{k,t}
=
\begin{bmatrix}
T_{k,t} &
H_{k,t} &
S_{k,t}
\end{bmatrix}^{\top},
\label{eq:zone_environment}
\end{equation}
where \(T_{k,t}\), \(H_{k,t}\), and \(S_{k,t}\) denote temperature, relative humidity, and soil moisture, respectively.

The multimodal model receives both global and zone-specific environmental information:
\begin{equation}
\vect{e}_t =
\begin{bmatrix}
T_{g,t} &
H_{g,t} &
S_{g,t} &
\vect{e}_{1,t}^{\top} &
\cdots &
\vect{e}_{K,t}^{\top}
\end{bmatrix}^{\top},
\label{eq:environment_vector}
\end{equation}
where \(T_{g,t}\), \(H_{g,t}\), and \(S_{g,t}\) are greenhouse-level aggregate measurements.

The \VLDA{} module predicts a proposed environmental reference vector
\begin{equation}
\hat{\vect{s}}_t =
\begin{bmatrix}
\hat{\vect{s}}_{1,t}^{\top} &
\cdots &
\hat{\vect{s}}_{K,t}^{\top}
\end{bmatrix}^{\top},
\qquad
\hat{\vect{s}}_{k,t}
=
\begin{bmatrix}
\hat{T}_{k,t} &
\hat{H}_{k,t} &
\hat{S}_{k,t}
\end{bmatrix}^{\top}.
\label{eq:setpoint_vector}
\end{equation}

These values are treated as proposed references rather than direct actuator commands. {\rev{The design specifies range, freshness, feasibility and safety admission before reference tracking; the present experiments do not establish implementation of this entire admission contract.}} For a controlled environmental variable \(y_{k,t}\) with reference \(r_{k,t}\), the tracking error is
\begin{equation}
e_{k,t}^{c}=r_{k,t}-y_{k,t},
\end{equation}
{and a representative feedback command can be written as}
\begin{equation}
u_{k,t}^{c}
=
K_{p}e_{k,t}^{c}
+
K_{i}\int_{0}^{t}e_{k,\tau}^{c}\,d\tau
+
K_{d}\frac{d e_{k,t}^{c}}{dt}.
\label{eq:pid_control}
\end{equation}

Depending on the regulated variable, \(u_{k,t}^{c}\) is converted into bounded irrigation, ventilation, heating/cooling, or humidity-control actuation. This design separates high-level environmental reasoning from low-level closed-loop regulation.

\paragraph{\rev{Quadruped Mobile Manipulator and Onboard Sensing}}
\label{subsec:robot_platform}

The mobile platform consists of a quadruped robot equipped with an onboard RGB camera, 3D LiDAR, IMU, local computing hardware, communication interfaces, and a manipulator-mounted spray end-effector. The quadruped configuration is intended for greenhouse environments that may contain narrow aisles, irregular floor surfaces, soil patches, crop-bed boundaries, irrigation infrastructure, and obstacles that can restrict conventional wheeled platforms.

The RGB camera provides crop-level visual context for inspection and intervention reasoning. LiDAR and IMU measurements support GNSS-denied localization, map construction, obstacle perception, and zone association. The current zone is determined from the local pose and greenhouse map:
\begin{equation}
z_t
=
\mathrm{ZoneAssoc}
\left(
\vect{x}_t,\mathcal{M}_t
\right).
\label{eq:zone_association}
\end{equation}

\rev{The intended inspection workflow acquires a closer crop observation after arrival and admits a predefined spray primitive only after target, base-stability and reachability checks. In the evaluated integrated missions the target and diagnosis are scenario-defined, not inferred autonomously from a measured crop condition; the implemented admission criteria are reported separately.}

The system therefore distinguishes two physical-action scales:

    \textbf{Zone-level regulation}, executed through fixed greenhouse actuators and PI/PID controllers; and
    \textbf{{Plant-level spray intervention}}, {executed locally through the quadruped-mounted manipulator \rev{in the virtual-greenhouse workflow}.}

\paragraph{\rev{Activity-State Representation}}
\label{subsec:activity_state}

To support synchronized reasoning, validation, execution, and post-mission analysis, \AgriSense{} maintains an activity-state record at each decision instant. The activity state is defined as
\begin{equation}
\vect{a}_{t}^{\mathrm{state}}
=
\begin{bmatrix}
\vect{x}_{t}^{\top} &
z_t &
\vect{e}_{t}^{\top} &
\vect{d}_{t}^{\top} &
\rho_t &
\eta_t &
\tau_t
\end{bmatrix}^{\top},
\label{eq:activity_state}
\end{equation}
\Needspace{14\baselineskip}
\begin{samepage}
where the compact decision vector is
\begin{equation}
\vect{d}_t=
\begin{bmatrix}
\hat{z}_t &
\hat{\vect{s}}_{t}^{\top} &
\hat{p}_t &
c_t
\end{bmatrix}^{\top}.
\label{eq:activity_decision_vector}
\end{equation}
Here, \(\hat{z}_t\) is the proposed target zone, \(\hat{\vect{s}}_t\) is the proposed environmental-reference vector, \(\hat{p}_t\) is the pesticide-intervention probability, \(c_t\) is the model confidence, \(\rho_t\) is the validation status, \(\eta_t\) is the current execution mode, and \(\tau_t\) is the packet timestamp.
\end{samepage}

The execution mode belongs to the finite set
\begin{equation}
\begin{aligned}
\eta_t &\in \mathcal{E},\\
\mathcal{E}
&=
\left\{
\begin{array}{ll}
\textsc{Monitor},   & \textsc{Navigate},\\
\textsc{Regulate},  & \textsc{Inspect},\\
\textsc{Intervene}, & \textsc{Fallback}
\end{array}
\right\}.
\end{aligned}
\label{eq:execution_modes}
\end{equation}

The validation variable is defined as
\begin{equation}
\rho_t =
\begin{cases}
1, & \text{valid proposal},\\
0, & \text{otherwise}.
\end{cases}
\label{eq:validation_status}
\end{equation}

\rev{The activity-state logging contract calls for synchronized inputs, outputs, validation outcomes and commands to support event reconstruction and traceability. This does not establish complete availability of those fields in every historical log. The confidence field is external summary metadata, not a fourth learned prediction head.}

\begin{algorithm}[!htbp]
\caption{\rev{Design specification: activity-state packet construction and dispatch}}
\label{alg:activity_state}
\begin{algorithmic}[1]
\Require RGB image \(I_t\), LiDAR cloud \(P_t\), IMU data \(\vect{m}_t\), environmental state \(\vect{e}_t\), local pose \(\vect{x}_t\), map \(\mathcal{M}_t\), and prompt \(\vect{q}_t\)
\State Associate the robot with the current zone:
\[
z_t \leftarrow \mathrm{ZoneAssoc}(\vect{x}_t,\mathcal{M}_t)
\]
\State Construct the multimodal inference packet:
\[
\mathcal{P}_t
\leftarrow
\{I_t,\vect{q}_t,\vect{e}_t,z_t,\vect{x}_t\}
\]
\State Run \VLDA{} inference to obtain
\(\hat{z}_t\), \(\hat{\vect{s}}_t\), \(\hat{p}_t\), and \(c_t\)
\State Check schema, confidence, freshness, numerical range, agronomic feasibility, localization consistency, navigation safety, and manipulator reachability
\If{all required validation conditions are satisfied}
    \State Set \(\rho_t \leftarrow 1\)
    \State Select execution mode \(\eta_t\)
    \State {Dispatch the accepted proposal to navigation, PI/PID reference tracking, or the intervention queue}
\Else
    \State Set \(\rho_t \leftarrow 0\)
    \State Reject the proposal and maintain the previous valid command or safe fallback
\EndIf
\State Store \(\vect{a}_{t}^{\mathrm{state}}\) and the associated sensor, decision, and execution records
\end{algorithmic}
\end{algorithm}

\paragraph{\rev{Original alternative validation-score specification}}
\rev{For traceability, the original design expressed a weighted MAE/F1 validation score as follows. It is retained as an alternative specification, not the selection rule of the saved checkpoint, which minimizes validation loss as documented above.}
\begin{equation}
\mathcal{J}_{\mathrm{val}}
=
\alpha
\frac{\mathrm{MAE}_{s}}{\gamma_s}
+
\beta
\left(
1-F1_{z}^{\mathrm{macro}}
\right)
+
\delta
\left(
1-F1_{p}
\right),
\label{eq:validation_score}
\end{equation}

where \(\gamma_s\) normalizes the setpoint MAE and lower values indicate better overall validation performance. \rev{The historical coefficients are not established for the saved run and no result is attributed to this alternative score.}

\subsection{\rev{Conceptual design: Proposal-validation and dispatch specification}}
\label{design:subsec:safe_dispatch}

The \VLDA{} outputs are treated as structured proposals rather than direct actuator commands. At time \(t\), the proposal packet is

\begin{equation}
\mathcal{D}_t
=
\left\{
\hat{\vect{s}}_t,
\boldsymbol{\pi}^{z}_t,
\hat{z}_t,
\hat{p}_t,
\tau_t
\right\},
\label{eq:decision_packet}
\end{equation}

where \(\tau_t\) is the inference timestamp. 

Let

\begin{equation}
\begin{aligned}
\chi_t^{\mathrm{common}}
={}&
\mathbb{I}\!\left[\mathcal{D}_t\ \text{is schema-valid}\right]
\mathbb{I}\!\left[t-\tau_t\leq\tau_{\max}\right]\\
&\times
\mathbb{I}\!\left[\ell_t=1\right].
\end{aligned}
\label{eq:common_validation}
\end{equation}

where \(\ell_t\) indicates that the current localization and map state are valid. Each action pathway then applies command-specific conditions.

For navigation, the target-zone proposal is accepted only if its class confidence exceeds \(c_z^{\min}\) and the corresponding map goal belongs to the safe-goal set:

\begin{equation}
\rho_t^{\mathrm{nav}}
=
\chi_t^{\mathrm{common}}
\mathbb{I}
\left[
\max_k \pi^{z}_{k,t}
\geq c_z^{\min}
\right]
\mathbb{I}
\left[
g(\hat{z}_t)\in\mathcal{G}_{\mathrm{safe}}
\right].
\label{eq:navigation_validation}
\end{equation}

For environmental regulation, the predicted references must lie within the admissible agronomic and actuator bounds:

\begin{equation}
\rho_t^{\mathrm{reg}}
=
\chi_t^{\mathrm{common}}
\mathbb{I}
\left[
\hat{\vect{s}}_t\in\mathcal{B}
\right]
\mathbb{I}
\left[
\left\|
\hat{\vect{s}}_t-\vect{s}_{t-1}^{\mathrm{acc}}
\right\|_{\infty}
\leq
\Delta s_{\max}
\right],
\label{eq:regulation_validation}
\end{equation}

where \(\vect{s}_{t-1}^{\mathrm{acc}}\) is the most recently accepted reference vector. The second condition prevents abrupt reference changes.

{For plant-level spray intervention, the proposal must exceed the action threshold and satisfy manipulator reachability:}

\begin{equation}
\rho_t^{\mathrm{spray}}
=
\chi_t^{\mathrm{common}}
\mathbb{I}
\left[
\hat{p}_t\geq\theta_p
\right]
\mathbb{I}
\left[
r_t=1
\right],
\label{eq:spray_validation}
\end{equation}

{where \(r_t\) is the reachability and execution-feasibility flag. Additional checks may reject the simulated intervention workflow when the robot has not reached a valid inspection pose or the target region cannot be localized reliably.}

Accepted outputs are dispatched through separate interfaces:

\begin{equation}
\begin{aligned}
\rho_t^{\mathrm{nav}}=1
&\Rightarrow
g(\hat{z}_t)
\rightarrow
\text{navigation-goal queue},
\\
\rho_t^{\mathrm{reg}}=1
&\Rightarrow
\hat{\vect{s}}_t
\rightarrow
{\text{PI/PID reference buffer}},
\\
\rho_t^{\mathrm{spray}}=1
&\Rightarrow
\text{spray primitive}
\rightarrow
\text{manipulator-action queue}.
\end{aligned}
\label{eq:dispatch_paths}
\end{equation}

\rev{This separation specifies that semantic outputs should not directly control motion, environmental actuators or manipulator joints; implementation coverage must be judged from the component-specific tests.}

\begin{algorithm}[!t]
\caption{\rev{Design specification: VLDA inference and command dispatch}}
\label{alg:vlda_dispatch}
\begin{algorithmic}[1]
\Require RGB image \(I_t\), prompt \(\vect{q}_t\), sensor state \(\vect{v}^{\mathrm{sen}}_t\), robot pose \(\vect{x}_t\), and local map \(\mathcal{M}_t\)
\State Normalize continuous sensor channels using training-set statistics
\State Compute \(\vect{h}^{I}_t\), \(\vect{h}^{Q}_t\), and \(\vect{h}^{E}_t\)
\State Fuse the multimodal features to obtain \(\vect{h}_t\)
\State Predict \(\hat{\vect{s}}_t\), \(\boldsymbol{\pi}^{z}_t\), \(\hat{z}_t\), and \(\hat{p}_t\)
\State Convert the regression outputs back to physical units
\State Construct the proposal packet \(\mathcal{D}_t\)
\State Check schema, numerical validity, freshness, and localization status
\If{common validation fails}
    \State Reject \(\mathcal{D}_t\) and retain the previous valid command or safe fallback
\Else
    \State Evaluate \(\rho_t^{\mathrm{nav}}\), \(\rho_t^{\mathrm{reg}}\), and \(\rho_t^{\mathrm{spray}}\)
    \If{\(\rho_t^{\mathrm{nav}}=1\)}
        \State Submit the validated target-zone goal to the navigation queue
    \EndIf
    \If{\(\rho_t^{\mathrm{reg}}=1\)}
        \State Submit the validated references to the PI/PID setpoint buffer
    \EndIf
    \If{\(\rho_t^{\mathrm{spray}}=1\)}
        \State Submit the bounded spray primitive to the intervention queue
    \EndIf
\EndIf
\State Log all predictions, validation outcomes, and dispatched commands
\end{algorithmic}
\end{algorithm}

\subsection{\rev{Conceptual design: Spatial grounding, reference execution and robotic interfaces}}
\label{design:sec:execution}
Greenhouse operation is \rev{represented} without GNSS access. \rev{Tightly} coupled LiDAR--IMU odometry \rev{estimates} motion relative to a local greenhouse map\rev{:} \rev{the} IMU provides high-rate short-term propagation, whereas LiDAR supplies geometric constraints that correct inertial \rev{drift}. The objective of this work is not to introduce a new LIO estimator\rev{. The new quantitative localization experiment uses actual Point-LIO \citep{he2023pointlio} in offline replay}; the \rev{robot} \rev{navigation} \rev{runs} \rev{continue} \rev{to} \rev{consume} \rev{simulator} \rev{pose}.

\rev{The Point-LIO input model uses 16 rings and 200 azimuth samples per ring, 10~Hz scans with 0.1~s rolling acquisition, a 100~Hz body-origin IMU, 35~m range, and a body-aligned LiDAR translation of $[0,0,0.33]$~m. Range, dropout, and inertial perturbations vary over five routes. Replay opens only LiDAR scans and IMU streams; ground truth is opened by a separate evaluator. Collision geometry supplies simulated returns, without a calibrated optical-return, moving-foliage, or physical sensor-error model. RGB and environmental values do not enter this estimator. The standard estimation derivation and retained historical trajectory records are moved to the supplement, together with an explicit distinction between their provenance and the new replay results.}

For downstream navigation, the six-degree-of-freedom robot pose is summarized
as

\begin{equation}
\vect{x}^{M}_t
=
\begin{bmatrix}
\vect{p}^{M\top}_{I,t} &
\vect{\vartheta}^{\top}_t
\end{bmatrix}^{\top},
\qquad
\vect{\vartheta}_t
=
\begin{bmatrix}
\phi_t & \theta_t & \psi_t
\end{bmatrix}^{\top},
\label{eq:local_pose}
\end{equation}

where \(\phi_t\), \(\theta_t\), and \(\psi_t\) denote roll, pitch, and yaw.

\paragraph{\rev{Management-Zone Association and Spatial Grounding}}
\label{design:subsec:zone_grounding}


The greenhouse floor is partitioned into \(K\) management regions
\(\mathcal{Z}_1,\ldots,\mathcal{Z}_K\), defined in the same local map frame
used by the LIO estimator. The horizontal projection of the estimated robot
pose is

\begin{equation}
\Pi(\vect{x}^{M}_t)
=
\begin{bmatrix}
p^{M}_{x,t} &
p^{M}_{y,t}
\end{bmatrix}^{\top}.
\label{eq:floor_projection}
\end{equation}

The current management-zone identifier is represented by \(\zeta_t\):

\begin{equation}
\zeta_t
=
\begin{cases}
k,
&
\Pi(\vect{x}^{M}_t)\in\mathcal{Z}_k,
\\[2mm]
0,
&
\Pi(\vect{x}^{M}_t)
\notin
\displaystyle\bigcup_{k=1}^{K}\mathcal{Z}_k .
\end{cases}
\label{eq:management_zone_membership}
\end{equation}

Here, \(\zeta_t=0\) denotes an entrance, corridor, boundary region, or
location outside the defined management zones. \rev{A boundary tolerance or hysteresis rule is a proposed means of limiting rapid zone switching; its tuning and efficacy are not quantified here.}

Each robot-view observation, environmental measurement, decision output, and
executed command is associated with its estimated pose, zone identifier,
local map, and timestamp:

\begin{equation}
\mathcal{G}_t
=
\left\{
I_t,
\vect{e}_t,
\vect{x}^{M}_t,
\zeta_t,
\mathcal{M}_t,
\tau_t
\right\}.
\label{eq:grounded_observation}
\end{equation}

This association connects the semantic decision layer to physical greenhouse
locations. It enables observations and environmental trends to be linked to
consistent management regions and provides the spatial context required to
validate target-zone navigation requests.

\paragraph{\rev{Validated Environmental-Reference Execution}}
\label{design:subsec:environmental_execution}

Environmental regulation is executed separately from robot locomotion and
manipulation. The \VLDA{} model proposes zone-specific temperature, humidity,
and soil-moisture references, but these references are applied only when the
regulation-validation flag \(\rho_t^{\mathrm{reg}}\), defined in
Section~\ref{design:subsec:safe_dispatch}, is satisfied.

For environmental variable \(j\) in zone \(k\), the accepted reference is

\begin{equation}
r_{k,j,t}
=
\begin{cases}
\hat{s}_{k,j,t},
&
\rho_t^{\mathrm{reg}}=1,
\\
r^{\mathrm{acc}}_{k,j,t-1},
&
\rho_t^{\mathrm{reg}}=0,
\end{cases}
\label{eq:accepted_reference}
\end{equation}

where \(r^{\mathrm{acc}}_{k,j,t-1}\) is the most recently accepted
reference. A rate limiter prevents abrupt changes:

\begin{equation}
\left|
r_{k,j,t}
-
r_{k,j,t-1}
\right|
\leq
\Delta r^{\max}_{k,j}.
\label{eq:reference_rate_limit}
\end{equation}

The tracking error is

\begin{equation}
e^{c}_{k,j,t}
=
r_{k,j,t}
-
y_{k,j,t},
\label{eq:environmental_error}
\end{equation}

where \(y_{k,j,t}\) is the measured environmental state. A PI or PID
controller computes the corresponding actuator request:

\begin{equation}
\tilde{u}^{c}_{k,j,t}
=
K_{p,j}e^{c}_{k,j,t}
+
K_{i,j}
\int_{0}^{t}
e^{c}_{k,j,\tau}\,d\tau
+
K_{d,j}
\frac{d e^{c}_{k,j,t}}{dt}.
\label{eq:environment_pid}
\end{equation}

The resulting physical command is constrained by the actuator limits:

\begin{equation}
u^{c}_{k,j,t}
=
\operatorname{sat}
\left(
\tilde{u}^{c}_{k,j,t};
u^{\min}_{k,j},
u^{\max}_{k,j}
\right).
\label{eq:bounded_environment_command}
\end{equation}

Depending on the controlled variable, this request is converted into bounded
irrigation, ventilation, heating/cooling, or humidity-regulation actuation.
Thus, \VLDA{} determines the desired high-level environmental condition,
whereas deterministic feedback controllers remain responsible for continuous
closed-loop execution.

\paragraph{\rev{Validated Zone Navigation and Quadruped Execution}}
\label{design:subsec:quadruped_navigation}

A target-zone proposal is converted into a navigation request only when
\(\rho_t^{\mathrm{nav}}=1\). In particular, the localization state must be
valid and fresh, the local map must be available, pose uncertainty must remain
within the configured acceptance bound, and the target-zone proposal must
satisfy the semantic and safe-goal checks defined in
Section~\ref{design:subsec:safe_dispatch}.

Each management zone is associated with one or more admissible inspection
goals:

\begin{equation}
\mathcal{G}_k
=
\left\{
\vect{g}_{k,1},
\ldots,
\vect{g}_{k,N_k}
\right\},
\label{eq:zone_goal_set}
\end{equation}

where every candidate goal lies within a traversable aisle or inspection
region. The selected goal is

\begin{equation}
\vect{g}^{\ast}_t
=
\arg\min_{\vect{g}\in\mathcal{G}_{\hat{z}_t}}
J
\left(
\vect{g},
\vect{x}^{M}_t,
\mathcal{M}_t
\right),
\label{eq:goal_selection}
\end{equation}

where \(J(\cdot)\) can account for path length, obstacle clearance, heading
alignment, and inspection-view quality.

The navigation layer computes a collision-free path

\begin{equation}
\Gamma_t
=
\left\{
\vect{w}_{0,t},
\vect{w}_{1,t},
\ldots,
\vect{w}_{N,t}
\right\},
\label{eq:planned_path}
\end{equation}

from the current robot pose to the selected inspection goal. The path is
accepted only when

\begin{equation}
\operatorname{dist}
\left(
\vect{w}_{n,t},
\mathcal{O}_t
\right)
\geq d_{\min},
\qquad
\forall\,\vect{w}_{n,t}\in\Gamma_t,
\label{eq:path_clearance}
\end{equation}

where \(\mathcal{O}_t\) denotes the obstacle and forbidden-region set and
\(d_{\min}\) is the required clearance.

The local planner converts the accepted path into linear and angular velocity
references,

\begin{equation}
\vect{v}^{\mathrm{cmd}}_t
=
\begin{bmatrix}
v^{x}_t &
v^{y}_t &
\omega^{z}_t
\end{bmatrix}^{\top},
\label{eq:velocity_command}
\end{equation}

which are tracked by the quadruped platform's low-level locomotion controller.
The present work evaluates the resulting localization and navigation behavior
at the trajectory level and does not claim a new quadruped gait-generation or
locomotion-control policy.

\rev{The original navigation contract calls for pausing or rejecting a request when localization is stale, the local map is unavailable, pose uncertainty or obstacle-clearance limits are violated, or the goal is unsafe. It specifies retaining a safe state, selecting another admissible inspection goal or requesting supervisory review. These requirements define future fault-injection and recovery experiments.}

\paragraph{\rev{Plant-Level Inspection and Localized Intervention}}
\label{design:subsec:plant_intervention}

\rev{The original workflow specifies a close robot-view observation and a new multimodal proposal after arrival at an inspection pose. This handoff is a future integration requirement.}

The current \VLDA{} implementation predicts whether a localized spray
intervention should be considered, but it does not independently produce a
pixel-level disease mask or a three-dimensional treatment point. Accordingly,
the simulation experiments use a scenario-defined or separately annotated
plant target associated with the inspection task.

Let

\begin{equation}
\vect{p}^{M}_{\mathrm{tar}}
\end{equation}

denote the target point in the local greenhouse/map frame. The point is
transformed into the robot-base frame according to

\begin{equation}
\bar{\vect{p}}^{B}_{\mathrm{tar}}
=
{}^{B}\!T_{M,t}
\bar{\vect{p}}^{M}_{\mathrm{tar}},
\label{eq:target_transform}
\end{equation}

where homogeneous coordinates are denoted by \(\bar{\vect{p}}\), and
\({}^{B}\!T_{M,t}\) is obtained from the current robot pose.

\rev{The original geometric reachability condition is specified as}

\begin{equation}
r_t
=
\mathbb{I}
\left[
\vect{p}^{B}_{\mathrm{tar}}
\in
\mathcal{W}_{\mathrm{arm}}
\right]
\mathbb{I}
\left[
\vect{q}^{\,\mathrm{arm}}_t
\in
\mathcal{Q}_{\mathrm{safe}}
\right],
\label{eq:manipulator_reachability}
\end{equation}

where \(\mathcal{W}_{\mathrm{arm}}\) denotes the reachable manipulator
workspace and \(\mathcal{Q}_{\mathrm{safe}}\) the admissible joint-state set.

Because reachability is already included in the spray-validation condition
\(\rho_t^{\mathrm{spray}}\), the final intervention authorization additionally
requires a sufficiently stationary mobile base:

\begin{equation}
\rho_t^{\mathrm{int}}
=
\rho_t^{\mathrm{spray}}
\,
\mathbb{I}
\left[
\left\|
\vect{v}^{\mathrm{base}}_t
\right\|
\leq
v_{\mathrm{stable}}
\right],
\label{eq:intervention_authorization}
\end{equation}

where \(v_{\mathrm{stable}}\) is the maximum permitted base velocity during
manipulation. \rev{This speed-only expression is a schematic admission condition; the evaluated dwell guard additionally checks upright posture, yaw rate and duration, as reported below.}

Once the intervention is authorized, the manipulator executes a predefined bounded primitive consisting of four stages: safe-posture initialization, target approach, directed spray execution, and return to the safe configuration.

The simulation evaluates command sequencing, geometric reachability, and
integration between navigation and manipulator execution. It does not
evaluate spray deposition, chemical-use reduction, crop response, disease
suppression, or biological treatment effectiveness.

\begin{algorithm}[!t]
\caption{\rev{Design specification: plant-level inspection and admitted intervention}}
\label{alg:plant_intervention}
\begin{algorithmic}[1]
\Require Validated target zone \(\hat{z}_t\), local map
\(\mathcal{M}_t\), robot pose \(\vect{x}^{M}_t\), RGB image \(I_t\),
spray-validation status \(\rho_t^{\mathrm{spray}}\), and target point
\(\vect{p}^{M}_{\mathrm{tar}}\)
\State Select an admissible inspection goal in the validated target zone
\State Navigate to the goal using the current local map
\If{localization is stale or the inspection goal is not reached}
    \State Reject the intervention and maintain a safe robot state
\Else
    \State Stop the mobile base and acquire a close crop-view image
    \State Update the \VLDA{} spray-intervention proposal
    \State Transform \(\vect{p}^{M}_{\mathrm{tar}}\) into the robot-base frame
    \State Check target validity, manipulator workspace, joint limits,
    and base stability
    \If{\(\rho_t^{\mathrm{spray}}=1\) and all execution checks pass}
        \State Move the manipulator from the safe posture to the target-approach pose
        \State Execute the bounded directed-spray primitive
        \State Return the manipulator to its safe posture
        \State Log target, zone, pose, timestamp, and execution status
    \Else
        \State Reject or defer the intervention for supervisory review
    \EndIf
\EndIf
\end{algorithmic}
\end{algorithm}

\paragraph{\rev{Execution Logging and Safe Fallback}}
\label{design:subsec:execution_logging}

\rev{The original logging contract associates accepted or rejected actions with localization, zone, environment, prediction, validation, reference and execution fields. The intended execution record is}

\begin{equation}
\mathcal{E}_t
=
\left\{
\vect{x}^{M}_t,
\zeta_t,
\vect{e}_t,
\mathcal{D}_t,
\rho_t^{\mathrm{nav}},
\rho_t^{\mathrm{reg}},
\rho_t^{\mathrm{spray}},
\vect{u}_t,
\eta_t,
\tau_t
\right\}.
\label{eq:execution_record}
\end{equation}

\rev{The design calls for retaining the last admitted environmental reference, pausing navigation on invalid spatial state, rejecting unsafe motion, and returning the manipulator to a safe configuration when possible. These intended fallback behaviors are distinct from the implemented stop/admission tests; full recovery and restart behavior is not established.}

\paragraph{\rev{Scenario-Instance Definition and Reproducible Execution}}
\label{subsec:scenario_specification}

\rev{Each virtual-greenhouse experiment} is represented as a scenario instance containing the initial robot state, mission request, environmental condition, visual condition, localization configuration, and expected system response. A scenario instance \(i\) is written as

\begin{equation}
\mathcal{S}_i =
\left(
\vect{x}_{0,i},
\vect{q}_i,
z_i^{\mathrm{tar}},
\vect{e}_{0,i},
v_i,
\ell_i,
a_i^{\mathrm{exp}},
r_i
\right),
\label{eq:scenario_instance}
\end{equation}
where \(\vect{x}_{0,i}\) is the initial robot pose, \(\vect{q}_i\) is the mission prompt, \(z_i^{\mathrm{tar}}\) is the requested or expected target zone, \(\vect{e}_{0,i}\) is the initial environmental state, \(v_i\) denotes the visual condition, \(\ell_i\) specifies the localization mode, \(a_i^{\mathrm{exp}}\) is the expected high-level response, and \(r_i\) is the scenario configuration or random seed.

\rev{The scenario schema calls for synchronized sensor streams, trajectories, maps, zone association, predictions, validation outcomes, references and execution status. Completeness depends on the individual retained experiment artifacts; the schema itself is not evidence that every field was logged in every historical scenario.}

\rev{Table~\ref{tab:scenario_instances} retains the scenario specification. The verified campaign protocols below identify the fields and evidence actually evaluated.}

\begin{table*}[!tb]
\centering
\caption{\rev{Virtual-Greenhouse Scenario-Instance Specification}\rev{. These are specification fields and representative values, not a claim that every combination was executed.}}
\label{tab:scenario_instances}
\small
\renewcommand{\arraystretch}{1.16}
\setlength{\tabcolsep}{4pt}

\begin{tabularx}{\textwidth}{
>{\raggedright\arraybackslash}p{2.7cm}
>{\raggedright\arraybackslash}p{4.0cm}
>{\raggedright\arraybackslash}p{4.6cm}
>{\raggedright\arraybackslash}X}
\toprule
\textbf{Field} &
\textbf{Meaning} &
\textbf{Representative values} &
\textbf{Evaluation output} \\
\midrule

Mission prompt
& High-level operator request
& Inspect Zone~2; inspect all zones; regulate dry soil; inspect and treat an affected crop
& Prompt interpretation and task proposal \\

Initial robot pose
& Starting pose in the local greenhouse frame
& Entrance, central aisle, or zone-specific starting location
& Localization initialization and mission repeatability \\

Target zone
& Requested or inferred management zone
& Zones~1--4 or complete-greenhouse inspection
& Target-zone prediction and navigation-goal consistency \\

Environmental condition
& Global and per-zone sensor state
& High humidity, low soil moisture, elevated temperature, or mixed disturbances
& Environmental-reference prediction and PI/PID response \\

Visual condition
& Crop and scene appearance in the RGB input
& Healthy, disease-like, occluded, low-light, or viewpoint-varied
& Inspection priority and intervention proposal \\

Localization mode
& Spatial-grounding configuration
& GNSS denied; LiDAR--IMU localization and local mapping
& Trajectory error, map consistency, and zone association \\

Expected response
& Intended high-level system behavior
& Patrol, navigate, regulate, inspect, spray, verify, or reject
& Validation status and selected execution mode \\

Configuration identifier
& Scenario version and repeatability information
& Mission identifier, disturbance profile, augmentation setting, and random seed
& Reproduction of data, decisions, and evaluation outputs \\

Recorded artifacts
& Synchronized evidence produced during execution
& RGB frames, point clouds, IMU, poses, environmental states, predictions, and command logs
& Offline metrics, trajectory analysis, event review, and failure diagnosis \\

\bottomrule
\end{tabularx}
\end{table*}

\rev{The recovered workflow specifies source sampling with replacement and seed 42, rather than documenting every executed invocation. The following equality retains the original intended metadata-inheritance contract for an augmented sample \(i'\) from parent \(i\); it must not be read as an observed property of all final records:}

\begin{equation}
\left(
\vect{q}_{i'},
z_{i'},
\vect{e}_{i'},
\vect{s}^{*}_{i'},
z^{*}_{i'}
\right)
=
\left(
\vect{q}_{i},
z_{i},
\vect{e}_{i},
\vect{s}^{*}_{i},
z^{*}_{i}
\right),
\label{eq:augmentation_metadata_inheritance}
\end{equation}

while the RGB observation is replaced by the controlled edited image
\(I_{i'}\).

\enlargethispage{\baselineskip}
\rev{The intended provenance schema records the parent, source path, augmentation type, class, prompt, model, quality, fidelity and creation time. These fields are not all recoverable for the final dataset; the empirical reconciliation below therefore takes precedence over the intended inheritance contract.} The original and augmented records are then
combined to form the multimodal dataset used for \VLDA{} model development
and evaluation.

\clearpage
\FloatBarrier

\end{document}